\documentclass[10pt,twocolumn,letterpaper]{article}

\usepackage[dvipsnames, svgnames]{xcolor}
\usepackage{tabularx}
\usepackage{animate}

\usepackage{iccv}
\usepackage{times}
\usepackage{epsfig}
\usepackage{graphicx}
\usepackage{amsmath}
\usepackage{amssymb}

\newcommand{\degree}[1]{${#1}^o$}
\def\360{\degree{360}}
\usepackage{xfrac}
\usepackage{hyphenat}

\DeclareMathOperator{\atantwo}{atan2}

\makeatletter
\DeclareRobustCommand{\atan}{%
  \operatorname{atan}%
  \@ifnextchar2{_}{}%
}
\makeatother

\usepackage{multirow}
\usepackage{graphicx}
\usepackage{booktabs}
\usepackage[caption=false]{subfig}

\usepackage{colortbl}
\definecolor{Yellow}{rgb}{1.0, 1.0, 0.6}
\definecolor{Orange}{rgb}{1.0, 0.8, 0.6}
\definecolor{Red}{rgb}{1.0, 0.6, 0.6}

\newcommand{\first}[1]{\textbf{#1} \cellcolor{Red}}
\newcommand{\second}[1]{#1 \cellcolor{Orange}}
\newcommand{\third}[1]{#1 \cellcolor{Yellow}}

\newcommand{\up}[1]{#1 $\uparrow$}
\newcommand{\down}[1]{#1 $\downarrow$}

\usepackage{float}
\usepackage{caption}

\usepackage[breaklinks=true,bookmarks=false]{hyperref}
\hypersetup{
    colorlinks = true,
    citecolor = {green},
    linkbordercolor = {green},
}

\iccvfinalcopy

\begin{document}

\title{Single-Shot Cuboids: Geodesics-based End-to-end Manhattan Aligned Layout Estimation from Spherical Panoramas}

\author{Nikolaos Zioulis\\
Centre for Research and Technology Hellas\\
Universidad Polit\'{e}cnica de Madrid\\
{\tt\small nzioulis@iti.gr}
\and
Federico Alvarez\\
Universidad Polit\'{e}cnica de Madrid\\
{\tt\small fag@gatv.ssr.upm.es}
\and
Dimitrios Zarpalas\qquad Petros Daras\\
Centre for Research and Technology Hellas\\
{\tt\small \{zarpalas,daras\}@iti.gr}
}
\maketitle

\begin{abstract}
It has been shown that global scene understanding tasks like layout estimation can benefit from wider field of views, and specifically spherical panoramas.
While much progress has been made recently, all previous approaches rely on intermediate representations and postprocessing to produce Manhattan-aligned estimates.
In this work we show how to estimate full room layouts in a single-shot, eliminating the need for postprocessing.
Our work is the first to directly infer Manhattan-aligned outputs.
To achieve this, our data-driven model exploits direct coordinate regression and is supervised end-to-end.
As a result, we can explicitly add quasi-Manhattan constraints, which set the necessary conditions for a homography-based Manhattan alignment module.
Finally, we introduce the geodesic heatmaps and loss and a boundary-aware center of mass calculation that facilitate higher quality keypoint estimation in the spherical domain. 
Our models and code are publicly available at \href{https://vcl3d.github.io/SingleShotCuboids/}{https://vcl3d.github.io/SingleShotCuboids/}.
\end{abstract}

\section{Introduction}
\begin{figure}
    \centering
    \includegraphics[width=\linewidth]{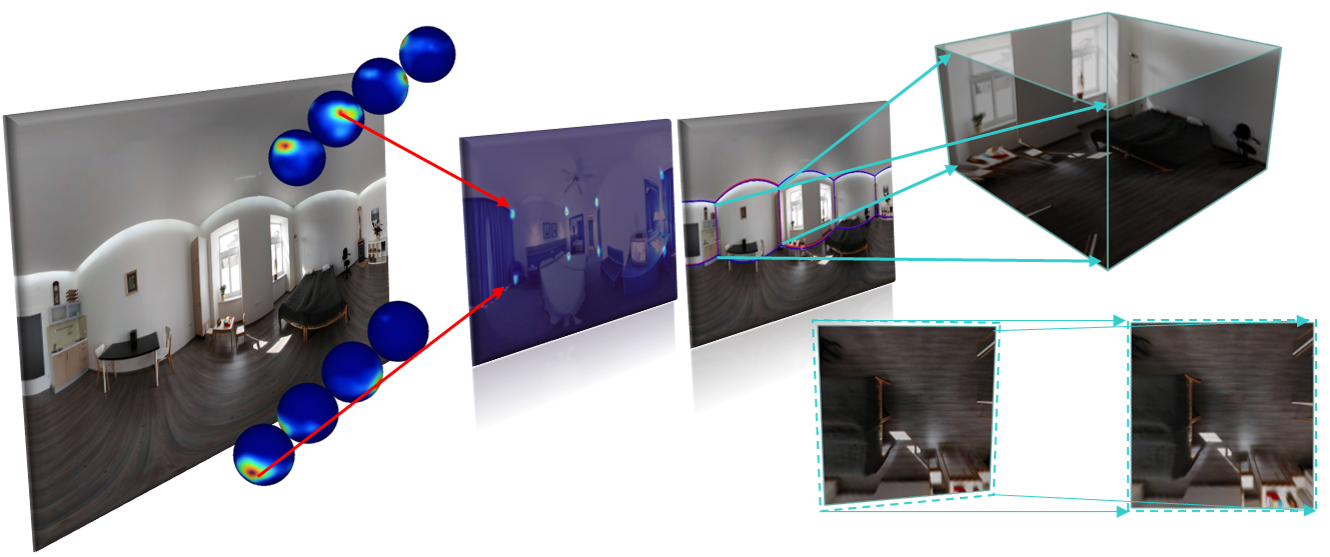}
    \caption{
    From a single indoor scene panorama input, we estimate a Manhattan aligned cuboid of the room's layout, in a single-shot.
    To achieve this, we rely on spherical coordinate localization using geodesic heatmaps.
    This explicit reasoning about the corner positions in the image, allows for the integration of vertical alignment constraints that drive a differentiable homography-based cuboid fitting module.
    }
    \label{fig:intro_concept}
\end{figure}

Modern hardware advances have commoditized spherical cameras\footnote{We will be using the adjective terms spherical, omnidirectional and \360 for cameras and images interchangeably.} which have evolved beyond elaborate optics and camera clusters.
Affordable handheld \360 cameras are finding widespread use in various applications, with the more prominent ones being real-estate, interior design and virtual tours, with recently introduced datasets following the same trends.
Realtor360 \cite{yang2019dula} contains panoramas acquired by a real-estate company, while Kujiale \cite{jin2020geometric} and Structured3D \cite{Structured3D} were rendered using a large corpus of computer-generated data from an interior design company.
Further, datasets containing spherical panoramas like Matterport3D \cite{chang2018matterport3d} and Stanford2D3D \cite{armeni2017joint}, were created using the Matterport camera, originally developed for virtual tours.
This signifies the importance of spherical panoramas for indoor 3D capturing, as they are (re-)used in multiple 3D vision tasks \cite{xia2018gibson, taira2018inloc, zioulis2018omnidepth}.

Spherical panoramas capture the entire scene context within their field\hyp{}of\hyp{}view (FoV), an important trait for scene understanding.
While humans can infer out of FoV information, the same cannot be said for machines, with view extrapolation methods \cite{song2018im2pano3d} using spherical data to address this.
Certain tasks like illumination or layout estimation implicitly extrapolate outside narrow FoVs.
Neural Illumination \cite{song2019neural} estimates a scene's lighting from a single perspective image employing a perspective-to-spherical completion intermediate task within their end-to-end model.
Estimating a scene's layout involves extrapolating structural information, and, thus, many works now resort to spherical panoramas to exploit their holistic structural and contextual information.

The seminal work of PanoContext \cite{zhang2014panocontext}, reconstructs an entire room into a 3D cuboid, fully exploiting the large FoV of omnidirectional panoramas .
Its complex formulation and weak priors resulted in high computational complexity, requiring several minutes for each panorama.
While modern deep priors produce higher quality results \cite{zou2018layoutnet, yang2019dula}, increasing the accuracy of their predictions and ensuring Manhattan-aligned layouts, requires postprocessing and hurts runtime efficiency.

Spherical panoramas necessitate higher resolution processing, and therefore, increased computational complexity, as evidenced by recent data-driven layout estimation models \cite{zou2018layoutnet, yang2019dula, sun2019horizonnet}.
More efficient alternatives \cite{fernandez2020corners} produce irregular (\textit{i.e.}~non-Manhattan) outputs, require parameter sensitive postprocessing, and increase efficiency by lowering spatial resolution, which comes at the cost of accuracy.
Moreover, data-driven spherical vision needs to address the distortion of the projective omnidirectional data formats.
But distortion mitigating convolutions add a significant computational overhead as reported in \cite{fernandez2020corners} and \cite{eder2019mapped}.

In this work, we present a single-shot spherical layout estimation model. 
As presented in Figure~\ref{fig:intro_concept}, we employ spherical-aware corner coordinate estimation and thus, add explicit constraints that facilitate vertically aligned corners.
Capitalizing on this, we further integrate full Manhattan alignment directly into the model, allowing for end-to-end training, lifting the postprocessing requirement.
\label{sec:intro}

\section{Related Work}
\label{sec:related}
\subsection{Layout Estimation}
\label{sec:layout}
While an excellent review regarding the 3D reconstruction of structured indoor environments exists \cite{pintore2020state}, our discussion will provide the necessary details for positioning our work.
We focus on monocular layout estimation and thus, refrain from discussing works using multiple panoramas \cite{pintore2016omnidirectional, pintore2018recovering, pintore20183d, pintore2019automatic}, interaction \cite{liu2018interactive}, other types of cameras \cite{lee2017roomnet,li20193d}.

PanoContext \cite{zhang2014panocontext} showcased the expressiveness of \360 panoramas in terms of structural and contextual information.
Prior to the maturation of deep data-driven methods, PanoContext relied on edge and line detection, Hough transform, and deformable part models to generate different room layout hypotheses.
Similarly, low-level line segments were used in an energy minimization formulation to estimate a scene's structural planes \cite{fukano2016room}.
In Panoramix \cite{yang2016efficient}, the line features were supplemented by superpixel facets, and embedded as vertices in a graph for a constrained least squares problem.

Hybrid data-driven methods \cite{fernandez2018layouts} used structural edge detection to improve the performance and runtime of \cite{zhang2014panocontext} when using fewer hypotheses.
Pano2CAD \cite{xu2017pano2cad} used a probabilistic formulation that relied on CNN object recognition and detection.
It generated a synthetic scene reconstruction but required several minutes of processing.
Its computational overhead largely comes from the fusion of narrow FoV predictions from perspective \360 crops.
This is common to all aforementioned methods relying on line segments and to \cite{yang2018automatic}, which runs various CNNs on all narrow FoV sub-views before merging them in \360.

PanoRoom \cite{fernandez2018panoroom} and LayoutNet \cite{zou2018layoutnet} were the first models to be trained on spherical panoramas.
They both modelled layout corner and structural edge estimation as a spatial probabilistic inference task.
While it is possible to extract the layout's corners by relying on heuristically or empirically parameterized peak detection, these estimations will most likely not deliver Manhattan-aligned outputs.
Consequently, joint optimization is performed using both sources of information to recover the final layout corner estimates.
LayoutNet requires several seconds to infer and optimize the layout on a CPU, but PanoRoom is much faster as it uses a greedy RANSAC approach.

DuLa-Net \cite{yang2019dula} employs a novel approach for \360 layout estimation.
The main insight is that spherical images can be projected in multiple ways, and different projections highlight different cues.
Specifically, DuLa-Net uses a `ceiling-view' that offers a more informative viewpoint with respect to the floor-plan, which is a projection of a Manhattan 3D layout.
It performs feature fusion across both the equirectangular and ceiling-view branches, using a height prediction to estimate the final 3D layout.
HorizonNet \cite{sun2019horizonnet} is yet another novel take at omnidirectional layout estimation.
Instead of image localised predictions, it encodes the boundaries and intersections in one-dimensional vectors, which are then used to reconstruct the scene's corners.
This allows HorizonNet to exploit the expressiveness of recurrent models (LSTM \cite{hochreiter1997long}) to offer globally coherent predictions.
After a postprocessing step involving peak detection and height optimization, the final Manhattan-aligned layout is computed.
A recent thorough comparison between LayoutNet, DuLa-Net and HorizonNet was presented in \cite{zou20193d}.
Unified encoding models and training scripts were used to fairly evaluate these approaches.
Their findings indicate that the PanoStretch data augmentation proposed in \cite{sun2019horizonnet}, as well as its heavier encoder backbone lead to improved performance for the other models as well.
The Corners-for-Layout (CFL) \cite{fernandez2020corners} model is currently the most efficient approach for \360 layout estimation in terms of runtime, but at the expense of accuracy and Manhattan alignment.
While an end-to-end model is discussed, an empirically or heuristically parameterized postprocessing image peak detection step is still required.

Compared to these approaches, our model is end-to-end trainable, producing Manhattan aligned corners in a single-shot.
We approach the layout estimation task as a keypoint localization one and use an efficiently designed spherical model.

\subsection{Learning on the Sphere}
\label{sec:sconv}
There are multiple representations for spherical images with the more straightforward being the cube-map.
Traditional CNN models can be applied to the cube faces \cite{monroy2018salnet360}, and then warped back to the sphere.
This was used in \cite{zhang2014panocontext} and \cite{yang2016efficient} to detect lines on each cube's faces \cite{von2012lsd}, while \cite{xu2017pano2cad} and \cite{yang2018automatic} used CNN inference on each face.
Still, cube-maps suffer from distortion as well, and additionally require face-specific padding \cite{cheng2018cube} to deal with the faces' discontinuities.
Yet, to capture the global context these approaches need to expand their receptive field to connect all faces continuously, which leads to inefficient models. 

A novel line of research pursues model adaptation from the perspective domain to the equirectangular one \cite{su2017learning}.
The follow-up work, Kernel Transformer Networks \cite{su2019kernel}, adapt traditional kernels to the spherical domain in a learned manner, also discussing two important aspects.
First, the accuracy-resolution trade-off for spherical images, which necessitates the user of higher resolutions.
Indeed, most aforementioned data-driven layout estimation methods from \360 images operate on $1024 \times 512$ images, which are unusually large for CNNs.
Only \cite{fernandez2020corners} is the exception to this rule, which further supports this point, taking into account its reduced performance.
The second point of discussion is related to the effect that non-linearities have, when combined with kernel projection methods like \cite{coors2018spherenet} and \cite{tateno2018distortion}.
It is shown that the assumption that needs to hold for no error to accumulate when using kernel projection, only holds for the first layers of the network, and as it deepens, the accumulated error becomes even larger.
Still, \cite{fernandez2020corners} shows that their EquiConv offer more robust predictions.
A generalization of this concept, Mapped Convolutions \cite{eder2019mapped}, decouple the sampling operation from the filtering one, and demonstrate increased performance in dense estimation tasks.
Still, runtime performance is greatly reduced as reported in both \cite{fernandez2020corners} and \cite{eder2019mapped}.

This is also the main drawback of frequency-based spherical convolutions as presented in the concurrent works of \cite{cohen2018spherical} and \cite{Esteves_2018_ECCV}.
They are also highly inefficient in terms of memory, allowing for training and inference in very low resolution images only.
DeepSphere \cite{deepsphere_ml} and \cite{khasanova2017graph} present another approach to handle distortion and discontinuity by leveraging graph convolutions and lifting the sphere representation to a graph.
Nonetheless, this requires a graph generation step and loses efficacy compared to traditional convolutions, whose implementations are highly optimized to exploit the memory regularity of image representations.

The most efficient way to handle the discontinuity is circular padding \cite{wang2018omnidirectional,sun2019horizonnet,da2019dense}, which is partly our approach as well, taking into account the inefficiency of distorted kernels.
It should also be noted that model adaptation methods would not transfer well for the layout estimation task.
While an object detection task parses a scene in a local manner, layout estimation requires to reason about the global context, with perspective methods typically needing to extrapolate the scene's structure.
However, as first proven by PanoContext \cite{zhang2014panocontext}, the availability of the entire scene is much more informative, and this would hinder the applicability of transferring models like RoomNet \cite{lee2017roomnet} to the \360 domain using such techniques \cite{su2017learning,su2019kernel}.

\subsection{Coordinate Regression}
\label{sec:coord}
Regressing coordinates in an image has been shown to be an intriguingly challenging problem \cite{liu2018intriguing}.
The proposed solution was to offer the coordinate information explicitly.
Yet, most keypoint estimation works in the literature initially used fully connected layers to regress coordinates.
The counter-intuition is that convolutions are inherently spatial, and should be more well-behaved in spatial prediction tasks.
This is how data-driven layout estimation models have addressed this problem up to now (\cite{zou2018layoutnet}, \cite{fernandez2020corners}), transforming coordinates into spatial configurations, using smoothing kernels to approximate coordinates, and leverage dense supervision.
Keypoint localisation tasks with semantic inter\hyp{}correlated structures, typically use one heatmap per keypoint.
However, an issue that has recently received attention \cite{zhang2020distribution}, is the way the final coordinate is estimated from each dense prediction.
Indeed the spatial maxima might not always best approximate the coordinate, and thus, heuristic approaches have persisted.
Specifically for layout estimation, where the corners are predicted on the same map, manually-set peak detection thresholds are used. 

The overlapping works of \cite{luvizon2019human}, \cite{sun2018integral} and \cite{nibali2018numerical} derive smooth operations to reduce a heatmap to single a coordinate.
Using the coordinate grid and a spatial \textit{softmax} function, they smoothly, and differentiably, transform a spatial probabilistic representation into a single location.
As shown in \cite{8892980}, all the above operations are treating pixels as particles with masses, and estimate their center of mass. 

\section{Single-Shot Cuboids}
\label{sec:method}
Unlike previous works, we approach layout estimation as a keypoint localisation task, alleviating the need for postprocessing and simultaneously ensure Manhattan aligned outputs.
Section~\ref{sec:scom} formulates our coordinate regression objective and its adaption to the spherical domain, Section~\ref{sec:geodesic} introduces the geodesic heatmaps and loss function and then, Section~\ref{sec:net} provide insights into our model's design, and the techniques to achieve end-to-end Manhattan alignment.

\subsection{Spherical Center of Mass}
\label{sec:scom}
The center of mass (CoM) $\mathbf{c}_{\mathcal{P}}$ for a collection of particles $\mathcal{P}: \{\mathbf{p}_0, \dots , \mathbf{p}_N\} \in \mathbb{R}^3$ is defined as:
\begin{equation}
\label{eq:com}
    \mathbf{c}_{\mathcal{P}} = \frac{\sum_i^N m_i \mathbf{p_i}}{M}, \qquad M = \sum_i^N m_i,
\end{equation}
with $m_i$ being the mass of particle $\mathbf{p}_i$ and $M$ the system's total mass.
The CoM $c_{\mathcal{P}}$ represents a concentration of the particle system's mass and does not necessarily lie on an existing particle.
This way, when considering a sparse keypoint estimation task in a structured grid, we can reformulate it as a dense prediction task by instead inferring the mass of each grid point.
Using Eq.~(\ref{eq:com}) we can directly supervise it with the keypoint coordinates, instead of relying on a surrogate objective as commonly done in pose estimation \cite{zhang2020distribution} or facial landmark detection \cite{feng2018wing}. 

For spherical layout estimation, the set of particles $\mathcal{P}$ for which we seek to individually estimate their per particle mass, lies on a sphere.
Each layout corner is considered as the CoM of a distinct particle system defined on the sphere.
Each particle $\mathbf{p} = (\phi, \theta)$ on the sphere is represented by its longitude $\phi$ and latitude $\theta$.
While there are ways for learning directly on the 2-sphere $S^2$ manifold, as explained in Section \ref{sec:sconv}, they are very inefficient.
Consequently, we consider the equirectangular projection of the sphere which preserves the angular parameterization of each particle.
The equirectangular projection is an equidistant planar projection of the sphere, where the pixels in the image domain $\Omega: (u, v) \in [0, W] \times [0, H]$ are linearly mapped to the angular domain\footnote{We transition between these terms flexibly given their linear mapping.} $\mathcal{A}: (\phi, \theta) \in [0, 2\pi] \times [0, \pi]$.
Nevertheless, this format necessitates a different approach to overcome its weaknesses, namely, image boundary discontinuity, and planar projection distortion.

\begin{figure}
    \centering
    \includegraphics[width=\linewidth]{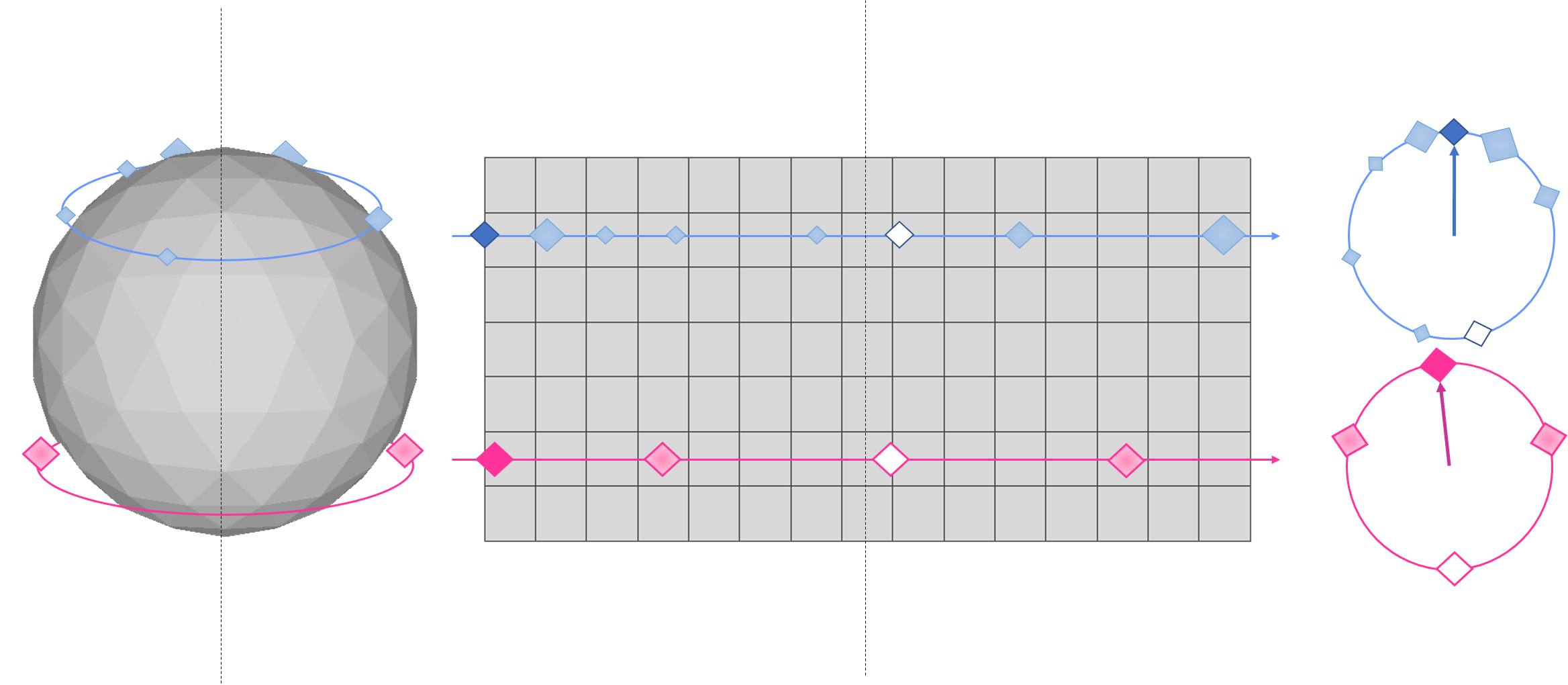}
    \caption{
    Spherical Center of Mass calculation.
    \textbf{Left}: Two sets of particles distributed on two circles of latitude (\textcolor{Cerulean}{blue} and \textcolor{HotPink}{pink}).
    \textbf{Middle}: Their equirectangular projection grid coordinates.
    \textbf{Right}: Lifting the problem to the unit circle allows for continuous CoM estimation.
    Darker points illustrate the CoMs calculated using our lifting approach, and white ones the erroneous estimates when directly estimating CoM on the grid.
    }
    \label{fig:scom}
\end{figure}

The discontinuity arises at the horizontal panorama boundary, where the particles, even though at the opposite sides of the image, are actually neighboring on the sphere.
For traditional images, the (normalized) grid coordinates are typically defined in $[0, 1]$ or $[-1, 1]$, and thus, the particles at the boundary would be maximally distant.
However, for spherical panoramas, the longitudinal coordinate $\phi$ is periodic and wraps around, with the particles at the boundaries being proximal (\textit{i.e.}~minimally distant).
To address this, we split the CoM calculation for the longitude and latitude coordinates, and adapt the former to consider each point as lying on a circle.
Therefore, for each panorama row, which represents a circle of (equal) latitude, we define new particles $\mathbf{r} \in \mathcal{R}$ with
\begin{equation}
\label{eq:periodic}
    \mathbf{r}(\phi) = (\lambda, \tau) = (\cos\phi, \sin\phi),
\end{equation}
while lie on a unit circle.
We can then calculate the CoM $\mathbf{c}_{\mathcal{R}}$:
\begin{equation}
\label{eq:pcom}
    \mathbf{c}_{\mathcal{R}} = (\bar{\lambda}, \bar{\tau}) = \frac{\sum_i^N m_i \mathbf{r}_i}{M}.
\end{equation}
This estimates, exactly and continuously, the CoM of the circle.
To map this back to the original domain, we extract the angle $\bar{\phi}$:
\begin{equation}
    \bar{\phi} = \atantwo(-\bar{\tau}, -\bar{\lambda}) + \pi,
\end{equation}
which represents the longitudinal CoM across the discontinuity.
Figure~\ref{fig:scom} shows a toy example of CoM calculations along two circles of latitude on the sphere, with the erroneous estimates acquired on the equirectangular projection and the correct ones when considering the boundary.

Although the equirectangular projection maps circles of latitude (longitude) to horizontal (vertical) lines of constant spacing, the same does not apply for its sampling density.
Indeed, while it samples the sphere with a constant density vertically, it stretches each circle of latitude to fit the same constant horizontal line.
Thus, its sphere sampling density is not uniform in all planar pixel locations.
The sampling density is $1 / \sin\theta$ \cite{sun2017weighted} and it approaches infinity near the pole singularities.
When calculating the CoM in the equirectangular domain, we need to compensate for it by re-weighting the contribution of each pixel $\mathbf{p}$ by $\sigma(\mathbf{p}) = \sin\theta$ \cite{zioulis2019spherical}.

Essentially, given a dense mass prediction $\mathbf{M}(\mathbf{p}), \mathbf{p} \in \mathcal{A}$, we calculate the spherical CoM by first estimating a three-dimensional coordinate $\mathbf{c}_a$:
\begin{equation}
\label{eq:scom}
    \mathbf{c}_a = (\bar{\lambda}, \bar{\tau}, \bar{\theta}) = \frac{\sum_\mathbf{p}^\mathcal{A} \, \mathbf{M}(\mathbf{p}) \, \sigma(\mathbf{p}) \, \mathbf{a}(\mathbf{p})}{\sum_\mathbf{p}^\mathcal{A} \, \mathbf{M}(\mathbf{p}) \,\sigma(\mathbf{p})}, 
\end{equation}
with $\mathbf{a}(\mathbf{p}) = (\mathbf{r}(\phi), \theta) = (\cos\phi, \sin\phi, \theta)$, and then drop it to the two-dimensions again to calculate the final CoM $\mathbf{c}_m = (\bar{\phi}, \bar{\theta}) = (\atantwo(-\bar{\tau}, -\bar{\lambda}) + \pi, \theta)$ of $\mathbf{M}$ in the equirectangular domain.

\subsection{Geodesic Heatmaps}
\label{sec:geodesic}
\begin{figure}
    \centering
    \subfloat{\includegraphics[width=0.19\linewidth]{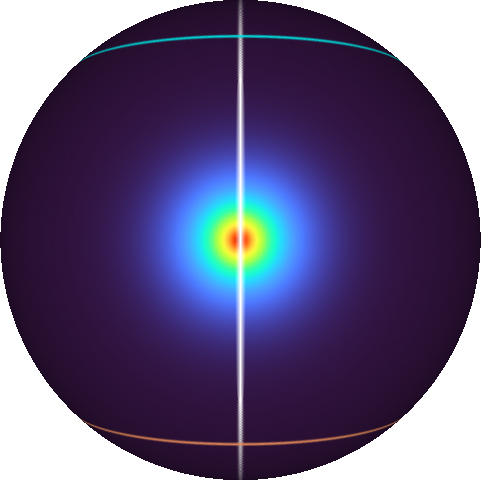}}
    \hfill
    \subfloat{\includegraphics[width=0.19\linewidth]{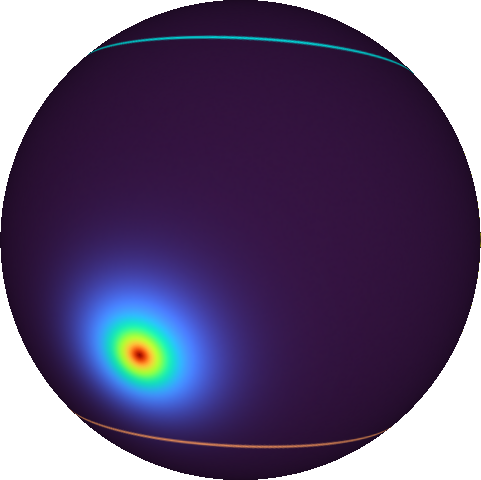}}
    \hfill
    \subfloat{\includegraphics[width=0.19\linewidth]{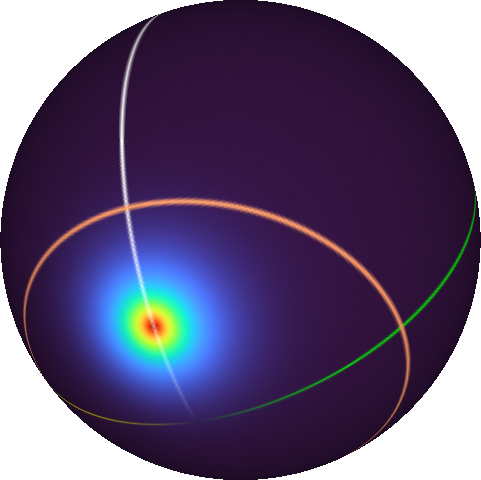}}
    \hfill
    \subfloat{\includegraphics[width=0.19\linewidth]{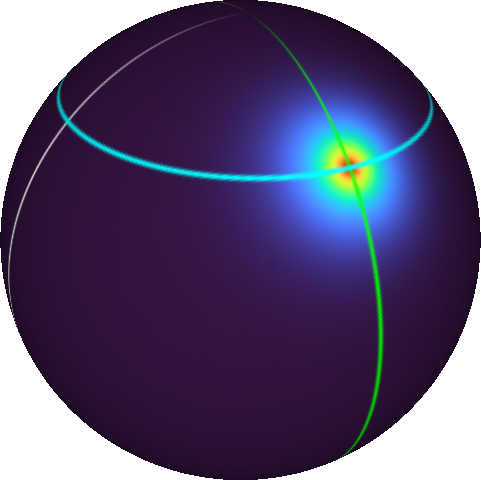}}
    \hfill
    \subfloat{\includegraphics[width=0.19\linewidth]{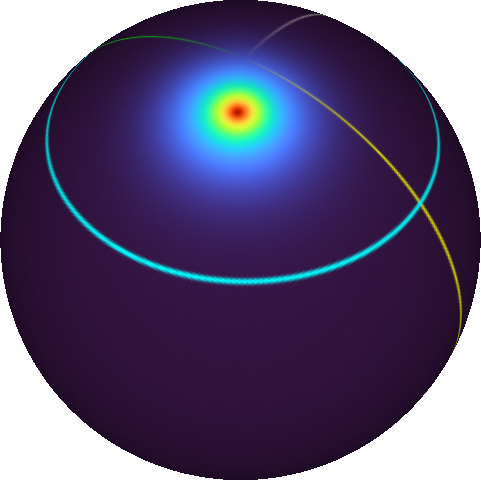}}
    \hfill
    \includegraphics[width=\linewidth]{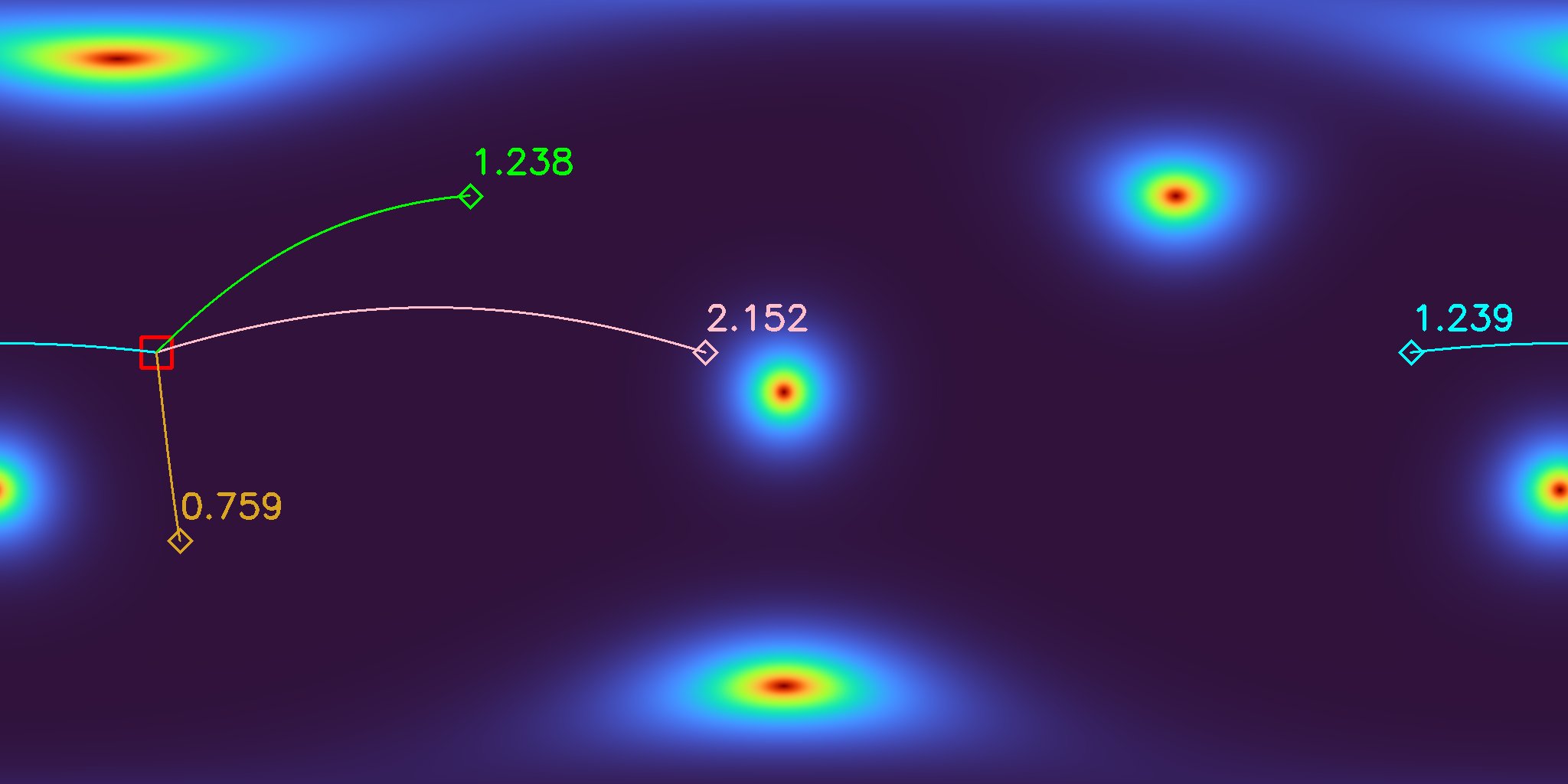}
    \caption{
    Geodesic heatmaps respect the horizontal boundary continuity and the equirectangular projection's distortion.
    Five normal distributions on the sphere centered around different coordinates but using the same angular standard deviation are presented on the top row.
    Their corresponding geodesic heatmaps are aggregated on the equirectangular image on the bottom row.
    In addition, the geodesic distance between the \textcolor{red}{red} square and the colorized diamond coordinates are also presented on the same image.
    The geodesic distance similarly respects the boundary and distortion of the equirectangular projection as seen by the great circles drawn on the image that correspond to each pair's angular distance.
    }
    \label{fig:geodesic}
\end{figure}

Accordingly, predicting the sparse coordinates of a corner comes down to predicting the dense mass map $\mathbf{M}$, or otherwise heatmap, which is the terminology we will be using hereafter.
Previous approaches complemented the sparse objective with a dense regularisation term \cite{nibali2018numerical}.
The reason was that CoM regression is not constrained in any way as to the shape of its dense prediction.
This was addressed by adding a distribution loss over the predicted heatmap and a Gaussian centered at the groundtruth coordinate.

Yet while extracting the CoM, as presented in Section \ref{sec:scom}, takes the spherical domain into account, traditional (flat) Gaussian heatmaps do not.
A spatial normal distribution $\mathcal{N}(\mathbf{c}, \mathbf{s})$ centered around a coordinate $\mathbf{c} = (u, v)$, using a standard deviation $\mathbf{s} = (s_x, s_y)$ would consider the equirectangular image as a flat one, with a discontinuous boundary and no distortion.

To overcome this, we construct geodesic heatmaps, which are reconstructed directly on the equirectangular domain using a shifted angular coordinate grid $\mathcal{A}_s$\footnote{$\phi$ and $\theta$ are shifted by $-\pi$ and $-\pi / 2$ respectively.} defined on the panorama:
\begin{equation}
\label{eq:geod_recon}
    \mathcal{G}(\mathbf{c}_m, \alpha) = \frac{1}{\alpha\sqrt{2\pi}}e^{\frac{-\mathit{g}(\mathbf{c}_m, \mathbf{p}_s)}{2\alpha^2}}, \mathbf{p}_s \in \mathcal{A}_s,
\end{equation}
where $\alpha$ is the angular standard deviation around the distribution's center $\mathbf{c}_m$, and $\mathit{g}(\cdot)$ is the geodesic distance:
\begin{equation}
\label{eq:geod_dist}
    \mathit{g}(\mathbf{p}_1, \mathbf{p}_2) = 2\arcsin\sqrt{\scriptstyle \sin^2\frac{\Delta\theta}{2} + \cos\theta_1\cos\theta_2\sin^2\frac{\Delta\phi}{2}},
\end{equation}
where $\Delta\phi = \phi_1 - \phi_2$ and $\Delta\theta = \theta_1 - \theta_2$.
As illustrated in Figure~\ref{fig:geodesic}, using the geodesic distance between two angular coordinates on the equirectangular panorama, we reconstruct geodesic heatmaps that simultaneously take into account both the continuous boundary, as well as the projection's distortion.

\subsection{End-to-end Manhattan Model}
\label{sec:net}
\begin{figure*}[!htbp]
    \includegraphics[width=\textwidth]{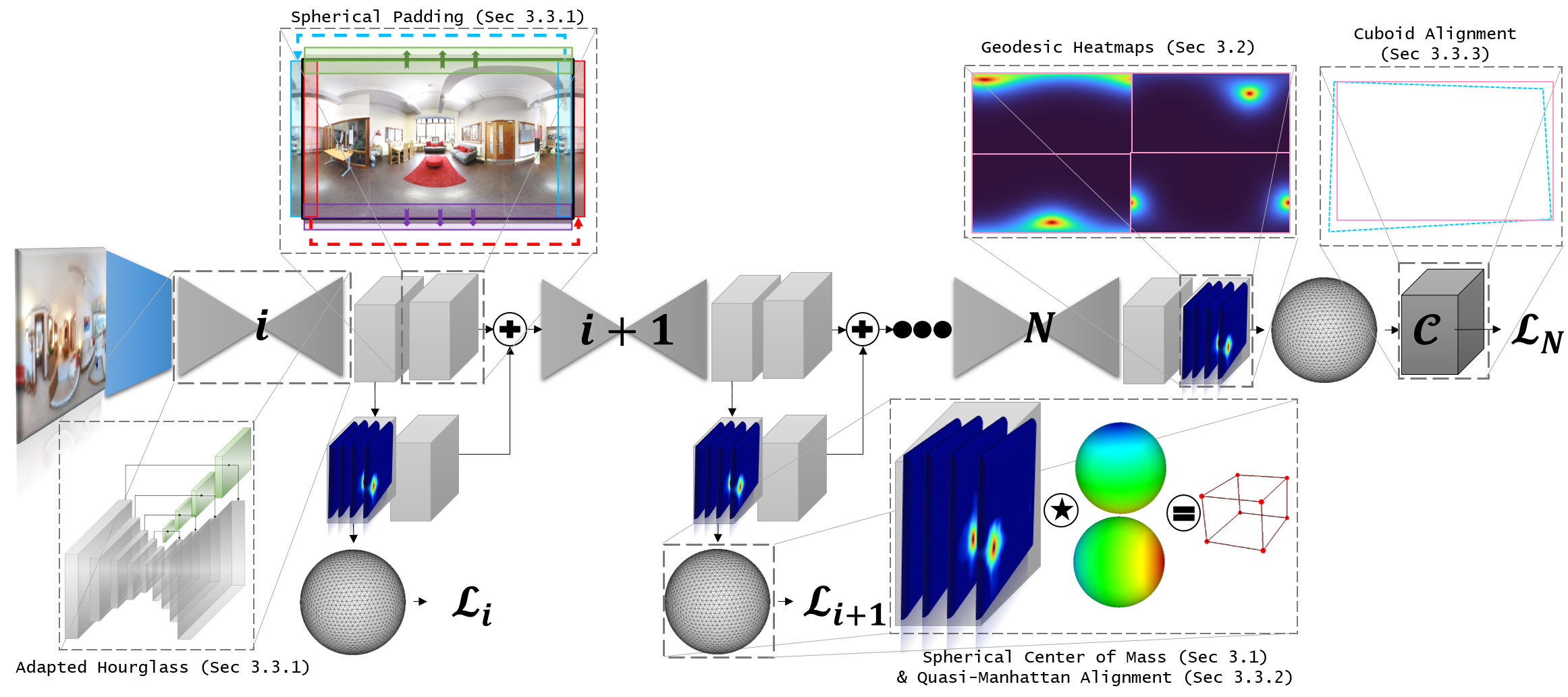}
    \caption{
    Our model stacks $N$ hourglasses which embed recently developed CNN modules for direct inter-hourglass information flow, spherically padded convolutions, and smoother multi-scale feature flow.
    The predicted geodesic heatmaps get transformed directly to panoramic layout coordinates through a spherical CoM module. %
    Since we regress coordinates, we explicitly enforce quasi-Manhattan alignment.
    This sets the ground for a homography-based cuboid alignment head that ensures the Manhattan alignment of our estimates.
    The $\bigstar$ symbol denotes a global multiply-accumulate operation, reducing the predicted dense representation to a set of sparse coordinates.
    Color-graded spheres indicate coordinate-based distance from the origin.
    }
    \label{fig:model}
\end{figure*}

Our model infers a set of heatmaps $\mathbf{M}^j$, one for each layout corner $j \in [1, J]$ (or junction, given that 3 planes intersect), with $J=8$ for cuboid layouts.
It operates in a single-shot manner, as these predictions are directly mapped into layout corners $\mathbf{c}_{m}^{j}$.
Apart from removing the post-processing step, another advantage of our single-shot approach is the sub-pixel level accuracy that it allows for, 
as the CoM of the particles is not necessarily one of the particles themselves.
This translates to a reduction of the input and working resolution of the model.

We choose a light-weight stacked hourglass (SH) architecture \cite{newell2016stacked}.
It is designed for multi-scale feature extraction and merging, that enables the effective capturing of spatial context.
It suits spherical layout estimation very well as it is a global scene understanding task that benefits from spatial context aggregation, which is achieved by lowering the spatial dimension of the features.
Still, it also requires precise localisation of specific keypoints, which needs higher spatial fidelity, (\textit{i.e.}~resolution) predictions.

\subsubsection{Stacked Hourglass Model Adaptation}
\label{sec:model_adaptation}
We made several modifications to the original SH model stemming mainly from recent advances made in the field.
While we preserve the original residual block \cite{he2016deep} in the feature preprocessing block, we replace the hourglass residual blocks with preactivated ones \cite{he2016identity}.
Essentially, this adds direct identity mappings between the stack of hourglasses, allowing for immediate information propagation from the output to the earlier hourglass modules.
We also use anti-aliased max-pooling \cite{zhang2019making}, which preserves shift equivariance and leads to smoother activations across downsampled layers.
Finally, unlike some state-of-the-art spherical layout estimation methods \cite{zou2018layoutnet, yang2019dula, zou20193d}, we address feature map discontinuity by using spherical padding.
For the horizontal image direction, we apply circular padding, as also done in \cite{wang2018omnidirectional} and \cite{sun2019horizonnet}, and for the vertical one at the pole singularities, we resort to replication padding.

\subsubsection{Quasi-Manhattan Alignment}
\label{sec:quasi_manhattan}
Since we are directly regressing coordinates, we can explicitly ensure quasi-Manhattan alignment during training and inference alike.
Previous approaches either use post-processing to ensure the Manhattan alignment of their predictions \cite{zou2018layoutnet, yang2019dula, sun2019horizonnet}, or simply forego it and produce non-Manhattan outputs \cite{fernandez2020corners}.
While this relaxation is sometimes presented as an advantage, most man-made environments are Manhattan-aligned, with walls being orthogonal to ceiling and floors, and therefore, same edge wall corners are vertically aligned.
For each wall-to-ceiling junction, there exists a wall-to-floor junction, effectively splitting our heatmaps in two groups, the \textit{top} $\mathbf{M}^j_t$ and \textit{bottom} $\mathbf{M}^j_b$ heatmaps (\textit{i.e.}~ceiling and floor junctions respectively). 
We enforce quasi\hyp{}Manhattan alignment by averaging the longitudinal coordinates of each wall's vertical edge, guaranteeing a consistent longitudinal coordinate for both the top and bottom junction.

\subsubsection{Homography-based Full Manhattan Alignment}
\label{sec:full_manhattan}
This quasi-Manhattan alignment ensures that wall edges are vertical to the floor, but does not enforce their orthogonality.
To achieve this, we introduce a differentiable operation that 
transforms the predicted corners so as to ensure the orthogonality between adjacent walls.
While the estimated corners are up-to-scale, with a single center-to-floor/ceiling measurement/assumption we can extract metric 3D coordinates for each corner as in \cite{zhang2014panocontext}\footnote{See the supplementary material \cite{xiao20123d}}, by fixing the ceiling/floor vertical distance to the corresponding average height.

\begin{figure}[h!]
    \includegraphics[width=\linewidth]{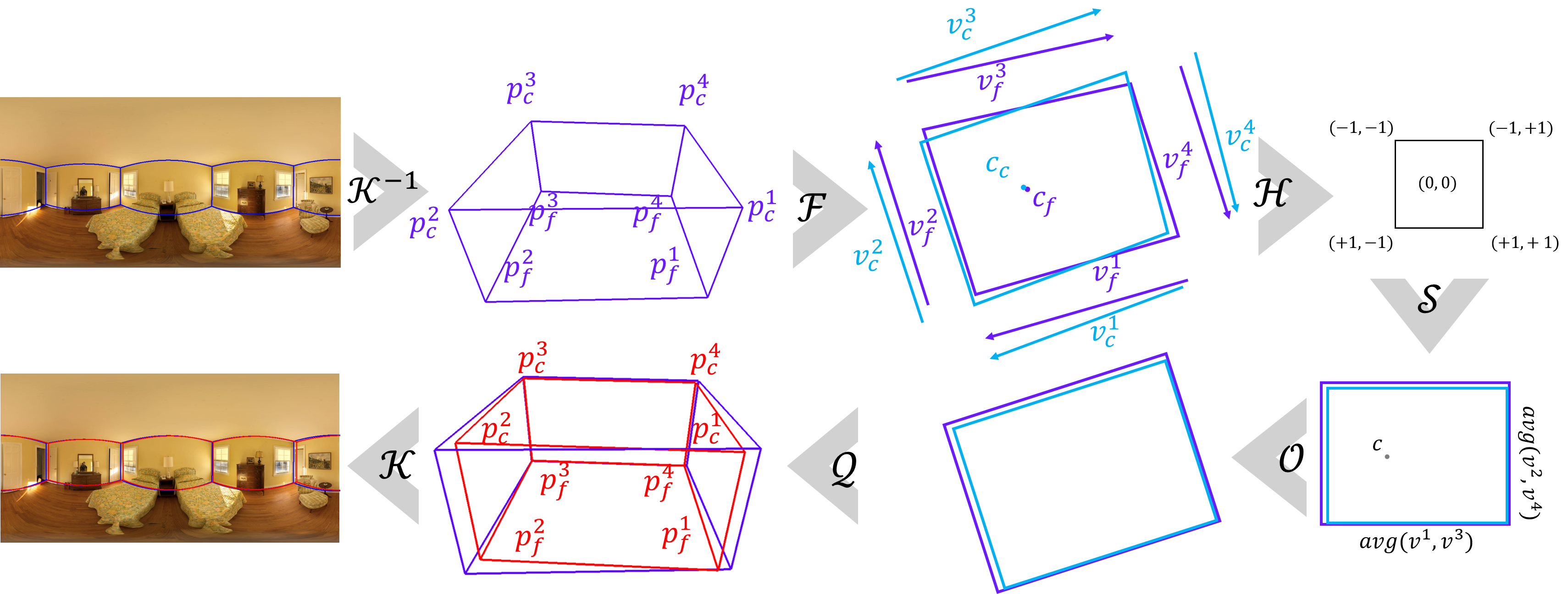}
    \caption{
    Starting from quasi-Manhattan corner estimates, these get first deprojected ($\mathcal{K}^{-1}$) to $3D$ coordinates.
    Then, keeping only the horizontal coordinates ($\mathcal{F}$), we get a floor view trapezoid, which depending on the measurement and coordinates (floor/ceiling) our projection operated on, is slightly different (\textcolor{cyan}{cyan} for the ceiling, and \textcolor{blue}{blue} for the floor).
    Using these floor view horizontal coordinates, we estimate a homography $\mathcal{H}$ to transform them to an axis aligned, unit square.
    This gets translated and scaled ($\mathcal{S}$) using the average opposite edge lengths and centroid of the original untransformed floor view coordinates.
    An orthogonal Procrustes analysis ($\mathcal{O}$) is used to align the rectangle to the trapezoid, which then gets lifted to a cuboid ($\mathcal{Q}$) using the original heights, taking into account the quasi-Manhattan alignment of our estimates.
    The cuboid's $3D$ coordinates then get projected ($\mathcal{K}$) back to equirectangular domain corners.
    Apart from the ceiling and floor starting corners, we also consider a joint approach where the horizontal floor view coordinates get averaged from both $3D$ estimates, before proceeding to estimate the homography.
    For this approach to work, we rescale the ceiling coordinates so that their camera to floor distances align, therefore removing any scale difference from the camera's position deviation from the true center.
    }
    \label{fig:homography}
\end{figure}

We extract the $\mathbf{f = (x, y)}$ horizontal coordinates coordinates, corresponding to an orthographic floor view projection, which comprise a general trapezoid.
This is transformed to a unit square by estimating the projective transformation $\mathcal{H}$ (planar homography) mapping the former to the latter \cite{hartley2003multiple}.
Using the trapezoid's edge norms $\| \mathbf{v} \|_2$, with $\mathbf{v} = \mathbf{f}^{j+1} - \mathbf{f}^j$, we calculate the average opposite edge distances and use them to scale the unit square to a rectangle, after translating it for their centroids to align.
Then, we rotate and translate the rectangle to align with the original trapezoid using orthogonal Procrustes analysis \cite{schonemann1966generalized}.
Finally, the rectangle gets lifted to a cuboid using the vertical ($z$) ceiling and floor coordinates.
The resulting cuboid vertices can be transformed back to angular coordinates for loss computation, with the overall process presented in Figure~\ref{fig:homography}.
We use this cuboid alignment transform $\mathcal{C}$ as the final block of our model to ensure full Manhattan alignment in an end-to-end manner.

We supervise the junction angular coordinates using the geodesic distance of Eq.(\ref{eq:geod_dist}):
\begin{equation}
    \mathcal{L}_G = \frac{1}{J}\sum_j \mathit{g}(\mathbf{c}_m^j, \hat{\mathbf{c}}_m^j),
\end{equation}
with $\mathbf{c}_m^j$ and $\hat{\mathbf{c}}_m^j$ being the groundtruth and predicted coordinates.
The geodesic distance smoothly handles the continuous boundary and provides a more appropriate distance metric on the sphere, instead of the equirectangular projection.
We additionally supervise the spatially normalized heatmaps $\mathbf{H}^j = spatial\_softmax(\mathbf{M}^j)$ predicted by our model with Kullback\- Leibler divergence:
\begin{equation}
\label{eq:KL}
    \mathcal{L}_D = \sum_{\mathcal{A_s}, j} \mathcal{D}_{KL}(\mathbf{H}^j, \mathcal{\tilde{G}}( \mathbf{c}^j_m)),
\end{equation}
where $\tilde{\mathcal{G}}(\cdot)$ is the spatially normalized geodesic heatmap $\mathcal{G}(\cdot)$.
Apart from regularizing the predicted heatmaps, this loss allows for stable end-to-end training with the cuboid alignment transform, as pure coordinate supervision destabilized the model during early training, which prevented convergence as a consequence of the double solve required in the homography and Procrustes analysis.
Our final loss is defined as:
\begin{equation}
\label{eq:total_loss}
    \mathcal{L} = \sum\limits_{n=1}^{N} \frac{\lambda_{G}}{N} \mathcal{L}_G^n + \frac{\lambda_{D}}{N} \mathcal{L}_D^n,
\end{equation}

with $\lambda_{G}$ and $\lambda_{D}$ being weighting factors between the geodesic distance and KL loss, applied on each of the $N$ hourglass predictions.

The higher level SH architecture allows for global processing without relying on heavy bottlenecks \cite{zou2018layoutnet}, computational expensive feature fusion \cite{yang2019dula} or recurrent models \cite{sun2019horizonnet}.
It also requires no post-processing as it can produce a Manhattan aligned layout in a single-shot with high accuracy albeit operating at lower than typical resolutions.

\section{Results}
\label{sec:results}
\begin{figure*}[!htbp]
    \centering
    \subfloat{\includegraphics[width=0.24\linewidth]{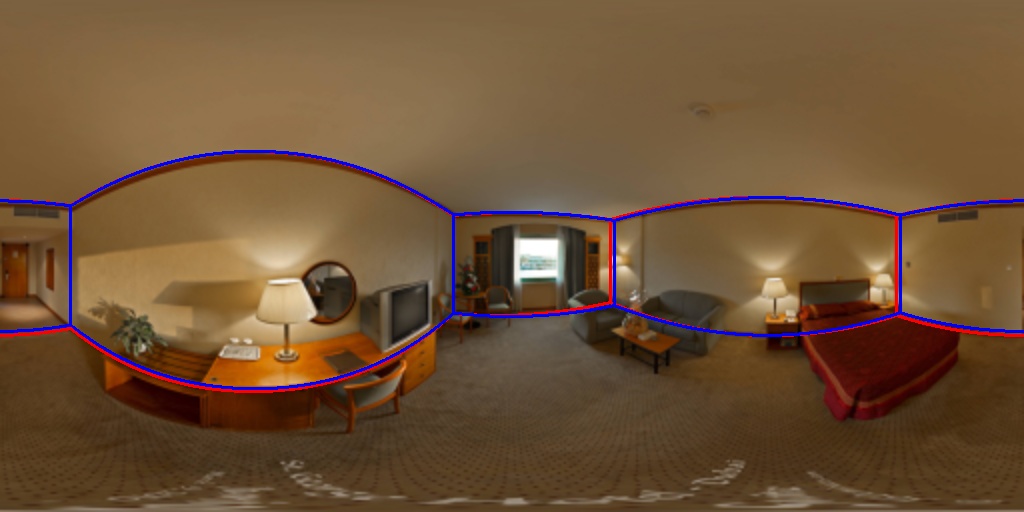}}
    \hfill
    \subfloat{\includegraphics[width=0.24\linewidth]{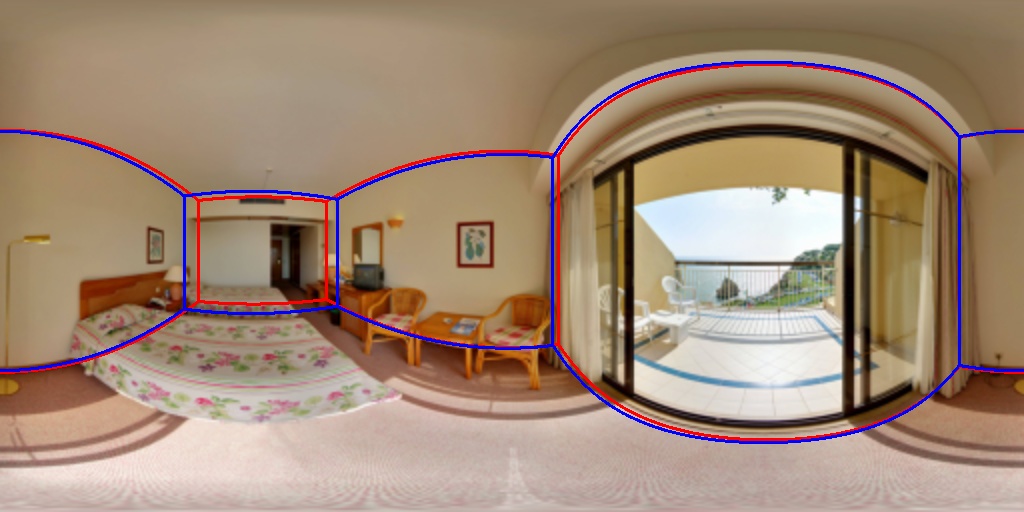}}
    \hfill
    \subfloat{\includegraphics[width=0.24\linewidth]{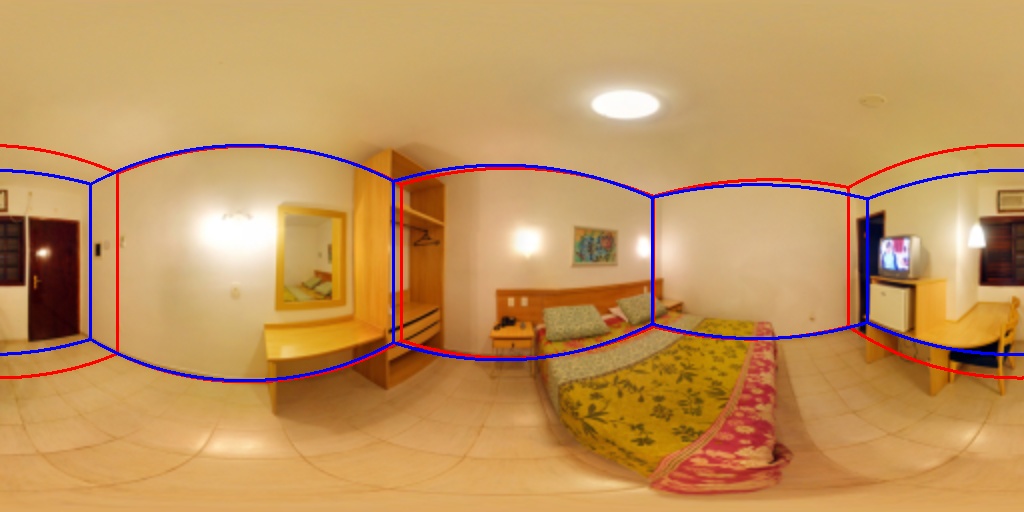}}
    \hfill
    \subfloat{\includegraphics[width=0.24\linewidth]{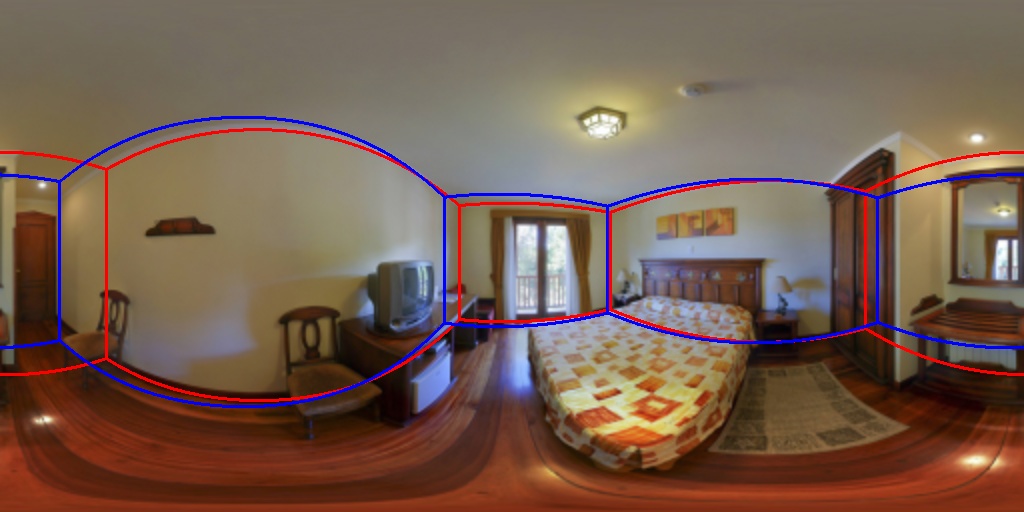}}
    \hfill
    \vspace{-0.3 cm}
    \subfloat{\includegraphics[width=0.24\linewidth]{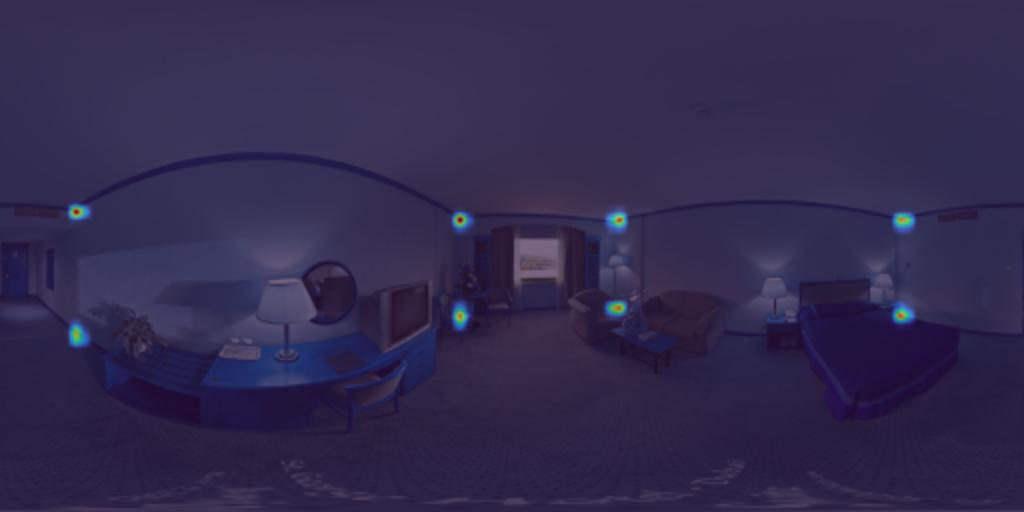}}
    \hfill
    \subfloat{\includegraphics[width=0.24\linewidth]{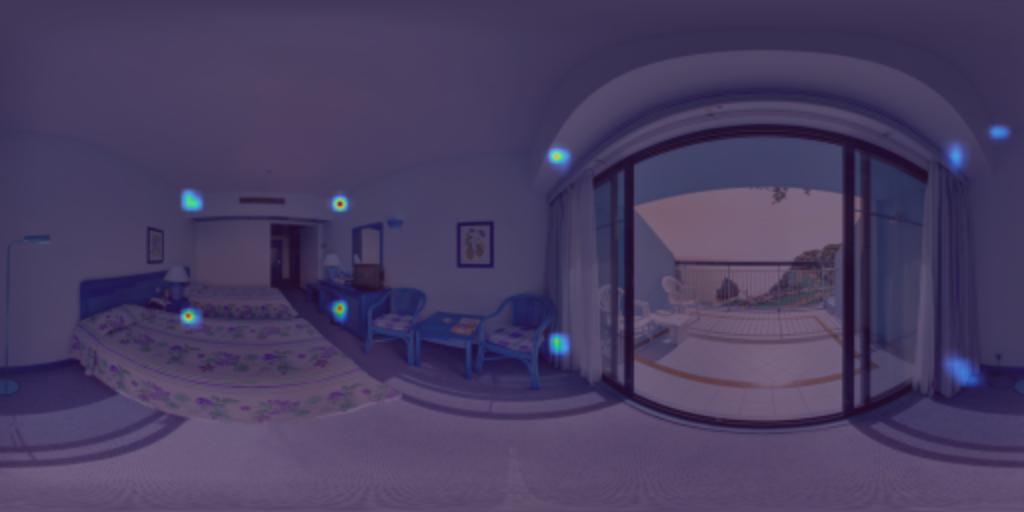}}
    \hfill
    \subfloat{\includegraphics[width=0.24\linewidth]{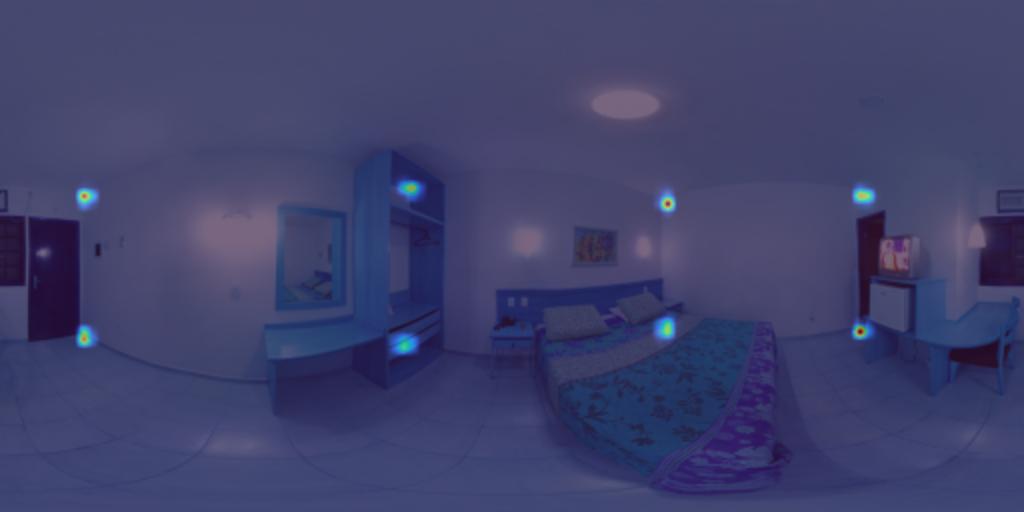}}
    \hfill
    \subfloat{\includegraphics[width=0.24\linewidth]{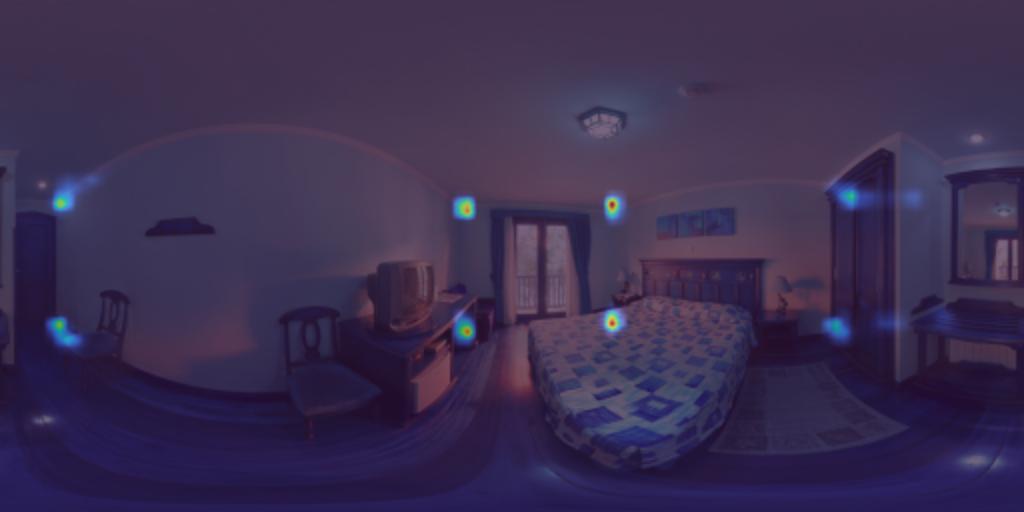}}
    \hfill
    \vspace{-0.35 cm}
    \subfloat{\includegraphics[width=0.24\linewidth,height=0.20\linewidth]{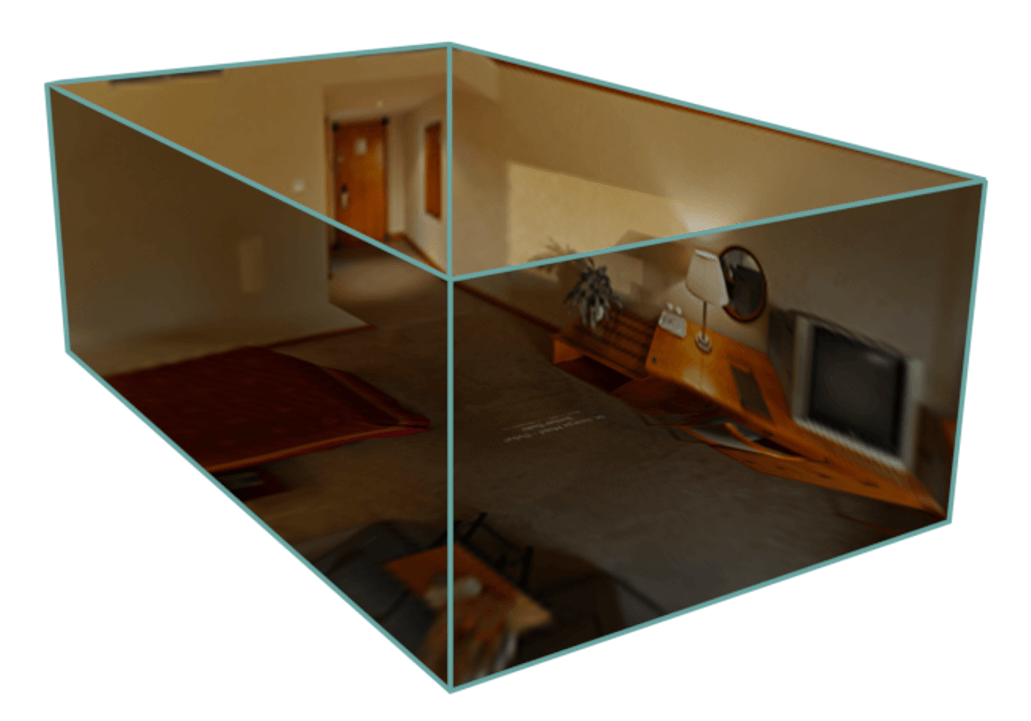}}
    \hfill
    \subfloat{\includegraphics[width=0.24\linewidth,height=0.20\linewidth]{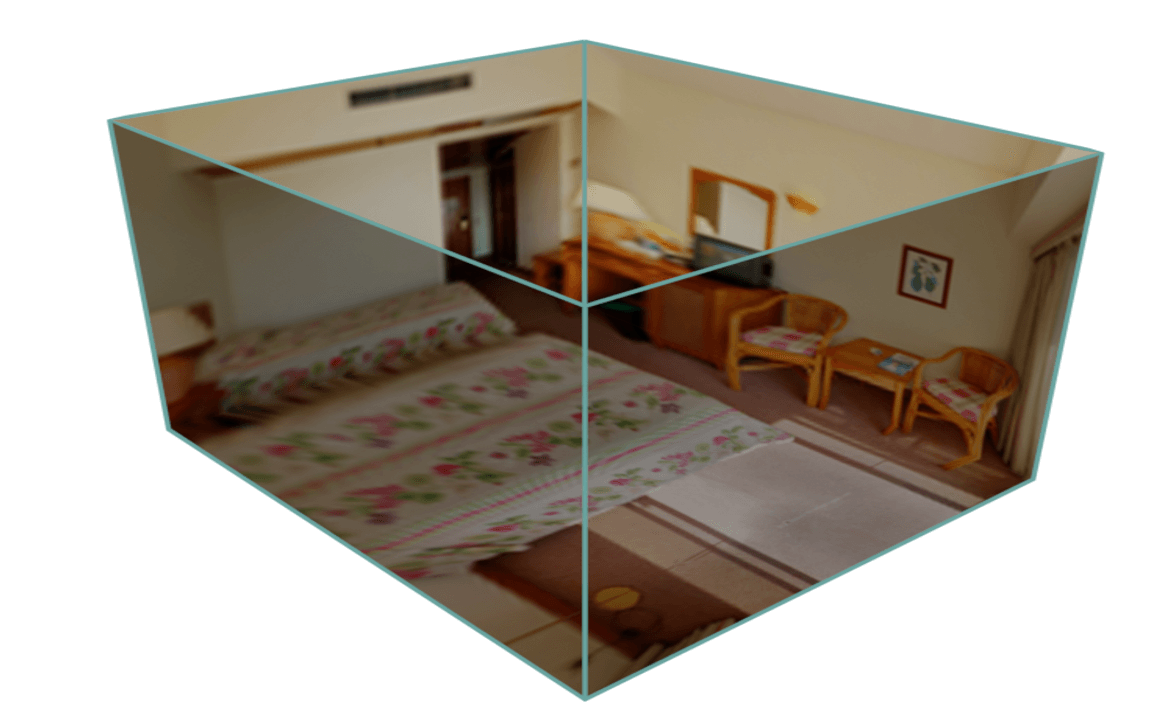}}
    \hfill
    \subfloat{\includegraphics[width=0.24\linewidth,height=0.20\linewidth]{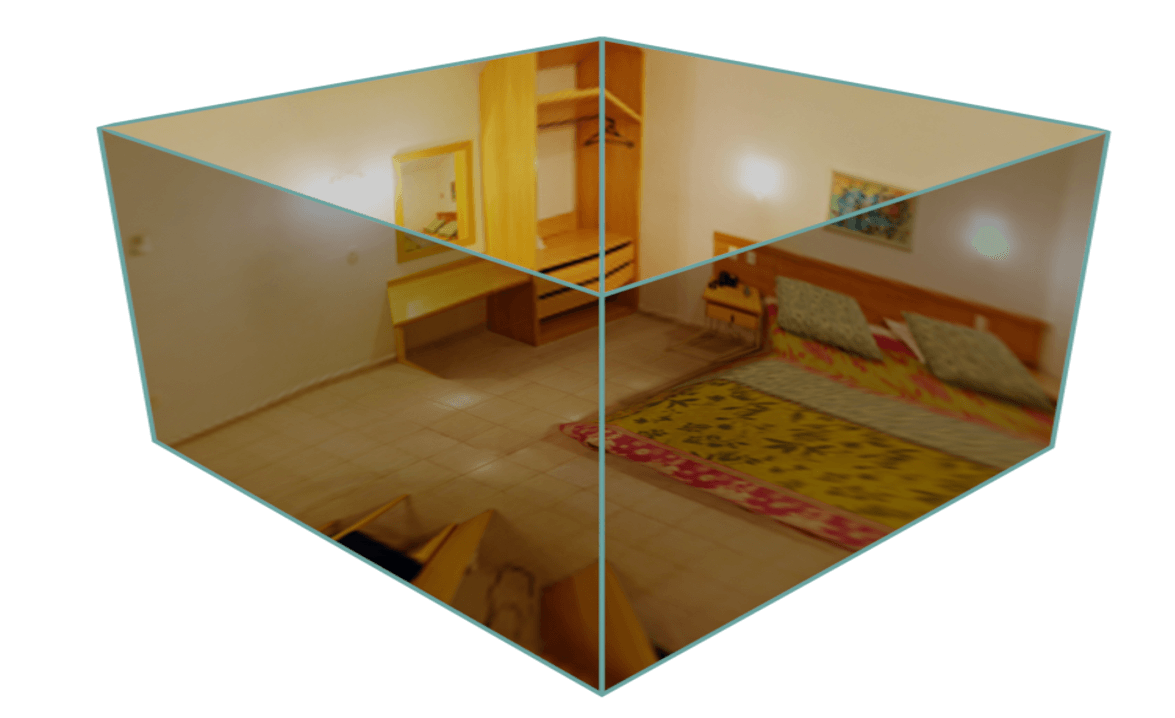}}
    \hfill
    \subfloat{\includegraphics[width=0.24\linewidth,height=0.20\linewidth]{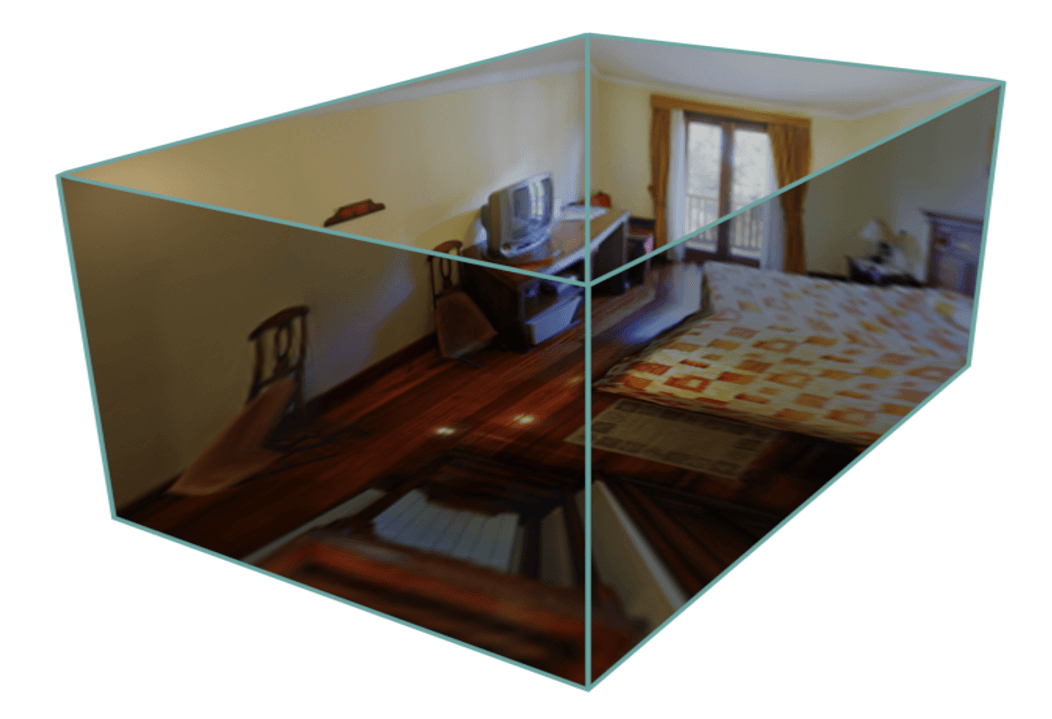}}
    \hfill
    \vspace{-0.35 cm}
    \subfloat{\includegraphics[width=0.12\linewidth, height=2.5cm]{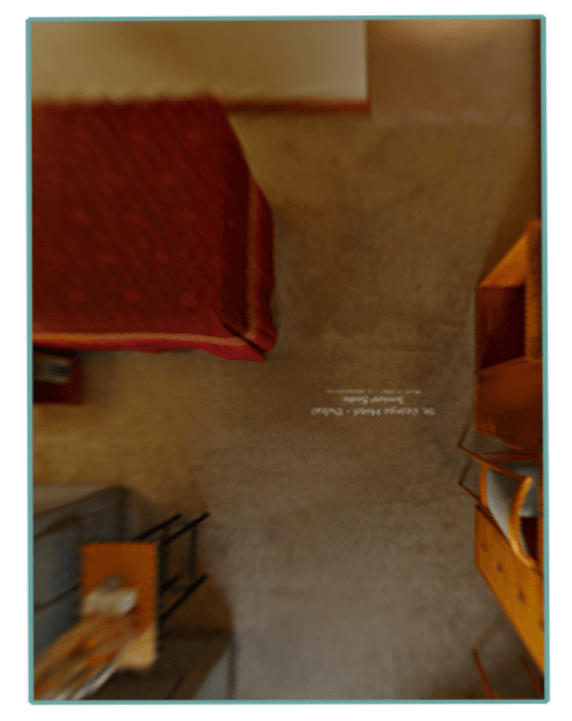}}
    \hfill
    \subfloat{\includegraphics[width=0.12\linewidth,height=2.5cm]{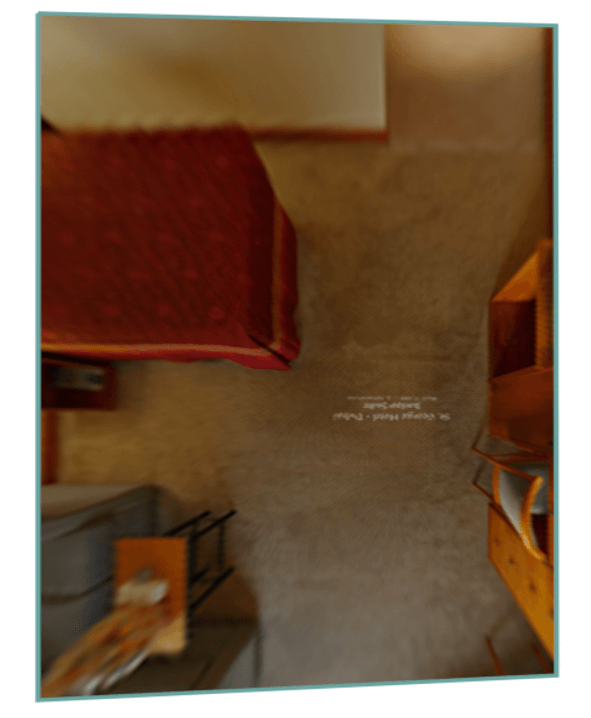}}
    \hfill
    \subfloat{\includegraphics[width=0.12\linewidth]{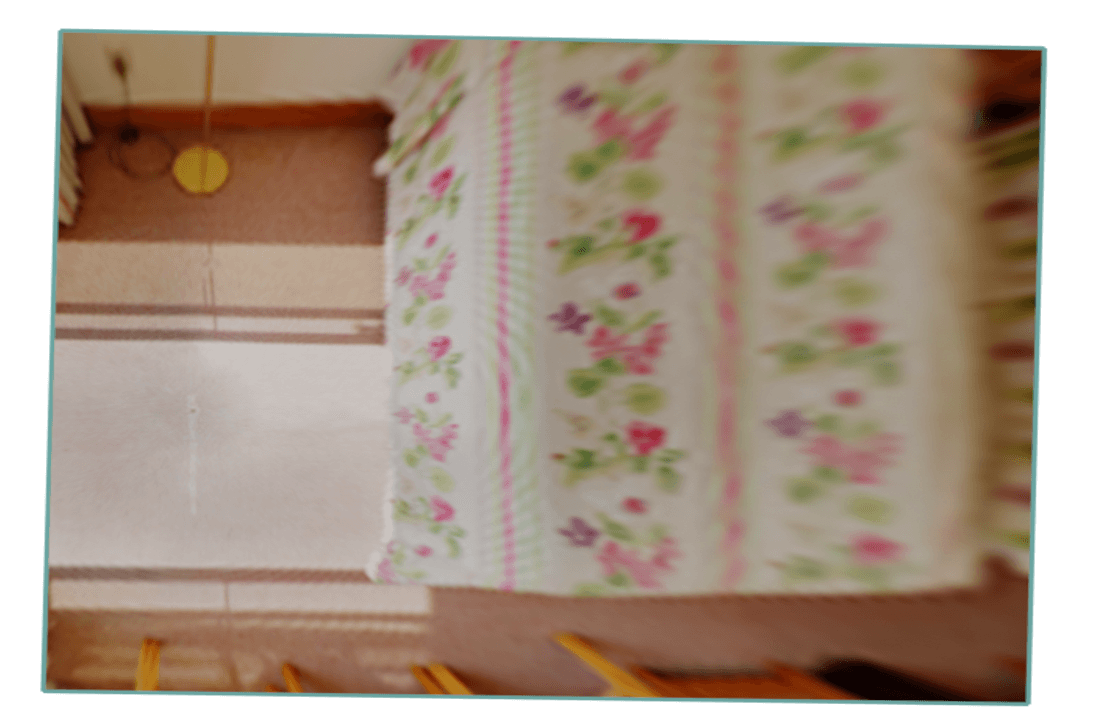}}
    \hfill
    \subfloat{\includegraphics[width=0.12\linewidth]{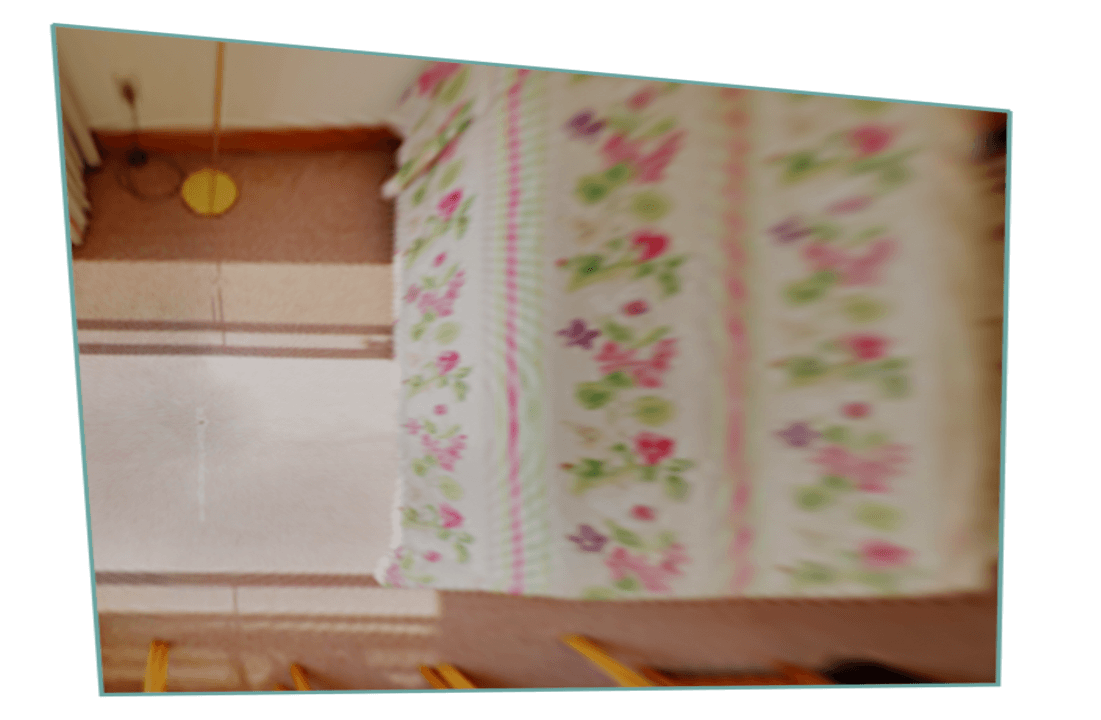}}
    \hfill
    \subfloat{\includegraphics[width=0.12\linewidth]{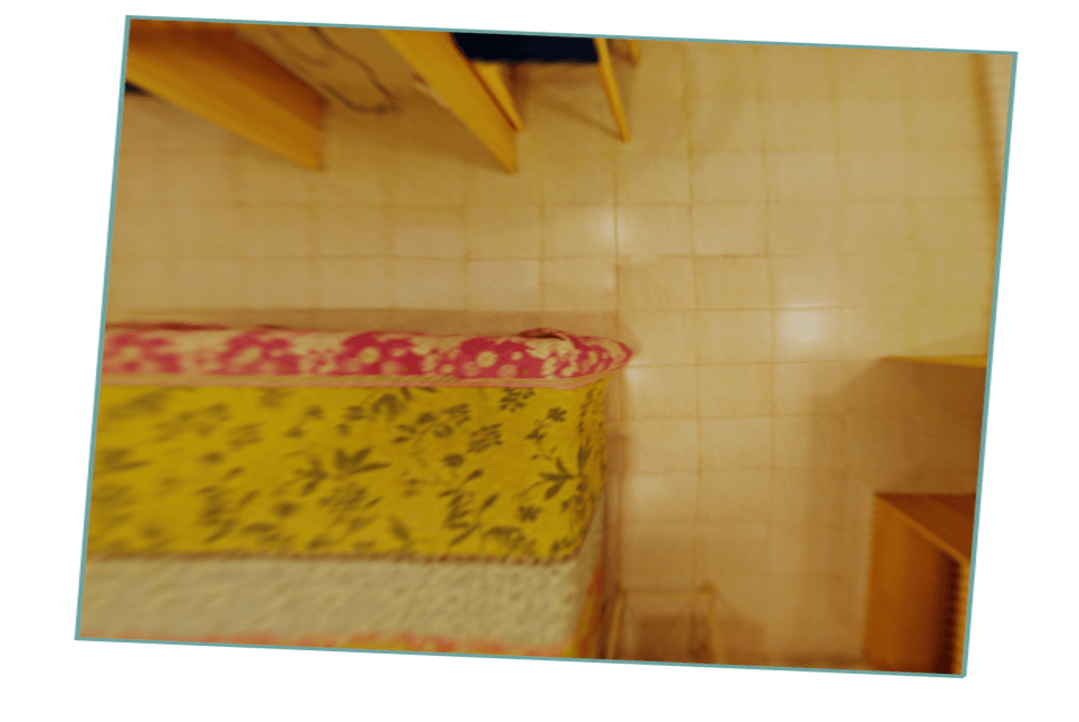}}
    \hfill
    \subfloat{\includegraphics[width=0.12\linewidth]{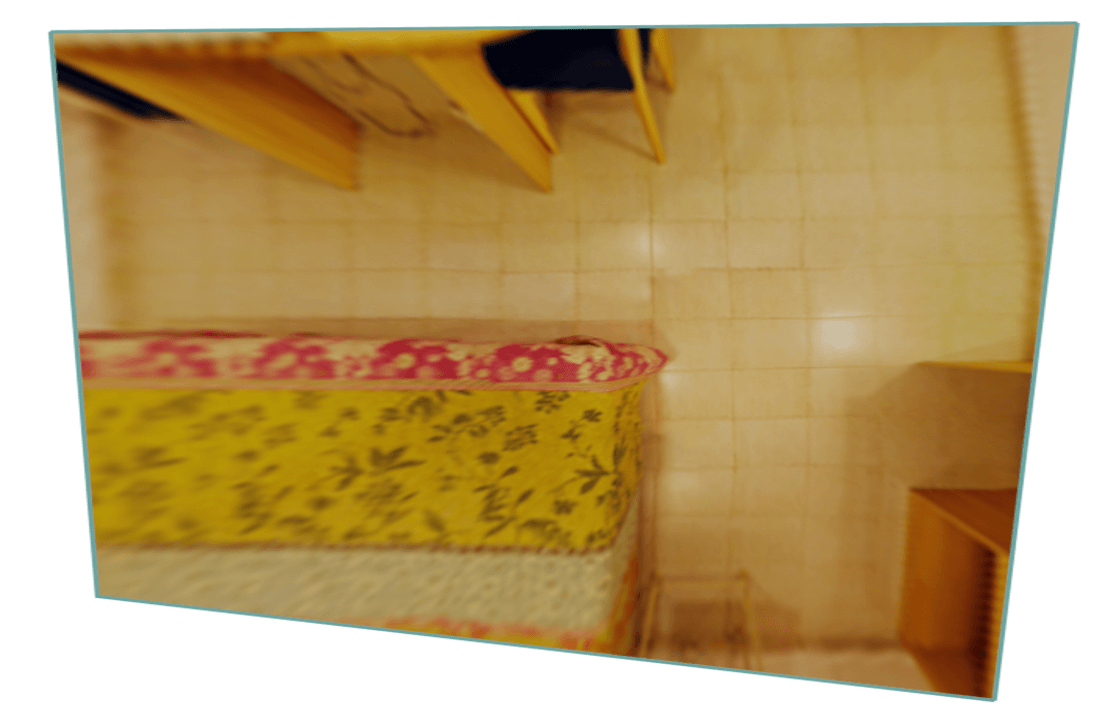}}
    \hfill
    \subfloat{\includegraphics[width=0.12\linewidth]{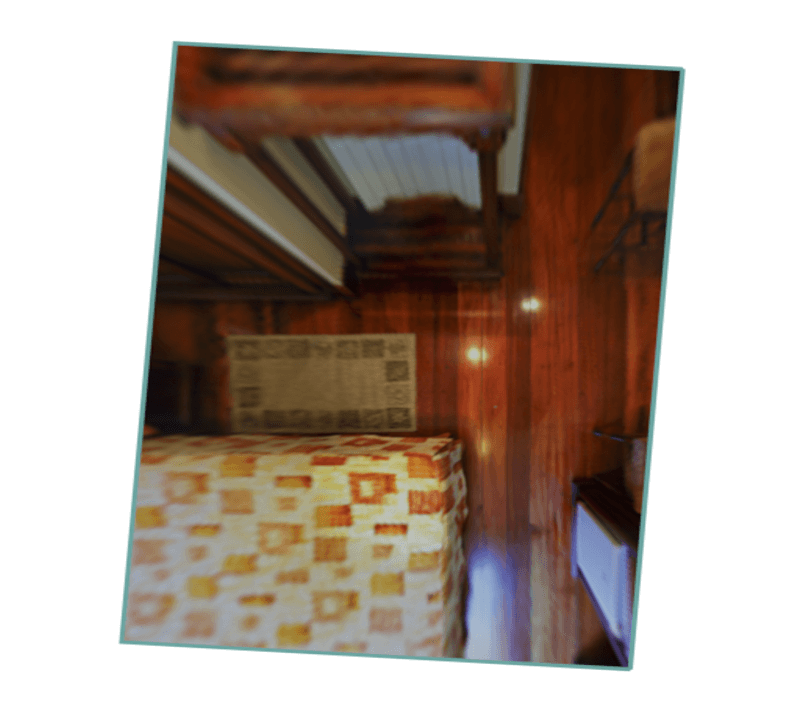}}
    \hfill
    \subfloat{\includegraphics[width=0.12\linewidth]{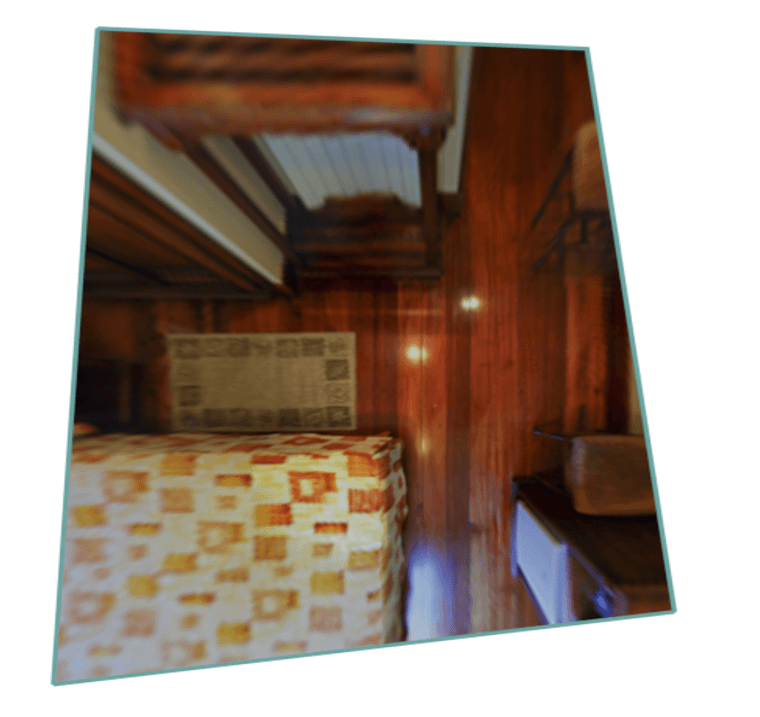}}
    \hfill
    \vspace{-0.3cm}
    \subfloat{\includegraphics[width=0.24\linewidth]{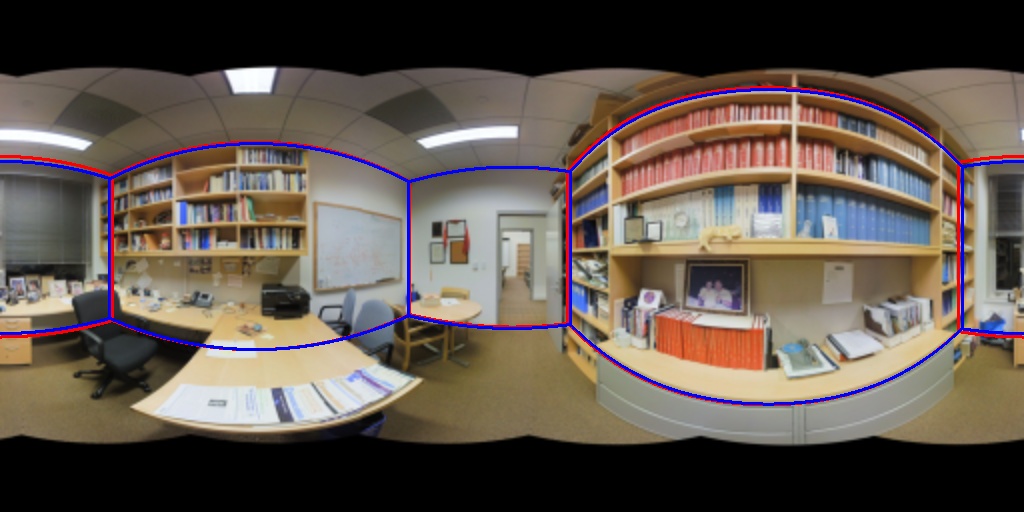}}
    \hfill
    \subfloat{\includegraphics[width=0.24\linewidth]{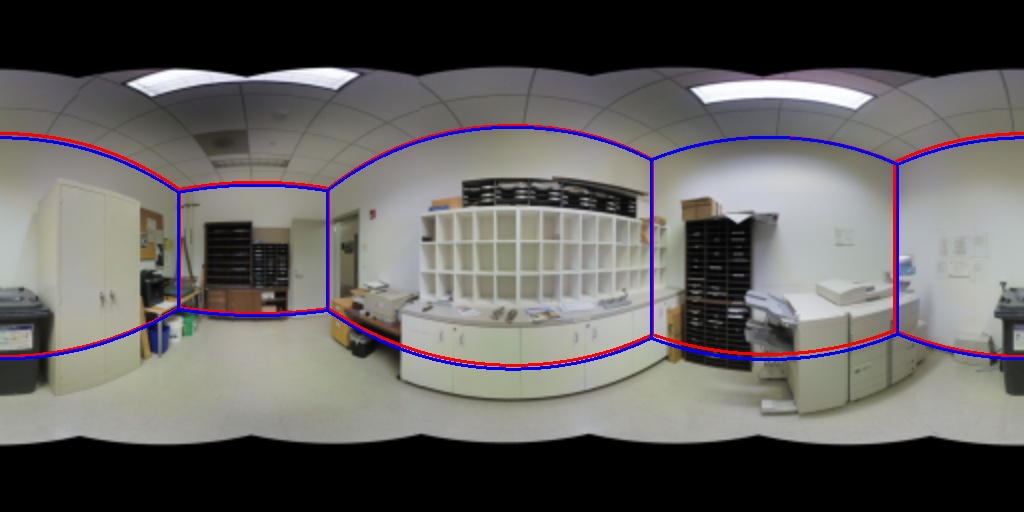}}
    \hfill
    \subfloat{\includegraphics[width=0.24\linewidth]{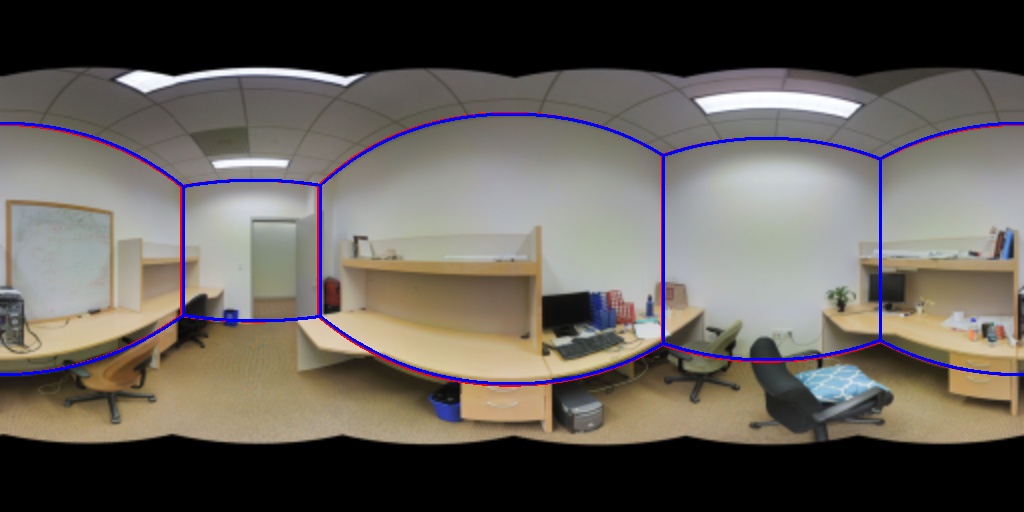}}
    \hfill
    \subfloat{\includegraphics[width=0.24\linewidth]{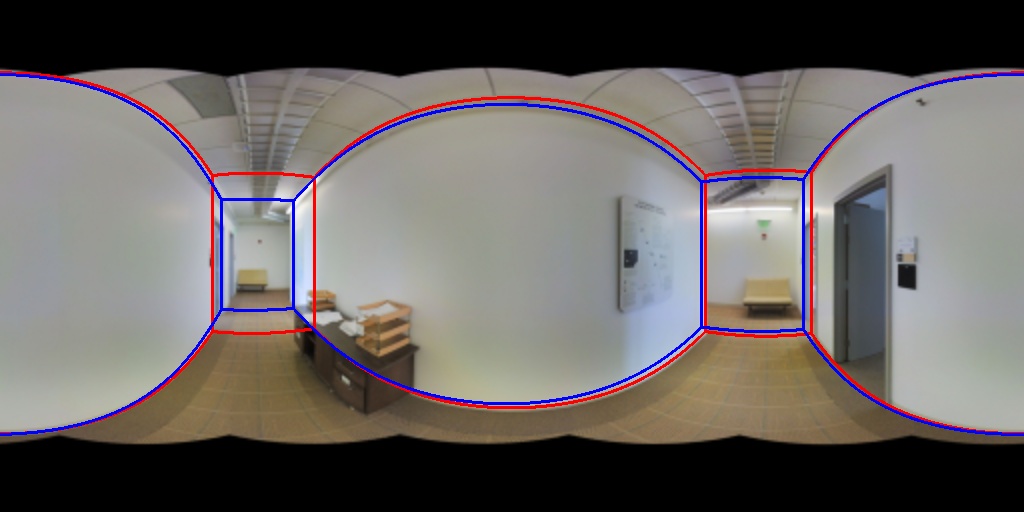}}
    \hfill
    \vspace{-0.3 cm}
    \subfloat{\includegraphics[width=0.24\linewidth]{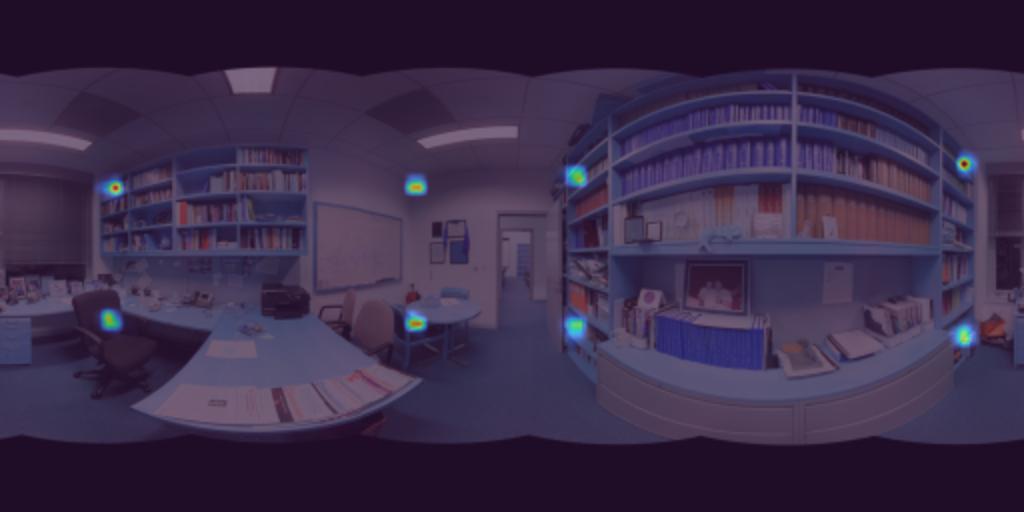}}
    \hfill
    \subfloat{\includegraphics[width=0.24\linewidth]{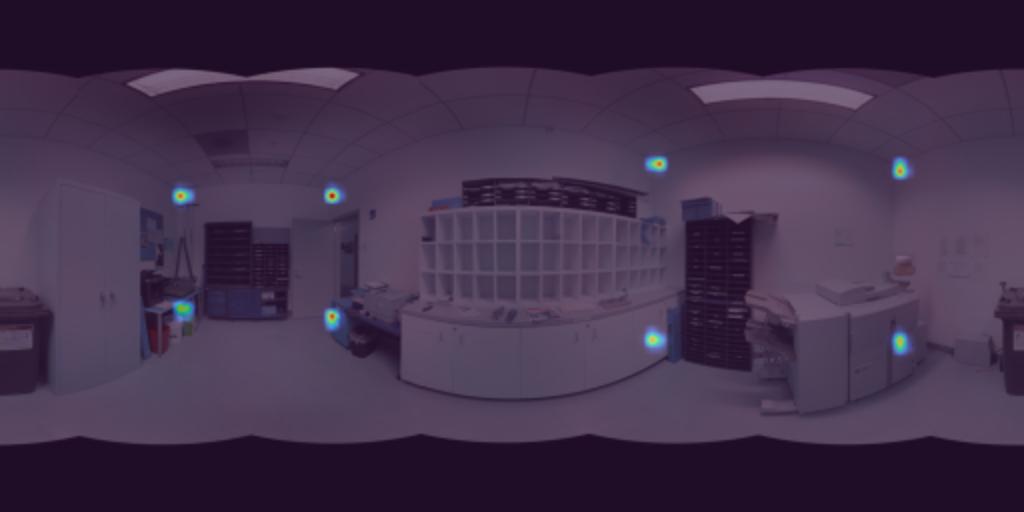}}
    \hfill
    \subfloat{\includegraphics[width=0.24\linewidth]{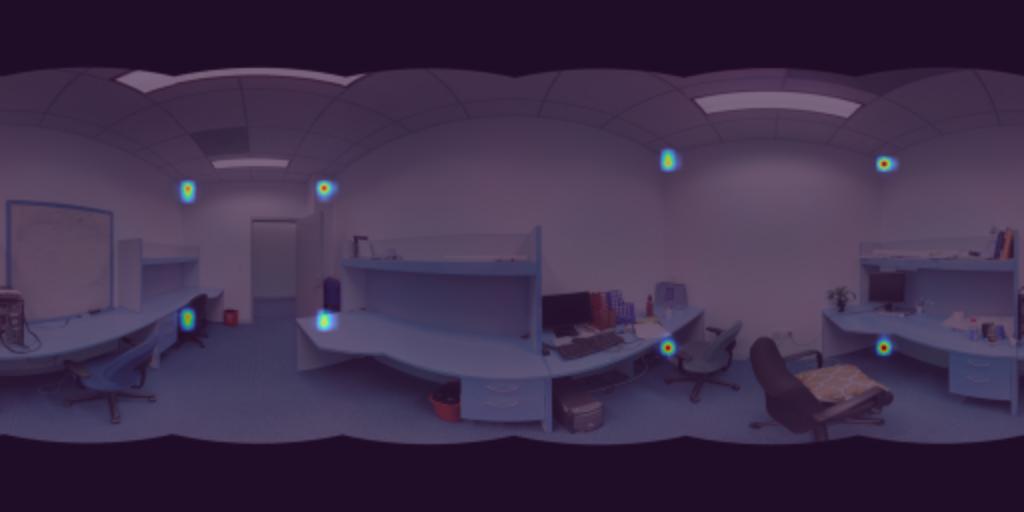}}
    \hfill
    \subfloat{\includegraphics[width=0.24\linewidth]{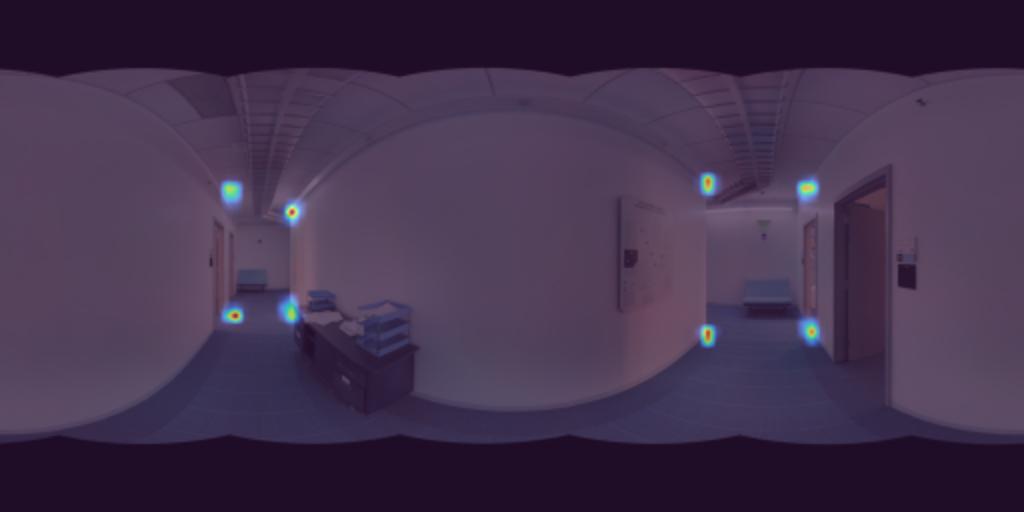}}
    \hfill
    \vspace{-0.35 cm}
    \subfloat{\includegraphics[width=0.24\linewidth,height=0.20\linewidth]{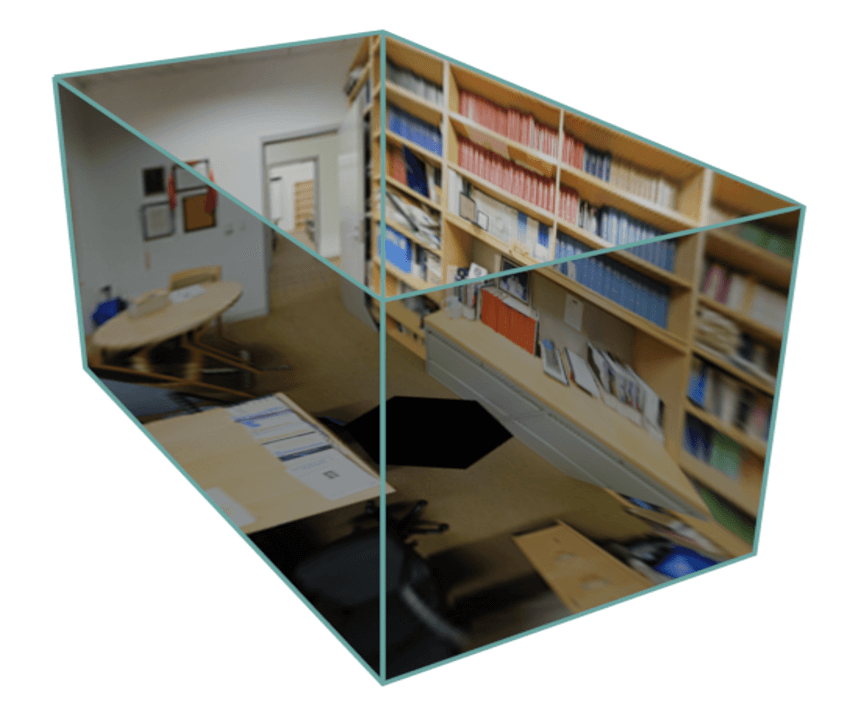}}
    \hfill
    \subfloat{\includegraphics[width=0.24\linewidth,height=0.20\linewidth]{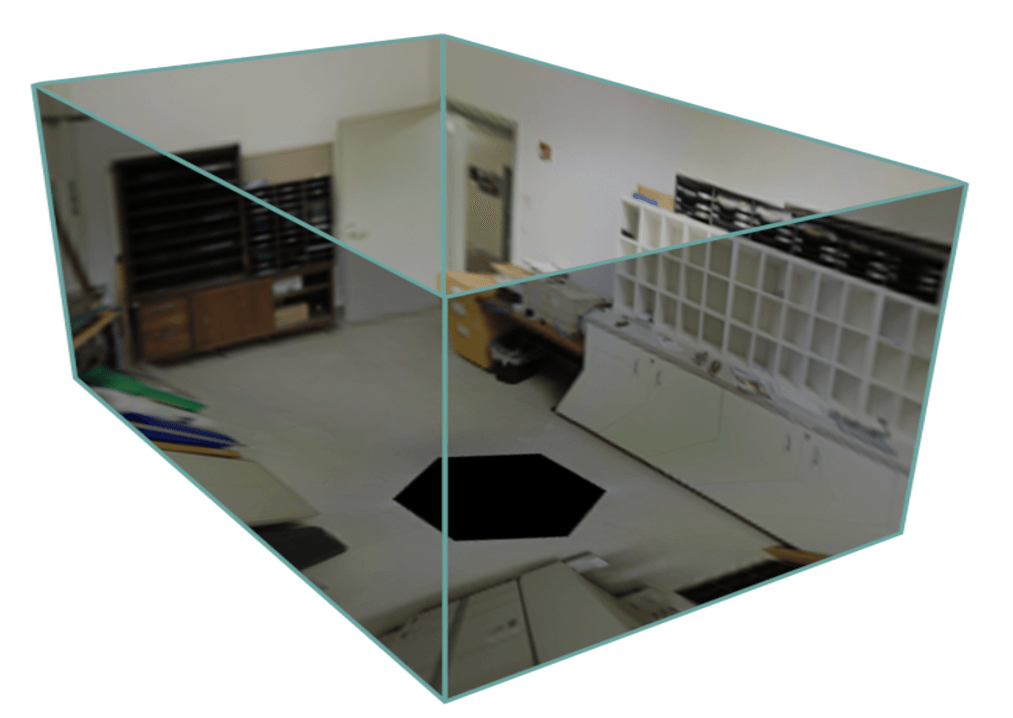}}
    \hfill
    \subfloat{\includegraphics[width=0.24\linewidth,height=0.20\linewidth]{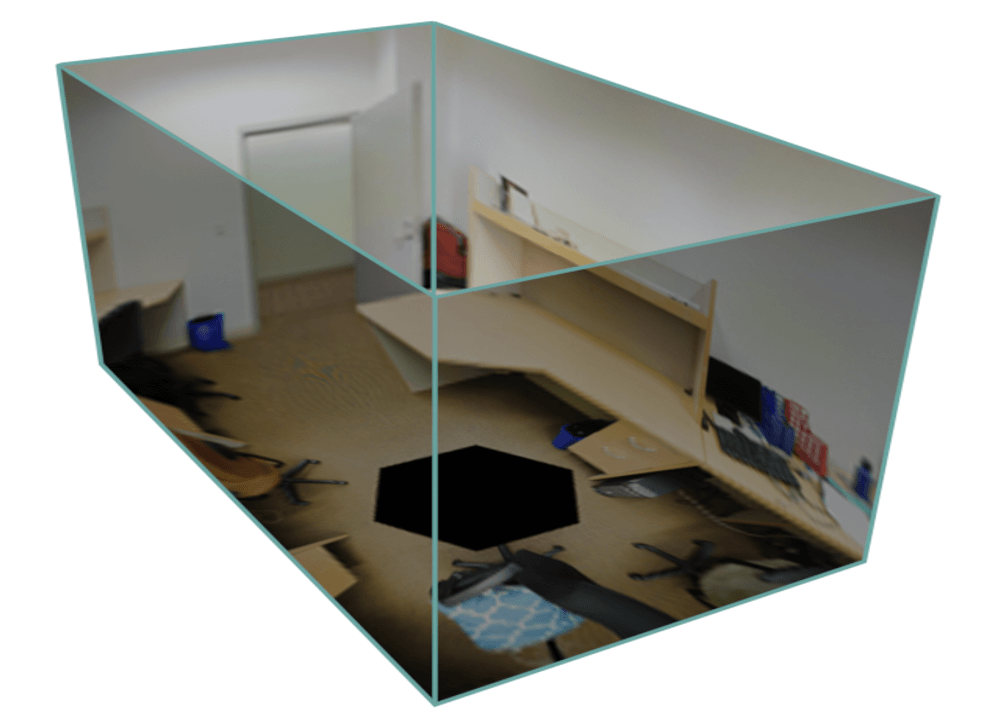}}
    \hfill
    \subfloat{\includegraphics[width=0.24\linewidth,height=0.20\linewidth]{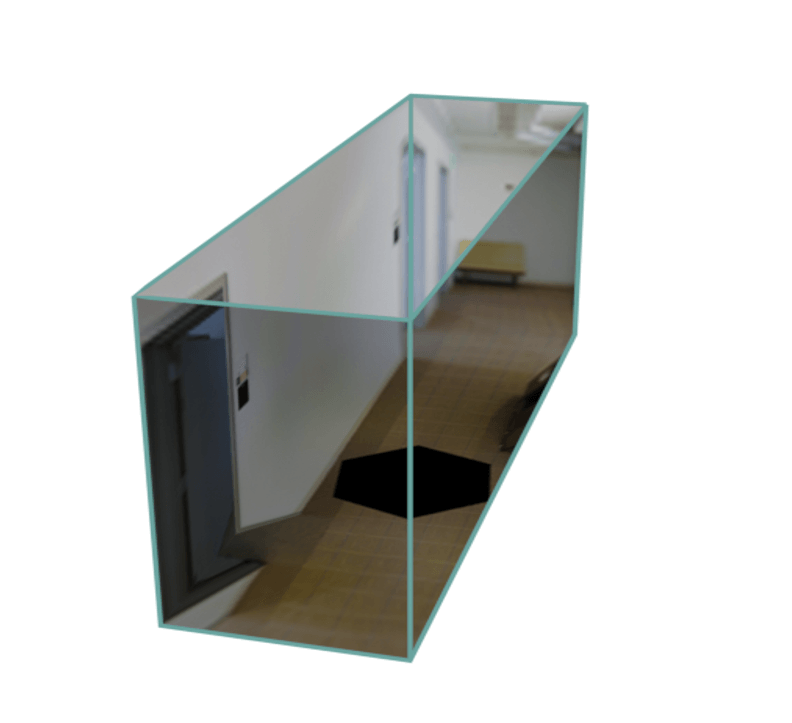}}
    \hfill
    \vspace{-0.35 cm}

    \subfloat{\includegraphics[width=0.105\linewidth]{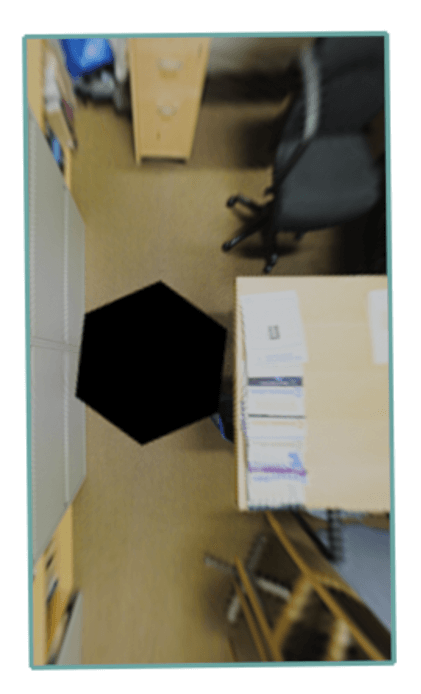}}
    \hfill
    \subfloat{\includegraphics[width=0.1\linewidth]{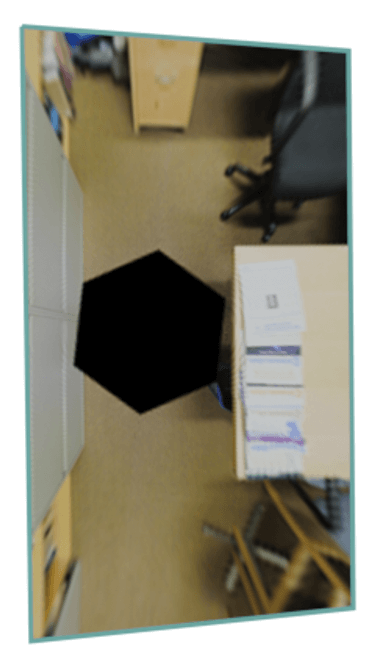}}
    \hfill
    \subfloat{\includegraphics[height=0.13\linewidth, angle=90]{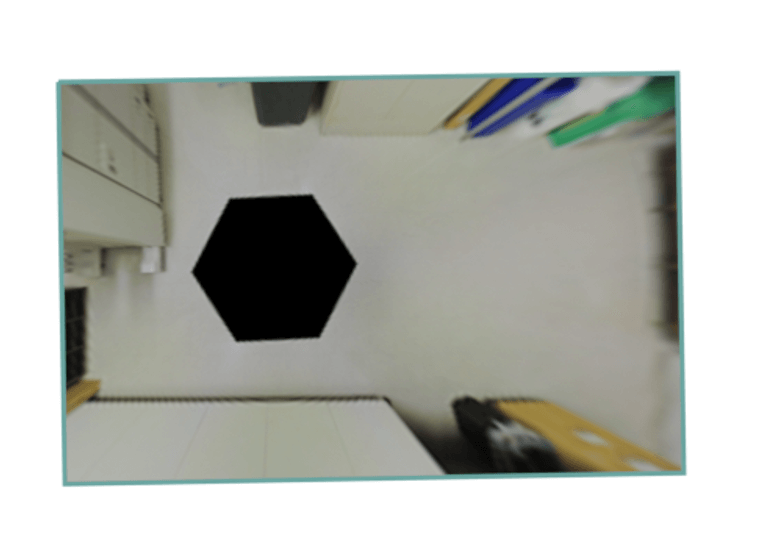}}
    \hspace{-0.1cm}
    \subfloat{\includegraphics[height=0.125\linewidth, angle=90]{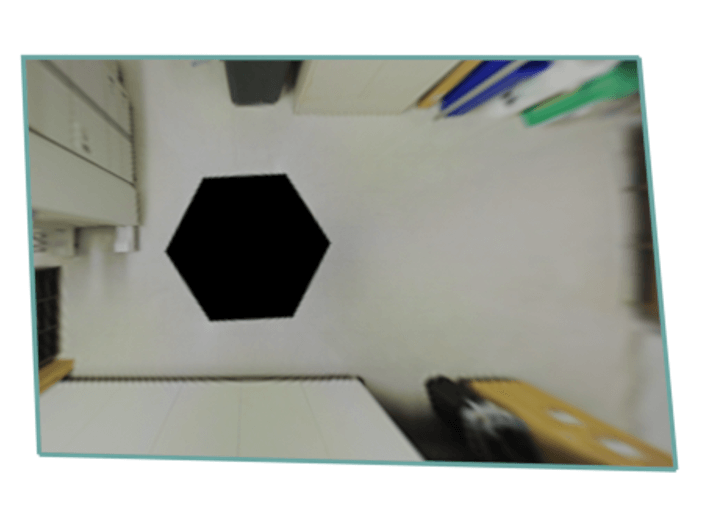}}
    \hfill
    \subfloat{\includegraphics[height=0.1\linewidth, angle=90]{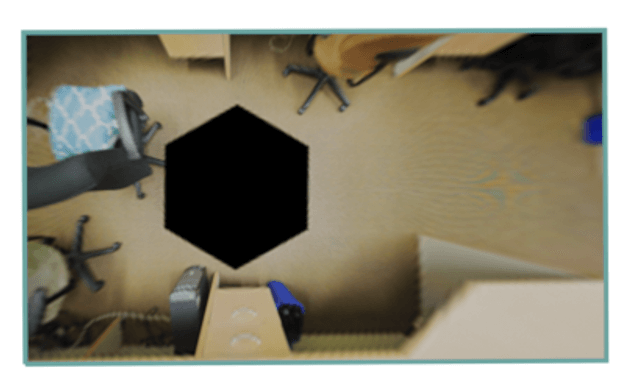}}
    \hfill
    \subfloat{\includegraphics[height=0.105\linewidth, angle=90]{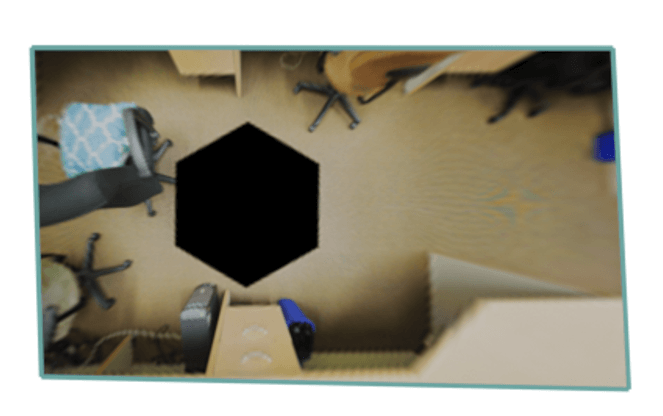}}
    \hspace{0.3cm}
    \subfloat{\includegraphics[height=0.06\linewidth, angle=90]{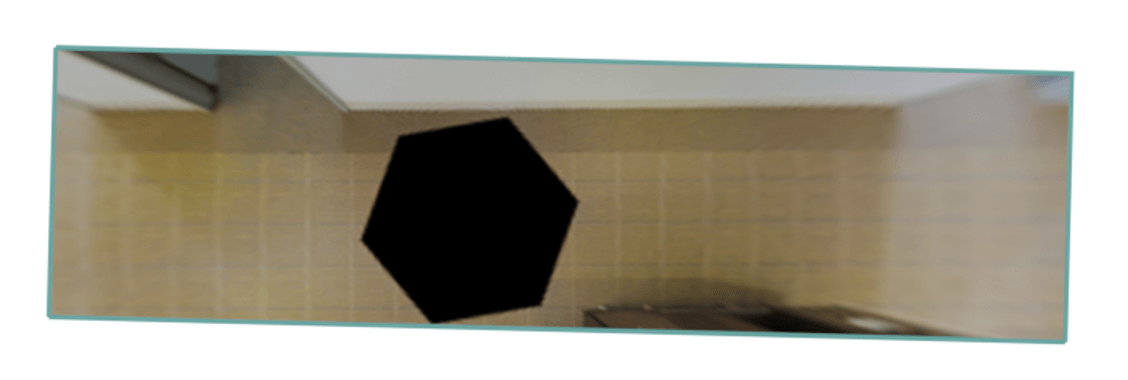}}
    \hspace{0.4cm}
    \subfloat{\includegraphics[height=0.06\linewidth, angle=90]{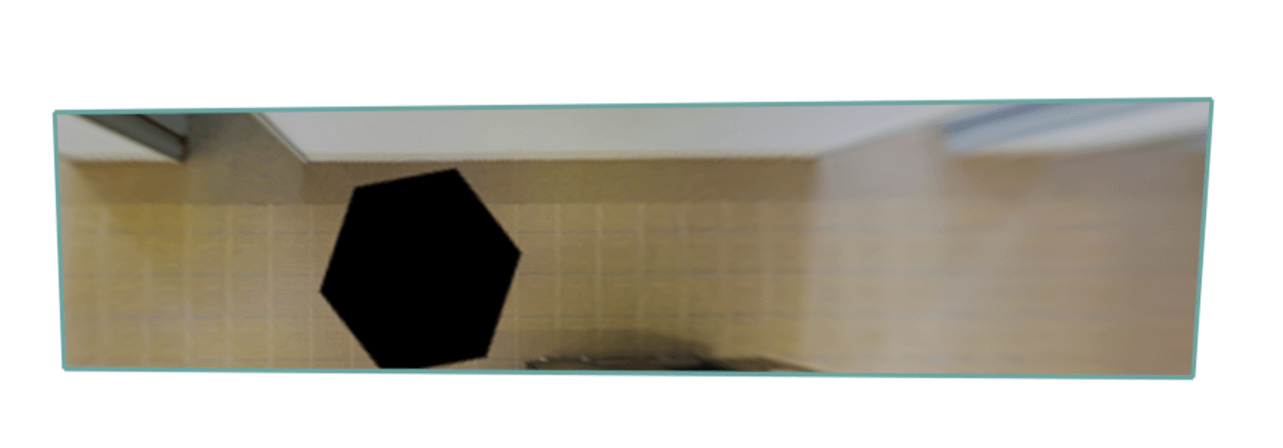}}
    \hspace{1.2cm}
    
    \caption{
        Qualitative results on the PanoContext (top) and Stanford2D3D (bottom) datasets.
        On each panorama, we overlay the reconstructed layout from the groundtruth \textcolor{red}{red} and predicted \textcolor{blue}{blue} junctions.
        The next row showcases the overlaid aggregated heatmap predictions, with the following one illustrating the resulting 3D mesh.
        Finally, two orthographic floor views are presented, showing the full Manhattan (left), and quasi-Manhattan aligned (right) estimations.
    }
\label{fig:real}
\end{figure*}

\subsection{Implementation Details}
\label{sec:implementation}
The input to our model is a single upright\footnote{Traditional \cite{zou2018layoutnet, jung2017robust}, or data-driven methods \cite{jung2019deep360up} can be used.}, \textit{i.e.}~horizontal floor, $512 \times 256$ spherical panorama.
We use $128$ features for each hourglass's residual block, with a $128 \times 64$ heatmap resolution, and initialize our SH model using \cite{he2015delving}.
We use the Adam \cite{kingma2014adam} optimizer with a learning rate of $0.002$ and default values for the other parameters, no weight decay, and a batch size of $8$.
Further, after an empirical greedy search, we use a fixed $\alpha = 2^o$ and $\mathbf{s} = (3.5, 3.5)$ for our Geodesic and Isotropic Gaussian distribution reconstructions respectively, which are created using the encoding of \cite{zhang2020distribution}, and set the loss weights to $\lambda_{G} = 1.0$ and $\lambda_D = 0.15$. 
For cuboid alignment we use the joint approach and use a floor distance of $-1.6m$.
We implement our models using PyTorch \cite{paszke2017automatic,falcon2019pytorch}, setting the same seed for all random number generators.
Further, each parameter update uses the gradients of $16$ samples.

We apply heavy data augmentation during training, as established in prior work \cite{zou20193d, sun2019horizonnet, fernandez2020corners}.
Apart from photometric augmentations (random brightness, contrast, and gamma \cite{buslaev2020albumentations}), following \cite{fernandez2020corners}, we further apply random erasing, with a uniform random selection between $1$ and $3$ blocks erased per sample.
We also probabilistically apply a set of \360 panorama specific augmentations in a cascaded manner: \textbf{i)} uniformly random horizontal rotations spanning the full angle range, \textbf{ii)} left-right flipping, and \textbf{iii)} PanoStretch augmentations \cite{sun2019horizonnet} using the default stretching ratio ranges.
All augmentation probabilities are set to $50\%$.

\subsection{Datasets}
\label{sec:datasets}
Prior work up to now has experimented with small scale datasets.
PanoContext \cite{zhang2014panocontext} manually annotated a total of $547$ panoramas from the Sun360 dataset \cite{xiao2012recognizing} as cuboids.
Additionally, LayoutNet manually annotated $552$ panoramas from the Stanford2D3D dataset \cite{armeni2017joint}, which are not complete spherical images as their vertical FoV is narrower.
Similar to previous works, we use the common train, test and validation splits as used in \cite{fernandez2020corners} and \cite{zou2018layoutnet} for the PanoContext and Stanford2D3D datasets respectively.
Taking into account their small scale, we jointly consider them as a single \textit{real} dataset and train all our models for $150$ epochs.

More recently, layout annotations have been provided in newer computer-generated datasets, the Kujiale dataset used in \cite{jin2020geometric} and the Structured3D dataset \cite{Structured3D}, totaling $3550$ and $21835$ annotated images respectively.
Albeit synthetic, they offer a much more expanded data corpus than what is currently available for real datasets.
Given their synthetic nature, these datasets offer different room styles for the same scene.
In particular, they provide empty rooms as well as rooms filled with furniture by interior designers.
For the Kujiale dataset we use both types of scenes, while for Structured3D we only use full scenes and follow their respective official dataset splits.
Our models are trained for $30$ and $125$ epochs respectively on Structured3D and Kujiale.

\begin{table*}[!htbp]
\centering
\caption{Quantitative results on the real domain datasets for each model variant.}
\label{tab:real}
\resizebox{\textwidth}{!}{%
\begin{tabular}{|ccc|rrr|ccc|ccc|}
\hline
\multicolumn{3}{|c|}{Model} &
  \multicolumn{3}{c|}{PanoContext} &
  \multicolumn{3}{c|}{Stanford2D3D} &
  \multicolumn{3}{c|}{Real (Combined)} \\ \hline
Name &
  Variant &
  \multicolumn{1}{l|}{\down{Parameters}} &
  \multicolumn{1}{c}{\down{CE}} &
  \multicolumn{1}{c}{\up{IoU3D}} &
  \multicolumn{1}{c|}{\down{PE}} &
  \down{CE} &
  \up{IoU3D} &
  \down{PE} &
  \down{CE} &
  \up{IoU3D} &
  \down{PE} \\ \hline
LayoutNet v2 &
  ResNet-18 &
  \second{15.57M} &
  \second{0.65\%} &
  \second{84.13\%} &
  \third{1.92\%} &
  0.77\% &
  83.53\% &
  2.30\% &
  \third{0.71\%} &
  83.83\% &
  \second{2.11\%} \\
LayoutNet v2 &
  ResNet-34 &
  25.68M &
  \first{0.63\%} &
  \first{85.02\%} &
  \second{1.79\%} &
  0.71\% &
  84.17\% &
  \second{2.04\%} &
  \second{0.67\%} &
  84.60\% &
  \first{1.92\%} \\
LayoutNet v2 &
  ResNet-50 &
  91.50M &
  0.75\% &
  82.44\% &
  2.22\% &
  0.83\% &
  82.66\% &
  2.59\% &
  0.79\% &
  82.55\% &
  2.41\% \\ \hline
DuLa-Net v2 &
  ResNet-18 &
  25.64M &
  0.83\% &
  82.43\% &
  2.55\% &
  0.74\% &
  84.93\% &
  2.56\% &
  0.79\% &
  83.68\% &
  2.56\% \\
DuLa-Net v2 &
  ResNet-34 &
  45.86M &
  0.82\% &
  83.41\% &
  2.54\% &
  \second{0.66\%} &
  \third{86.45\%} &
  2.43\% &
  0.74\% &
  \third{84.93\%} &
  2.49\% \\
DuLa-Net v2 &
  ResNet-50 &
  57.38M &
  0.81\% &
  83.77\% &
  2.43\% &
  \third{0.67\%} &
  \second{86.6\%} &
  2.48\% &
  0.74\% &
  \second{85.19\%} &
  2.46\% \\ \hline
HorizonNet &
  ResNet-18 &
  \third{23.49M} &
  0.83\% &
  80.27\% &
  2.44\% &
  0.82\% &
  80.59\% &
  2.72\% &
  0.83\% &
  80.43\% &
  2.58\% \\
HorizonNet &
  ResNet-34 &
  33.59M &
  0.76\% &
  81.30\% &
  2.22\% &
  0.78\% &
  80.44\% &
  2.65\% &
  0.77\% &
  80.87\% &
  2.44\% \\
HorizonNet &
  ResNet-50 &
  81.57M &
  \third{0.74\%} &
  82.63\% &
  2.17\% &
  0.69\% &
  82.72\% &
  \third{2.27\%} &
  0.72\% &
  82.68\% &
  \third{2.22\%} \\ \hline
\cellcolor[HTML]{ECF4FF}SSC &
  \cellcolor[HTML]{ECF4FF}HG-3 &
  \first{6.35M} &
  \multicolumn{1}{c}{\first{0.63\%}} &
  \multicolumn{1}{c}{\third{83.97\%}} &
  \multicolumn{1}{c|}{\first{1.78\%}} &
  \first{0.51\%} &
  \first{87.80\%} &
  \first{1.62\%} &
  \first{0.57\%} &
  \first{85.89\%} &
  \first{1.70\%} \\ \hline
\end{tabular}%
}
\end{table*}

\subsection{Metrics}
\label{sec:metrics}
For the quantitative assessment of our approach against prior works we use a set of standard metrics found in the literature \cite{zou20193d}, complemented by another set of accuracy metrics.
The standard metrics include 2D and 3D intersection over union (IoU2D and IoU3D), normalized corner error (CE), pixel error (PE), and the depth-based RMSE and $\delta_1$ accuracy \cite{eigen2014depth}.
For all 3D calculations a fixed floor distance at $-1.6m$ is used.
We also use junction ($J_d$) and wireframe ($W_d$) accuracy metrics, defined as correct when the closest groundtruth junction or line segment respectively is within a pixel threshold $d$.
More specifically, we use the thresholds $d = [5, 10, 15]$.
Finally, since we regress sub-pixel coordinates, all metric calculations are evaluated on a $1024 \times 512$ panorama resolution, and the arrows next to each metric denote the direction of better performance.

\begin{figure*}[!htbp]
    \centering
    \subfloat{\includegraphics[width=0.24\linewidth]{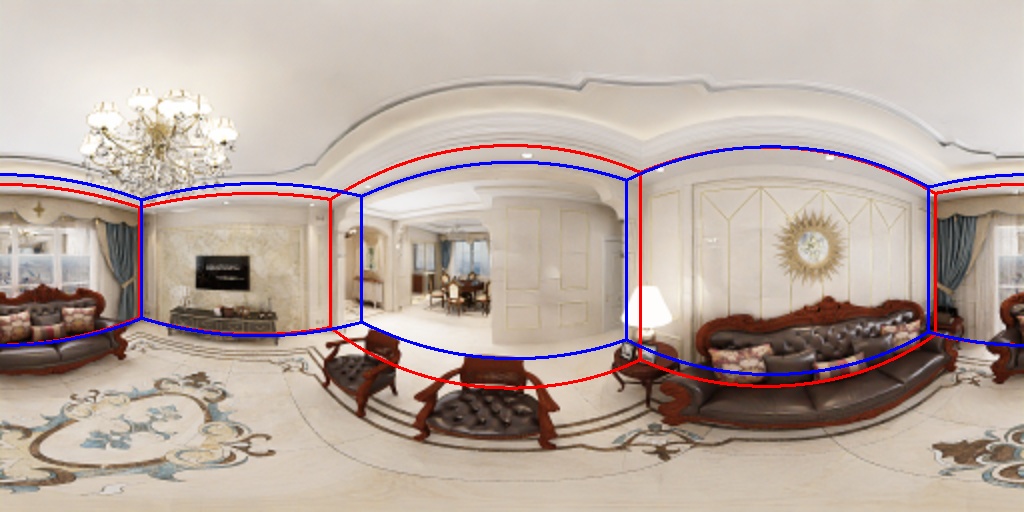}}
    \hfill
    \subfloat{\includegraphics[width=0.24\linewidth]{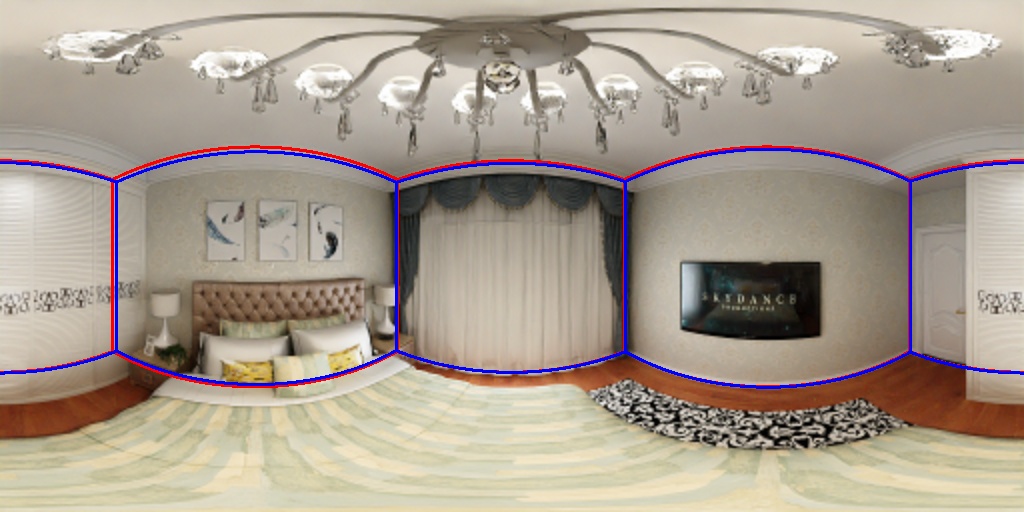}}
    \hfill
    \subfloat{\includegraphics[width=0.24\linewidth]{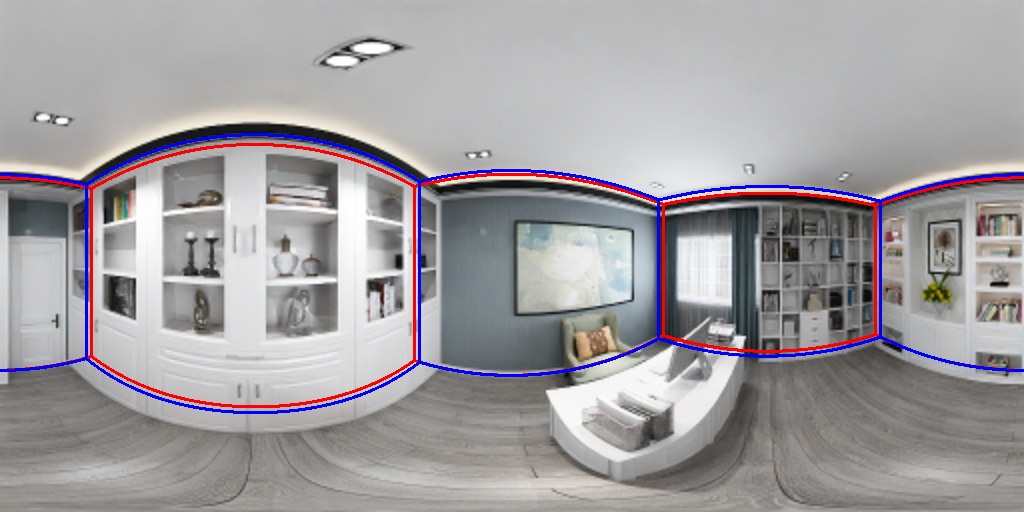}}
    \hfill
    \subfloat{\includegraphics[width=0.24\linewidth]{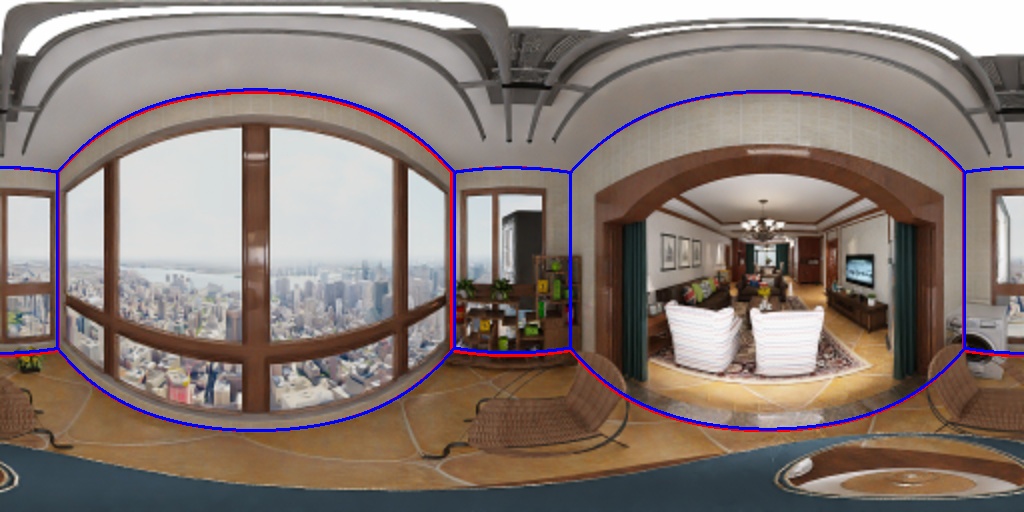}}
    \hfill
    \vspace{-0.3 cm}
    \subfloat{\includegraphics[width=0.24\linewidth]{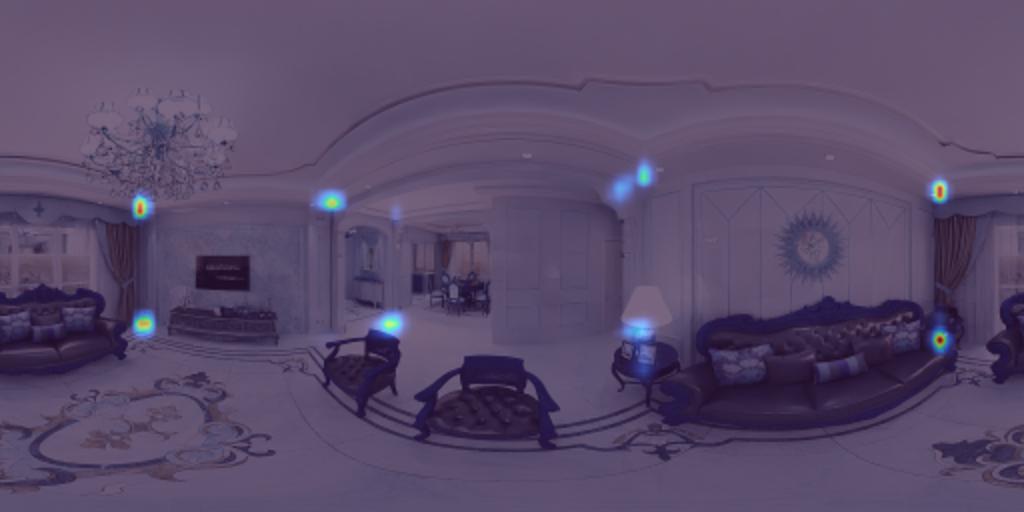}}
    \hfill
    \subfloat{\includegraphics[width=0.24\linewidth]{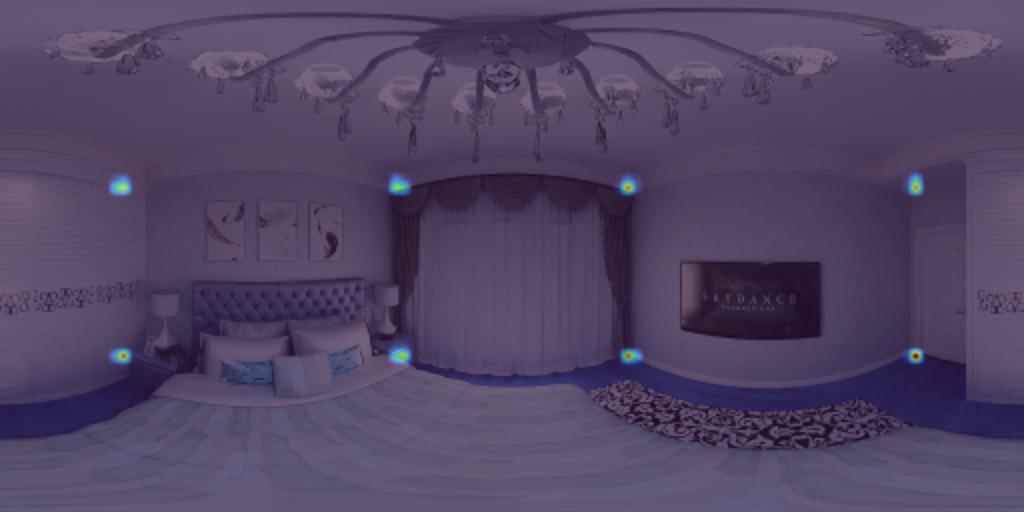}}
    \hfill
    \subfloat{\includegraphics[width=0.24\linewidth]{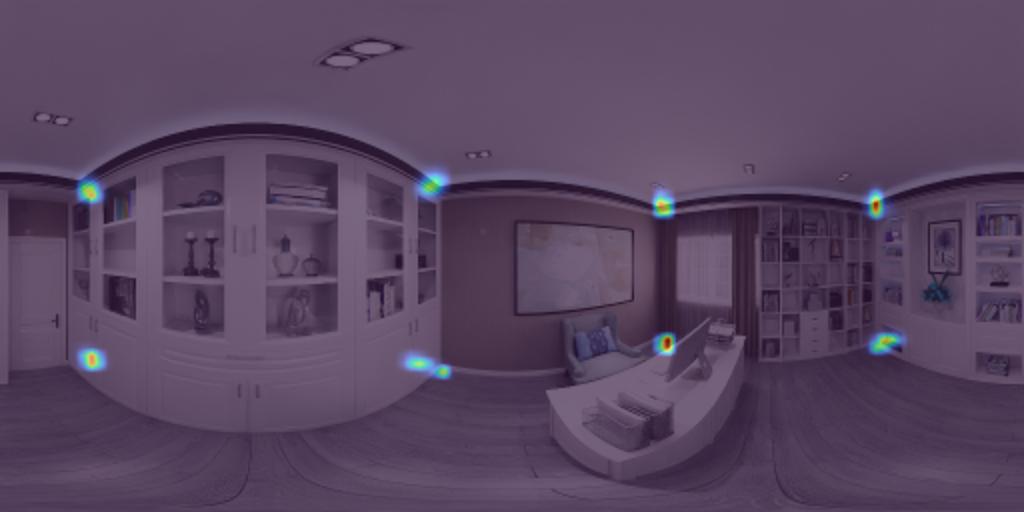}}
    \hfill
    \subfloat{\includegraphics[width=0.24\linewidth]{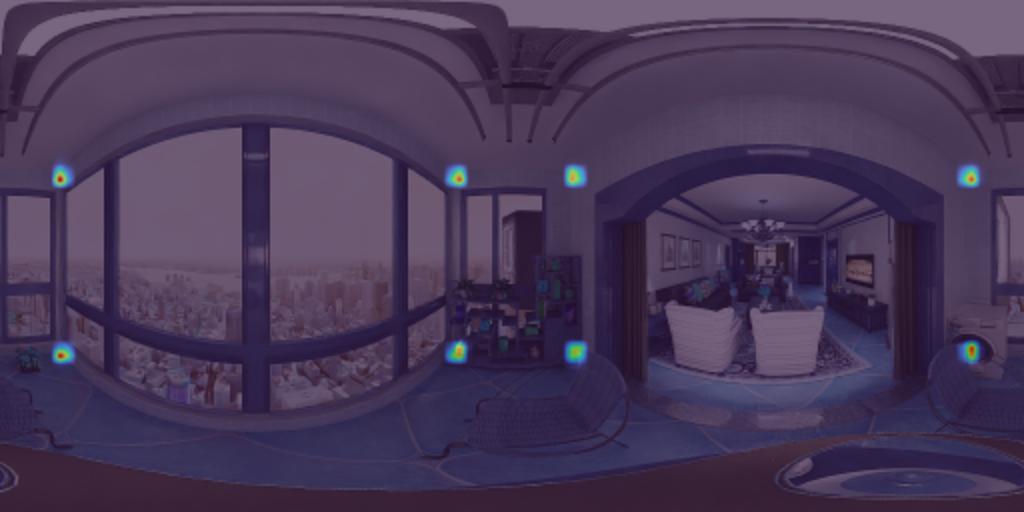}}
    \hfill
    \vspace{-0.35 cm}
    \subfloat{\includegraphics[width=0.24\linewidth,height=0.20\linewidth]{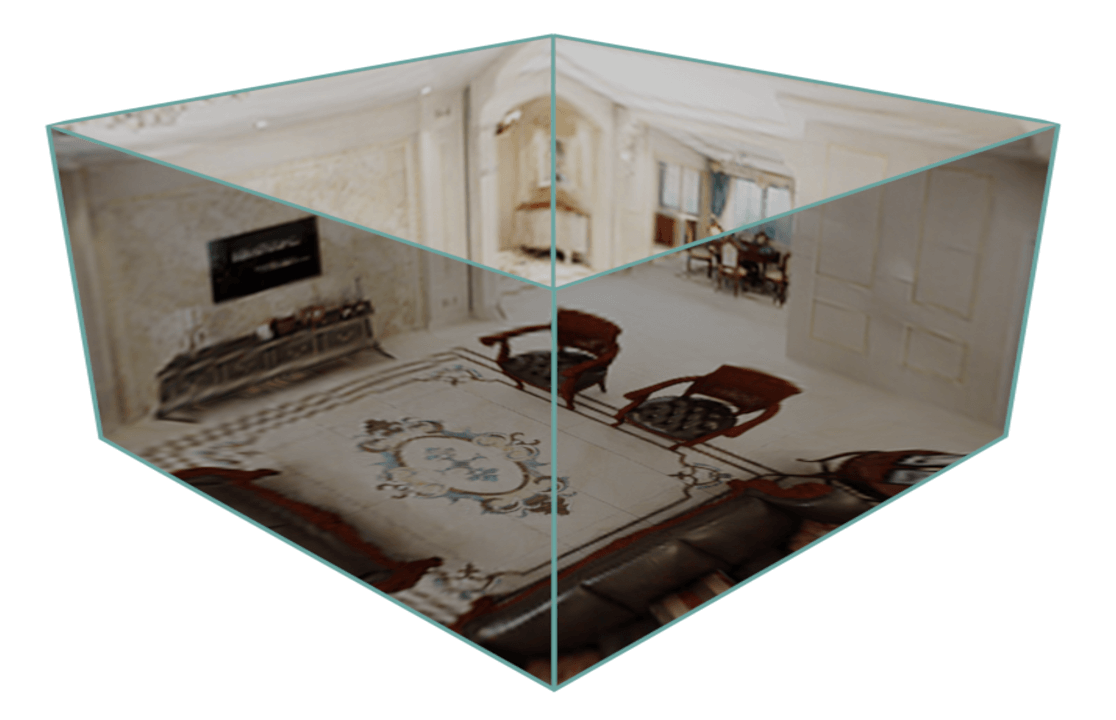}}
    \hfill
    \subfloat{\includegraphics[width=0.24\linewidth,height=0.20\linewidth]{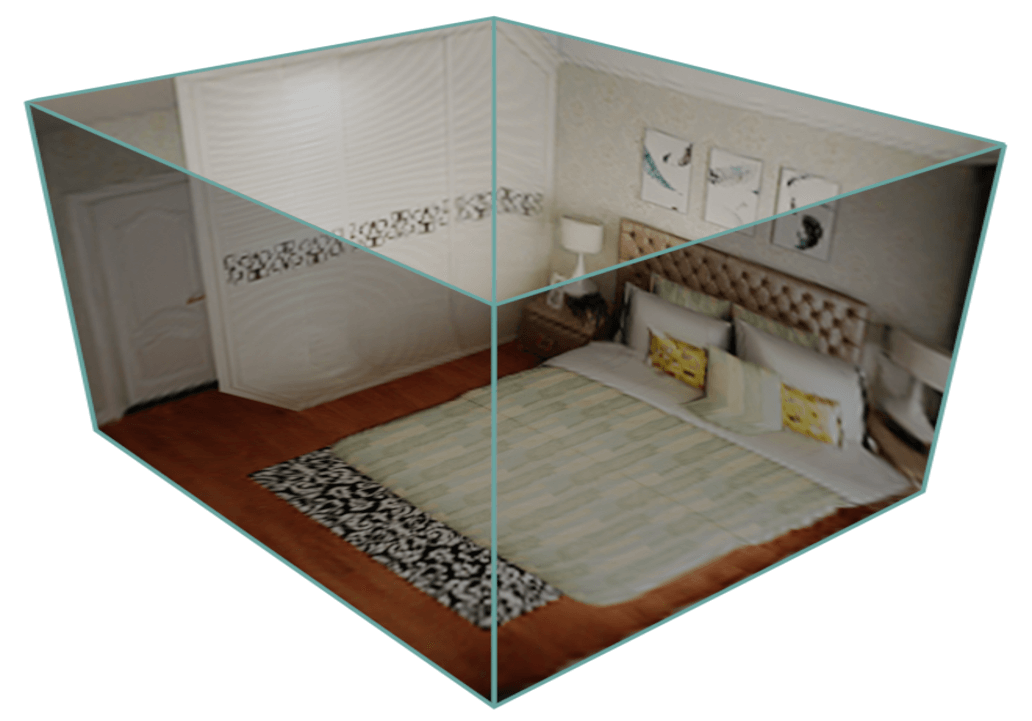}}
    \hfill
    \subfloat{\includegraphics[width=0.24\linewidth,height=0.20\linewidth]{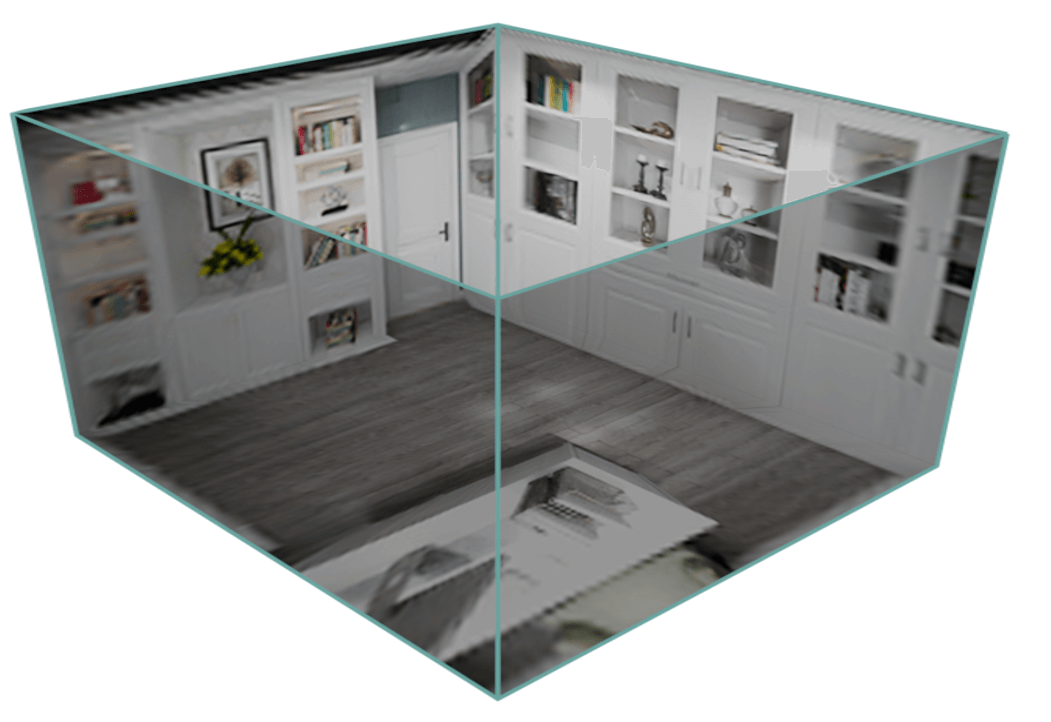}}
    \hfill
    \subfloat{\includegraphics[width=0.24\linewidth,height=0.20\linewidth]{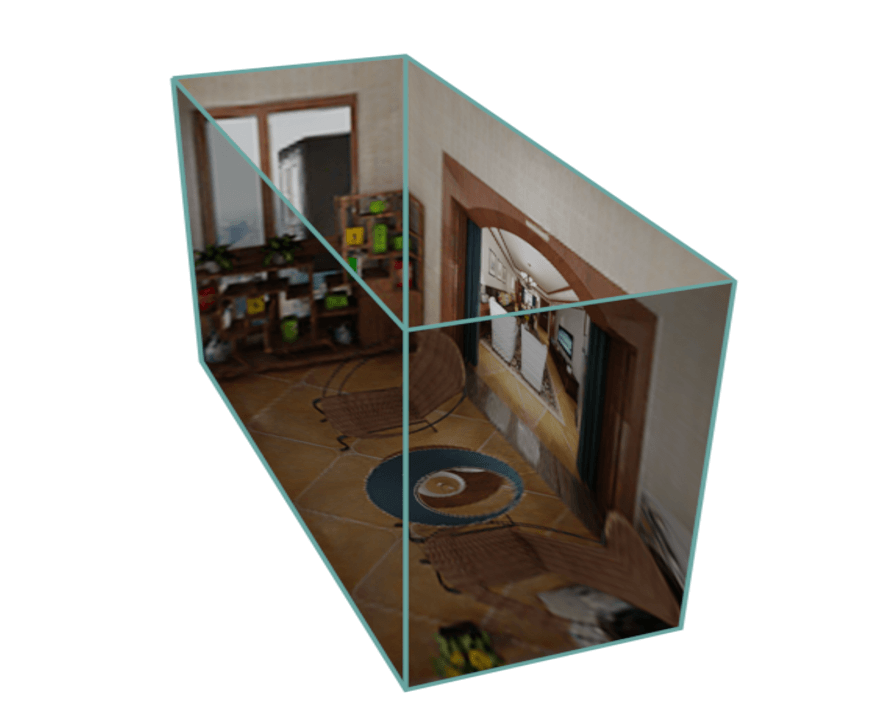}}
    \hfill
    \vspace{-0.35 cm}
    \subfloat{\includegraphics[width=0.12\linewidth, height=2.5cm]{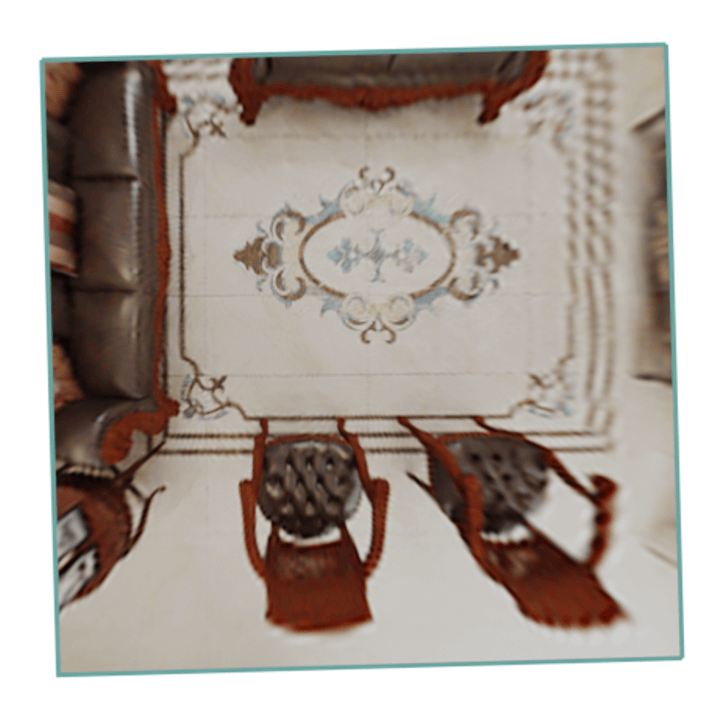}}
    \hfill
    \subfloat{\includegraphics[width=0.13\linewidth, height=2.5cm]{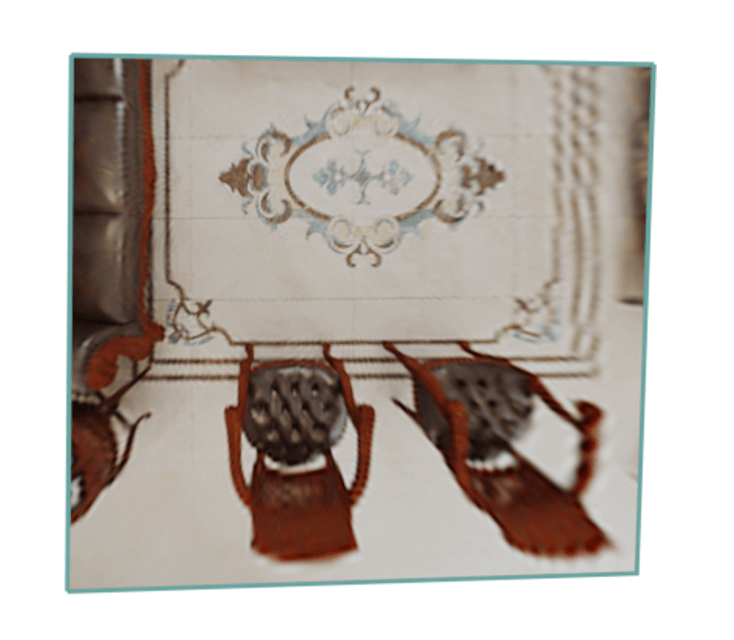}}
    \hfill
    \subfloat{\includegraphics[width=0.12\linewidth, height=2.5cm]{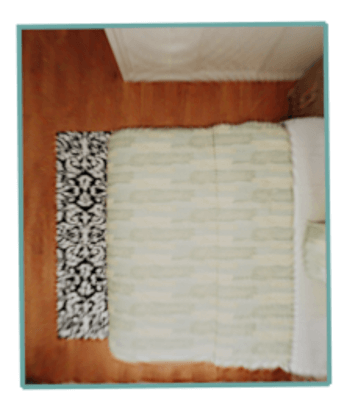}}
    \hfill
    \subfloat{\includegraphics[width=0.12\linewidth, height=2.5cm]{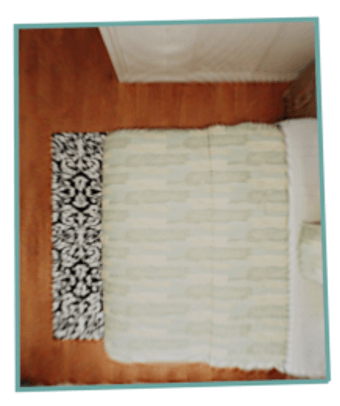}}
    \hfill
    \subfloat{\includegraphics[width=0.12\linewidth, height=2.5cm]{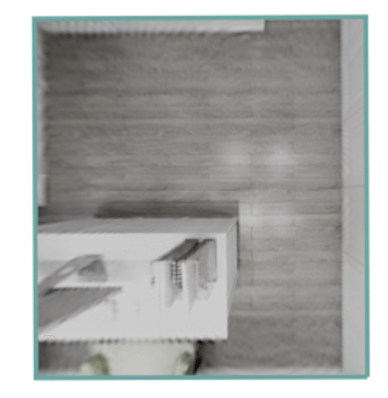}}
    \hfill
    \subfloat{\includegraphics[width=0.14\linewidth, height=2.5cm]{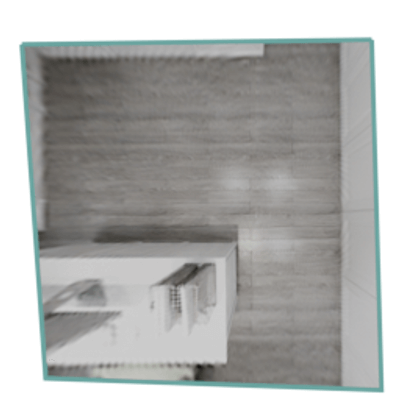}}
    \hfill
    \subfloat{\includegraphics[width=0.12\linewidth, height=2.5cm]{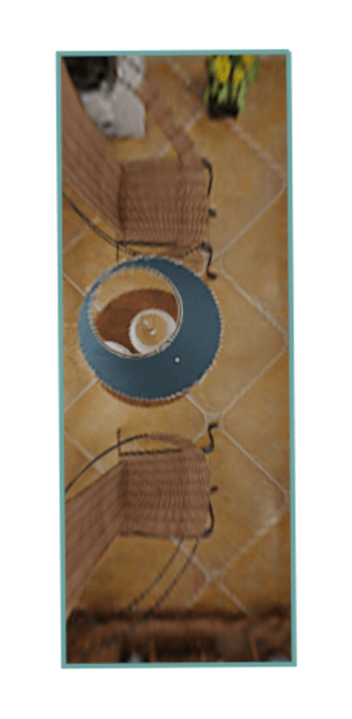}}
    \hfill
    \subfloat{\includegraphics[width=0.12\linewidth, height=2.5cm]{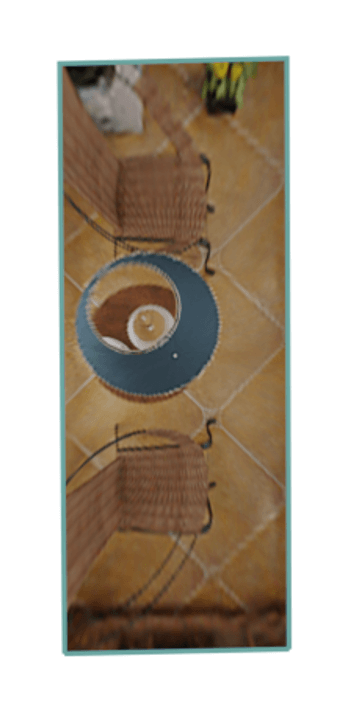}}
    \hfill
    \vspace{-0.3cm}
    \subfloat{\includegraphics[width=0.24\linewidth]{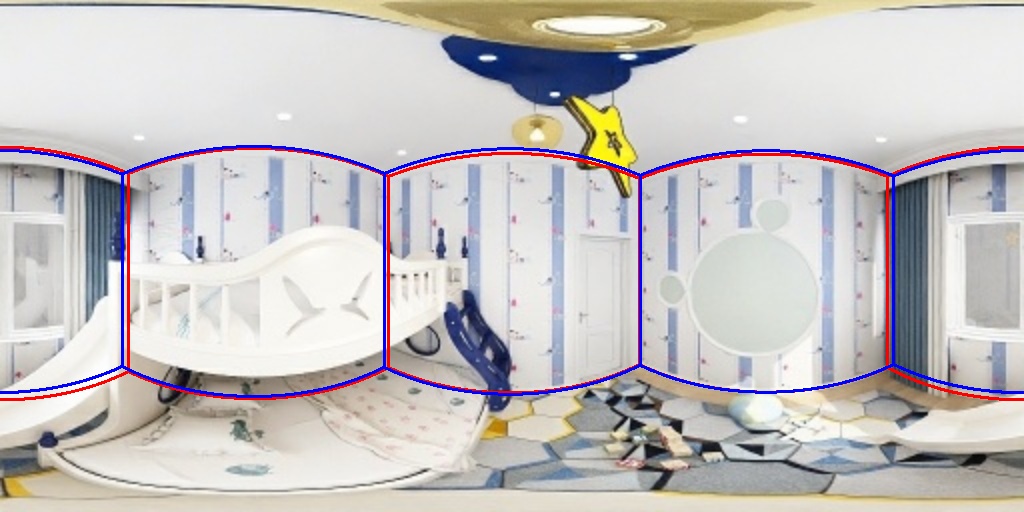}}
    \hfill
    \subfloat{\includegraphics[width=0.24\linewidth]{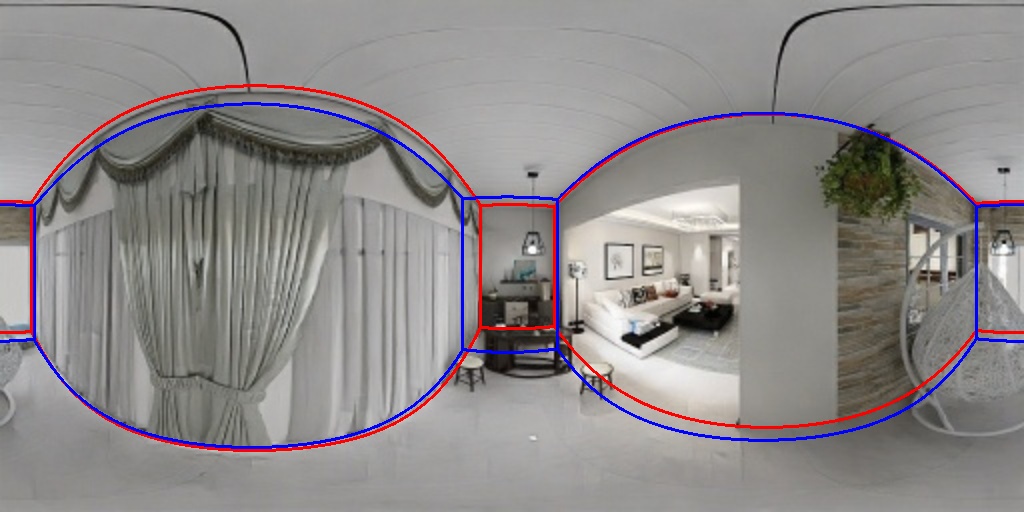}}
    \hfill
    \subfloat{\includegraphics[width=0.24\linewidth]{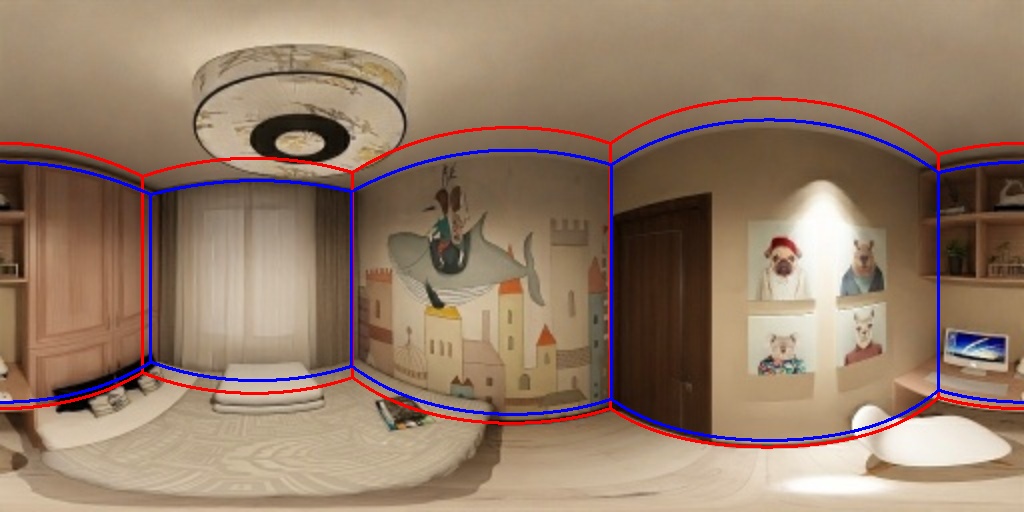}}
    \hfill
    \subfloat{\includegraphics[width=0.24\linewidth]{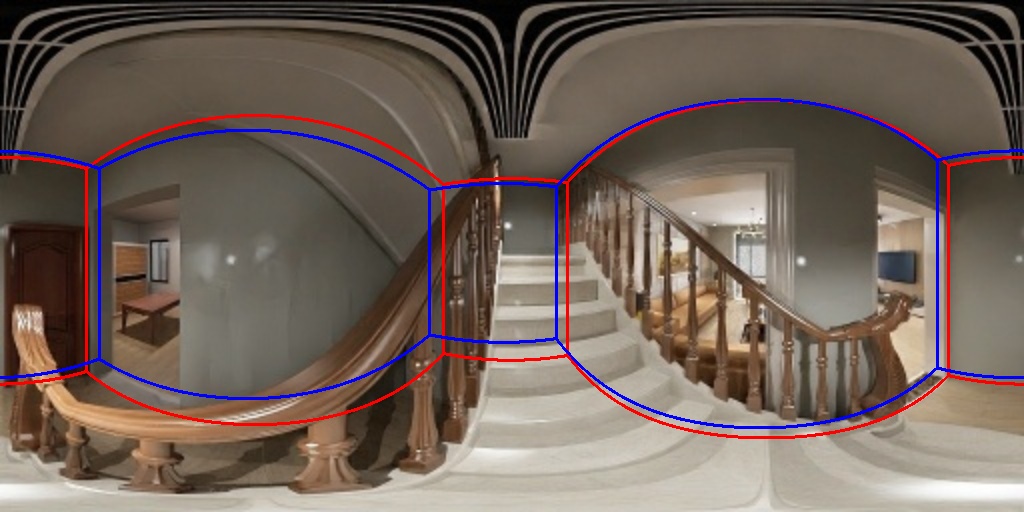}}
    \hfill
    \vspace{-0.3 cm}
    \subfloat{\includegraphics[width=0.24\linewidth]{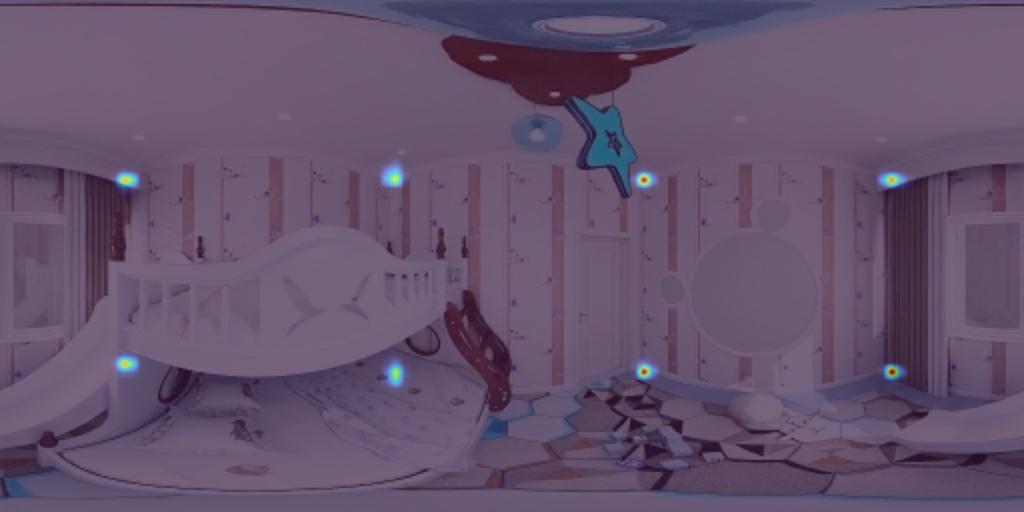}}
    \hfill
    \subfloat{\includegraphics[width=0.24\linewidth]{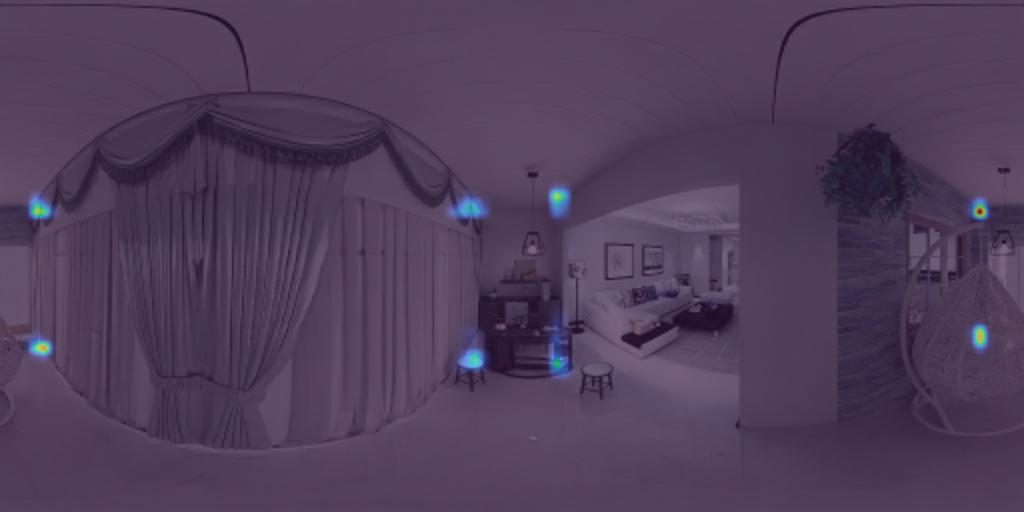}}
    \hfill
    \subfloat{\includegraphics[width=0.24\linewidth]{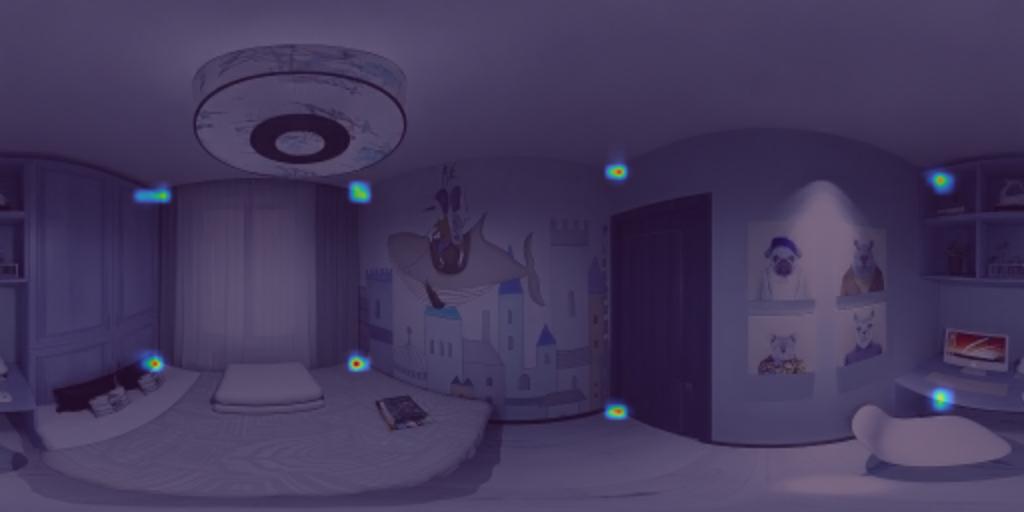}}
    \hfill
    \subfloat{\includegraphics[width=0.24\linewidth]{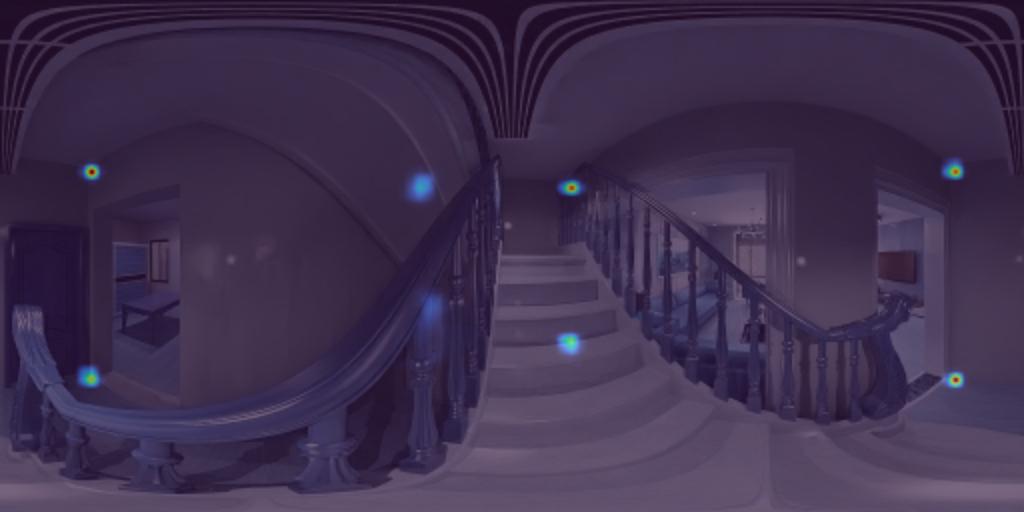}}
    \hfill
    \vspace{-0.35 cm}
    \subfloat{\includegraphics[width=0.24\linewidth,height=0.20\linewidth]{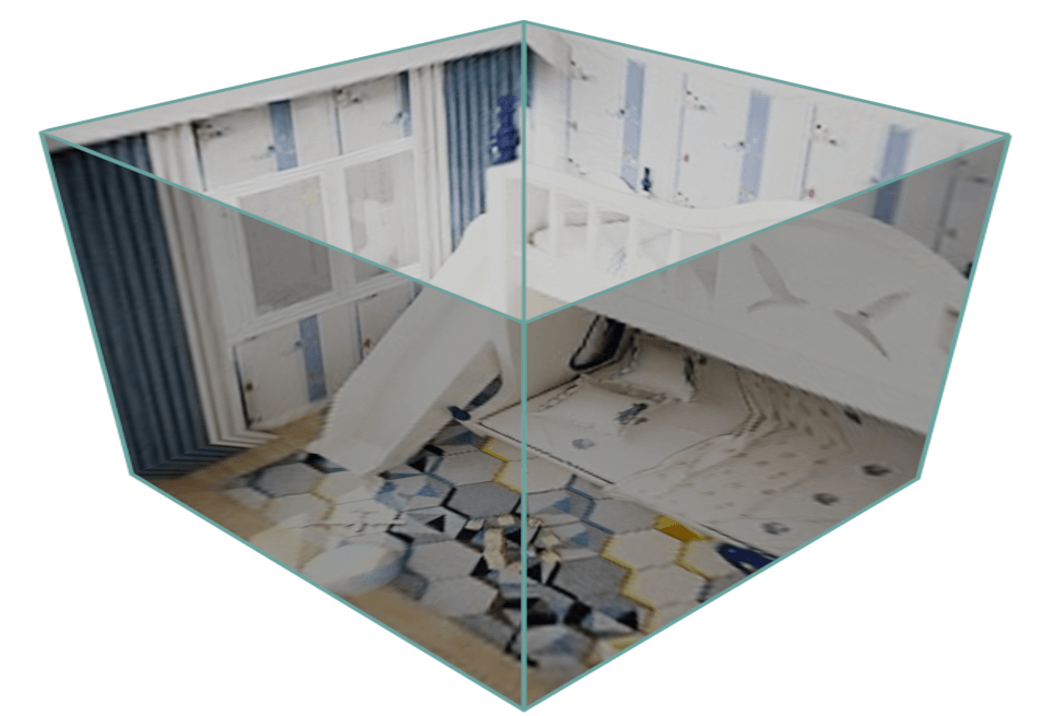}}
    \hfill
    \subfloat{\includegraphics[width=0.24\linewidth,height=0.20\linewidth]{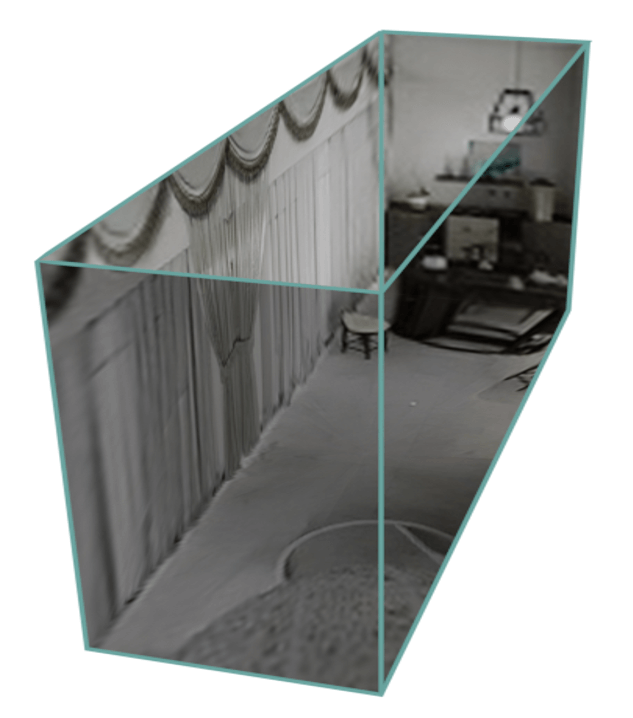}}
    \hfill
    \subfloat{\includegraphics[width=0.24\linewidth,height=0.20\linewidth]{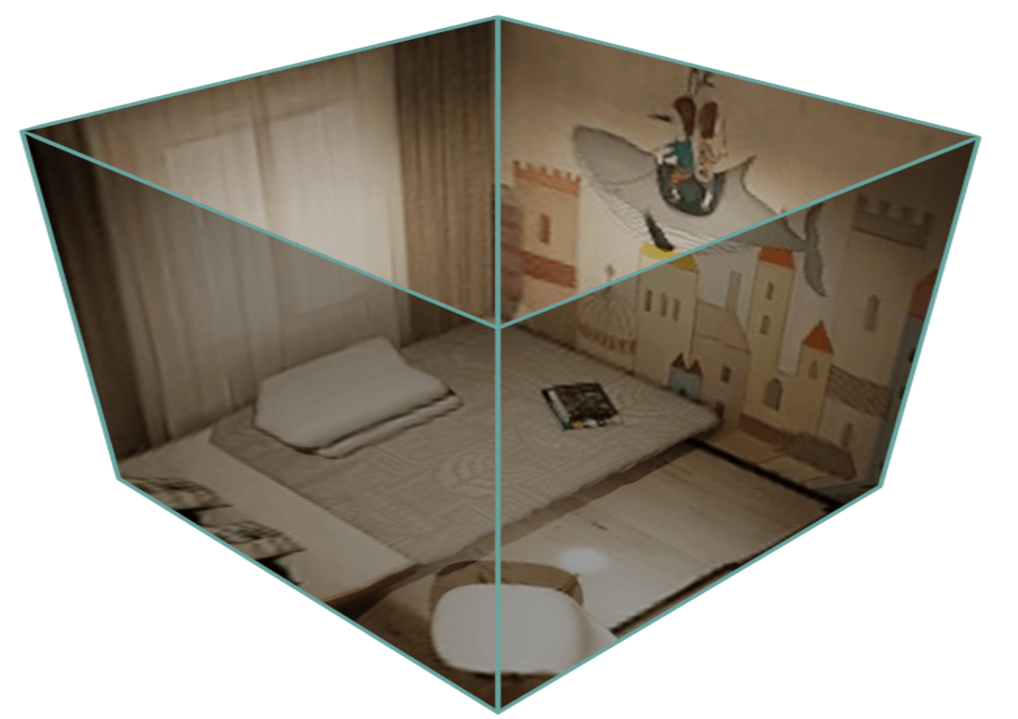}}
    \hfill
    \subfloat{\includegraphics[width=0.24\linewidth,height=0.20\linewidth]{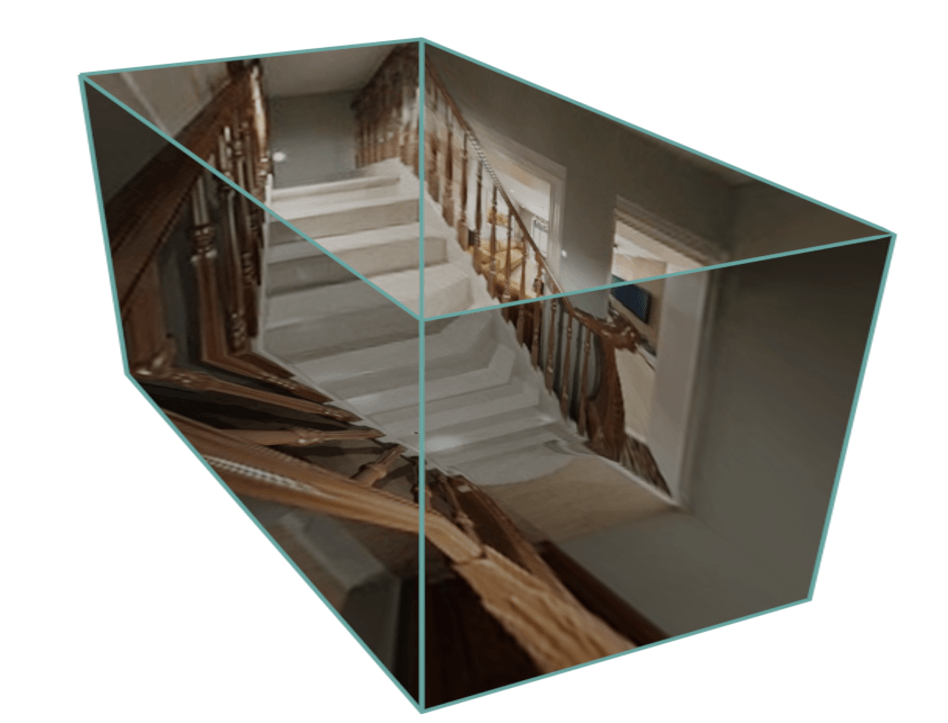}}
    \hfill
    \vspace{-0.35 cm}
    \subfloat{\includegraphics[width=0.11\linewidth, angle=90]{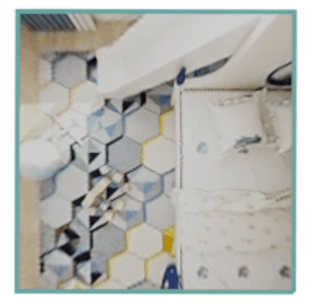}}
    \hfill
    \subfloat{\includegraphics[width=0.115\linewidth, angle=90]{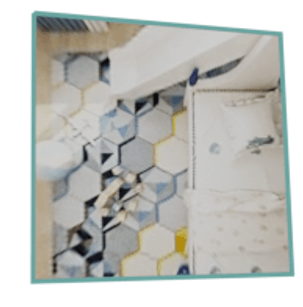}}
    \hspace{0.8cm}
    \subfloat{\includegraphics[width=0.06\linewidth]{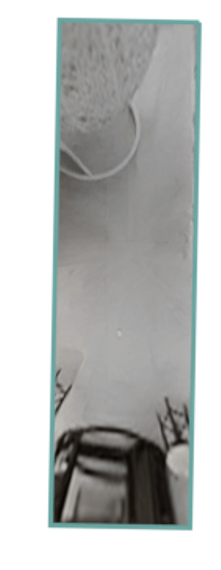}}
    \hfill
    \subfloat{\includegraphics[width=0.05\linewidth]{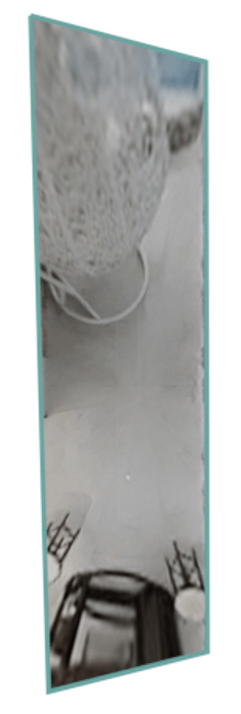}}
    \hspace{0.9cm}
    \subfloat{\includegraphics[width=0.11\linewidth]{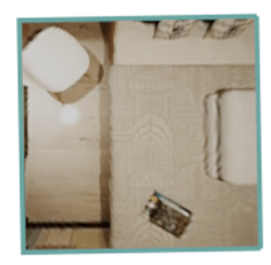}}
    \hfill
    \subfloat{\includegraphics[width=0.12\linewidth]{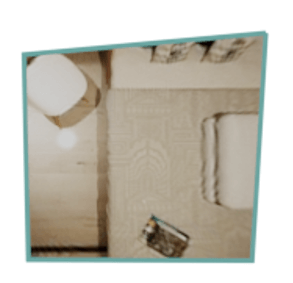}}
    \hspace{0.5cm}
    \subfloat{\includegraphics[width=0.105\linewidth]{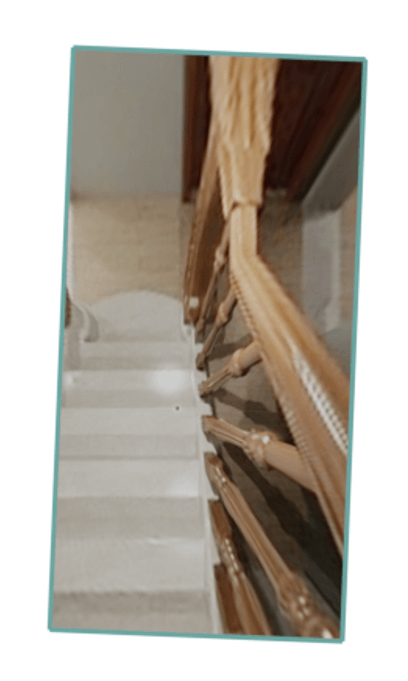}}
    \hfill
    \subfloat{\includegraphics[height=0.19\linewidth]{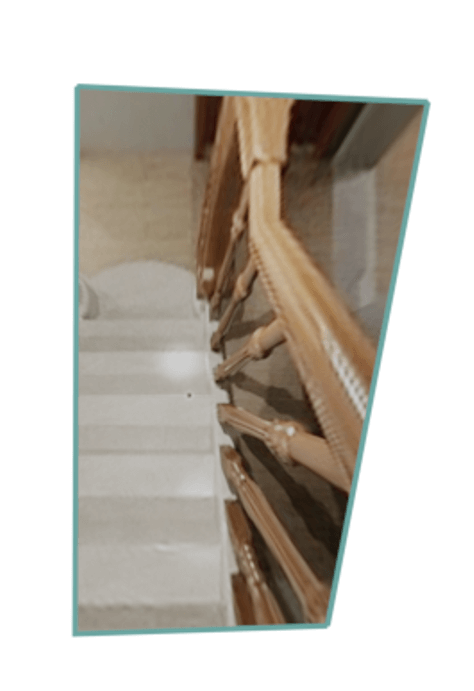}}
    \hfill
    
    \caption{
        Qualitative results on the Structured3D (top) and Kujiale (bottom) datasets.
        Same scheme as Figure~\ref{fig:real} applies.
    }
\label{fig:syn}
\end{figure*}

\begin{table*}[!htbp]
\centering
\caption{Quantitative results and ablation on the synthetic Structured3D synthetic dataset.}
\label{tab:s3d}
\resizebox{\textwidth}{!}{%
\begin{tabular}{|cc|cccc|cccccc|cc|}
\hline
Model &
  Variant &
  \down{CE} &
  \up{IoU2D} &
  \up{IoU3D} &
  \down{PE} &
  \up{$J_5$} &
  \up{$J_{10}$} &
  \up{$J_{15}$} &
  \up{$W_5$} &
  \up{$W_{10}$} &
  \up{$W_{15}$} &
  \down{RMSE} &
  \up{$\delta_1$} \\ \hline
\multirow{2}{*}{HNet} &
  Single-Shot &
  0.57\% &
  93.10\% &
  91.17\% &
  1.53\% &
  \second{78.20\%} &
  90.69\% &
  95.09\% &
  \first{56.67\%} &
  \second{77.50\%} &
  \second{86.53\%} &
  0.0712 &
  97.84\% \\
 &
  Postprocessed &
  0.75\% &
  93.49\% &
  91.82\% &
  1.46\% &
  \first{78.61\%} &
  \first{91.64\%} &
  \second{95.75\%} &
  \first{56.67\%} &
  \first{78.22\%} &
  \first{87.57\%} &
  0.0756 &
  98.58\% \\ \hline
\multirow{5}{*}{\begin{tabular}[c]{@{}c@{}}SSC\\ HG-3\end{tabular}} &
  Quasi-Manhattan &
  \first{0.39\%} &
  \second{93.97\%} &
  \third{92.00\%} &
  \second{1.25\%} &
  75.18\% &
  \second{90.96\%} &
  \first{95.82\%} &
  \second{49.27\%} &
  74.22\% &
  \third{86.29\%} &
  \third{0.0667} &
  \first{98.80\%} \\
 &
  w/ Homography (joint) &
  \second{0.40\%} &
  \first{94.27\%} &
  \first{92.33\%} &
  \third{1.26\%} &
  \third{75.35\%} &
  \third{90.90\%} &
  \third{95.74\%} &
  48.16\% &
  \third{74.35\%} &
  85.68\% &
  \first{0.0626} &
  \second{98.76\%} \\
 &
  w/o Geodesics &
  \first{0.39\%} &
  \third{93.94\%} &
  \second{92.03\%} &
  \second{1.25\%} &
  73.95\% &
  90.14\% &
  95.35\% &
  \third{49.26\%} &
  73.20\% &
  84.98\% &
  0.0671 &
  98.71\% \\
 &
  w/o Model Adaptation &
  \third{0.45\%} &
  93.15\% &
  91.04\% &
  1.44\% &
  71.10\% &
  88.19\% &
  94.35\% &
  43.01\% &
  69.59\% &
  82.74\% &
  0.0800 &
  98.20\% \\
 &
  w/o Quasi-Manhattan &
  \first{0.39\%} &
  93.89\% &
  \third{92.00\%} &
  \first{1.23\%} &
  73.43\% &
  90.42\% &
  95.63\% &
  47.38\% &
  72.63\% &
  85.00\% &
  \second{0.0651} &
  \third{98.73\%} \\ \hline
\end{tabular}%
}
\end{table*}
\begin{table*}[!htbp]
\centering
\caption{Cross-validation results on the Kujiale dataset using the Structured3D trained model.}
\label{tab:s3d2kuj}
\resizebox{\textwidth}{!}{%
\begin{tabular}{@{}|cccccccccccccc|@{}}
\toprule
Model &
  Variant &
  \down{CE} &
  \up{IoU2D} &
  \up{IoU3D} &
  \down{PE} &
  \up{$J_5$} &
  \up{$J_{10}$} &
  \up{$J_{15}$} &
  \up{$W_5$} &
  \up{$W_{10}$} &
  \up{$W_{15}$} &
  \down{RMSE} &
  \up{$\delta_1$} \\ \midrule
\multirow{2}{*}{HNet} &
  \multicolumn{1}{c|}{Single-Shot} &
  \third{0.61\%} &
  \third{91.68\%} &
  \third{89.53\%} &
  \multicolumn{1}{c|}{1.83\%} &
  \first{72.27\%} &
  \first{87.91\%} &
  \second{94.09\%} &
  \first{46.27\%} &
  \first{71.30\%} &
  \multicolumn{1}{c|}{\second{81.88\%}} &
  \third{0.0899} &
  98.02\% \\
 &
  \multicolumn{1}{c|}{Postprocessed} &
  1.04\% &
  90.97\% &
  88.96\% &
  \multicolumn{1}{c|}{\third{1.82\%}} &
  \third{71.18\%} &
  86.59\% &
  92.95\% &
  \second{45.45\%} &
  \third{69.67\%} &
  \multicolumn{1}{c|}{\first{81.39\%}} &
  0.0967 &
  \third{98.29\%} \\ \midrule
\multirow{2}{*}{\begin{tabular}[c]{@{}c@{}}SSC\\ HG-3\end{tabular}} &
  \multicolumn{1}{c|}{Quasi-Manhattan} &
  \second{0.45\%} &
  \second{92.83\%} &
  \second{90.55\%} &
  \multicolumn{1}{c|}{\second{1.46\%}} &
  70.95\% &
  \third{86.86\%} &
  \third{93.41\%} &
  41.55\% &
  68.58\% &
  \multicolumn{1}{c|}{\second{81.88\%}} &
  \second{0.0811} &
  \second{98.40\%} \\
 &
  \multicolumn{1}{c|}{w/ Homography (joint)} &
  \first{0.42\%} &
  \first{93.37\%} &
  \first{91.21\%} &
  \multicolumn{1}{c|}{\first{1.38\%}} &
  \second{71.82\%} &
  \second{87.36\%} &
  \first{94.86\%} &
  \third{44.12\%} &
  \second{70.73\%} &
  \multicolumn{1}{c|}{\first{82.06\%}} &
  \first{0.0706} &
  \first{98.45\%} \\ \bottomrule
\end{tabular}%
}
\end{table*}

\subsection{Performance Analysis}
\label{sec:performance}
First, we focus on the latest results reported in \cite{zou20193d}, where three data-driven cuboid panoramic layout estimation methods (\cite{zou2018layoutnet, yang2019dula, sun2019horizonnet}) were adapted for fairer comparison.
Similar to \cite{zou20193d}, we train a $3$ stack (HG-3) single-shot cuboid (SSC) model using the real dataset.
We present results tested on real (combined and single) datasets in Table~\ref{tab:real} where our model compares favorably with the state-of-the-art\footnote{Best three performances are denoted with bold red, orange and yellow.}, offering robust performance and end-to-end Manhattan aligned estimates, a trait no other state-of-the-art method offers currently.
For these results, we report the same metrics as those reported in \cite{zou20193d}.
Furthermore, Figure \ref{fig:real} presents a set of qualitative results for our HG-3 model on these two datasets.

With the recent availability of large scale synthetic datasets, we additionally train a model using Structured3D \cite{Structured3D}.
Since only HorizonNet offers a pretrained model using the same data, we present results on the Structured3D test dataset for two HorizonNet variants and our model in Table~\ref{tab:s3d}.
Apart from the standard model that includes postprocessing, we also assess a single-shot variant of HorizonNet.
For this, we only perform peak detection on the predicted wall-to-wall boundary vector and directly sample the heights at the detected peaks to reconstruct the layout.
While this saves an amount of processing, the postprocessing scheme used by HorizonNet improves the results when applied to Structured3D's test set.
On the other hand, our model produces accurate layout corner estimates without any postprocessing.
While SSC outperforms HorizonNet in the established metrics, HorizonNet offers higher accuracy in the junction and wireframe metrics.
This is also the case for the cross-validation experiment that we present in Table~\ref{tab:s3d2kuj}.
We test the models trained using Structured3D on the test set of Kujiale, using only the full rooms.
The difference is this setting is that the single-shot variant of HorizonNet provides more accurate layout estimates than the postprocessed one.
This exposes the weakness of postprocessing approaches, which require empiric or heuristic tuning.
Nonetheless, this HorizonNet model is trained for general layout estimation, and the performance deviation might be related to this extra trait.
Qualitative results for our end-to-end model for both synthetic datasets are presented in Figure \ref{fig:syn}.

\subsection{Ablation Study}
\label{sec:ablation}
We perform an ablation study across all datasets.
Tables~ \ref{tab:real_ablation}, \ref{tab:s3d} and \ref{tab:kuj} present the results on the real and synthetic datasets\footnote{Our supplement offers results for each of the real datasets.}.
Our baseline is the model as presented in Section~\ref{sec:net} without the end-to-end Manhattan alignment homography module (Section~\ref{sec:full_manhattan}), but with the quasi-Manhattan alignment (Section~\ref{sec:quasi_manhattan}) offered by aligning the longitude of top and bottom corners.
Apart from adding the end-to-end Manhattan alignment module, we also ablate the effect of the geodesic heatmap and loss (Section~\ref{sec:geodesic}), the SH model adaptation (spherical padding, pre-activated residual blocks and anti-aliased maxpooling - Section~\ref{sec:model_adaptation}), and the quasi-Manhattan alignment itself by training a model with unrestricted, traditional (\textit{i.e.}not spherical as presented in Section~\ref{sec:scom}) CoM calculation for each corner.

These offer a number of insights.
While the end-to\hyp{}end model provides the more robust performance across all datasets, its performance is uncontested in the IoU and depth related metrics. 
However, on the remaining projective metrics, the unrestricted coordinate regression approaches usually perform better.
This is reasonable as the homography fits a cuboid on the predictions, while the un-/semi-constrained approaches can freely localise the corners, even though at the expense of unnatural/Manhattan outputs, which manifests at an IoU3D drop.
Overall, we observe that the additional of explicit Manhattan constraints (quasi and homography-based) offer increased performance compared to directly regressing the corners.
The same applies to spherical (periodic CoM and geodesics) and model adaptation that consistently increase performance.

We also ablate the three approaches (floor/ceiling/joint) that use different starting coordinates for the homography estimation in Tables~\ref{tab:real_ablation} and \ref{tab:kuj}.
We find that the joint approach produces higher quality results, as it enforces both the top and bottom predictions to be consistent between them.
This way, the cuboid misalignment errors are backpropagated to all corner estimates through the homography.

\begin{table*}[!htbp]
\centering
\caption{Ablation study on the real dataset.}
\label{tab:real_ablation}
\resizebox{\textwidth}{!}{%
\begin{tabular}{|c|cccc|cccccc|cc|}
\hline
Variant &
  \down{CE} &
  \up{IoU2D} &
  \up{IoU3D} &
  \down{PE} &
  \up{$J_5$} &
  \up{$J_{10}$} &
  \up{$J_{15}$} &
  \up{$W_5$} &
  \up{$W_{10}$} &
  \up{$W_{15}$} &
  \down{RMSE} &
  \up{$\delta_1$} \\ \hline
Quasi-Manhattan &
  \first{0.55\%} &
  \third{87.90\%} &
  \third{85.02\%} &
  \second{1.74\%} &
  \first{61.54\%} &
  \first{84.04\%} &
  \third{91.02\%} &
  \first{30.61\%} &
  \first{57.97\%} &
  \first{74.93\%} &
  0.1734 &
  96.14\% \\ \hline
\cellcolor[HTML]{ECF4FF}w/ Homography (joint) &
  \second{0.57\%} &
  \first{88.39\%} &
  \second{85.89\%} &
  \first{1.70\%} &
  \third{55.01\%} &
  \third{80.62\%} &
  \first{91.75\%} &
  \third{20.98\%} &
  \second{52.63\%} &
  \second{71.07\%} &
  \first{0.1557} &
  \first{97.93\%} \\
w/ Homography (floor) &
  0.68\% &
  \second{88.25\%} &
  \first{85.97\%} &
  1.91\% &
  47.01\% &
  76.66\% &
  88.65\% &
  16.30\% &
  43.11\% &
  63.24\% &
  \second{0.1591} &
  \third{97.52\%} \\
w/ Homography (ceil) &
  0.63\% &
  87.63\% &
  85.25\% &
  1.88\% &
  52.79\% &
  \second{81.58\%} &
  \second{91.43\%} &
  18.50\% &
  \third{51.64\%} &
  69.73\% &
  \third{0.1671} &
  \second{97.64\%} \\ \hline
w/o Geodesics &
  0.79\% &
  84.40\% &
  81.08\% &
  2.31\% &
  33.78\% &
  70.05\% &
  86.80\% &
  4.66\% &
  26.33\% &
  53.39\% &
  0.2233 &
  95.48\% \\
w/o Model Adaptation &
  0.65\% &
  86.60\% &
  82.94\% &
  1.98\% &
  \second{55.41\%} &
  78.16\% &
  88.55\% &
  \second{22.73\%} &
  50.21\% &
  66.49\% &
  0.2033 &
  95.61\% \\
w/o Quasi-Manhattan &
  \third{0.61\%} &
  87.24\% &
  84.09\% &
  \third{1.81\%} &
  54.15\% &
  80.21\% &
  91.00\% &
  16.26\% &
  50.04\% &
  \third{69.62\%} &
  0.1874 &
  96.31\% \\ \hline
\end{tabular}%
}
\end{table*}
\begin{table*}[!htbp]
\centering
\caption{Ablation study on the synthetic Kujiale dataset.}
\label{tab:kuj}
\resizebox{\textwidth}{!}{%
\begin{tabular}{|c|cccc|cccccc|cc|}
\hline
Variant &
  \down{CE} &
  \up{IoU2D} &
  \up{IoU3D} &
  \down{PE} &
  \up{$J_5$} &
  \up{$J_{10}$} &
  \up{$J_{15}$} &
  \up{$W_5$} &
  \up{$W_{10}$} &
  \up{$W_{15}$} &
  \down{RMSE} &
  \up{$\delta_1$} \\ \hline
Quasi-Manhattan &
  \first{0.53\%} &
  \third{91.13\%} &
  88.43\% &
  \third{1.74\%} &
  \first{65.00\%} &
  \second{82.14\%} &
  \second{90.50\%} &
  \first{37.27\%} &
  \first{62.30\%} &
  \first{74.91\%} &
  0.0979 &
  96.99\% \\ \hline
w/ Homography (joint) &
  \first{0.53\%} &
  \first{91.40\%} &
  \first{89.01\%} &
  \first{1.70\%} &
  \second{63.36\%} &
  \first{82.55\%} &
  \first{90.55\%} &
  \second{35.91\%} &
  \second{61.79\%} &
  \second{74.24\%} &
  \first{0.0872} &
  \third{97.09\%} \\
w/ Homography (floor) &
  0.57\% &
  \second{91.28\%} &
  \second{88.72\%} &
  1.79\% &
  62.09\% &
  80.55\% &
  89.68\% &
  34.24\% &
  58.97\% &
  72.42\% &
  \third{0.0945} &
  \first{97.44\%} \\
w/ Homography (ceil) &
  \second{0.56\%} &
  90.92\% &
  \third{88.55\%} &
  1.78\% &
  61.68\% &
  80.82\% &
  89.59\% &
  \third{35.55\%} &
  \third{60.42\%} &
  72.48\% &
  \second{0.0925} &
  \second{97.13\%} \\ \hline
w/o Geodesics &
  0.59\% &
  90.81\% &
  88.31\% &
  1.81\% &
  59.55\% &
  79.36\% &
  89.64\% &
  27.64\% &
  56.39\% &
  71.24\% &
  0.0998 &
  96.92\% \\
w/o Model Adaptation &
  0.59\% &
  90.42\% &
  87.52\% &
  1.82\% &
  61.36\% &
  79.14\% &
  88.68\% &
  29.61\% &
  57.27\% &
  70.82\% &
  0.1026 &
  96.65\% \\
w/o Quasi-Manhattan &
  \second{0.54\%} &
  90.92\% &
  88.42\% &
  \second{1.73\%} &
  \third{62.59\%} &
  \third{80.91\%} &
  \third{90.36\%} &
  33.03\% &
  59.33\% &
  \third{73.36\%} &
  0.0962 &
  97.07\% \\ \hline
\end{tabular}%
}
\end{table*}

\section{Conclusion}
\label{sec:discussion}
Our work has focused on keypoint estimation on the sphere and in particular on layout corner estimation.
Through coordinate regression we integrate explicit constraints in our model.
Moreover, while we have also shown that end-to-end single-shot layout estimation is possible, our approach is rigid as it is based on a frequent and logical assumption, that the underlying room is, or can be approximated by, a cuboid.
Nonetheless, this rigidity comes from the structured predictions that CNN enforce, with the number of heatmaps that will be predicted being strictly defined at the design phase.
Future work should try to address this limitation to fully exploit the potential that single-shot approaches offer, mainly stemming from end-to-end supervision.
Finally, as with all prior layout estimation works, predictions are up to a scale, which hinders applicability.
Even so, structured scene layout estimation is an important task that can even be used as an intermediate task to improve other tasks, as shown in \cite{jin2020geometric}.
With metric scale inference, it has the potential for significant interplay with other 3D vision tasks like depth or surface estimation.

\section*{Supplement}
Supplementary material including additional ablation experiments and qualitative results are appended after the references.

\section*{Acknowledgements}
This work was supported by the EC funded H2020 project ATLANTIS [GA 951900].

{\small
\bibliographystyle{ieee_fullname}
\bibliography{egbib}
}

\clearpage
\appendix
\section{Supplementary Material}
\begin{figure}[!htbp]
    \centering
    \includegraphics[width=0.7\linewidth]{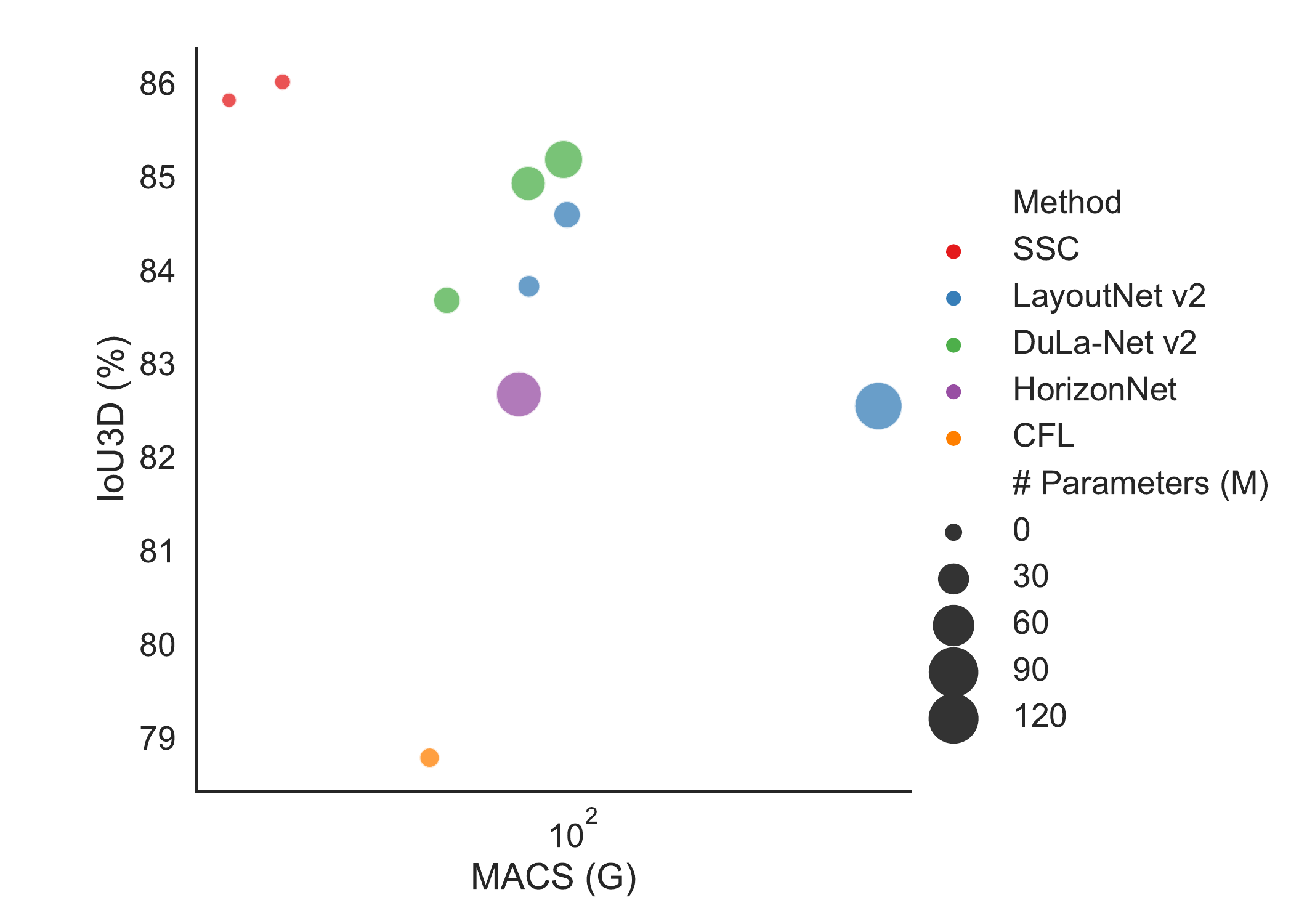}
    \caption{\textbf{Model Size vs Accuracy vs Complexity.}
    Visual comparison of spherical layout estimation models in terms of parameters (denoted by each bullet's size), computational complexity ($x$ axis, in $\log$ scale, billions of multiply-accumulate operations) and accuracy ($y$ axis, average IoU3D accuracy).
    Our model (SSC) is the most light-weight and offers a good comprise between complexity and accuracy, surpassing most other approaches.
    It also provides an end-to-end layout prediction in a single-shot, compared to all other approaches that require postprocessing.
    Different variants of each model are depicted.
    The exact data of this plot can be found in Table~\ref{tab:efficiency}.
    }
    \label{fig:efficiency}
\end{figure}

\begin{figure*}[!htbp]
    \centering

    \subfloat{\includegraphics[width=0.24\linewidth]{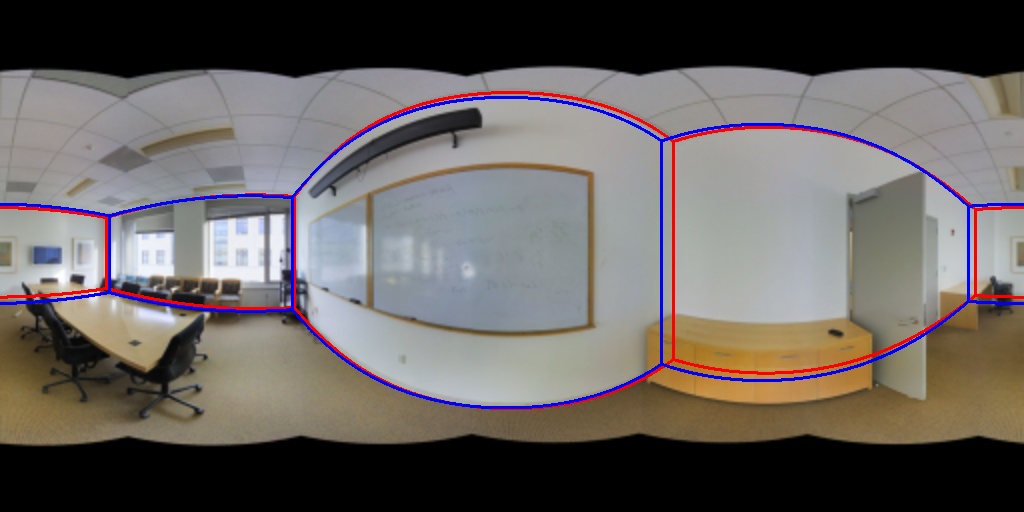}}
    \hfill
    \subfloat{\includegraphics[width=0.24\linewidth]{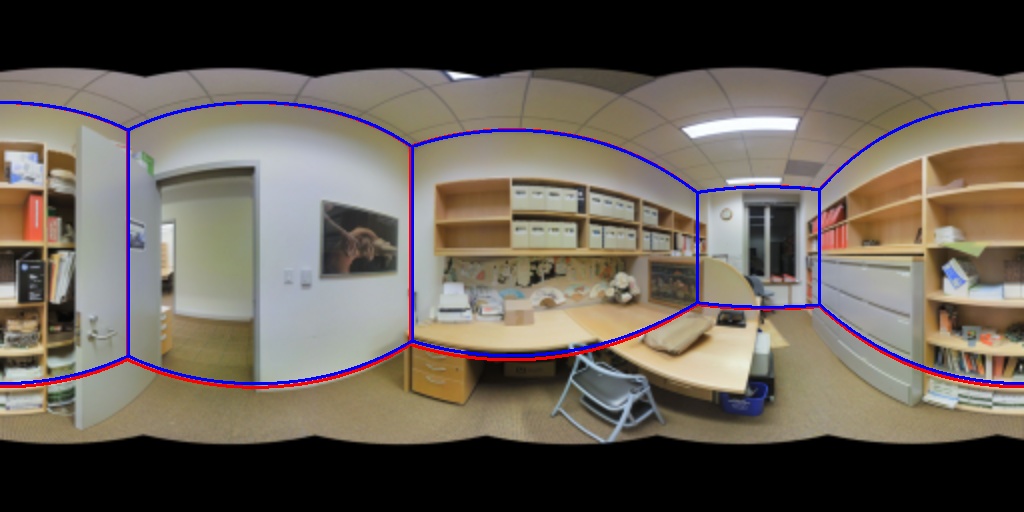}}
    \hfill
    \subfloat{\includegraphics[width=0.24\linewidth]{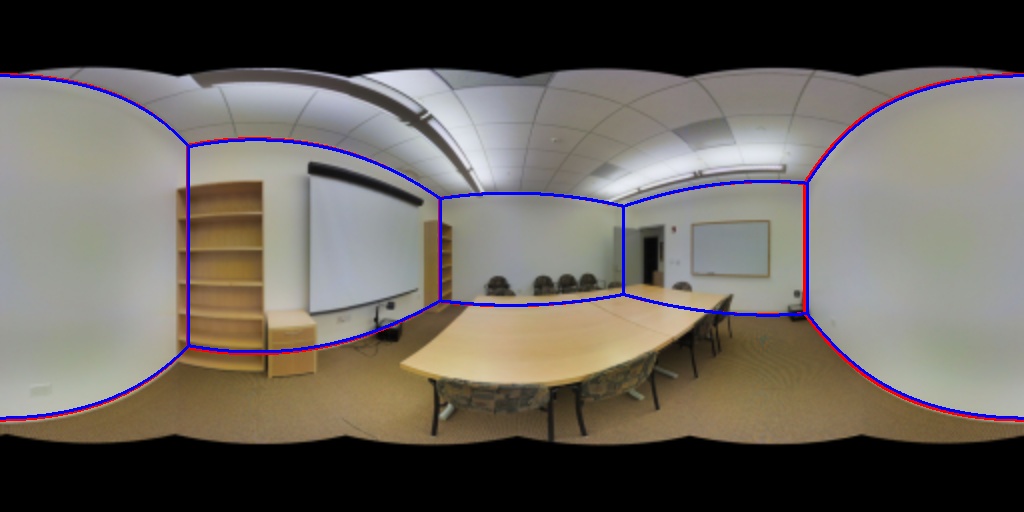}}
    \hfill
    \subfloat{\includegraphics[width=0.24\linewidth]{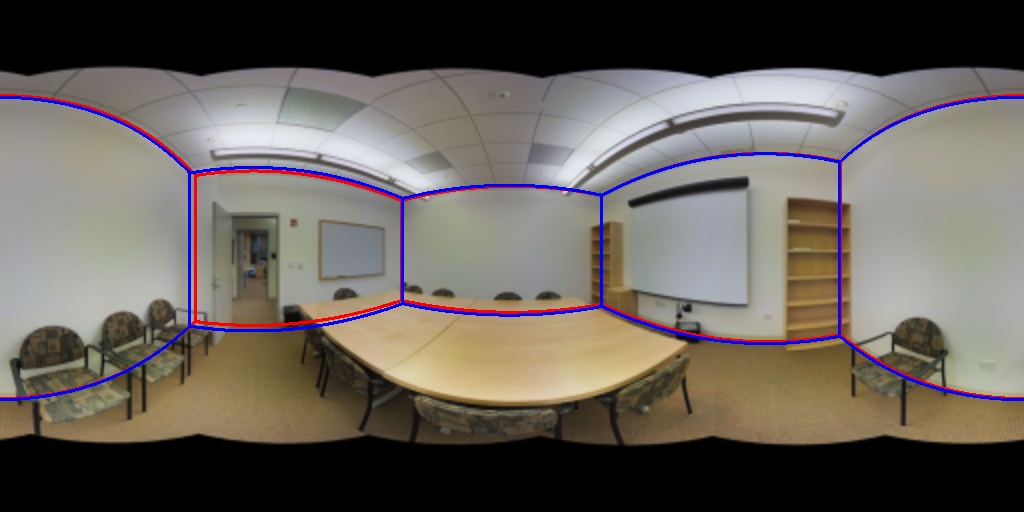}}
    \hfill
    
    \subfloat{\includegraphics[width=0.24\linewidth]{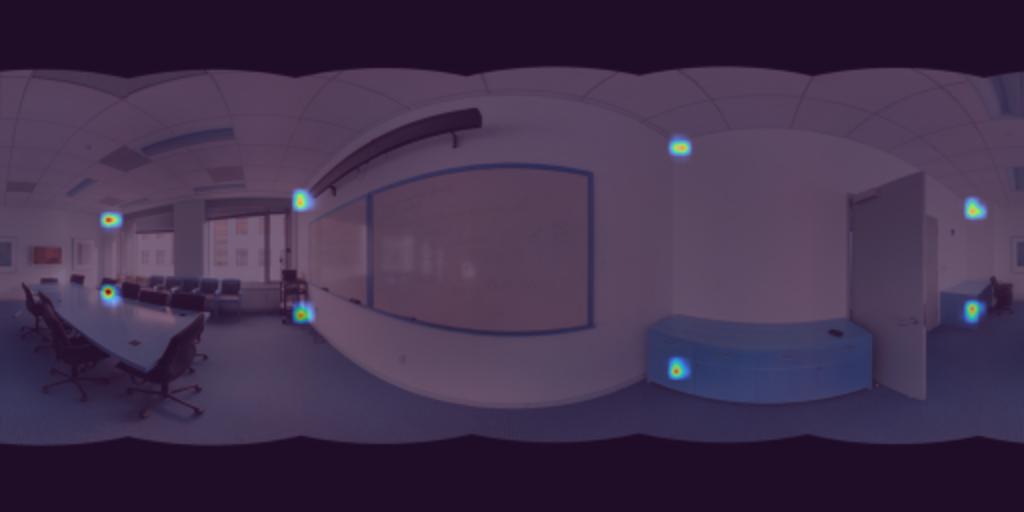}}
    \hfill
    \subfloat{\includegraphics[width=0.24\linewidth]{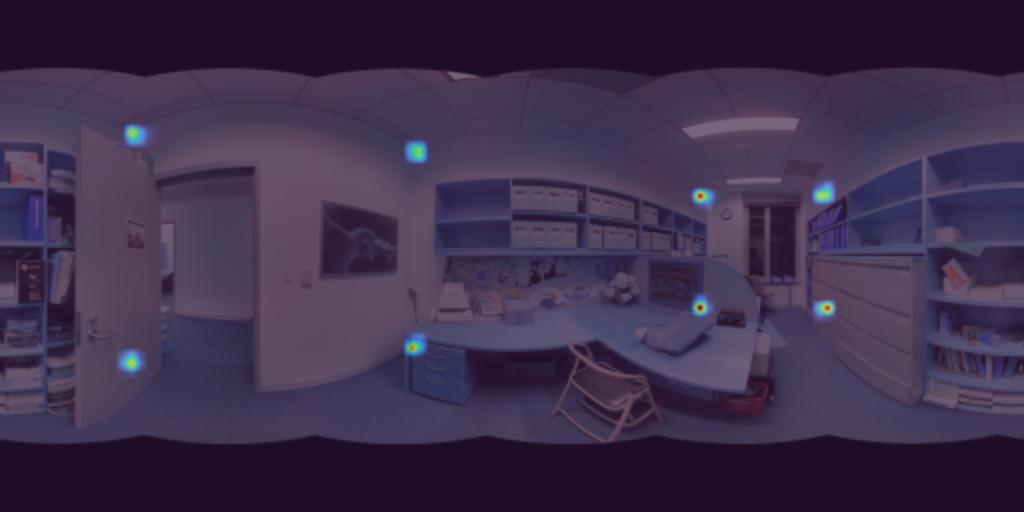}}
    \hfill
    \subfloat{\includegraphics[width=0.24\linewidth]{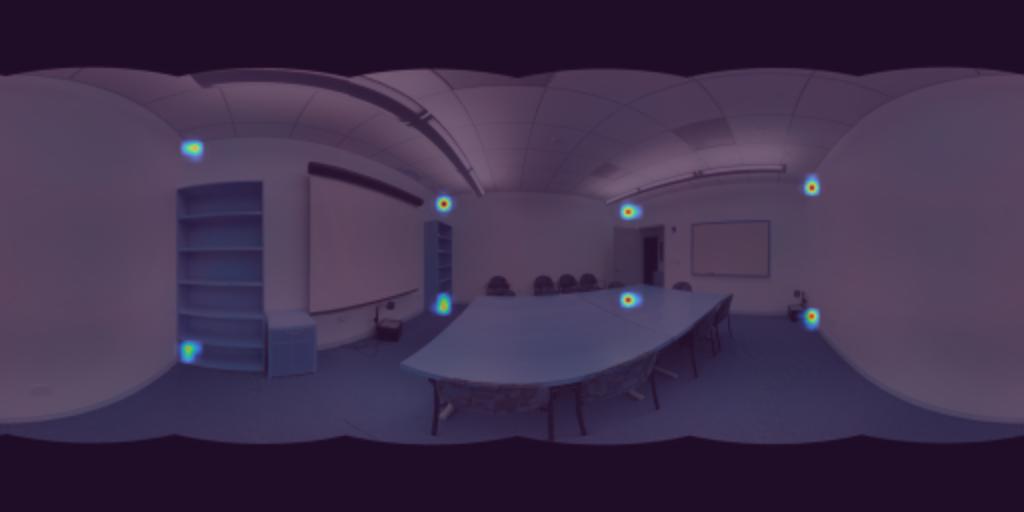}}
    \hfill
    \subfloat{\includegraphics[width=0.24\linewidth]{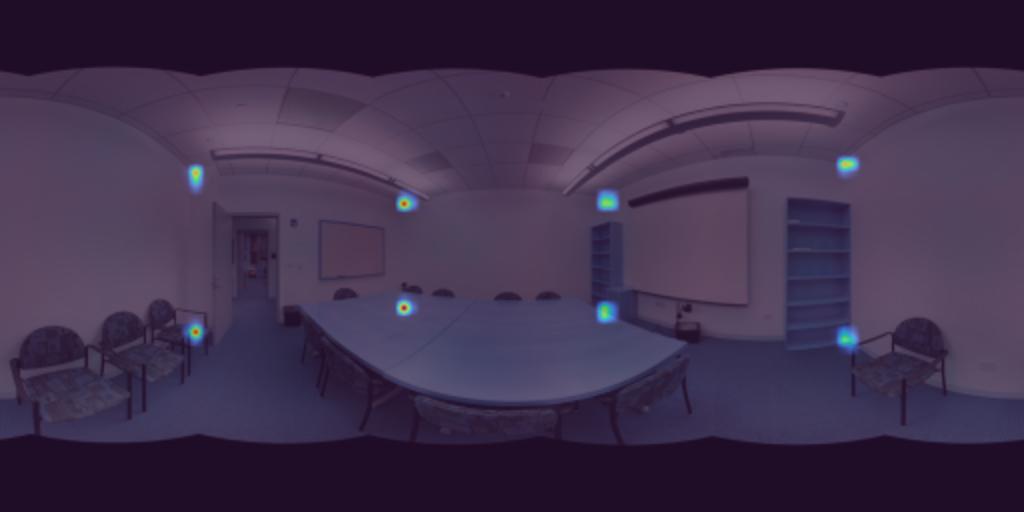}}
    \hfill
    
    \subfloat{\includegraphics[width=0.24\linewidth]{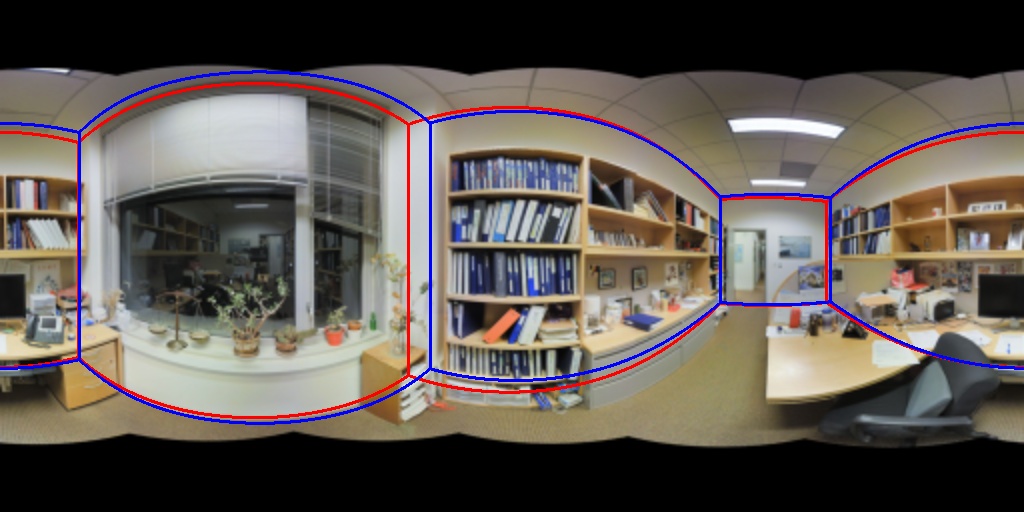}}
    \hfill
    \subfloat{\includegraphics[width=0.24\linewidth]{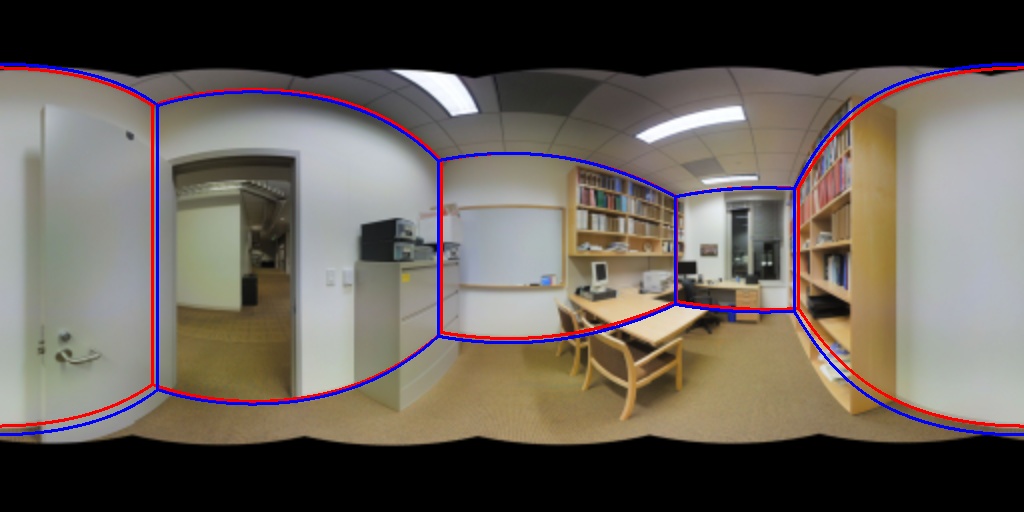}}
    \hfill
    \subfloat{\includegraphics[width=0.24\linewidth]{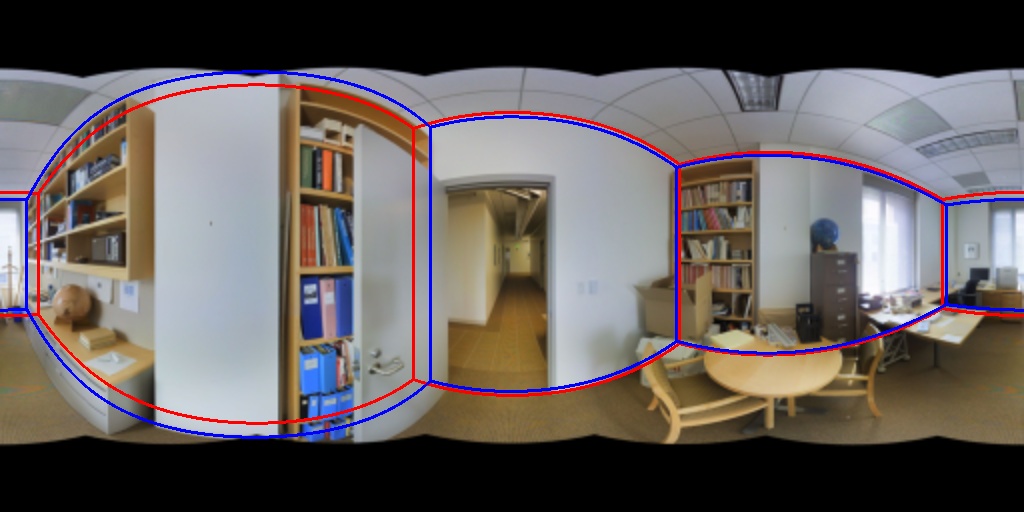}}
    \hfill
    \subfloat{\includegraphics[width=0.24\linewidth]{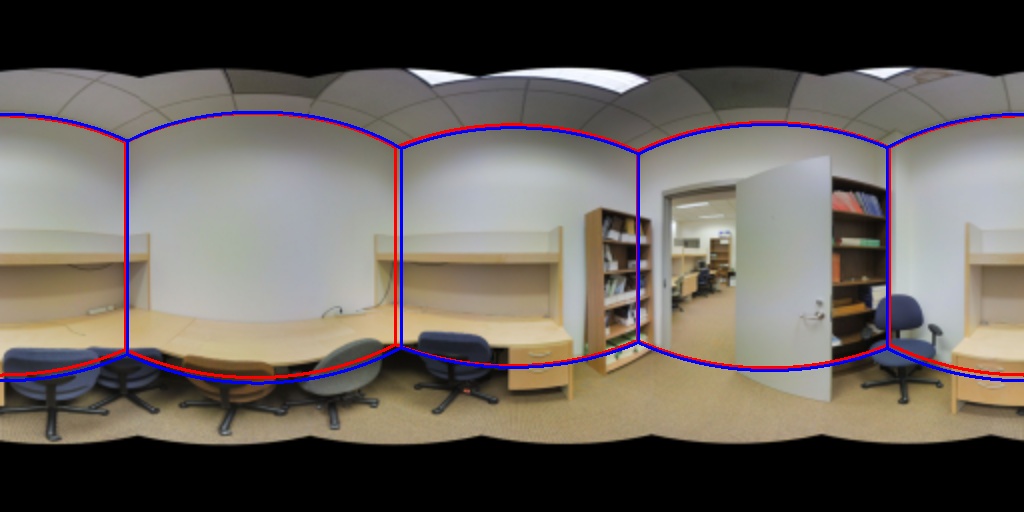}}
    \hfill
    
    \subfloat{\includegraphics[width=0.24\linewidth]{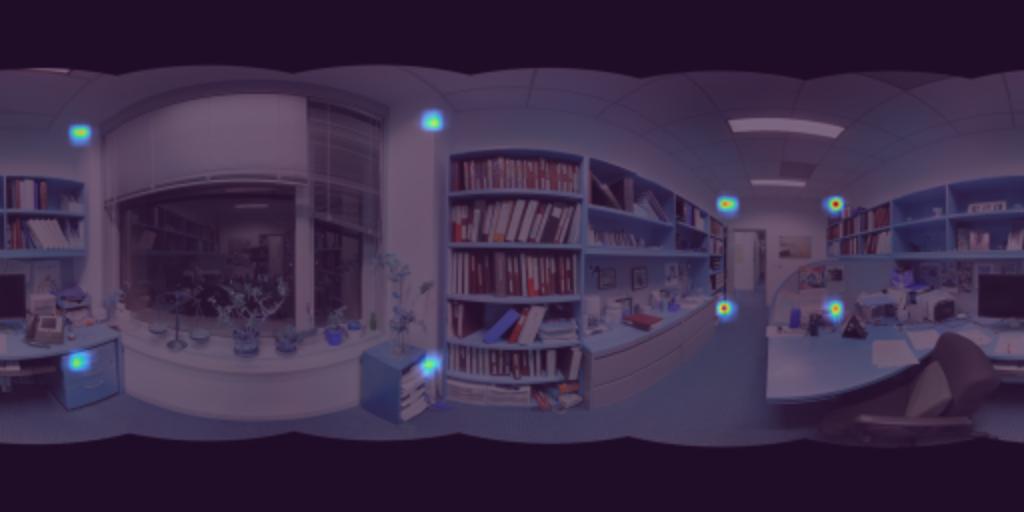}}
    \hfill
    \subfloat{\includegraphics[width=0.24\linewidth]{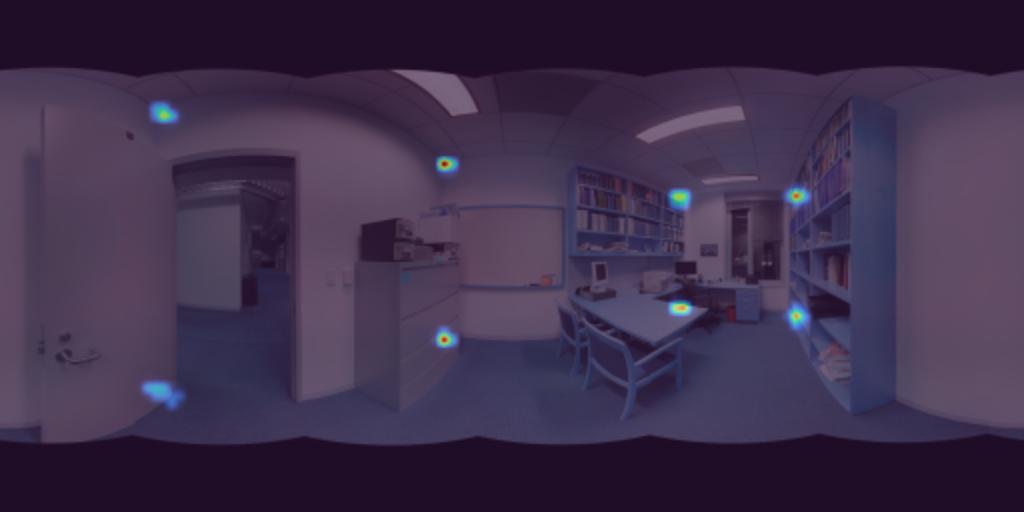}}
    \hfill
    \subfloat{\includegraphics[width=0.24\linewidth]{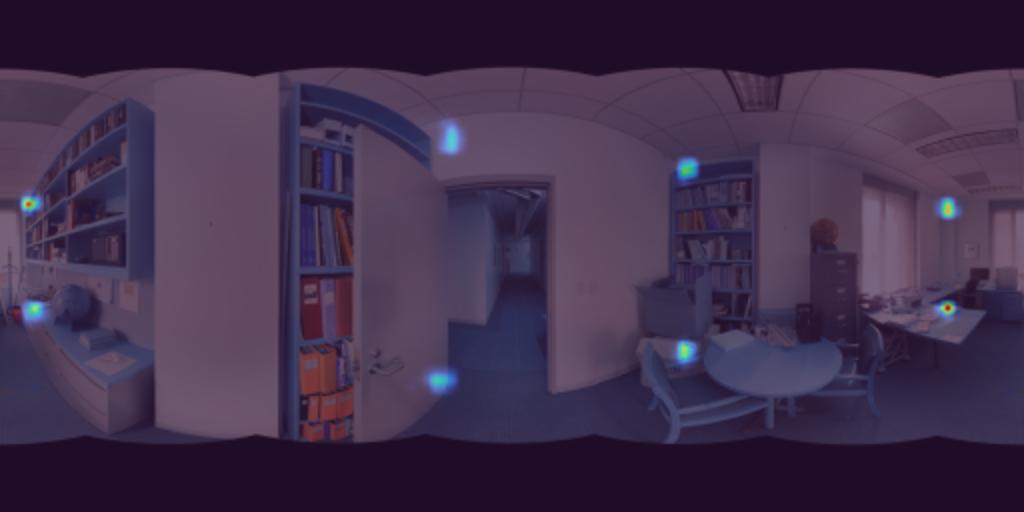}}
    \hfill
    \subfloat{\includegraphics[width=0.24\linewidth]{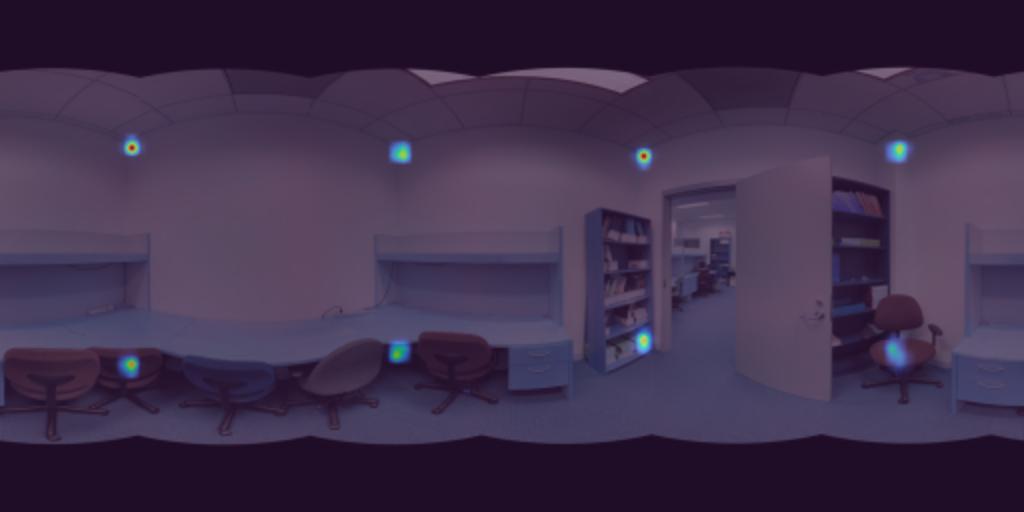}}
    \hfill
    
    \subfloat{\includegraphics[width=0.24\linewidth]{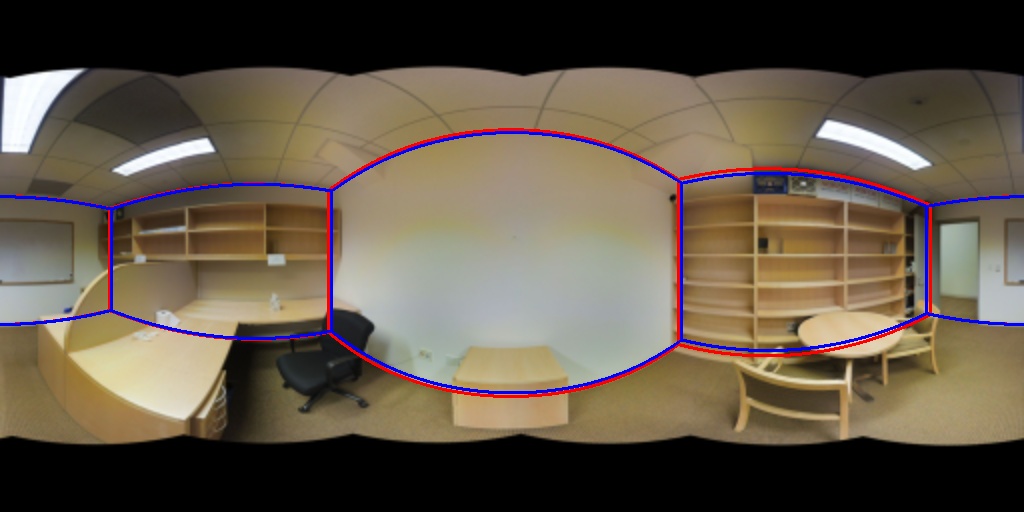}}
    \hfill
    \subfloat{\includegraphics[width=0.24\linewidth]{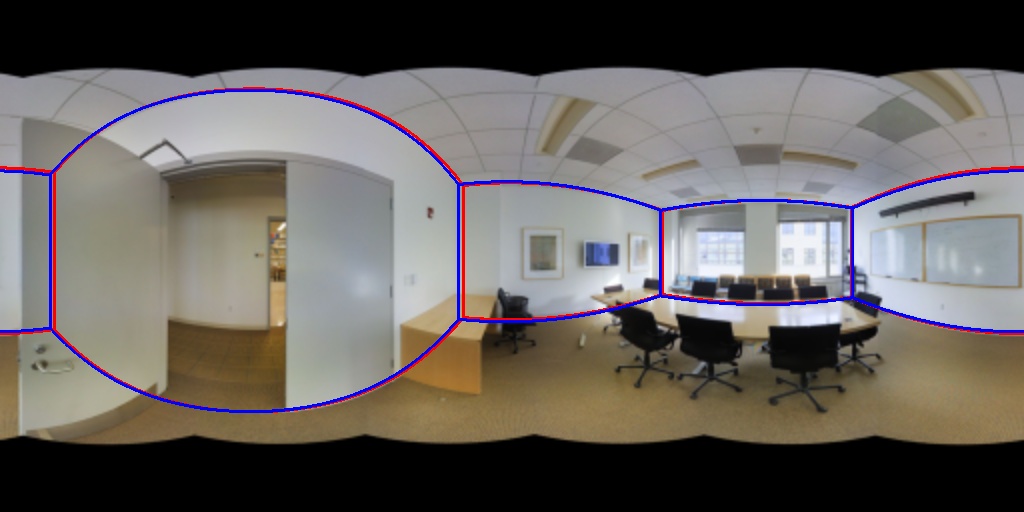}}
    \hfill
    \subfloat{\includegraphics[width=0.24\linewidth]{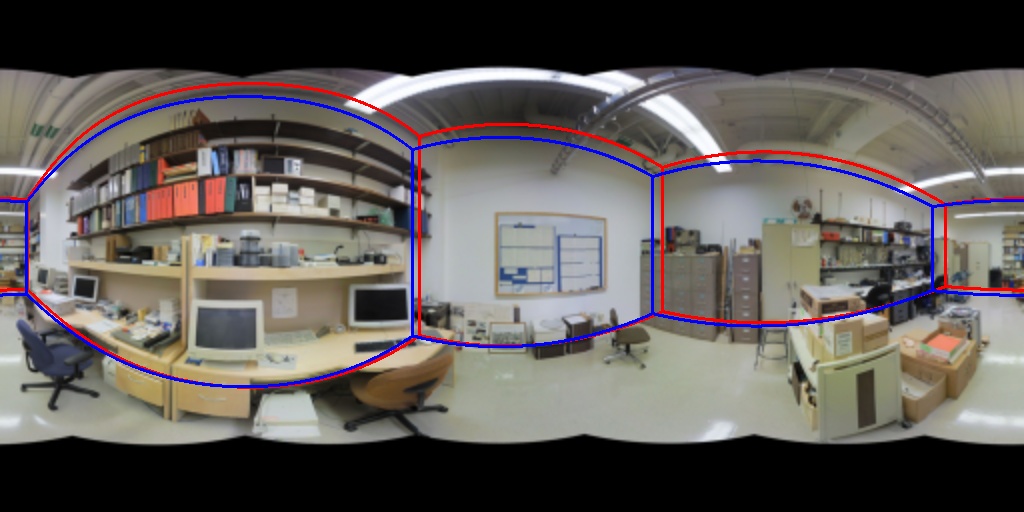}}
    \hfill
    \subfloat{\includegraphics[width=0.24\linewidth]{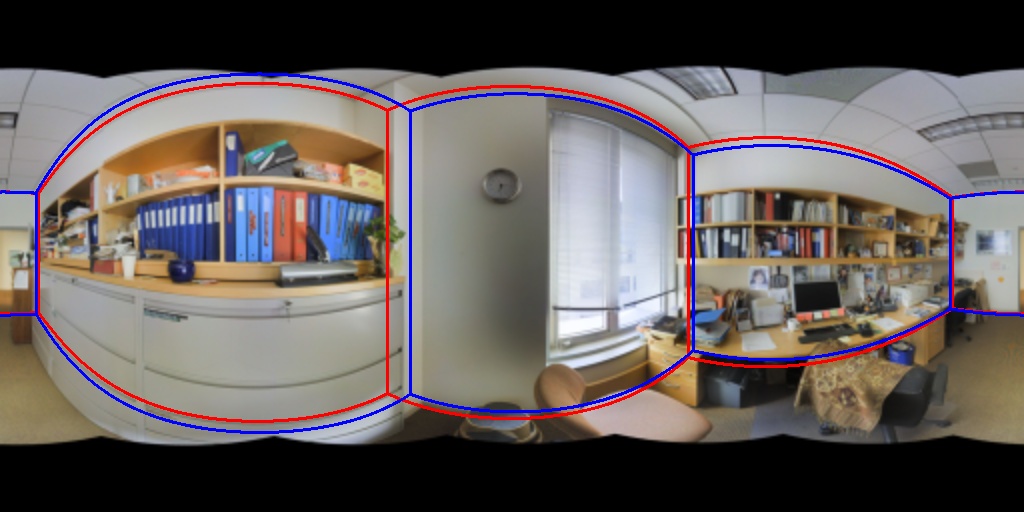}}
    \hfill
    
    \subfloat{\includegraphics[width=0.24\linewidth]{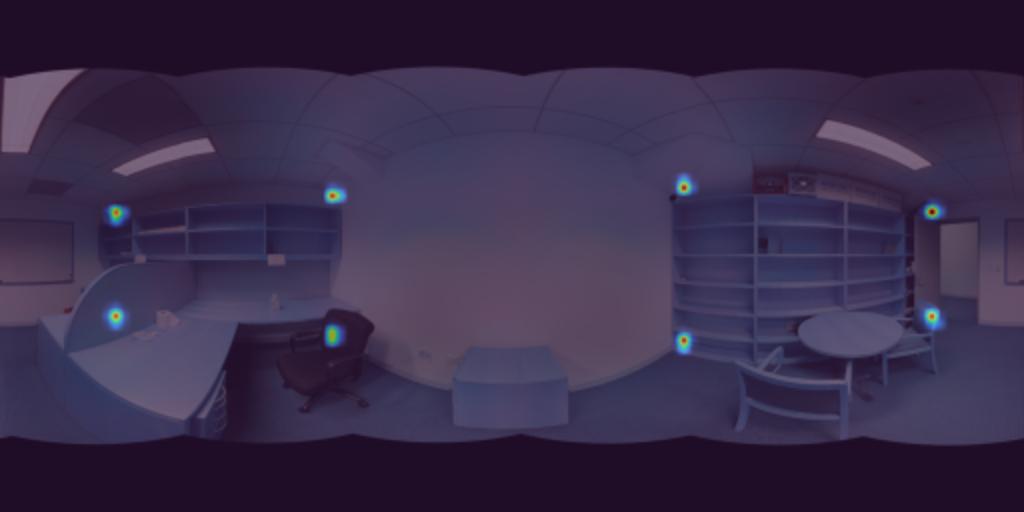}}
    \hfill
    \subfloat{\includegraphics[width=0.24\linewidth]{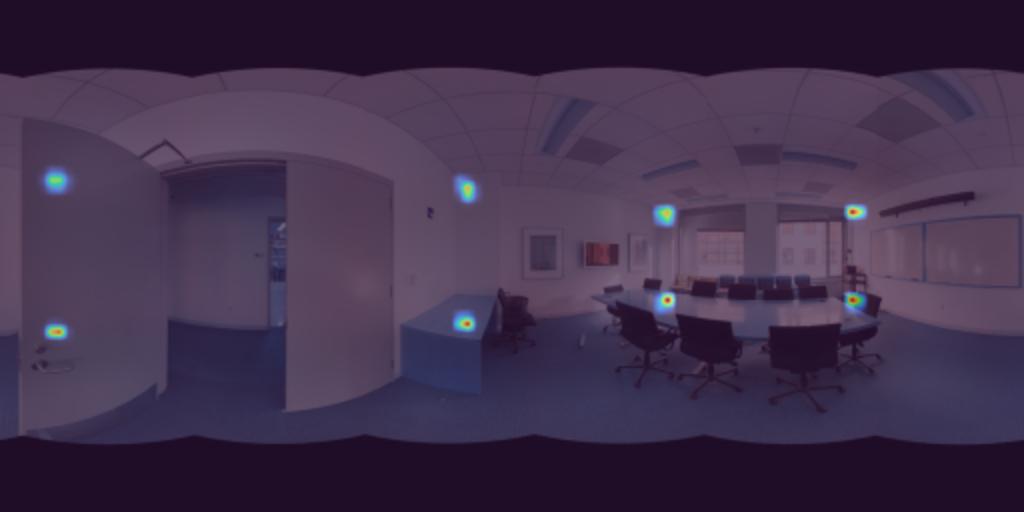}}
    \hfill
    \subfloat{\includegraphics[width=0.24\linewidth]{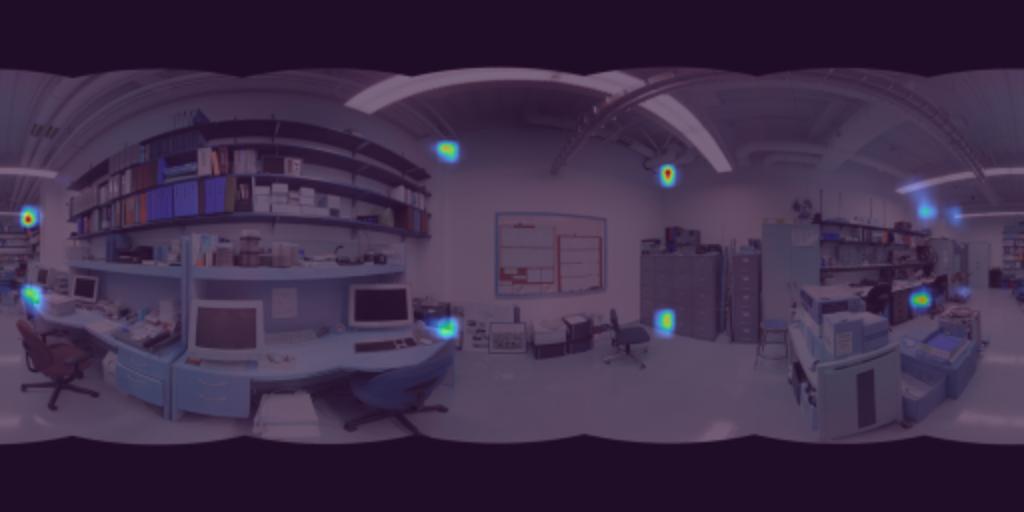}}
    \hfill
    \subfloat{\includegraphics[width=0.24\linewidth]{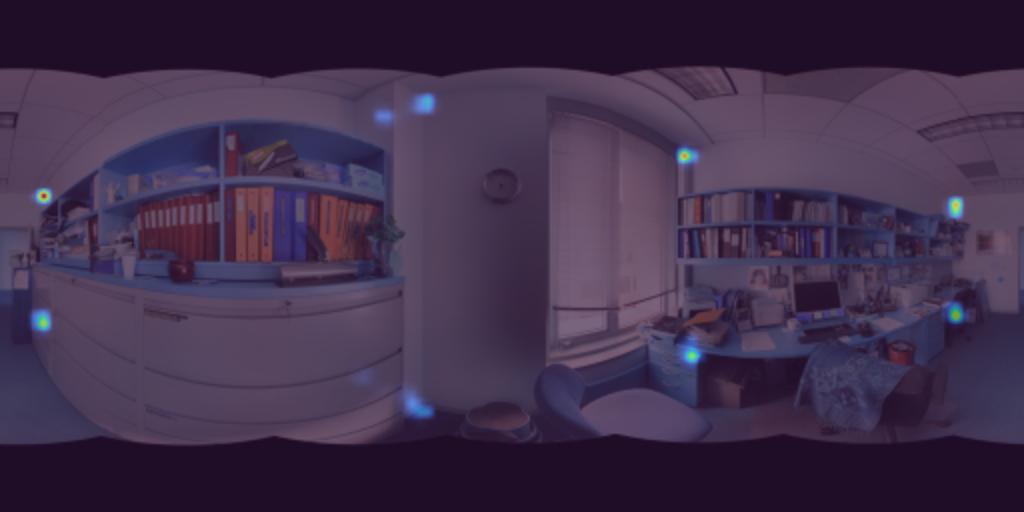}}
    \hfill
    
    \subfloat{\includegraphics[width=0.24\linewidth]{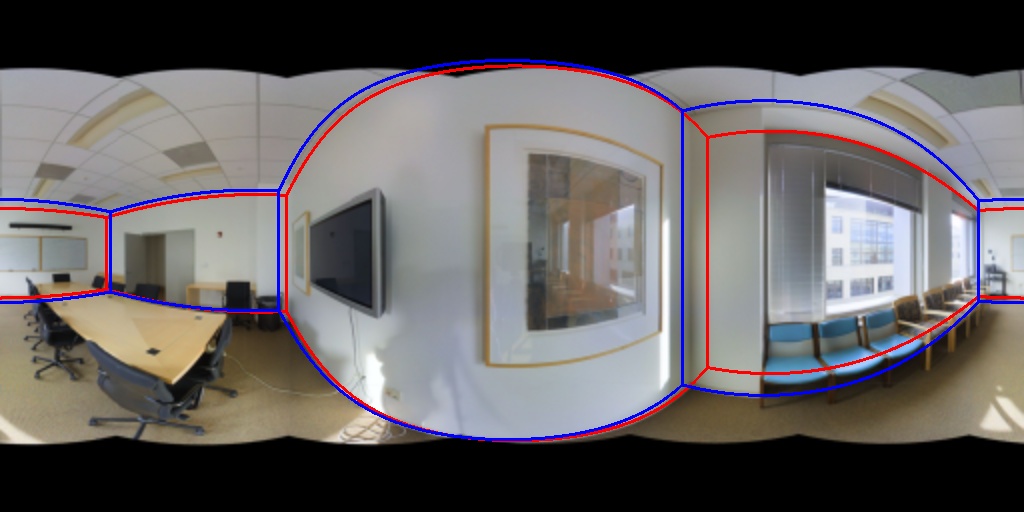}}
    \hfill
    \subfloat{\includegraphics[width=0.24\linewidth]{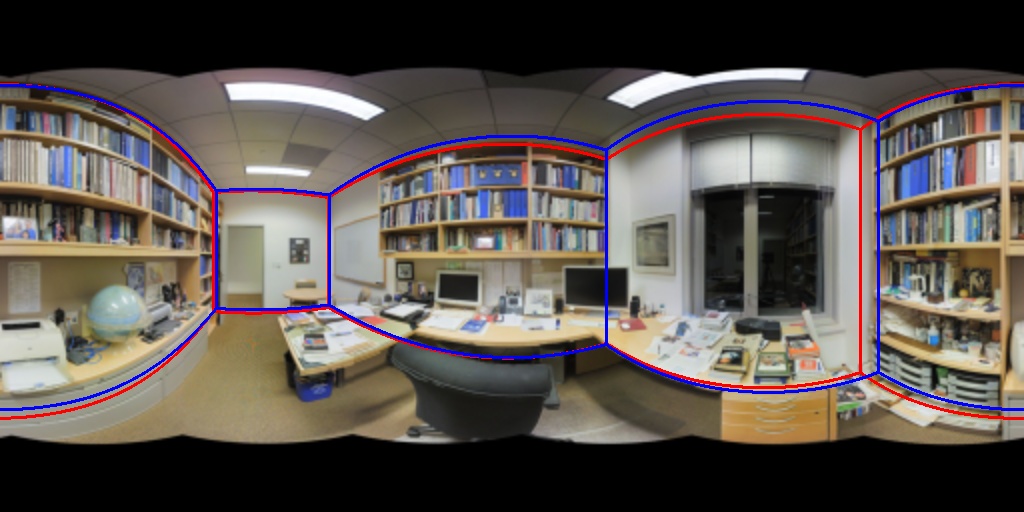}}
    \hfill
    \subfloat{\includegraphics[width=0.24\linewidth]{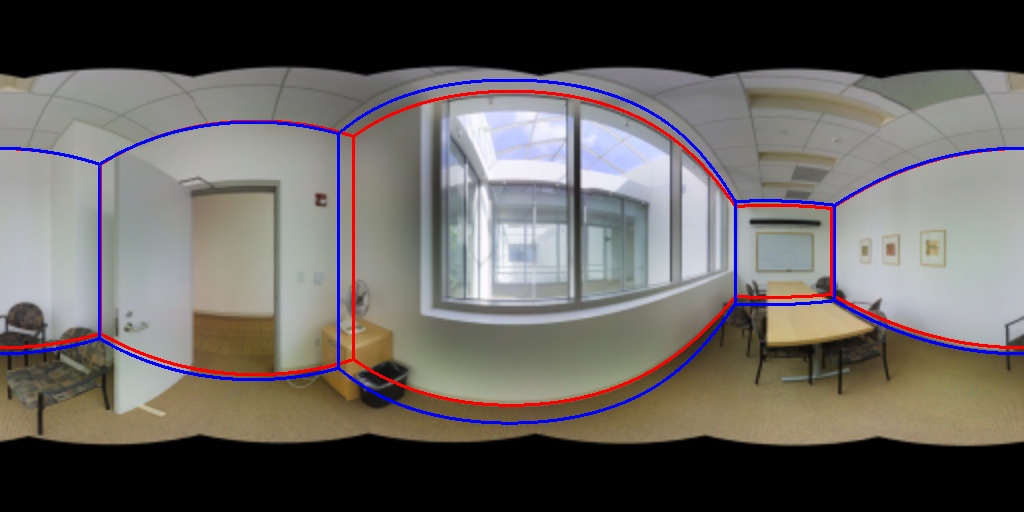}}
    \hfill
    \subfloat{\includegraphics[width=0.24\linewidth]{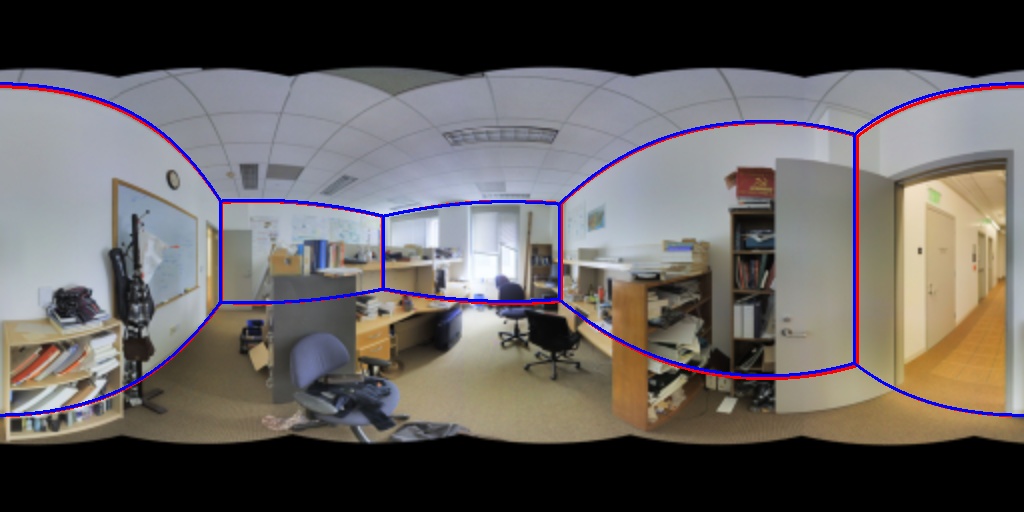}}
    \hfill
    
    \subfloat{\includegraphics[width=0.24\linewidth]{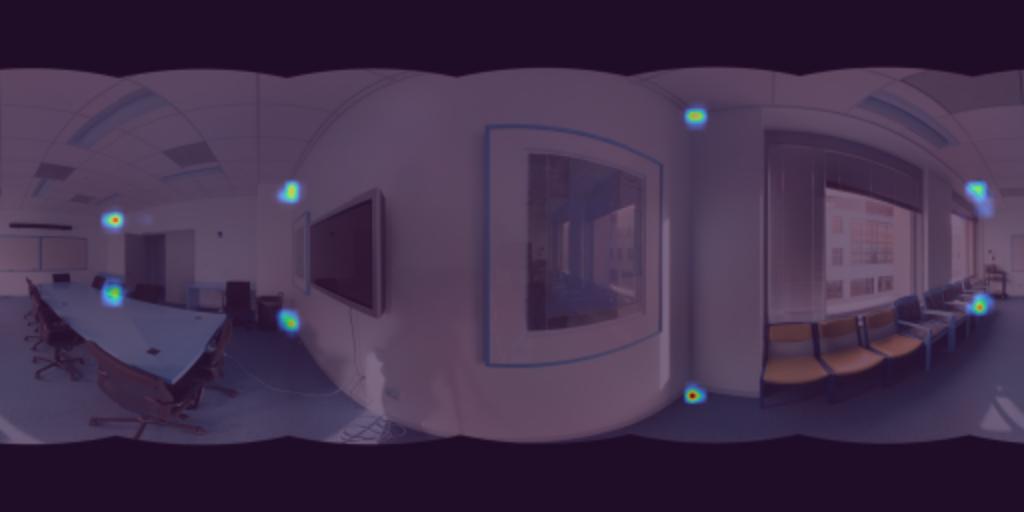}}
    \hfill
    \subfloat{\includegraphics[width=0.24\linewidth]{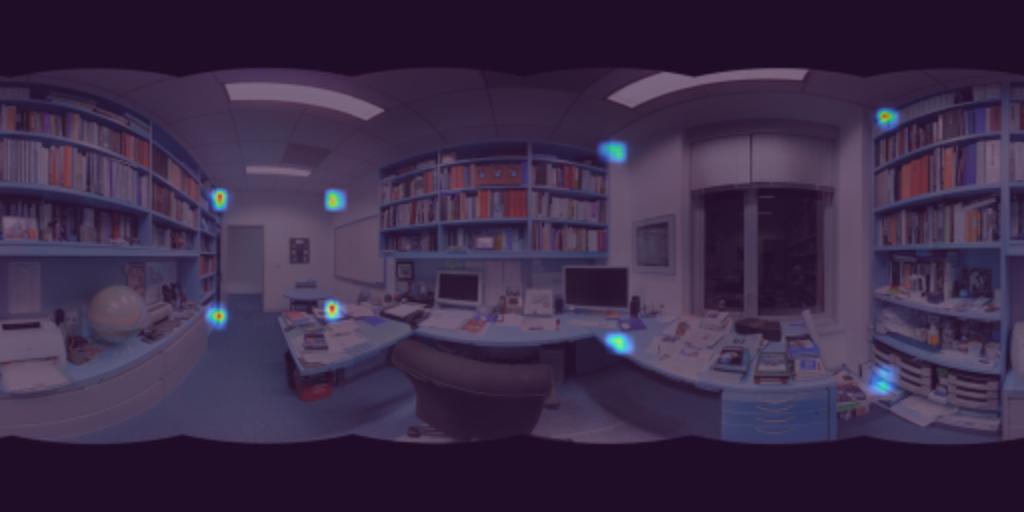}}
    \hfill
    \subfloat{\includegraphics[width=0.24\linewidth]{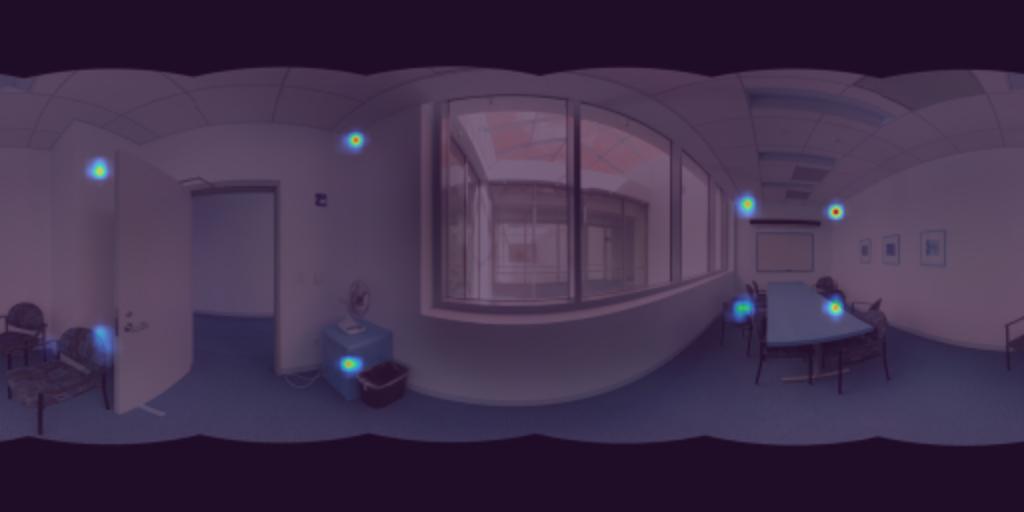}}
    \hfill
    \subfloat{\includegraphics[width=0.24\linewidth]{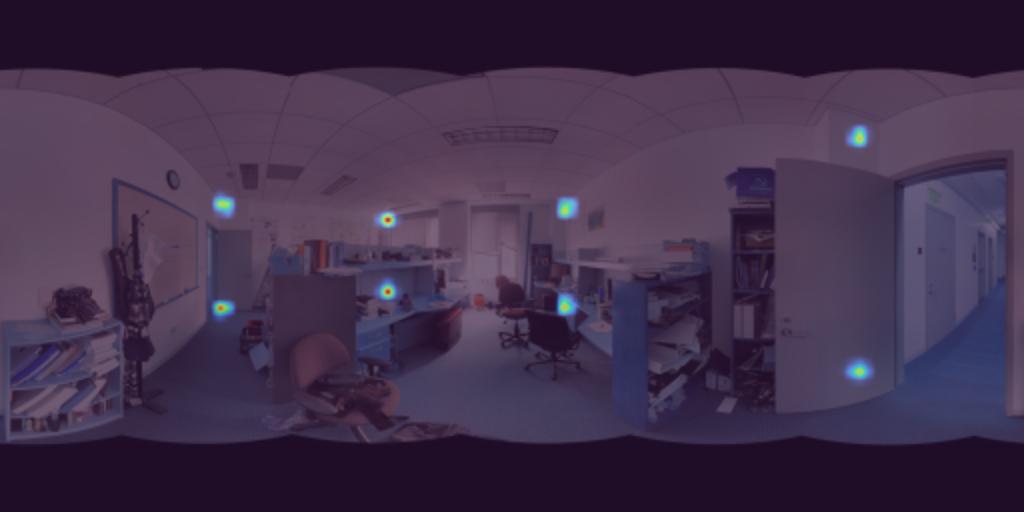}}
    \hfill
    
    \caption{
        Additional qualitative results on the Stanford2D3D dataset.
    }
\label{fig:s2d3d}
\end{figure*}

\begin{figure*}[!htbp]
    \centering

    \subfloat{\includegraphics[width=0.24\linewidth]{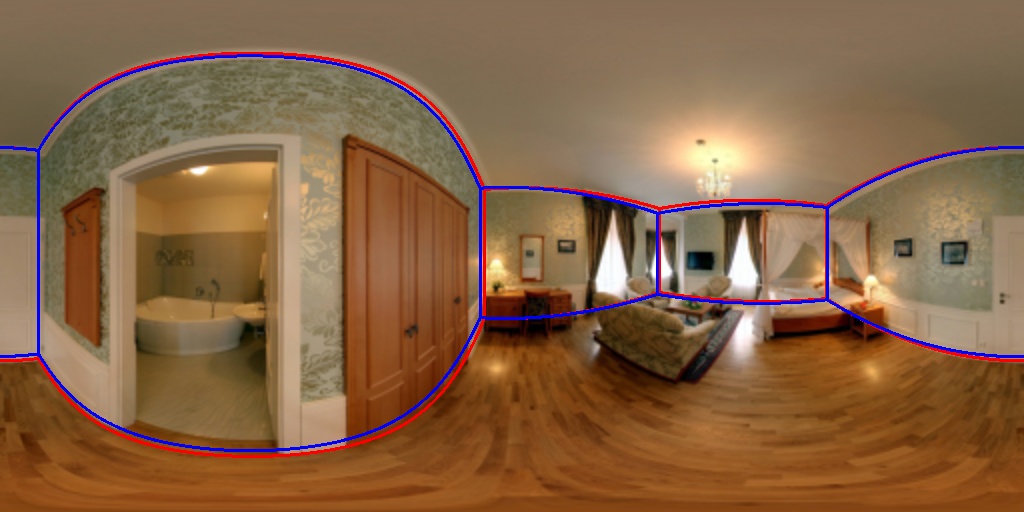}}
    \hfill
    \subfloat{\includegraphics[width=0.24\linewidth]{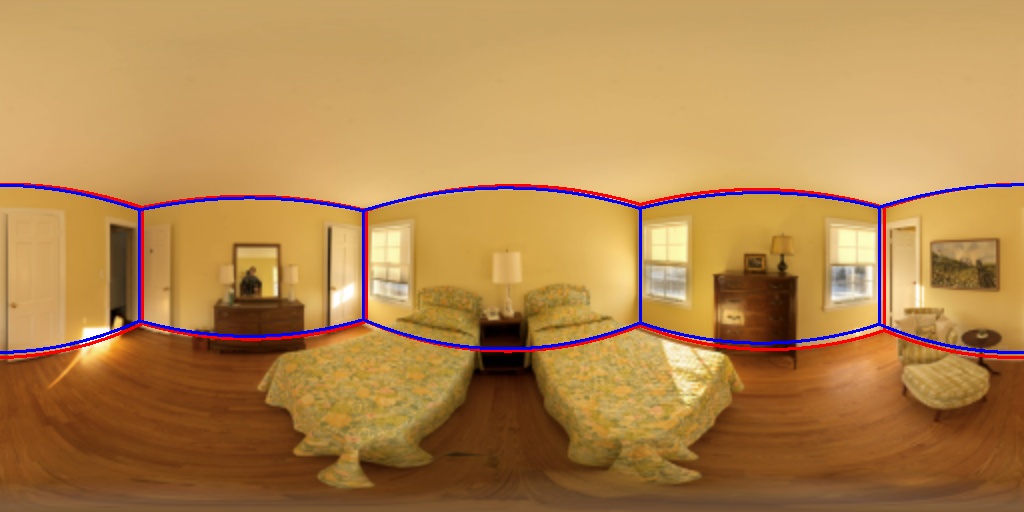}}
    \hfill
    \subfloat{\includegraphics[width=0.24\linewidth]{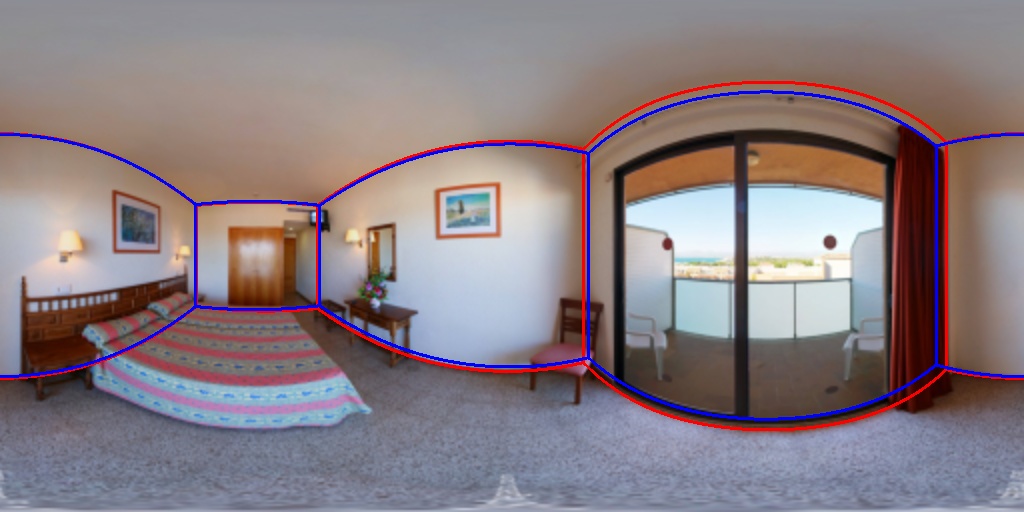}}
    \hfill
    \subfloat{\includegraphics[width=0.24\linewidth]{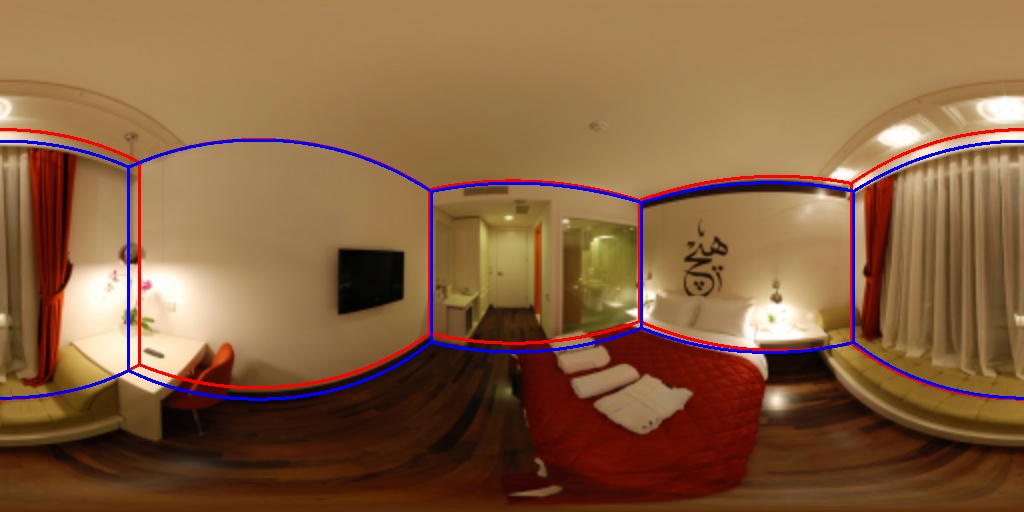}}
    \hfill

    \subfloat{\includegraphics[width=0.24\linewidth]{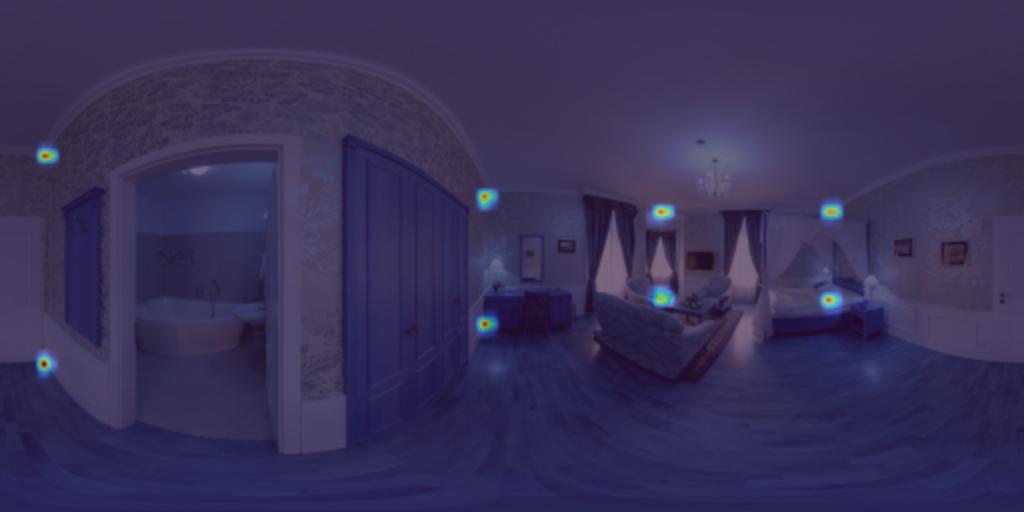}}
    \hfill
    \subfloat{\includegraphics[width=0.24\linewidth]{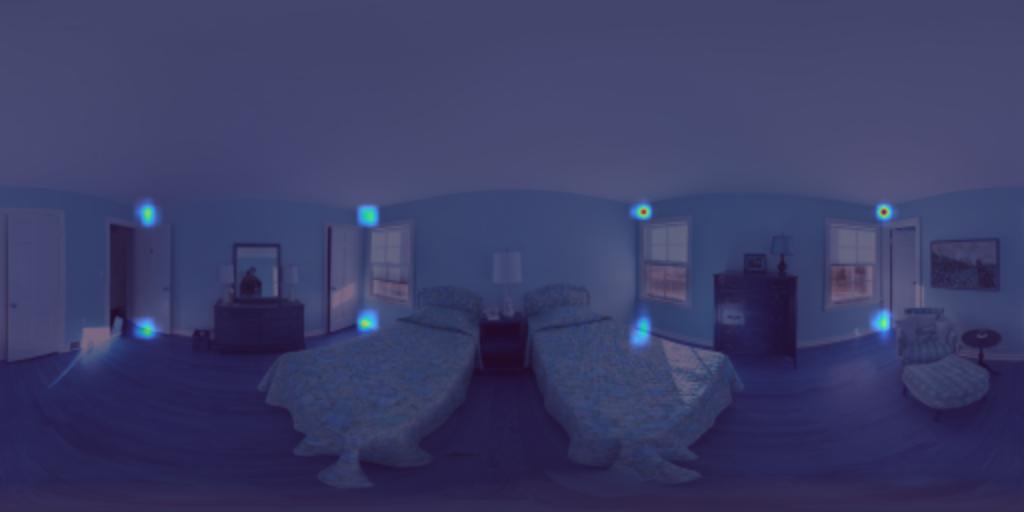}}
    \hfill
    \subfloat{\includegraphics[width=0.24\linewidth]{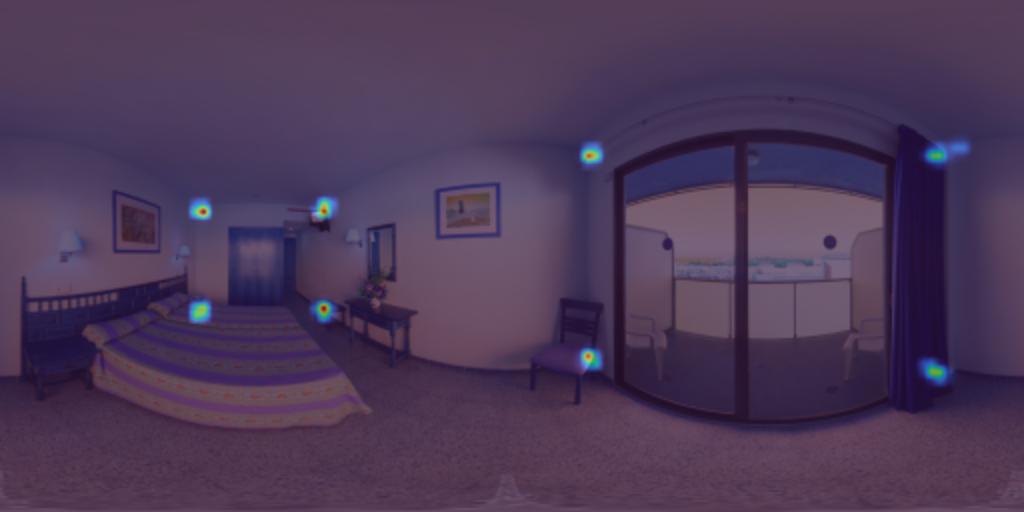}}
    \hfill
    \subfloat{\includegraphics[width=0.24\linewidth]{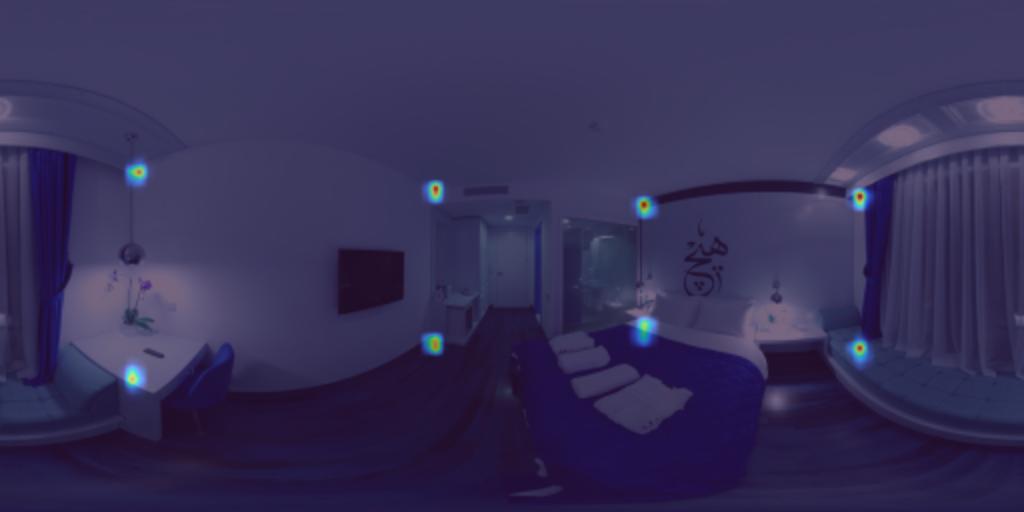}}
    \hfill
    
    \subfloat{\includegraphics[width=0.24\linewidth]{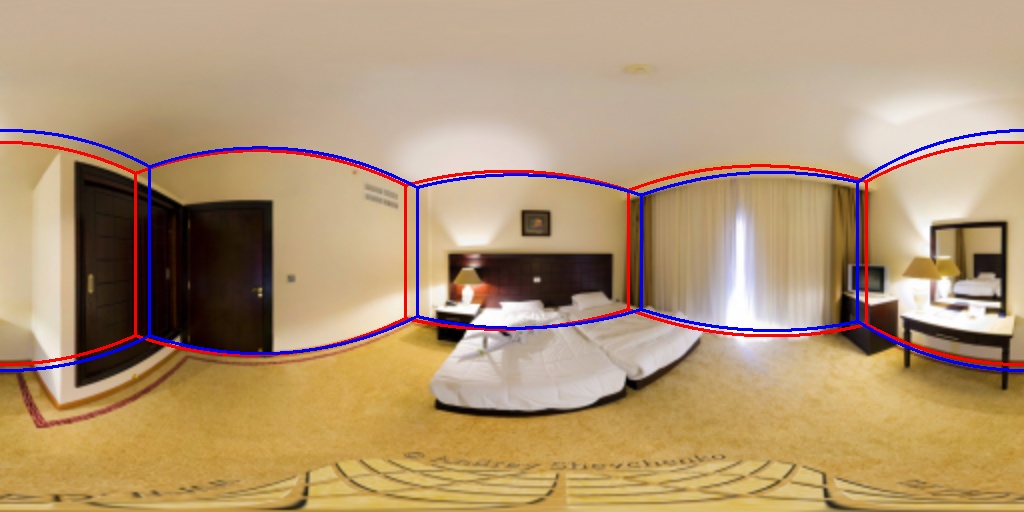}}
    \hfill
    \subfloat{\includegraphics[width=0.24\linewidth]{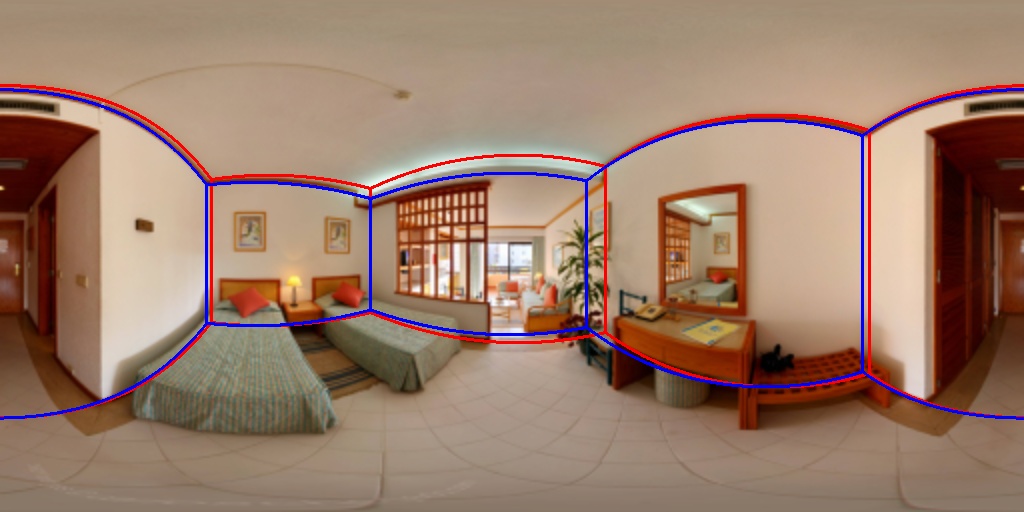}}
    \hfill
    \subfloat{\includegraphics[width=0.24\linewidth]{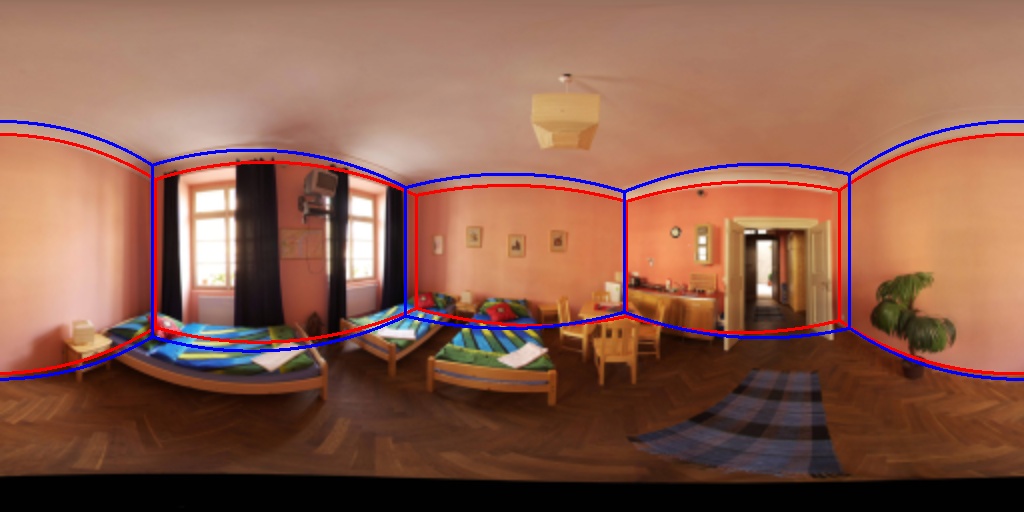}}
    \hfill
    \subfloat{\includegraphics[width=0.24\linewidth]{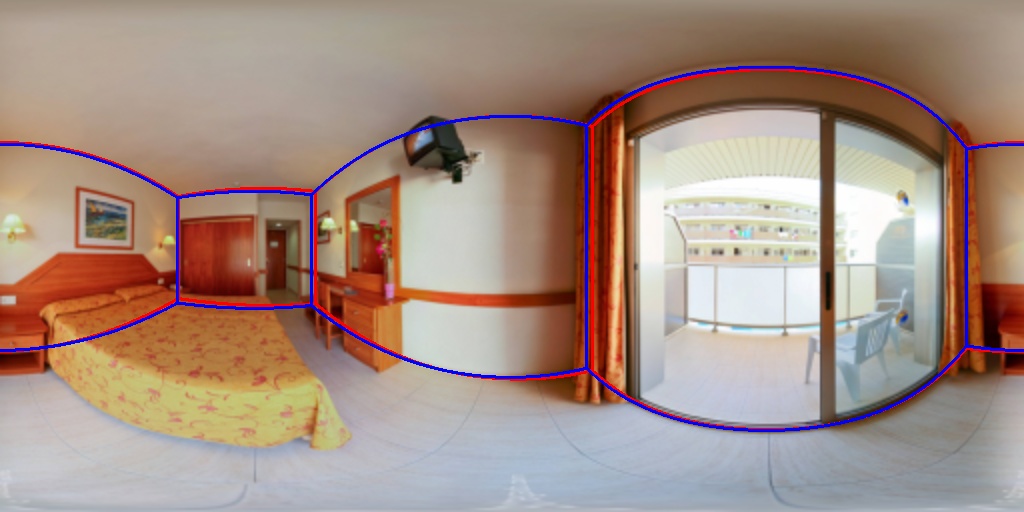}}
    \hfill
    
    \subfloat{\includegraphics[width=0.24\linewidth]{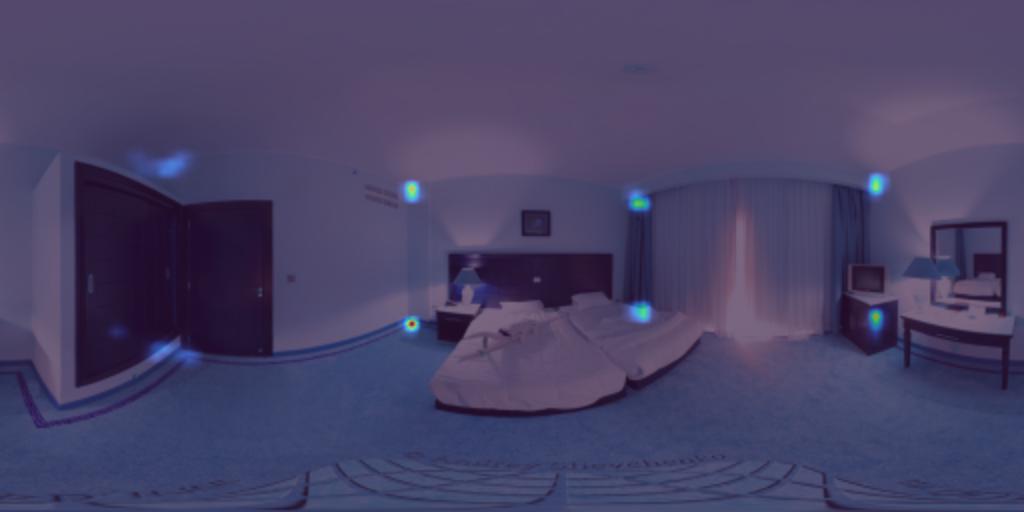}}
    \hfill
    \subfloat{\includegraphics[width=0.24\linewidth]{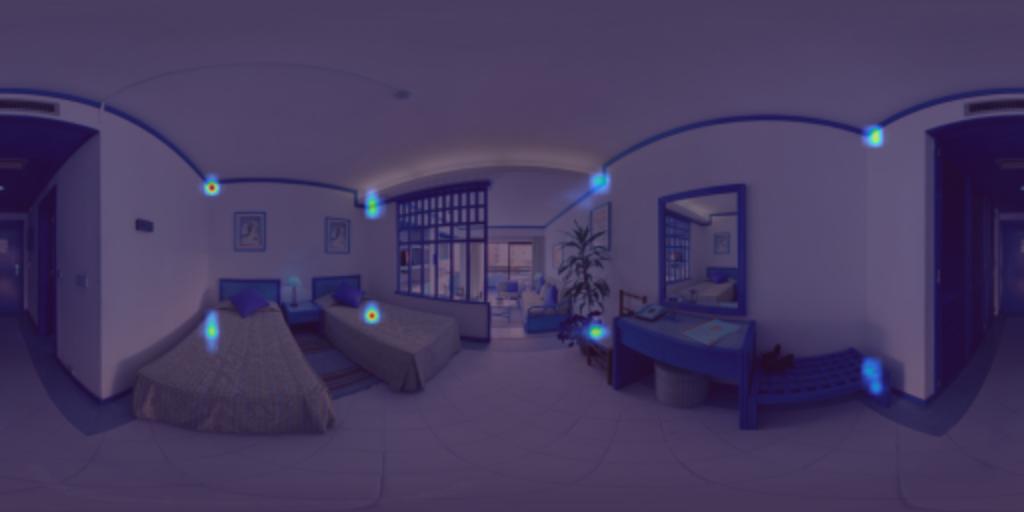}}
    \hfill
    \subfloat{\includegraphics[width=0.24\linewidth]{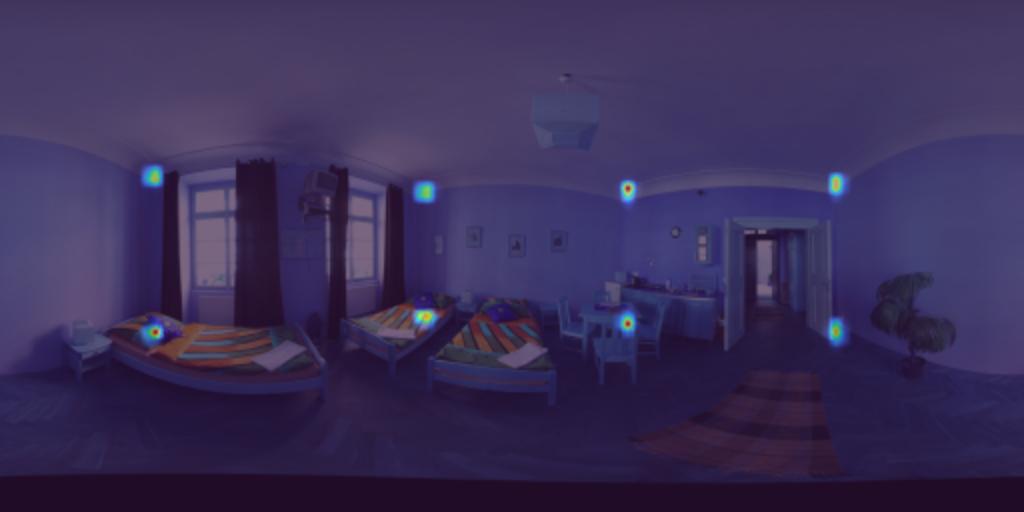}}
    \hfill
    \subfloat{\includegraphics[width=0.24\linewidth]{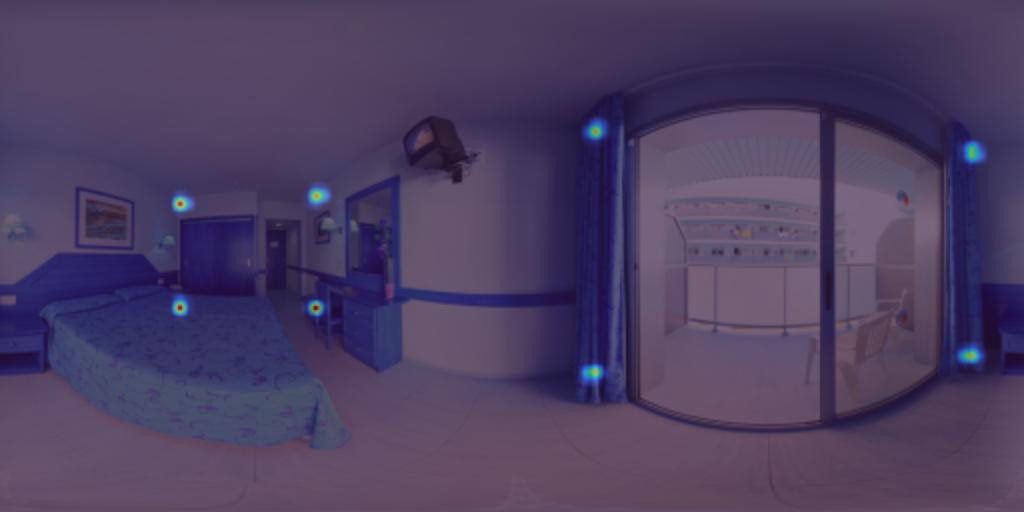}}
    \hfill
    
    \subfloat{\includegraphics[width=0.24\linewidth]{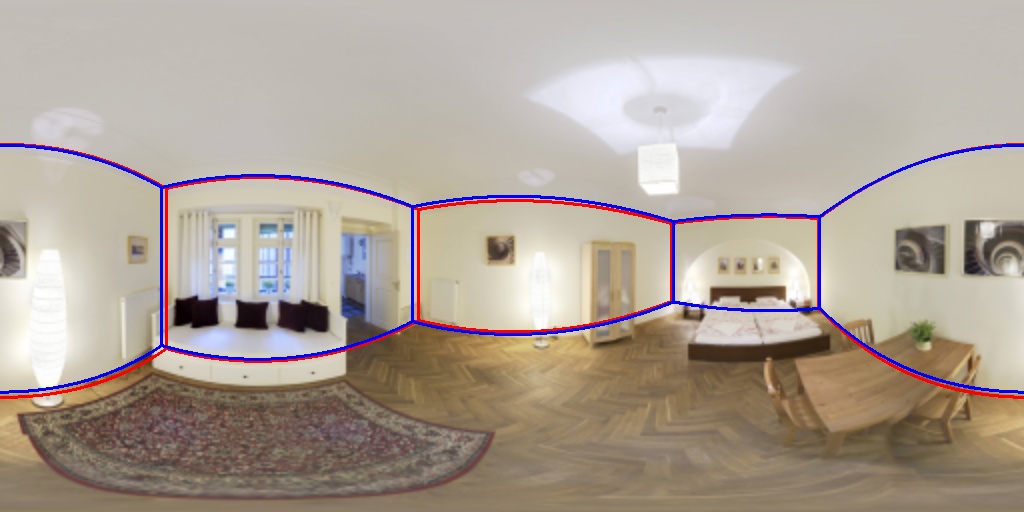}}
    \hfill
    \subfloat{\includegraphics[width=0.24\linewidth]{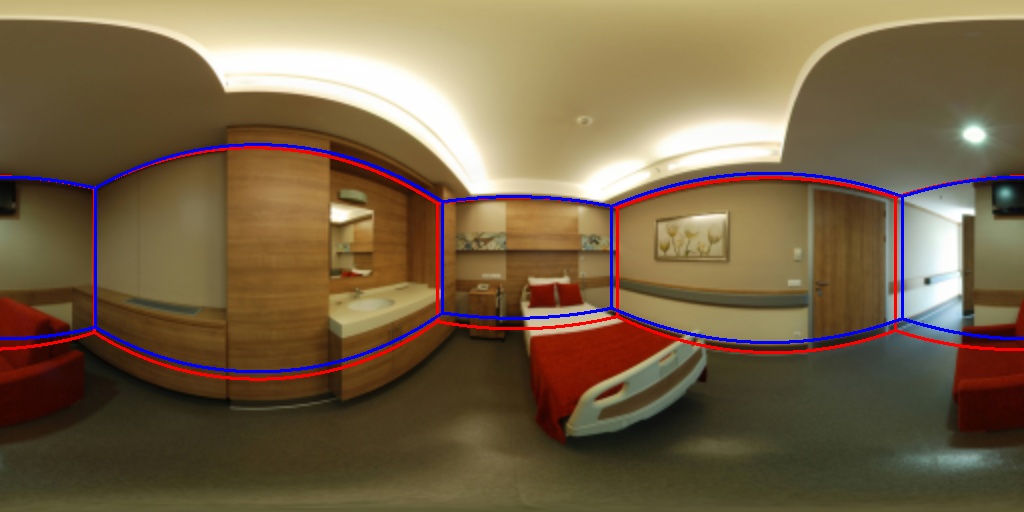}}
    \hfill
    \subfloat{\includegraphics[width=0.24\linewidth]{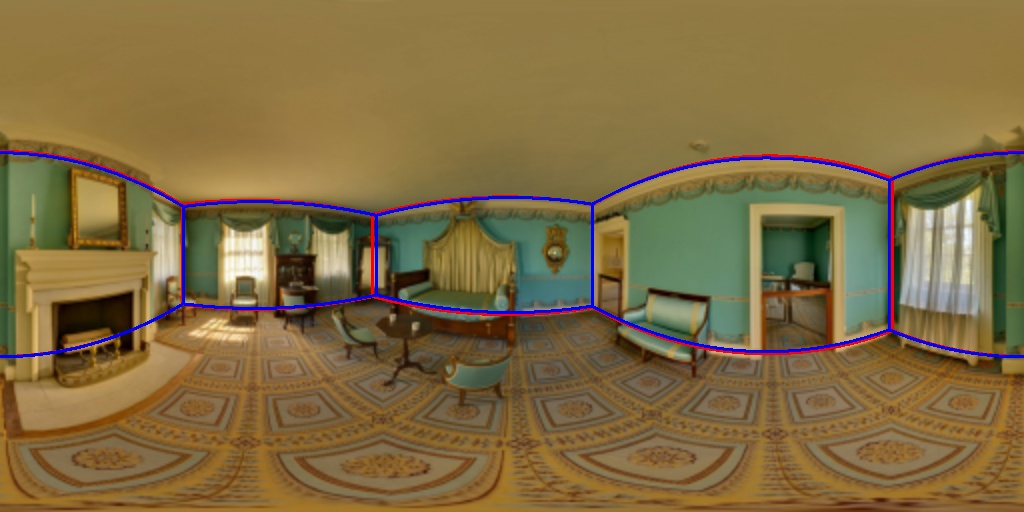}}
    \hfill
    \subfloat{\includegraphics[width=0.24\linewidth]{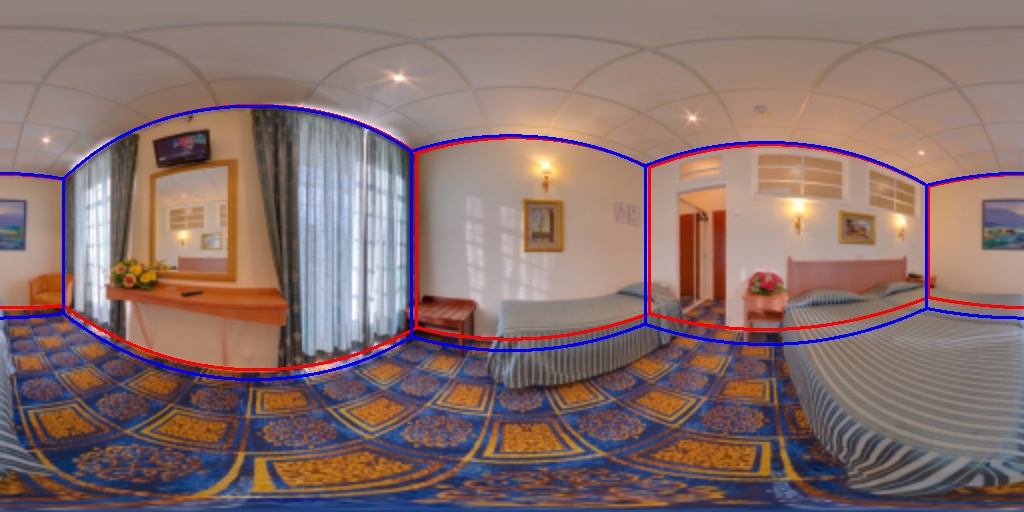}}
    \hfill
    
    \subfloat{\includegraphics[width=0.24\linewidth]{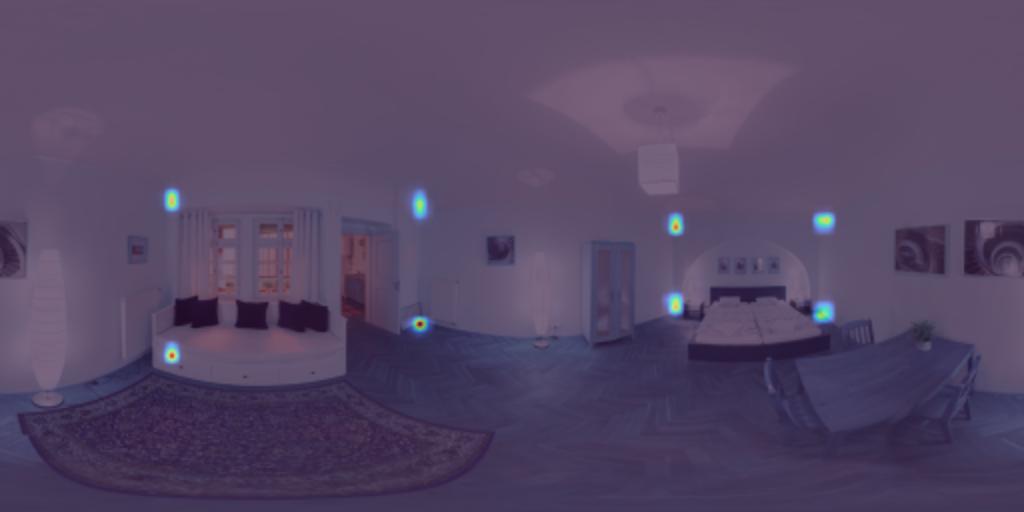}}
    \hfill
    \subfloat{\includegraphics[width=0.24\linewidth]{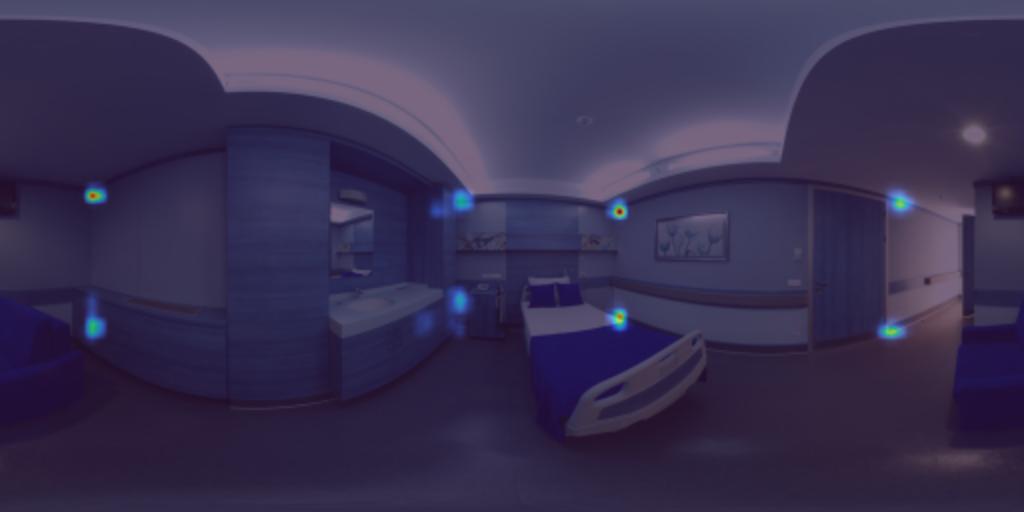}}
    \hfill
    \subfloat{\includegraphics[width=0.24\linewidth]{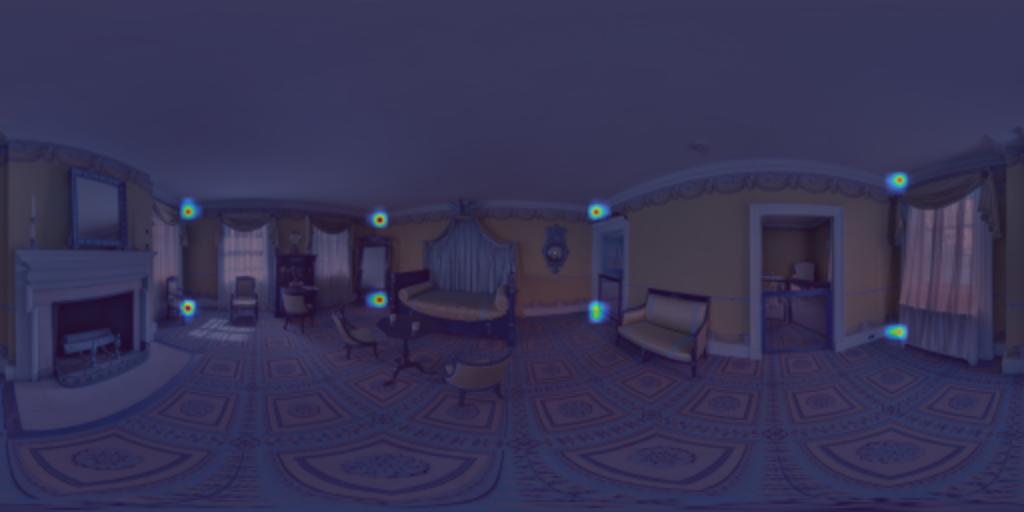}}
    \hfill
    \subfloat{\includegraphics[width=0.24\linewidth]{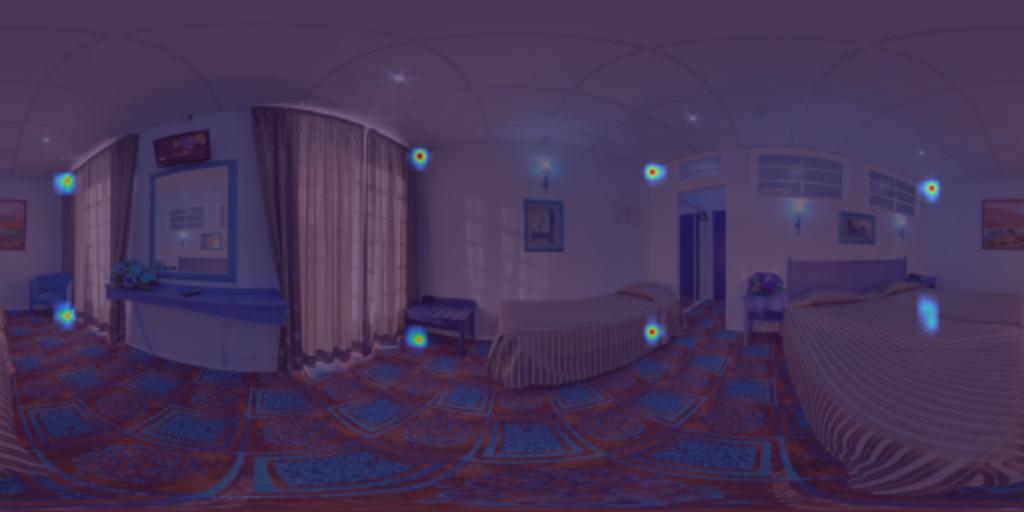}}
    \hfill
    
    \subfloat{\includegraphics[width=0.24\linewidth]{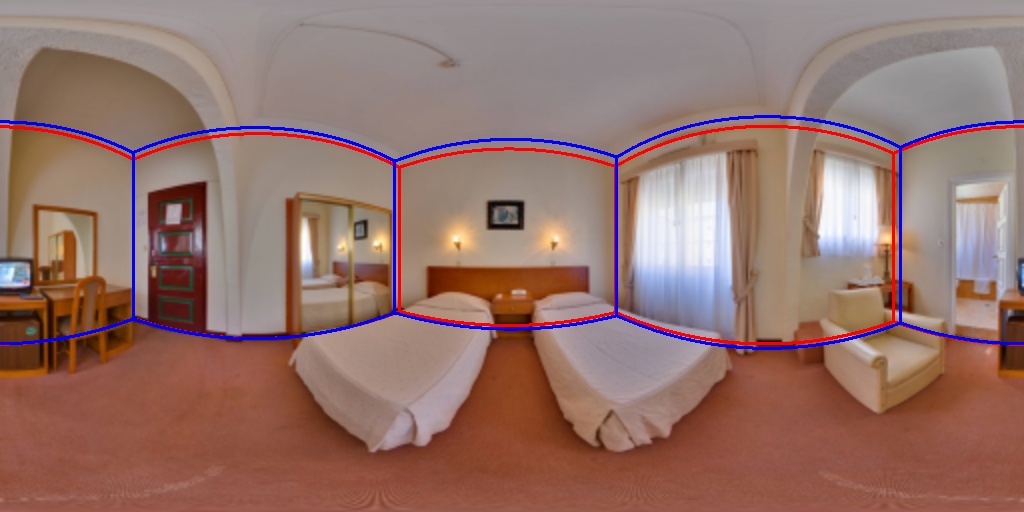}}
    \hfill
    \subfloat{\includegraphics[width=0.24\linewidth]{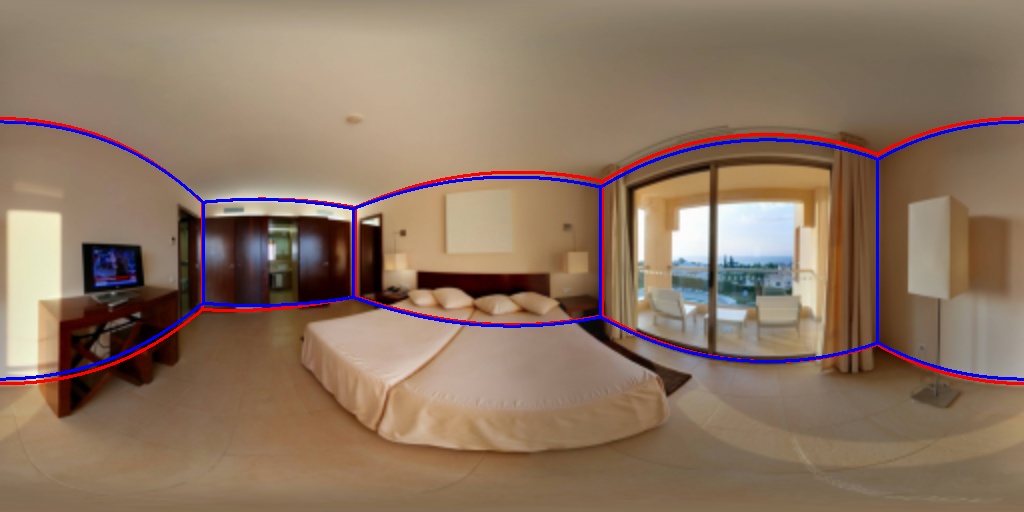}}
    \hfill
    \subfloat{\includegraphics[width=0.24\linewidth]{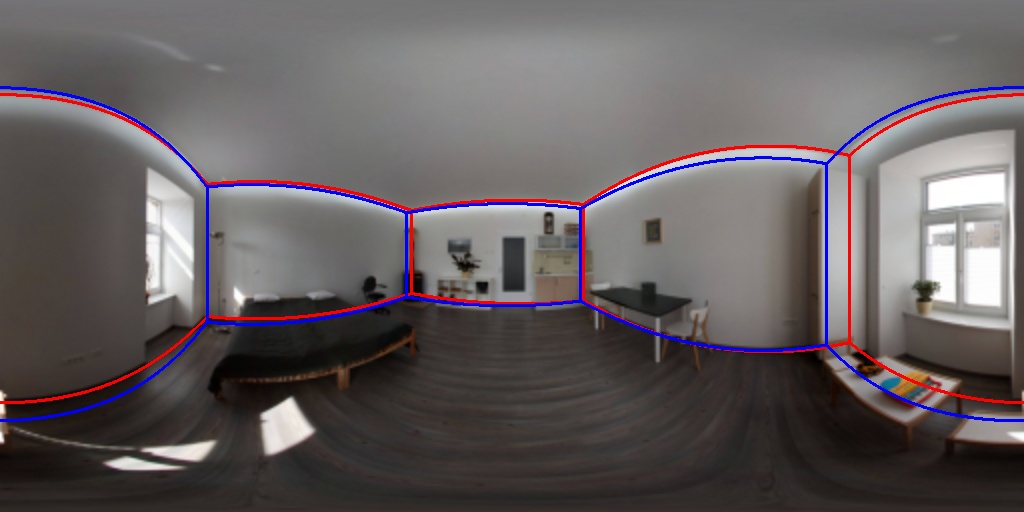}}
    \hfill
    \subfloat{\includegraphics[width=0.24\linewidth]{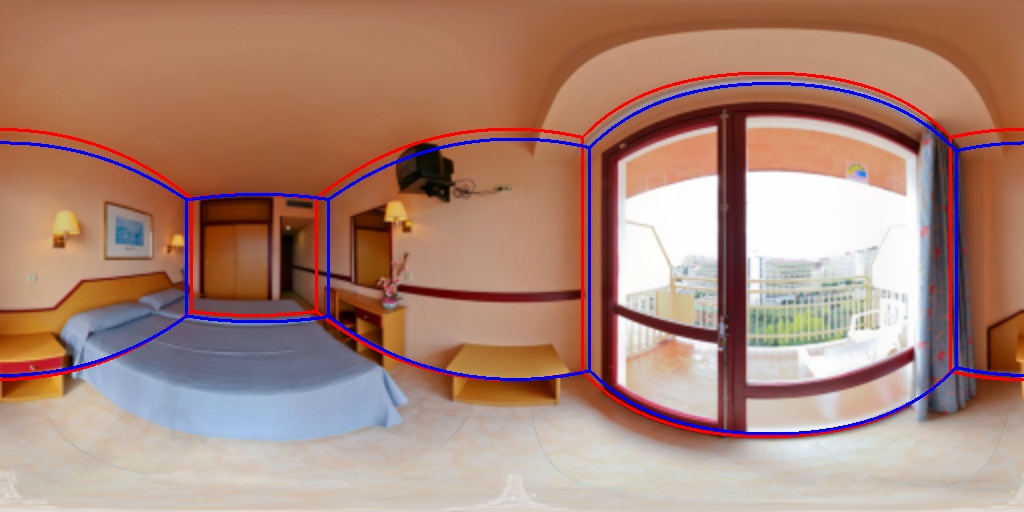}}
    \hfill
    
    \subfloat{\includegraphics[width=0.24\linewidth]{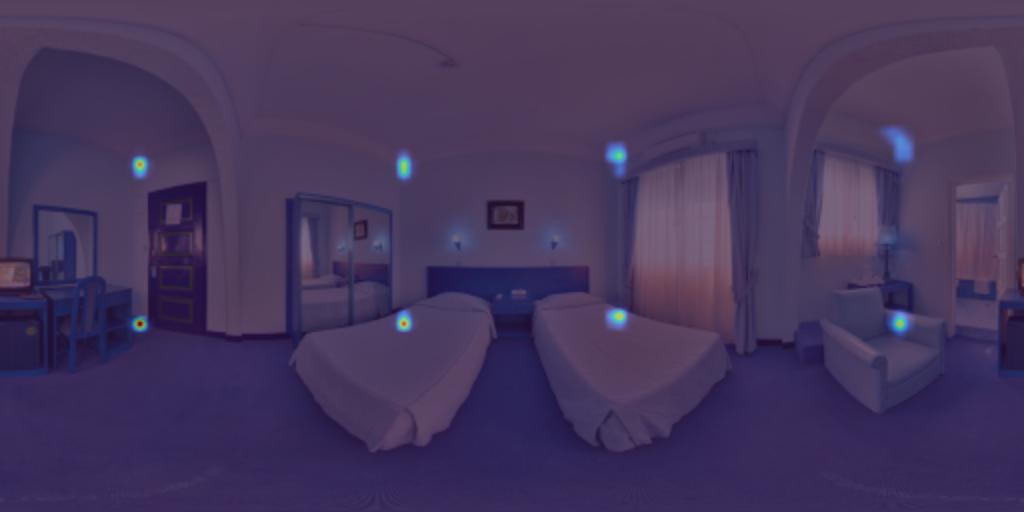}}
    \hfill
    \subfloat{\includegraphics[width=0.24\linewidth]{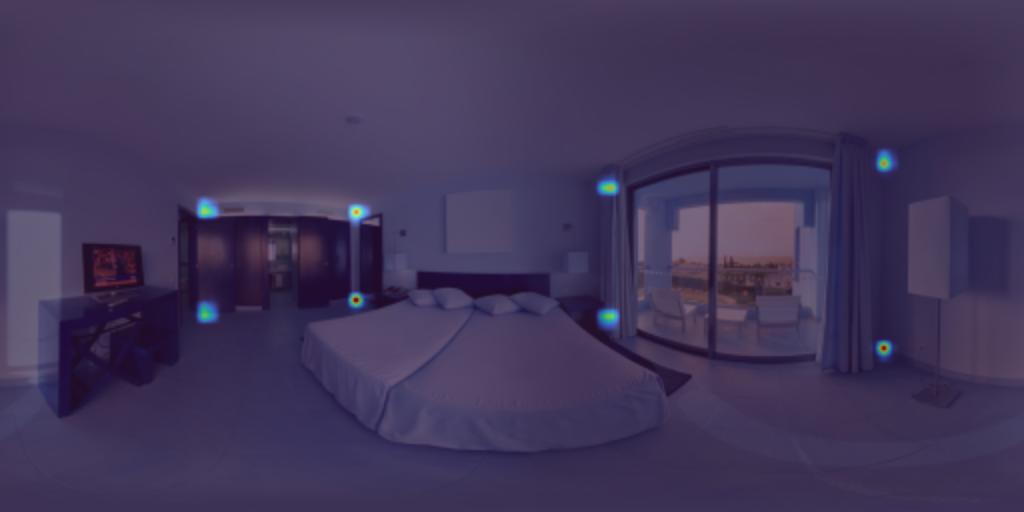}}
    \hfill
    \subfloat{\includegraphics[width=0.24\linewidth]{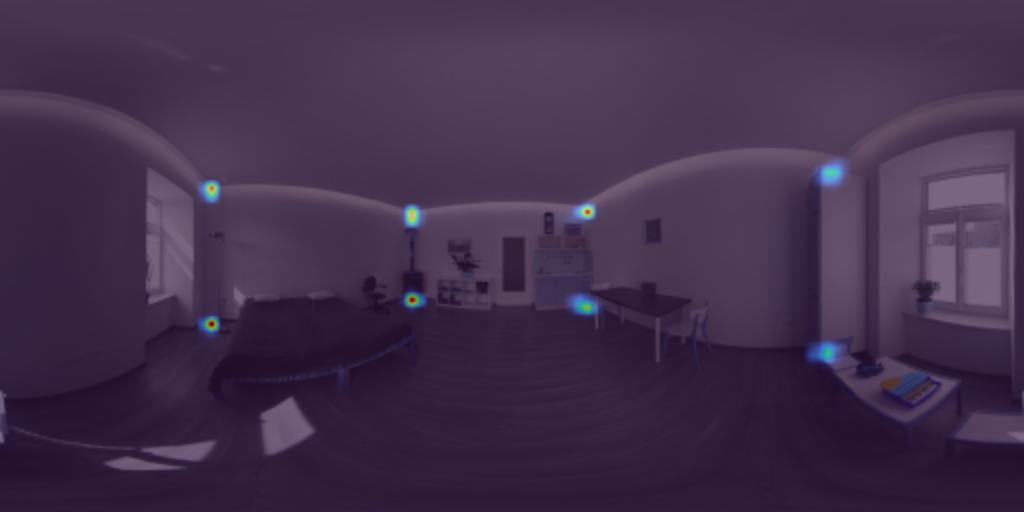}}
    \hfill
    \subfloat{\includegraphics[width=0.24\linewidth]{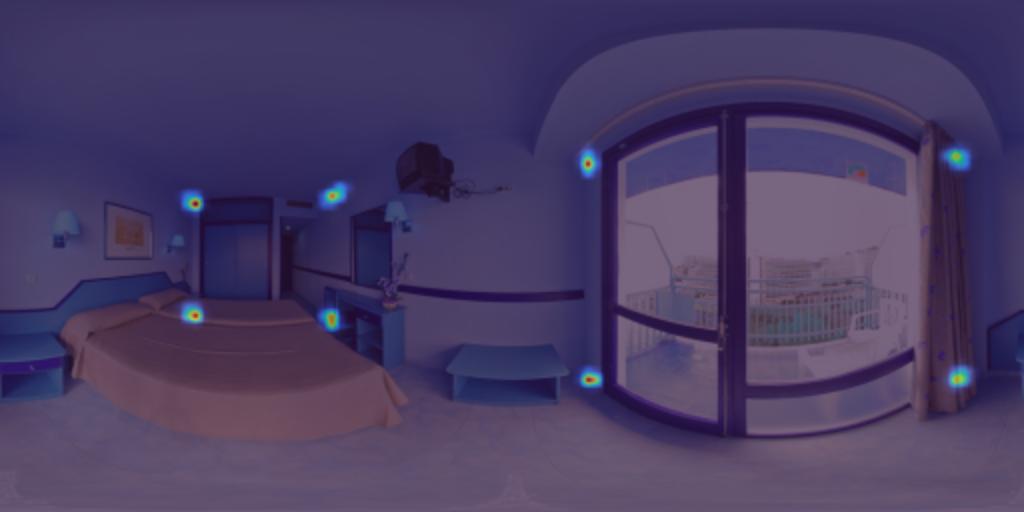}}
    \hfill
    
    \caption{
        Additional qualitative results on the PanoContext dataset.
    }
\label{fig:pc}
\end{figure*}

\begin{figure*}[!htbp]
    \centering

    \subfloat{\includegraphics[width=0.24\linewidth]{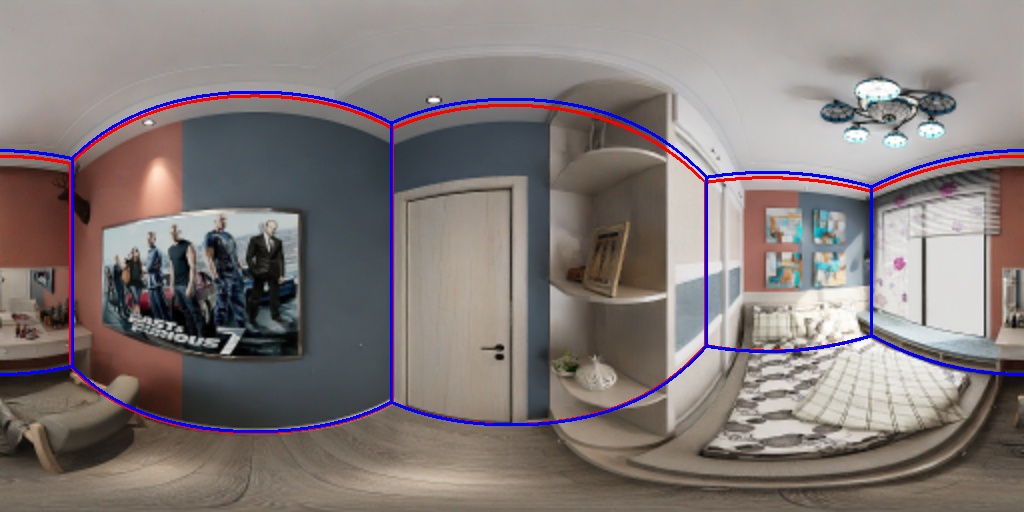}}
    \hfill
    \subfloat{\includegraphics[width=0.24\linewidth]{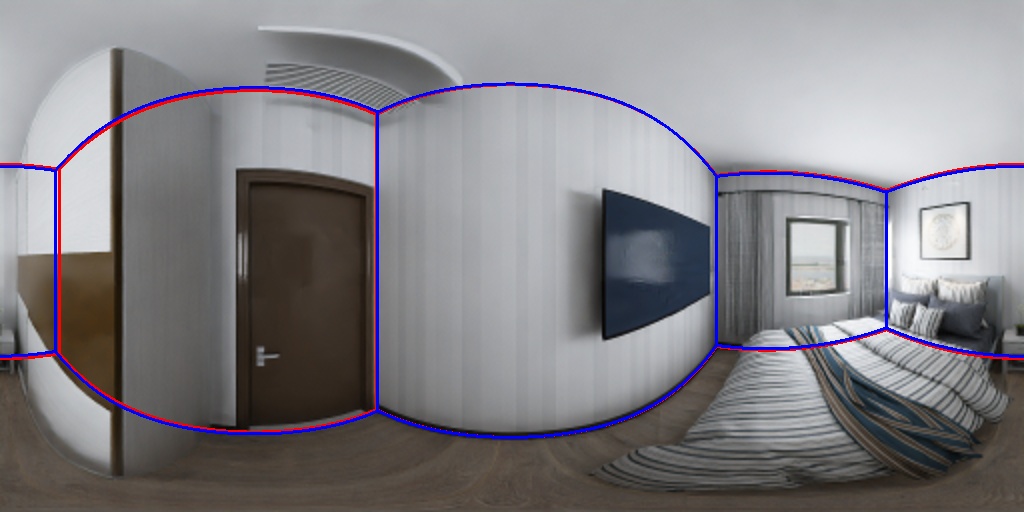}}
    \hfill
    \subfloat{\includegraphics[width=0.24\linewidth]{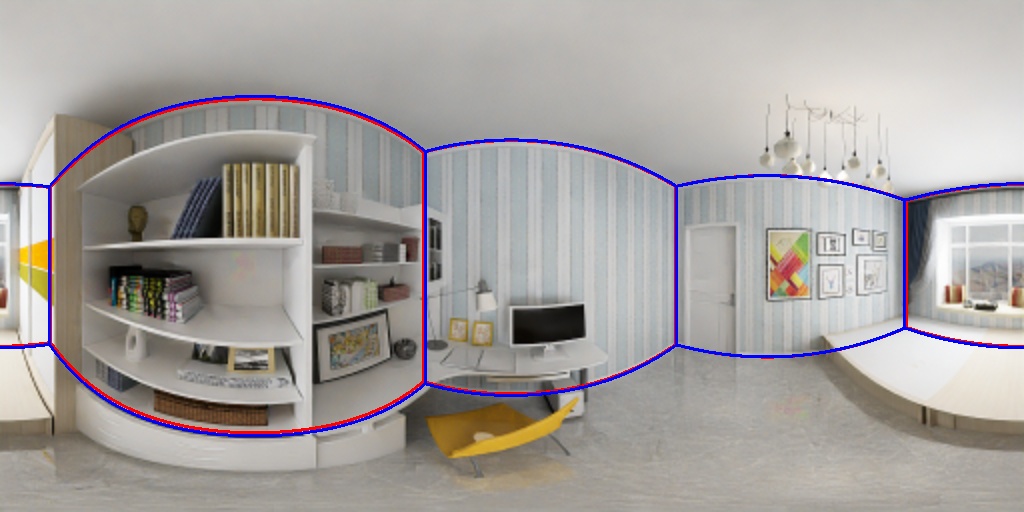}}
    \hfill
    \subfloat{\includegraphics[width=0.24\linewidth]{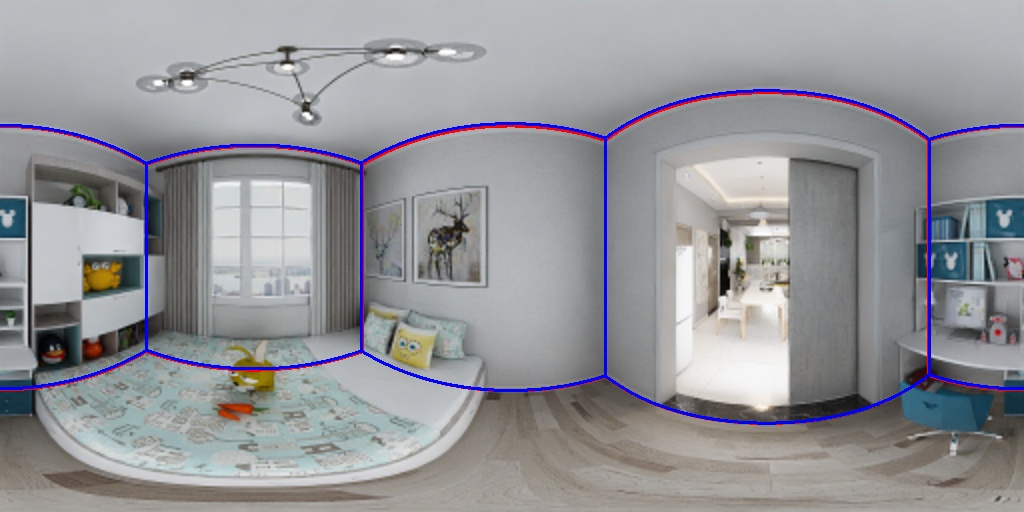}}
    \hfill
    
    \subfloat{\includegraphics[width=0.24\linewidth]{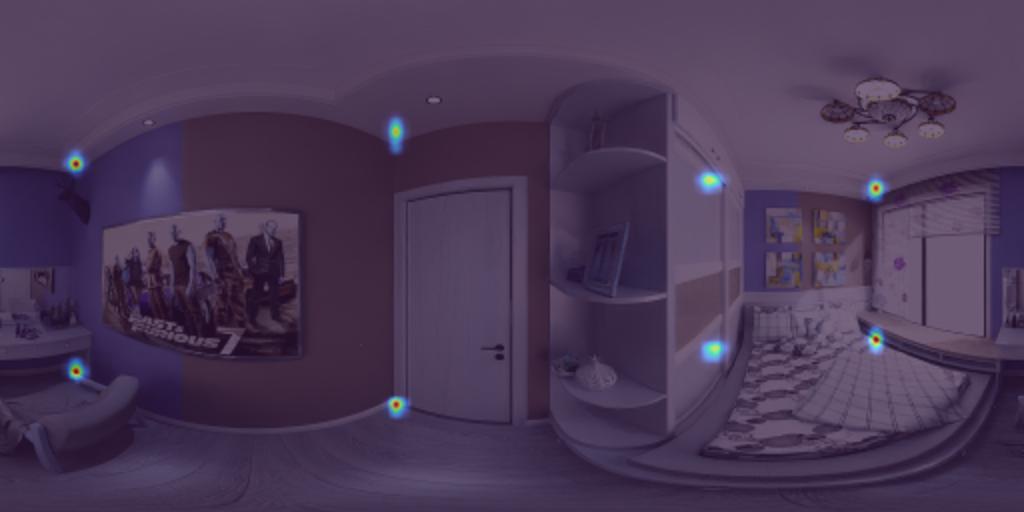}}
    \hfill
    \subfloat{\includegraphics[width=0.24\linewidth]{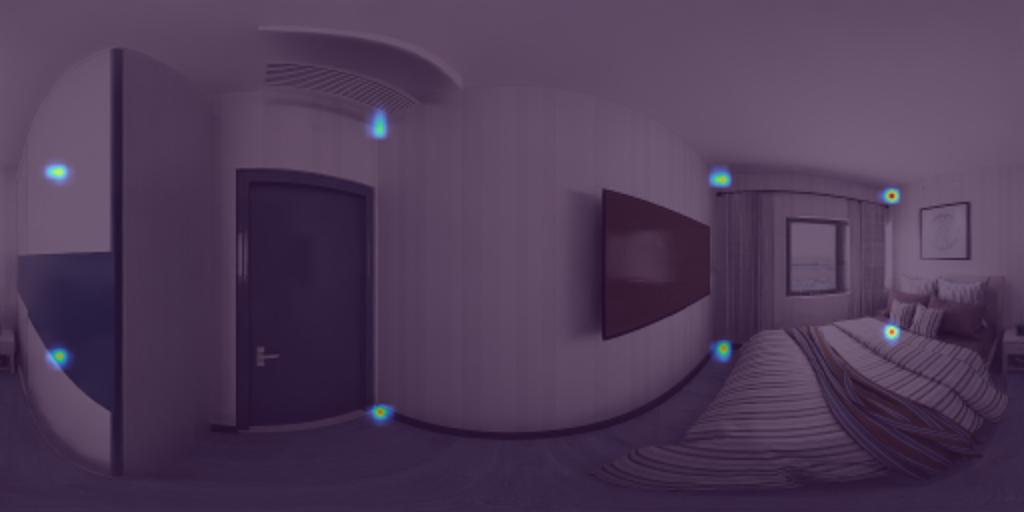}}
    \hfill
    \subfloat{\includegraphics[width=0.24\linewidth]{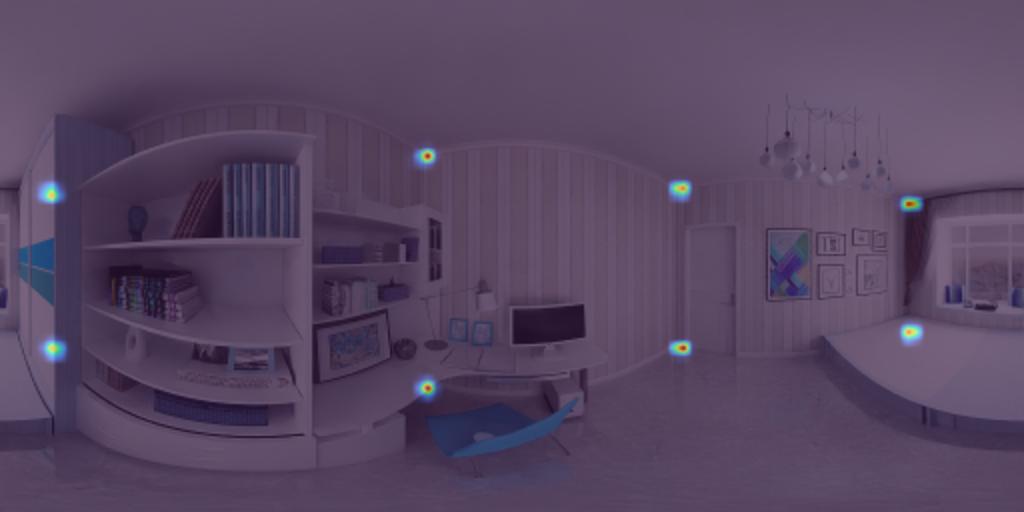}}
    \hfill
    \subfloat{\includegraphics[width=0.24\linewidth]{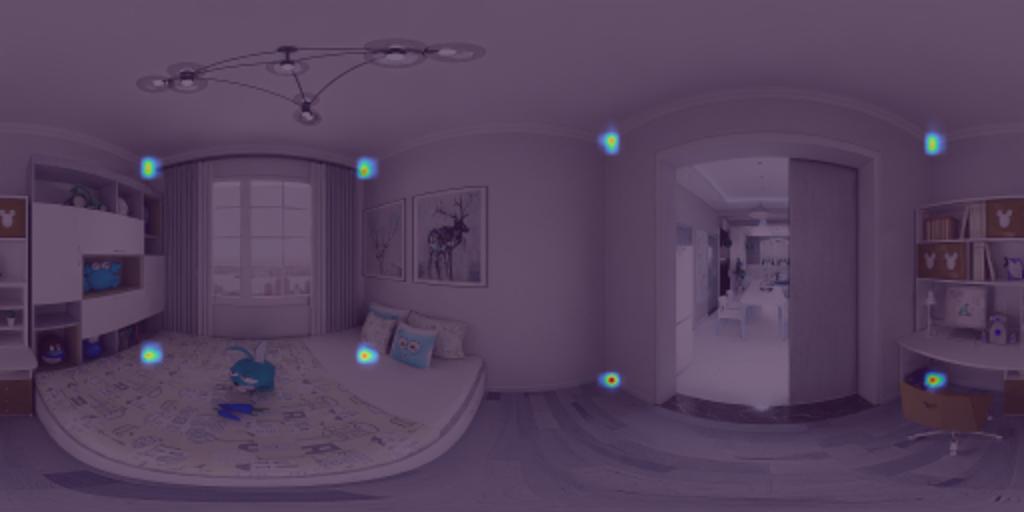}}
    \hfill
    
    \subfloat{\includegraphics[width=0.24\linewidth]{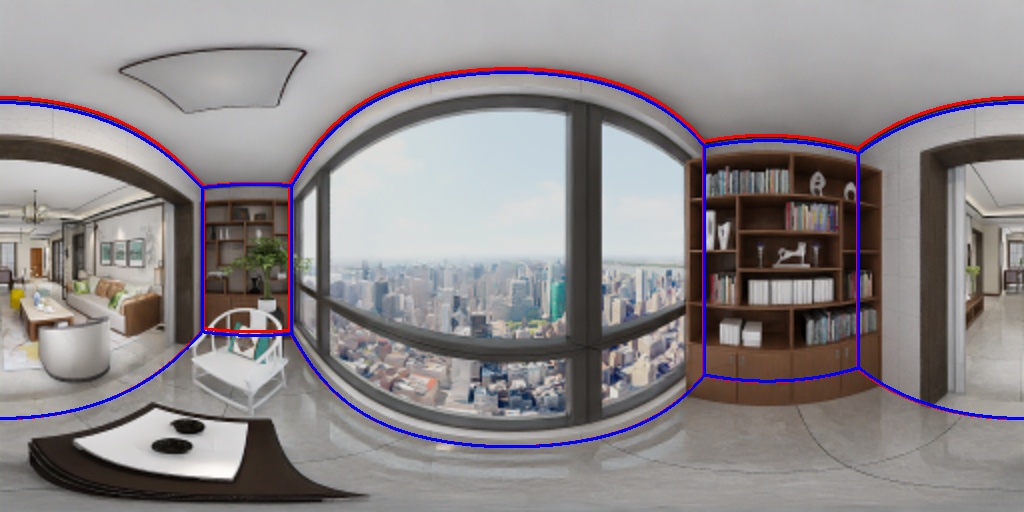}}
    \hfill
    \subfloat{\includegraphics[width=0.24\linewidth]{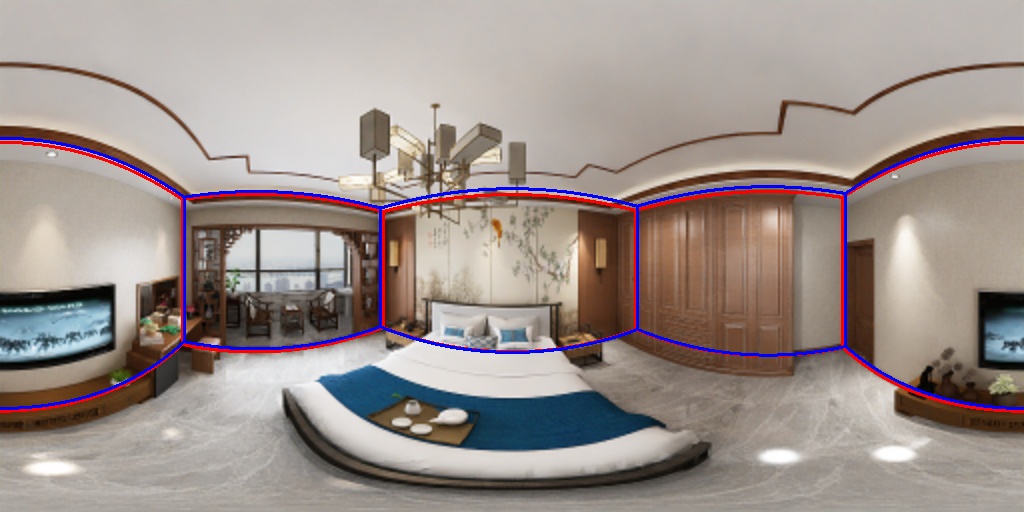}}
    \hfill
    \subfloat{\includegraphics[width=0.24\linewidth]{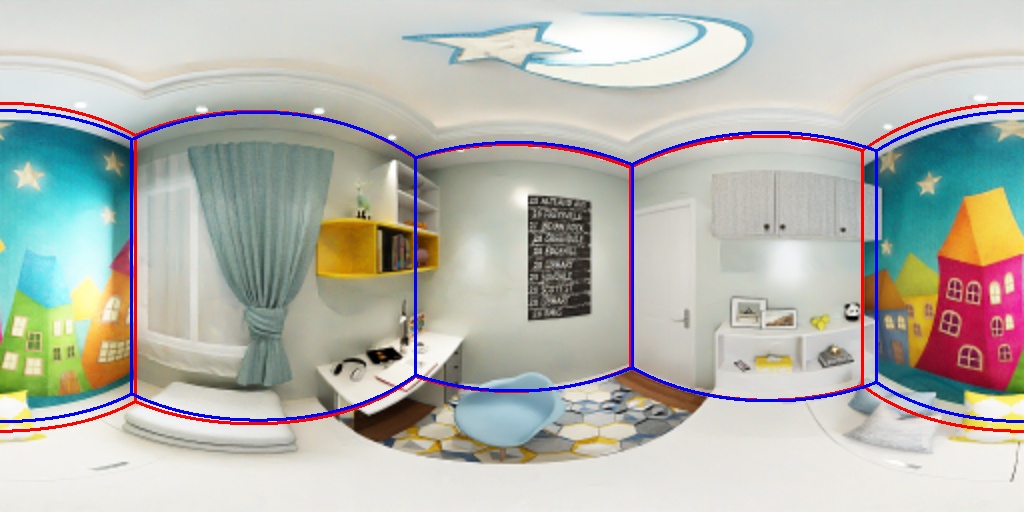}}
    \hfill
    \subfloat{\includegraphics[width=0.24\linewidth]{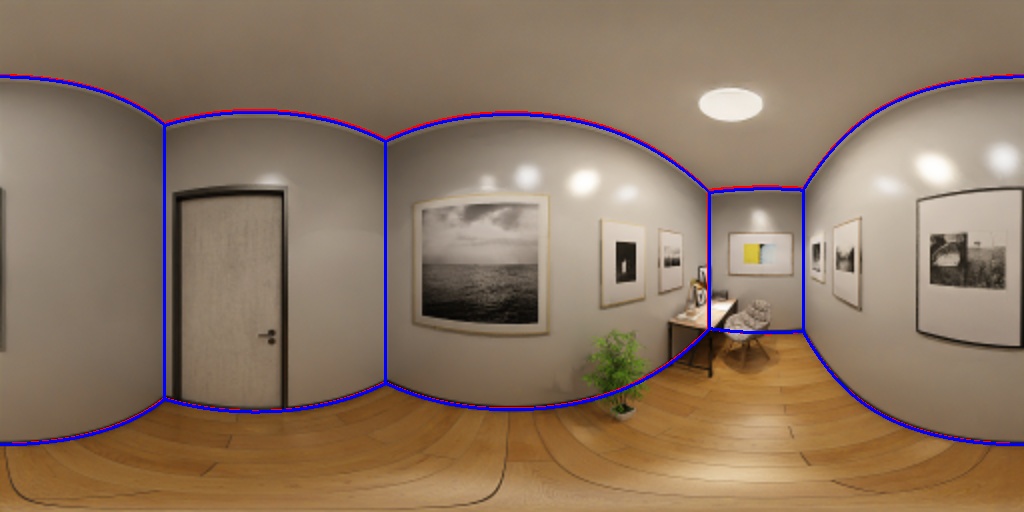}}
    \hfill

    \subfloat{\includegraphics[width=0.24\linewidth]{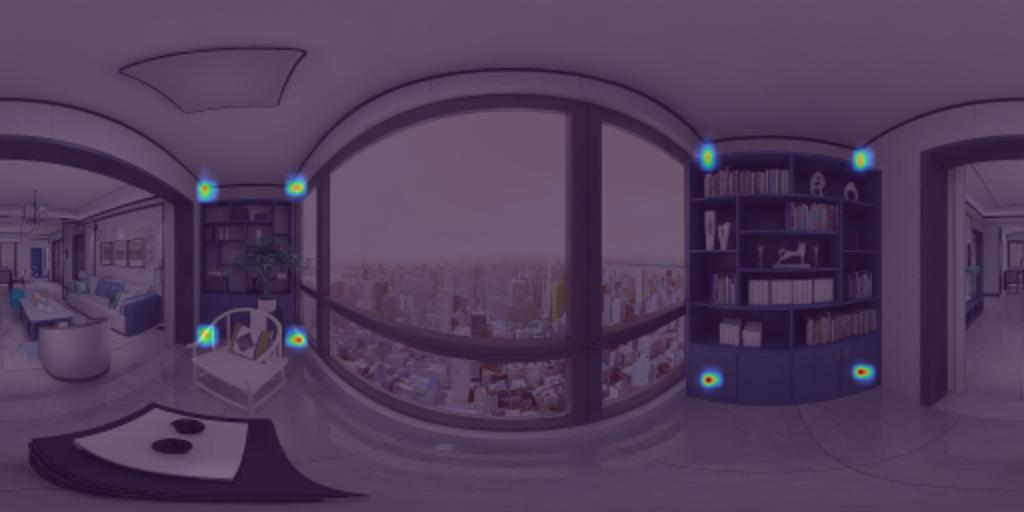}}
    \hfill
    \subfloat{\includegraphics[width=0.24\linewidth]{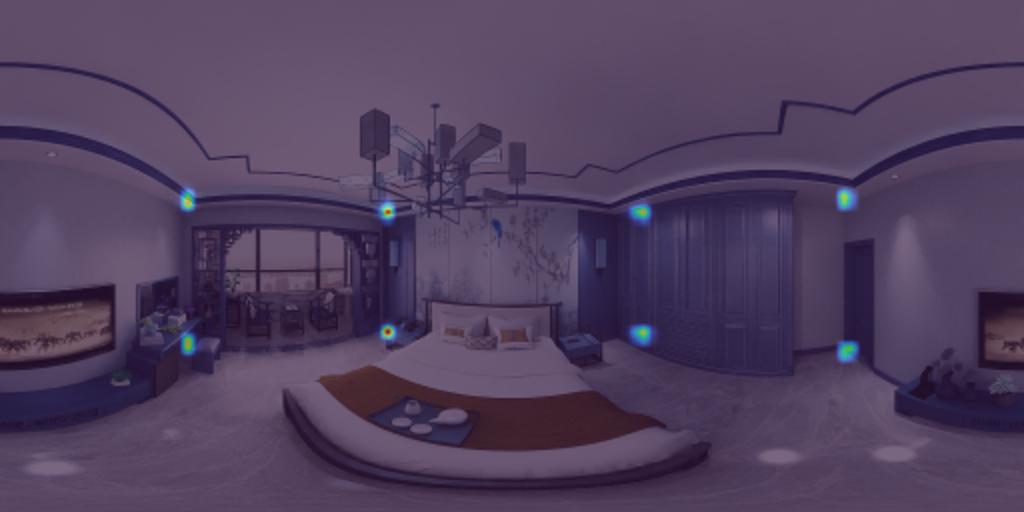}}
    \hfill
    \subfloat{\includegraphics[width=0.24\linewidth]{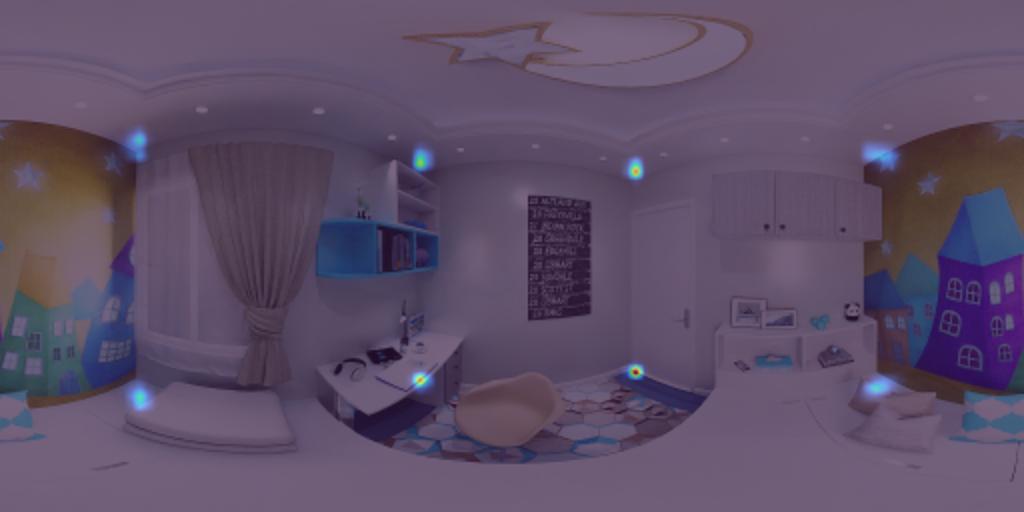}}
    \hfill
    \subfloat{\includegraphics[width=0.24\linewidth]{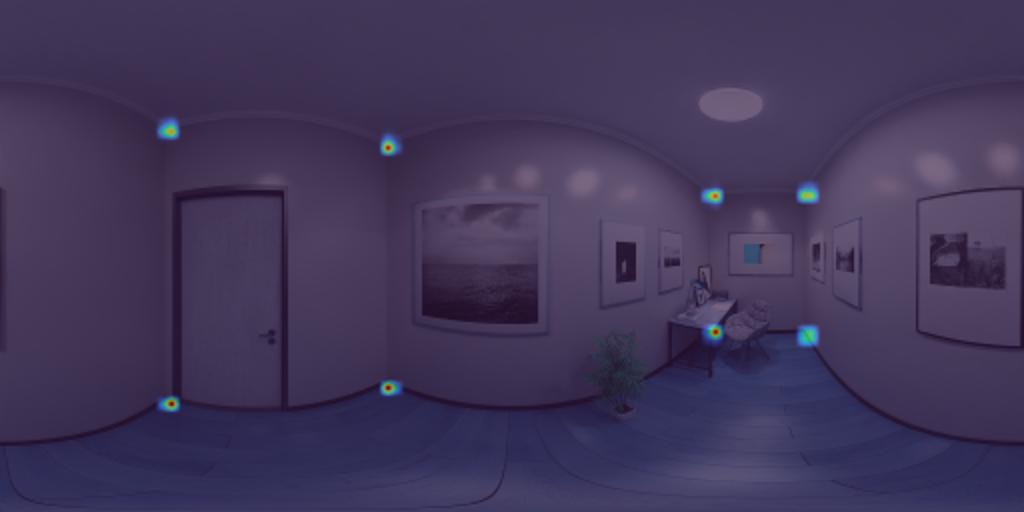}}
    \hfill
    
    \subfloat{\includegraphics[width=0.24\linewidth]{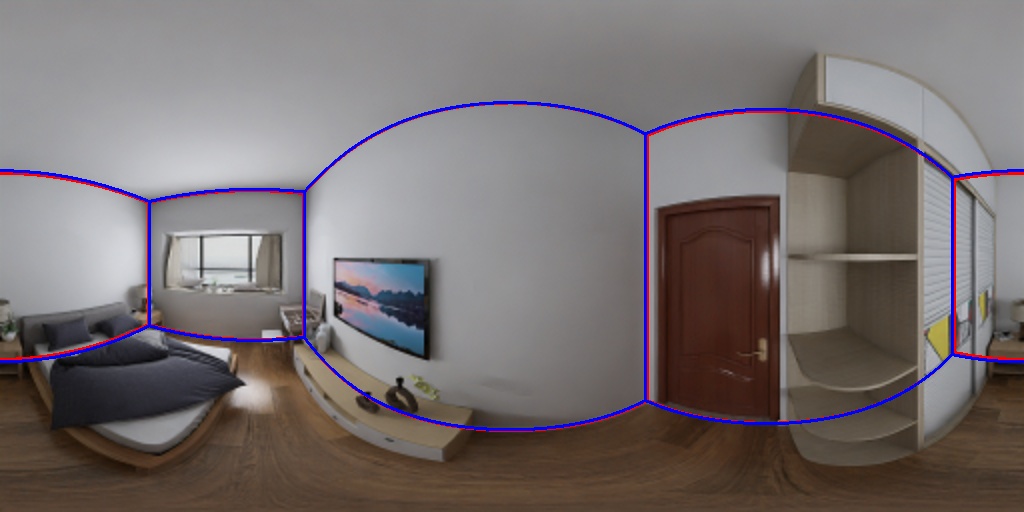}}
    \hfill
    \subfloat{\includegraphics[width=0.24\linewidth]{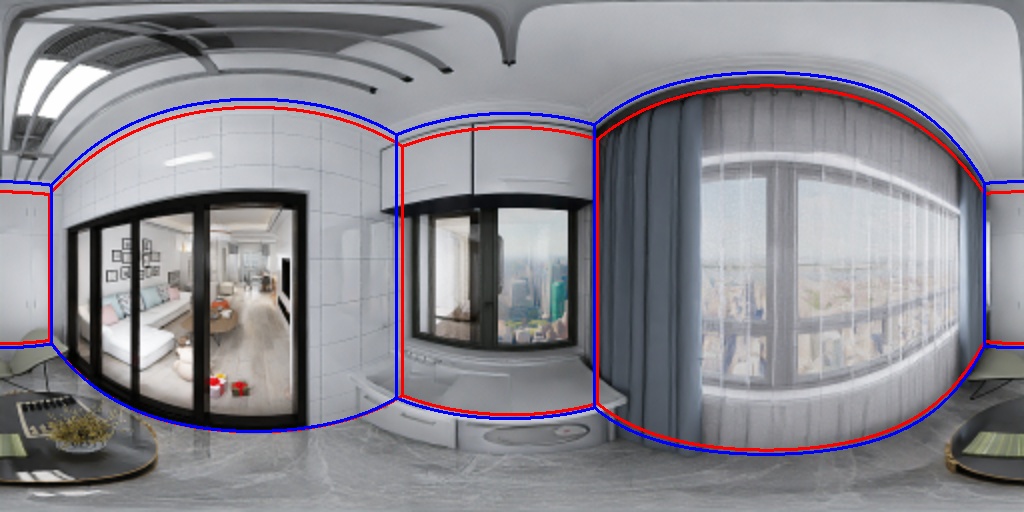}}
    \hfill
    \subfloat{\includegraphics[width=0.24\linewidth]{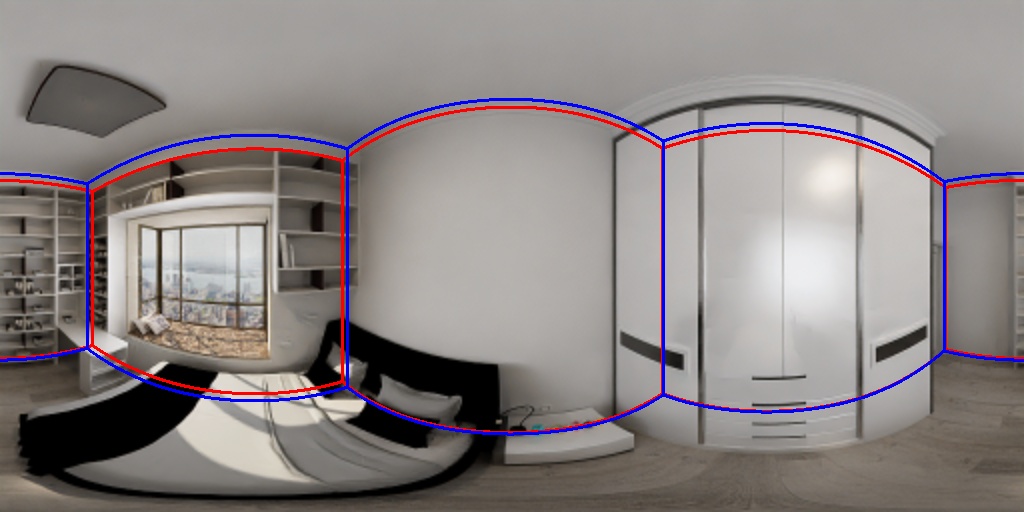}}
    \hfill
    \subfloat{\includegraphics[width=0.24\linewidth]{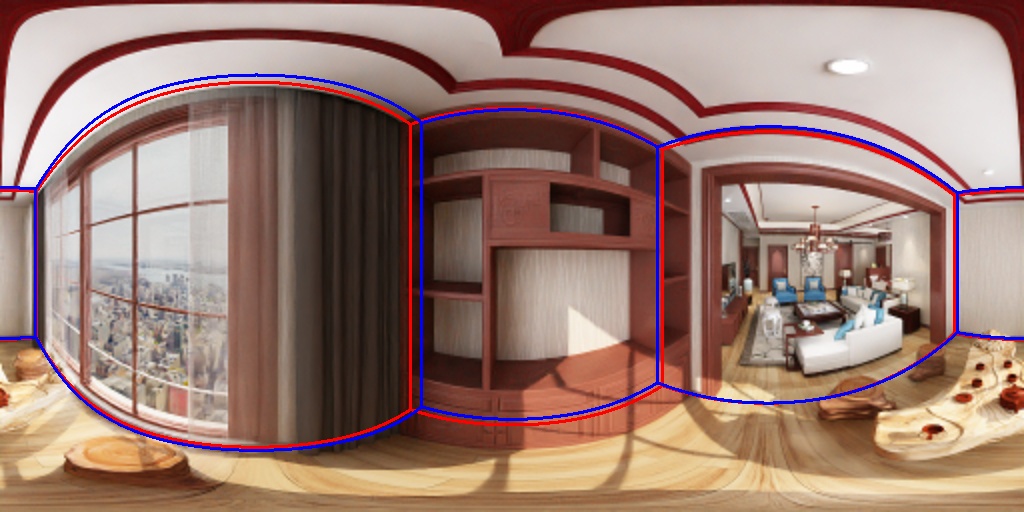}}
    \hfill

    \subfloat{\includegraphics[width=0.24\linewidth]{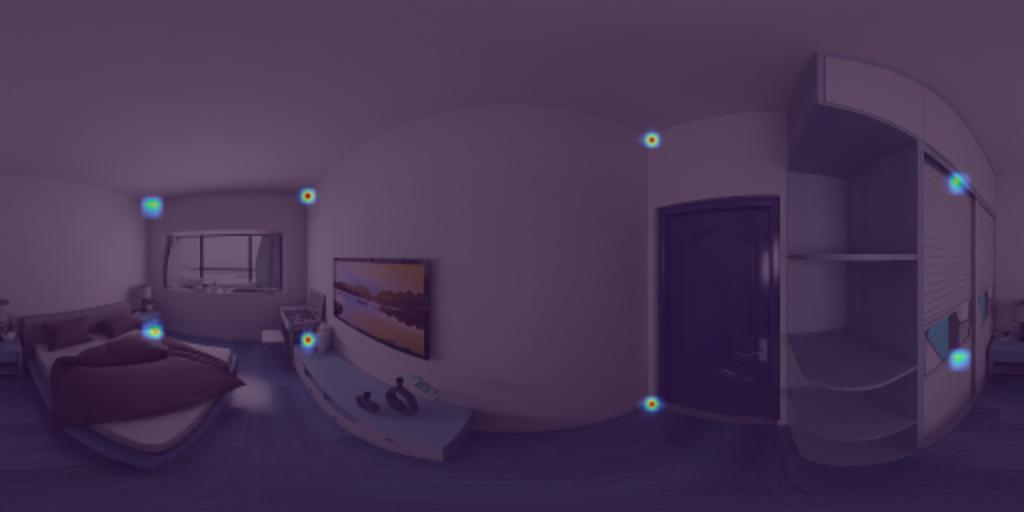}}
    \hfill
    \subfloat{\includegraphics[width=0.24\linewidth]{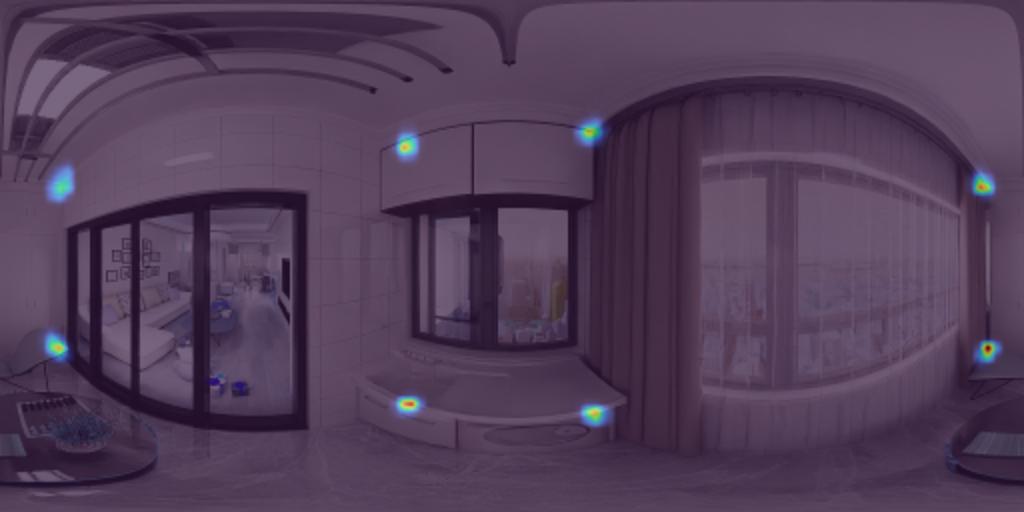}}
    \hfill
    \subfloat{\includegraphics[width=0.24\linewidth]{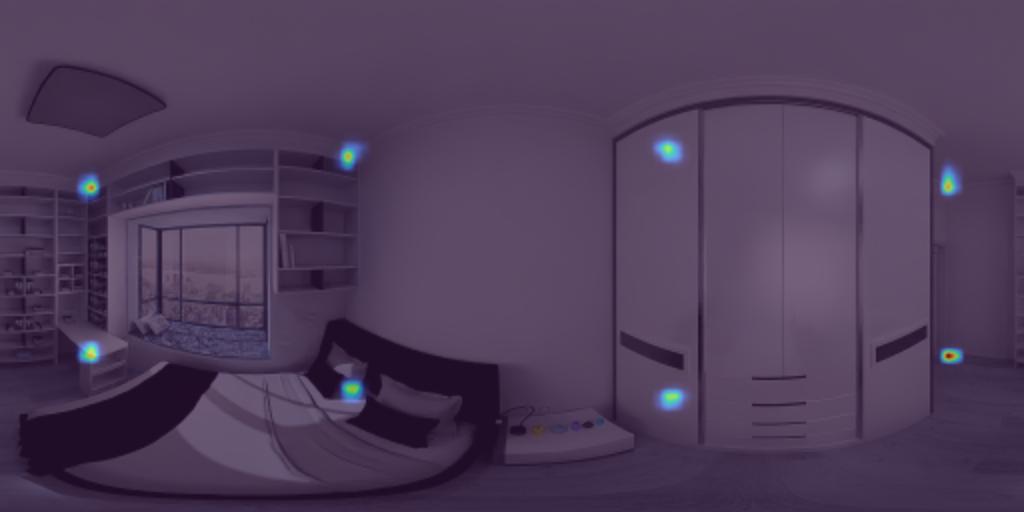}}
    \hfill
    \subfloat{\includegraphics[width=0.24\linewidth]{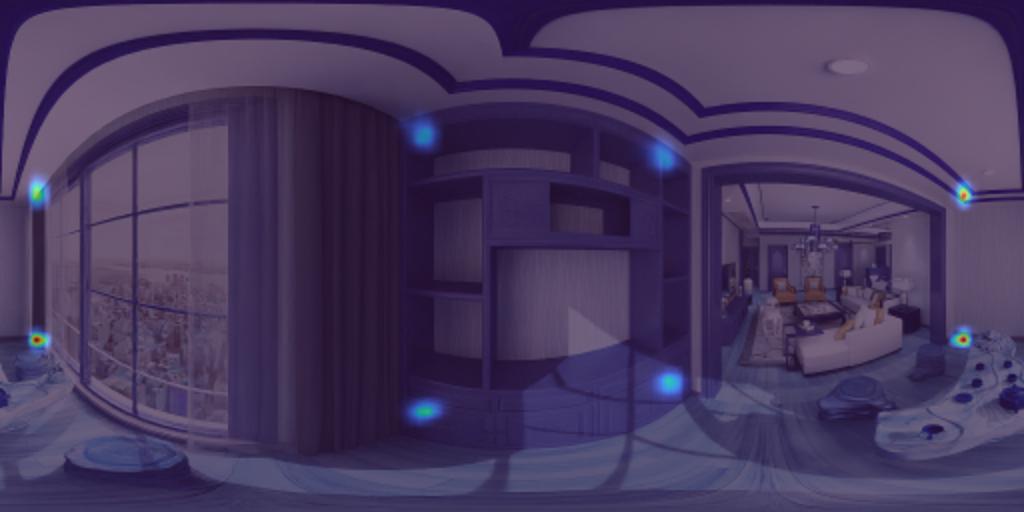}}
    \hfill
    
    \subfloat{\includegraphics[width=0.24\linewidth]{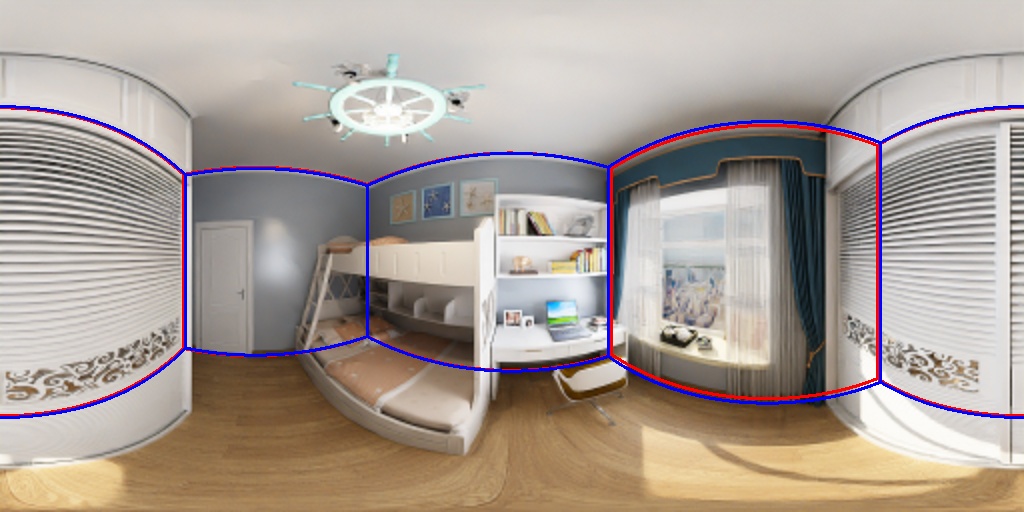}}
    \hfill
    \subfloat{\includegraphics[width=0.24\linewidth]{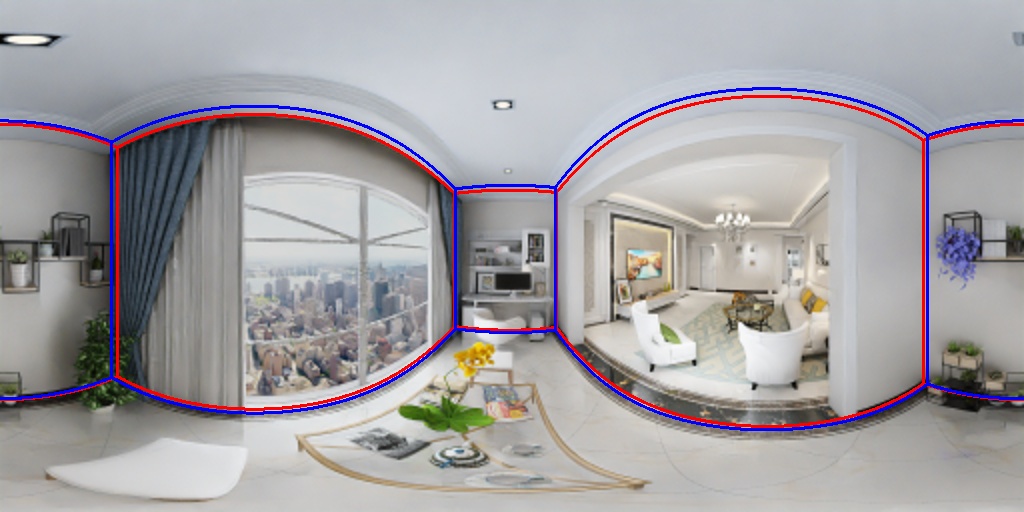}}
    \hfill
    \subfloat{\includegraphics[width=0.24\linewidth]{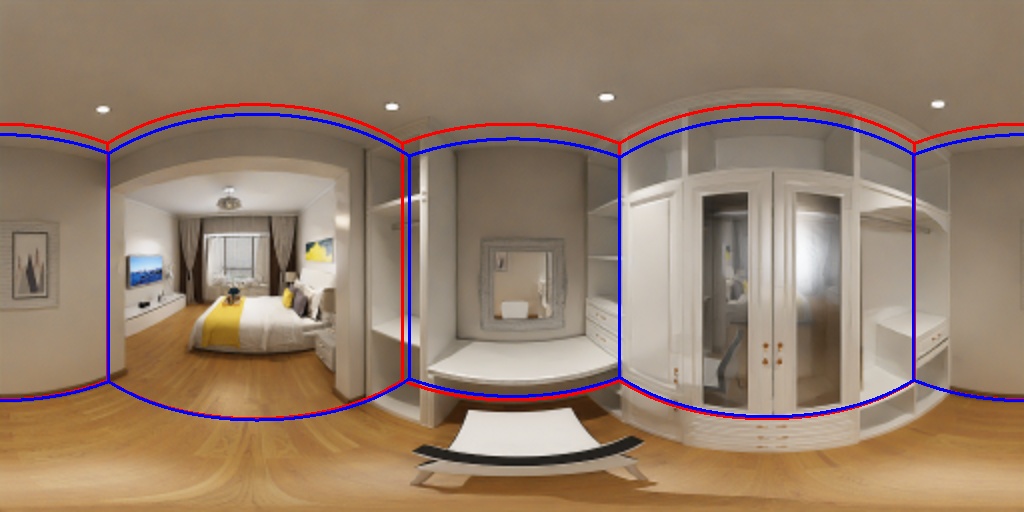}}
    \hfill
    \subfloat{\includegraphics[width=0.24\linewidth]{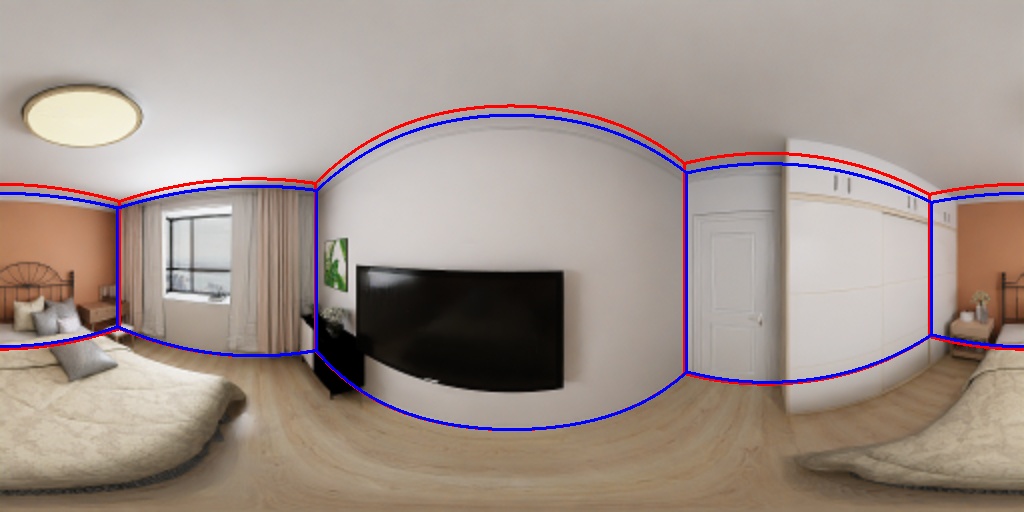}}
    \hfill
    
    \subfloat{\includegraphics[width=0.24\linewidth]{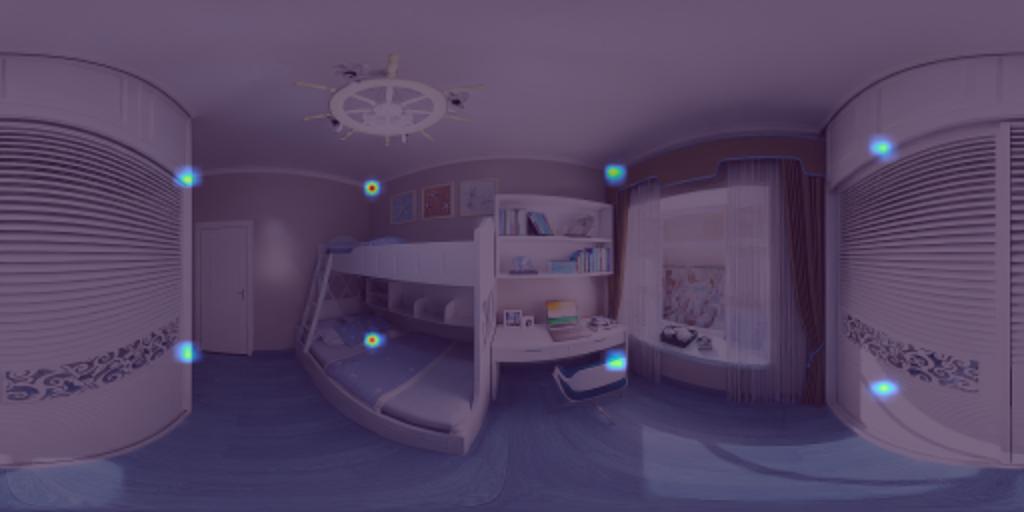}}
    \hfill
    \subfloat{\includegraphics[width=0.24\linewidth]{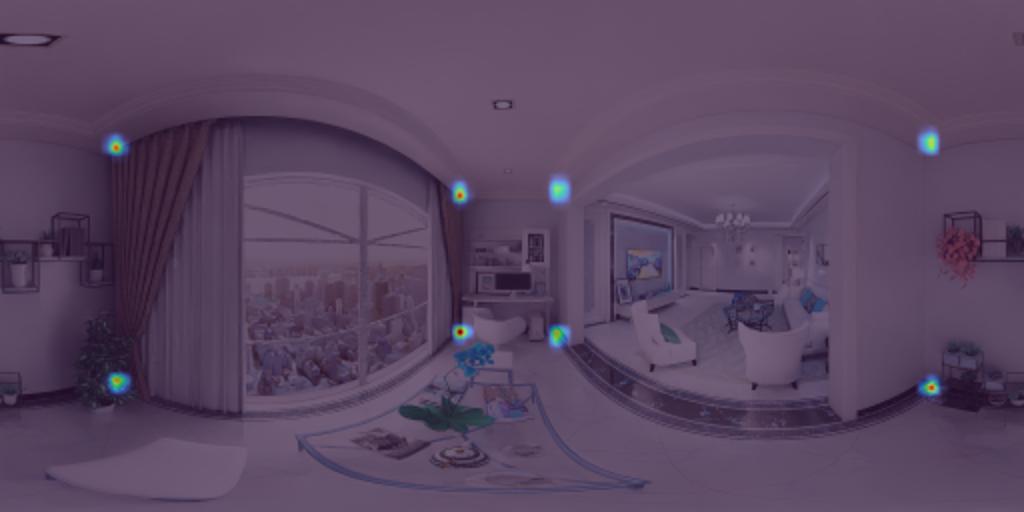}}
    \hfill
    \subfloat{\includegraphics[width=0.24\linewidth]{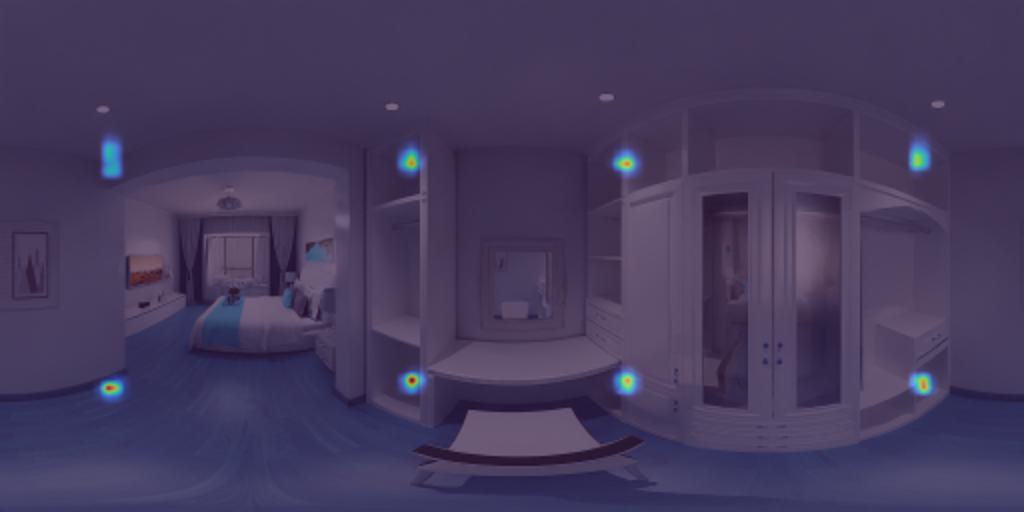}}
    \hfill
    \subfloat{\includegraphics[width=0.24\linewidth]{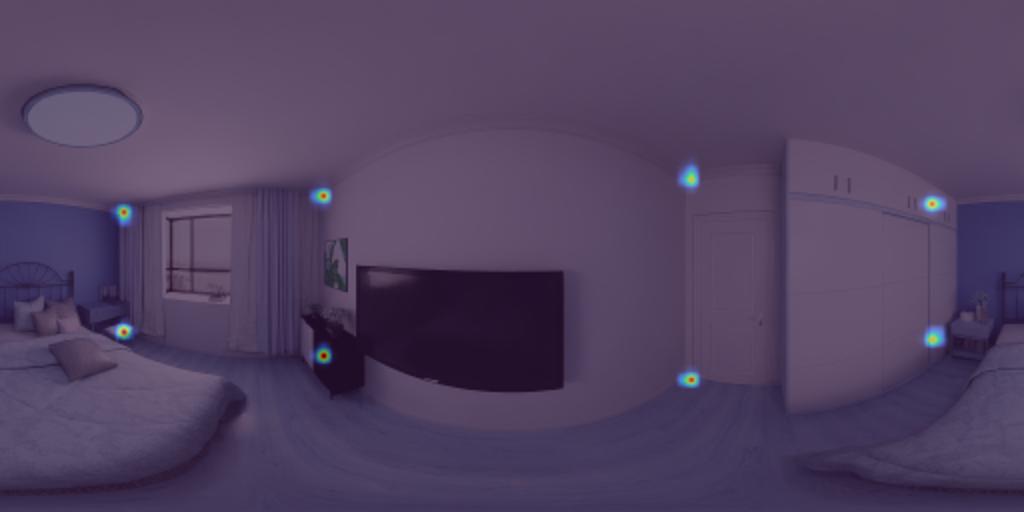}}
    \hfill
    
    \caption{
        Additional qualitative results on the Structured3D dataset.
    }
\label{fig:s3d}
\end{figure*}

\begin{figure*}[!htbp]
    \centering

    \subfloat{\includegraphics[width=0.24\linewidth]{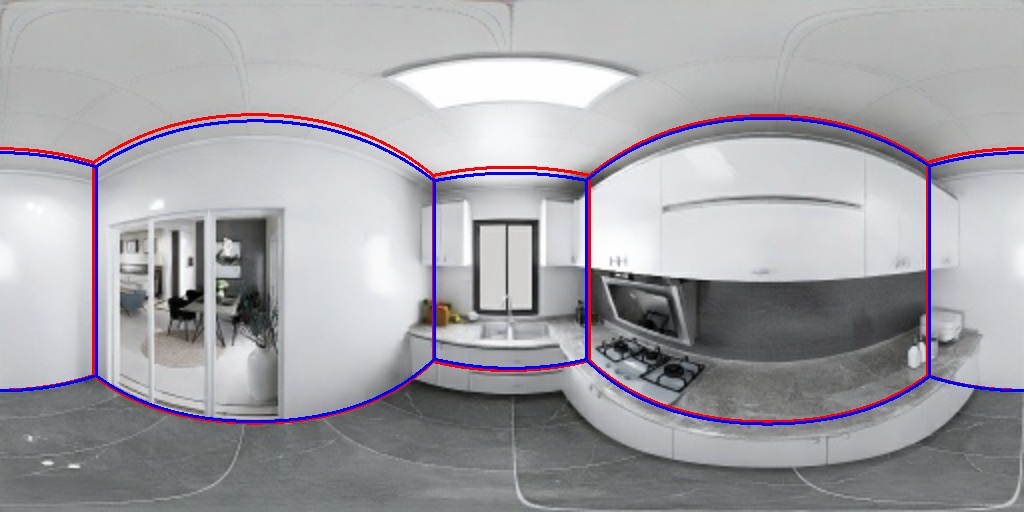}}
    \hfill
    \subfloat{\includegraphics[width=0.24\linewidth]{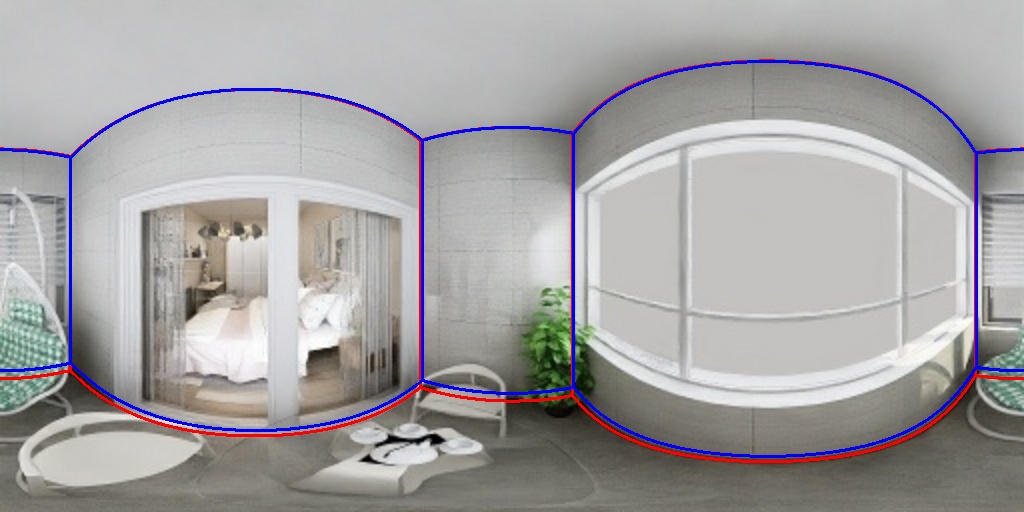}}
    \hfill
    \subfloat{\includegraphics[width=0.24\linewidth]{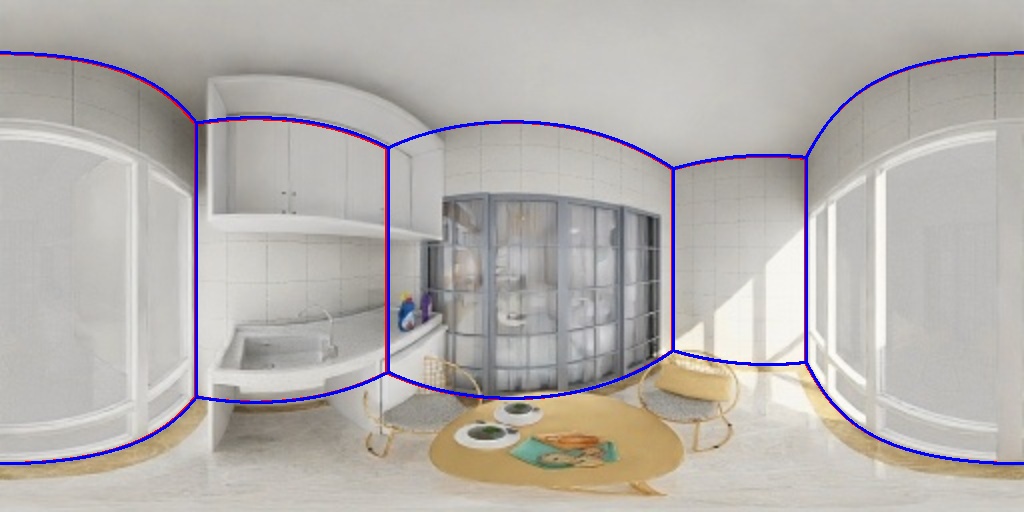}}
    \hfill
    \subfloat{\includegraphics[width=0.24\linewidth]{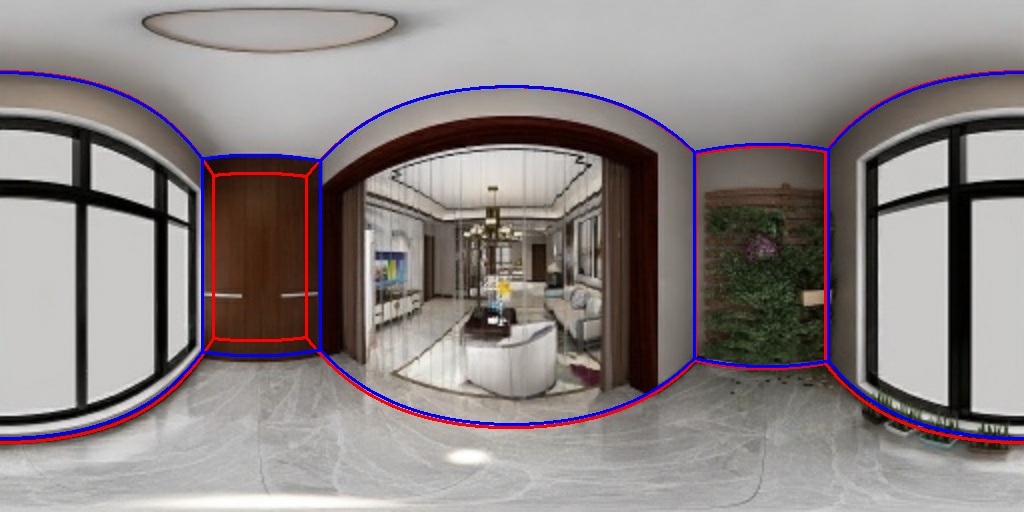}}
    \hfill
    
    \subfloat{\includegraphics[width=0.24\linewidth]{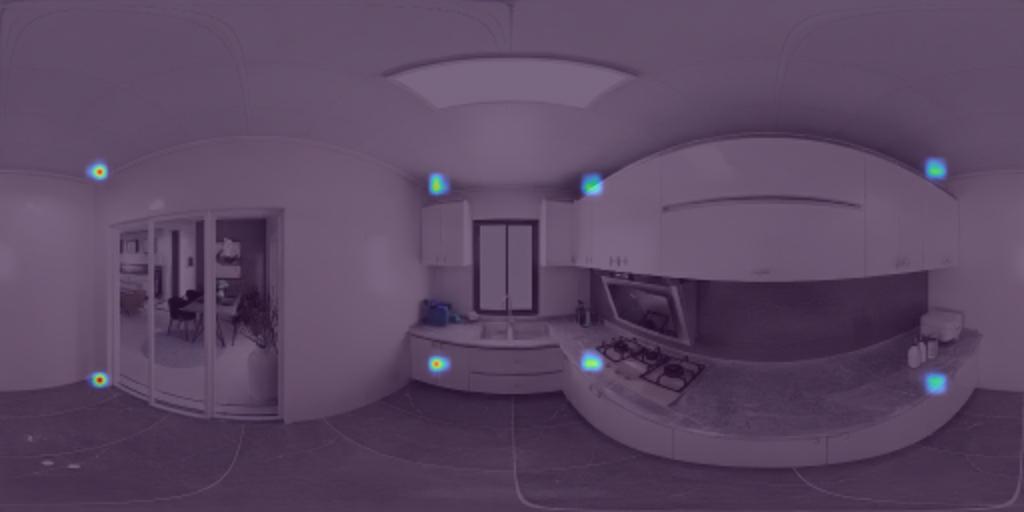}}
    \hfill
    \subfloat{\includegraphics[width=0.24\linewidth]{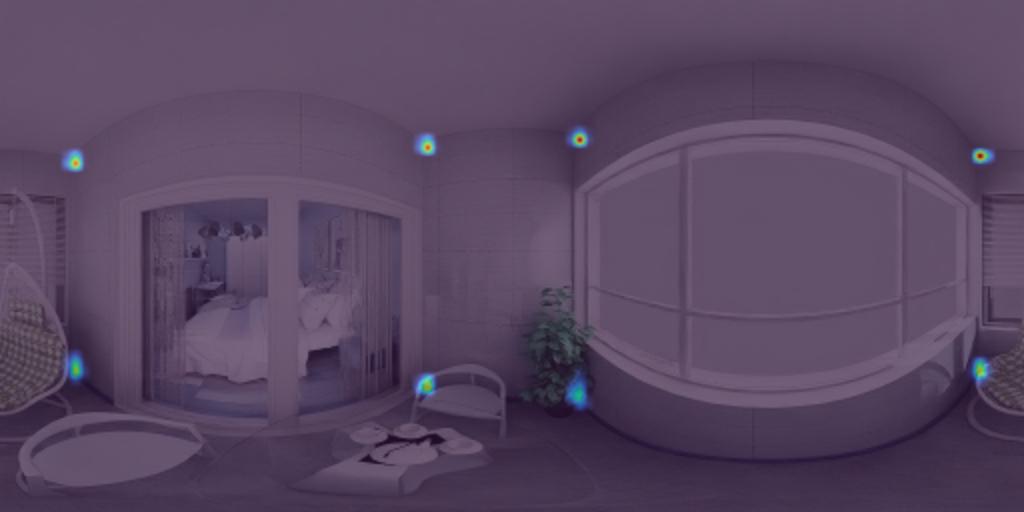}}
    \hfill
    \subfloat{\includegraphics[width=0.24\linewidth]{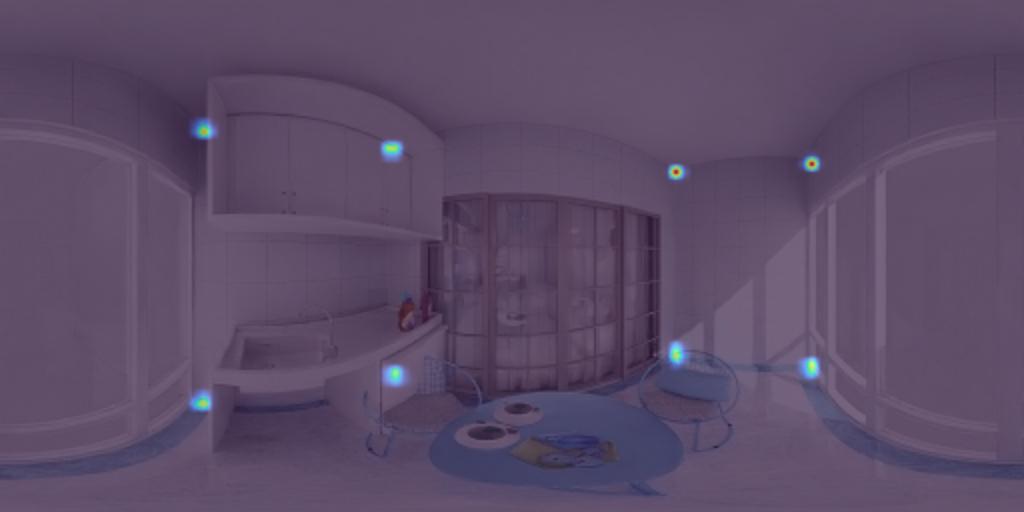}}
    \hfill
    \subfloat{\includegraphics[width=0.24\linewidth]{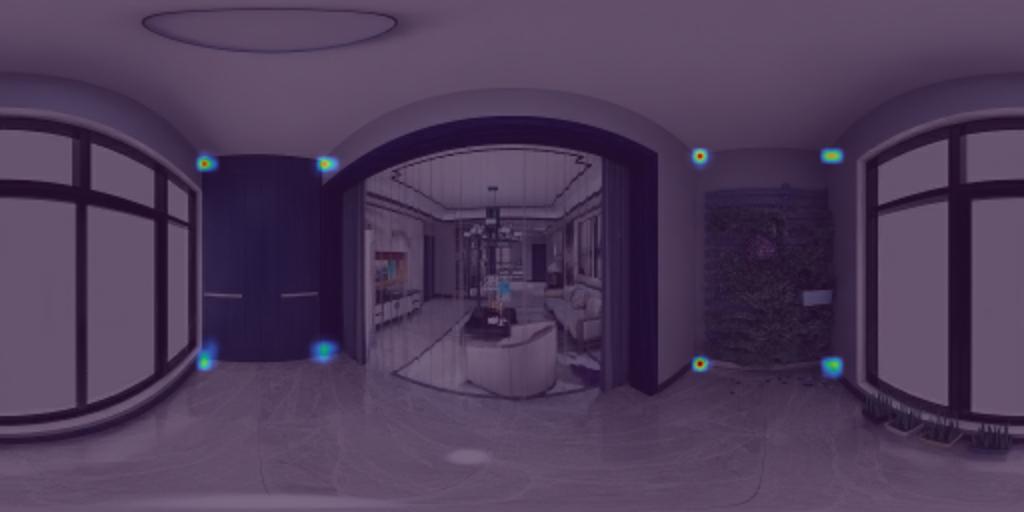}}
    \hfill
    
    \subfloat{\includegraphics[width=0.24\linewidth]{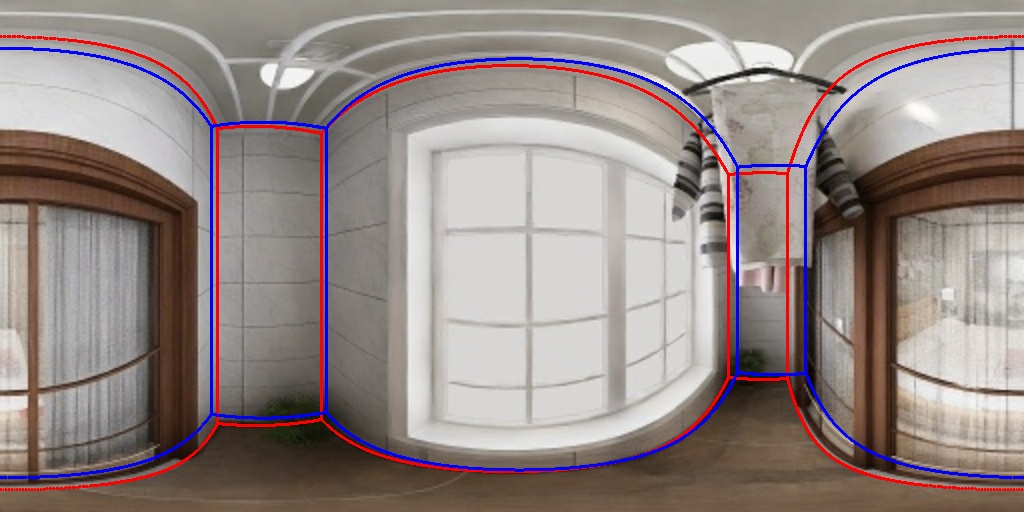}}
    \hfill
    \subfloat{\includegraphics[width=0.24\linewidth]{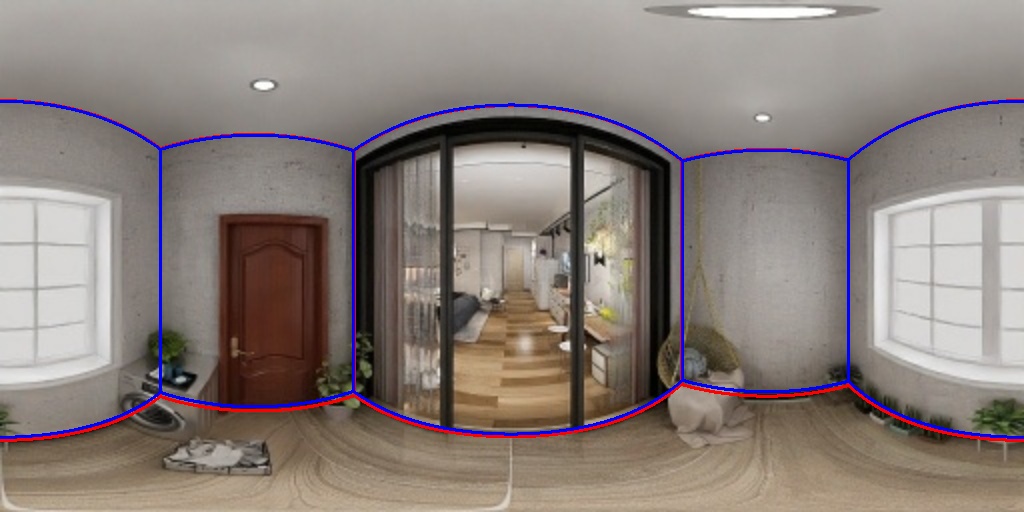}}
    \hfill
    \subfloat{\includegraphics[width=0.24\linewidth]{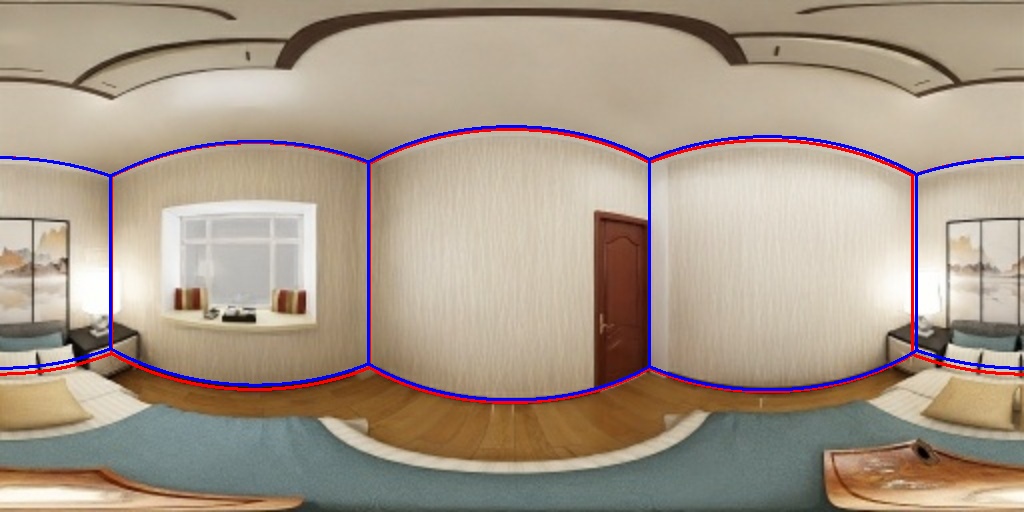}}
    \hfill
    \subfloat{\includegraphics[width=0.24\linewidth]{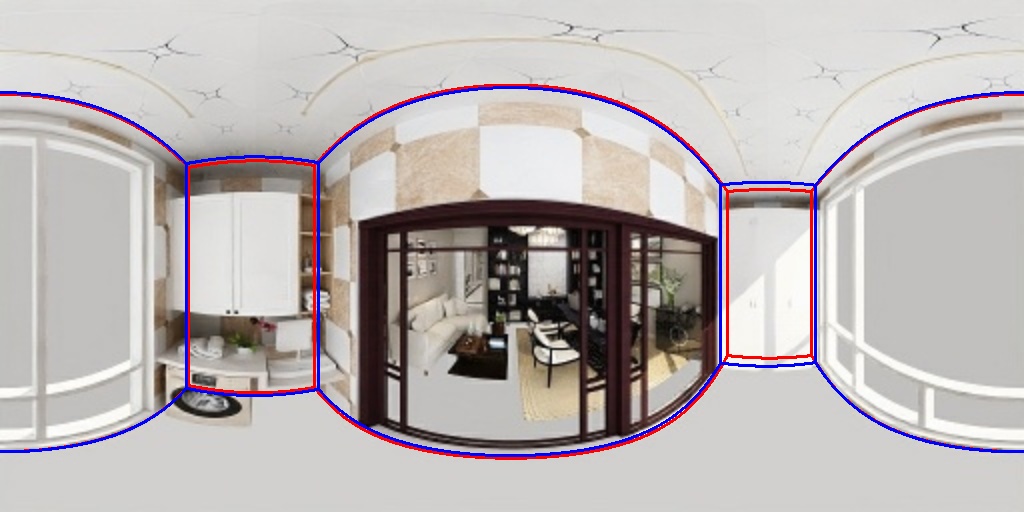}}
    \hfill
   
    \subfloat{\includegraphics[width=0.24\linewidth]{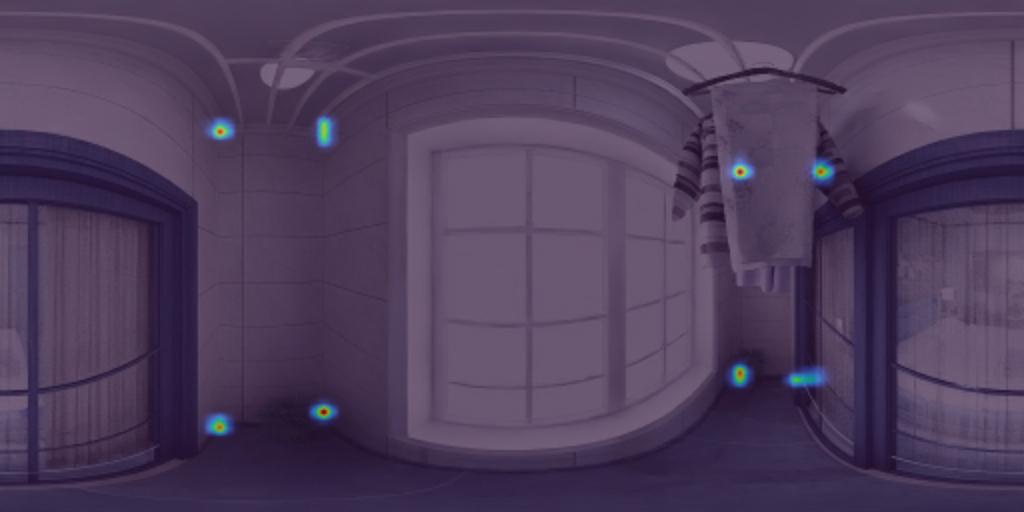}}
    \hfill
    \subfloat{\includegraphics[width=0.24\linewidth]{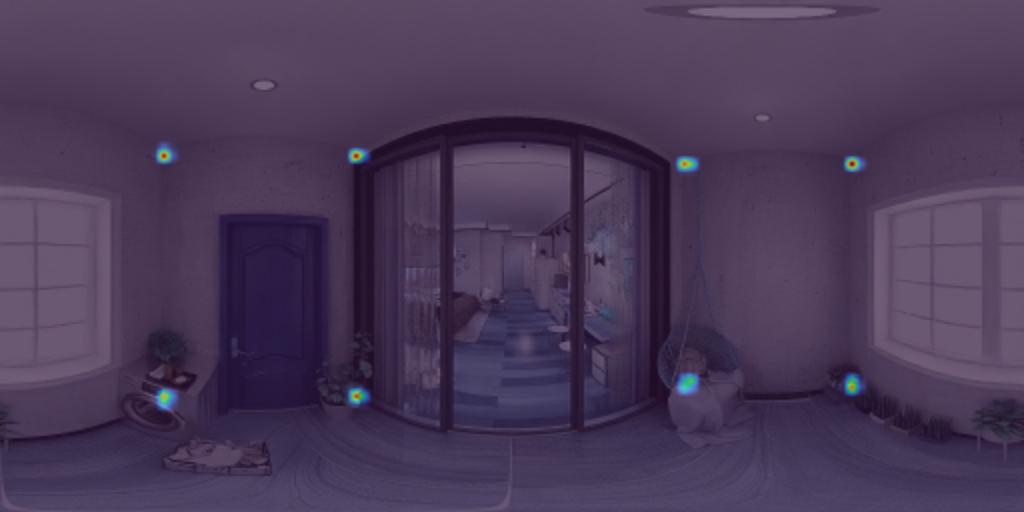}}
    \hfill
    \subfloat{\includegraphics[width=0.24\linewidth]{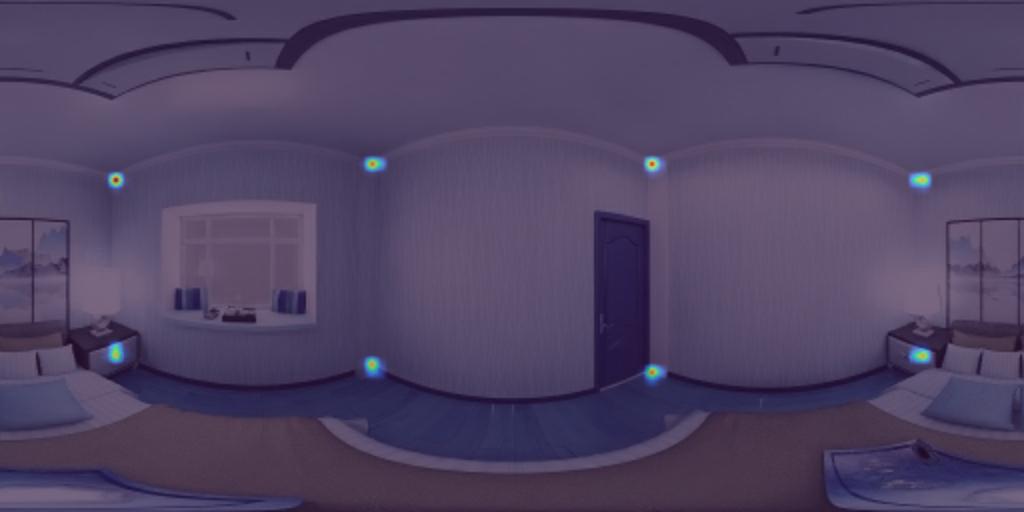}}
    \hfill
    \subfloat{\includegraphics[width=0.24\linewidth]{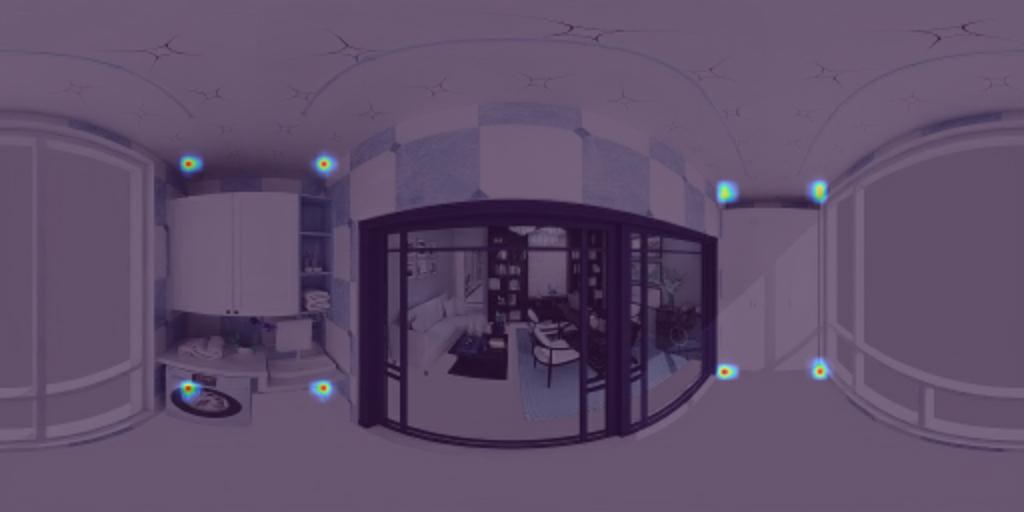}}
    \hfill
    
    \subfloat{\includegraphics[width=0.24\linewidth]{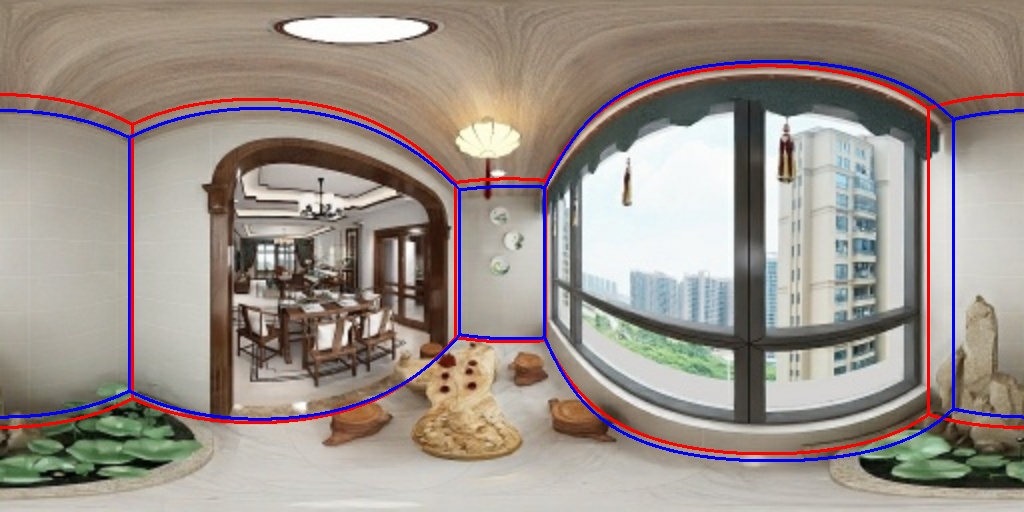}}
    \hfill
    \subfloat{\includegraphics[width=0.24\linewidth]{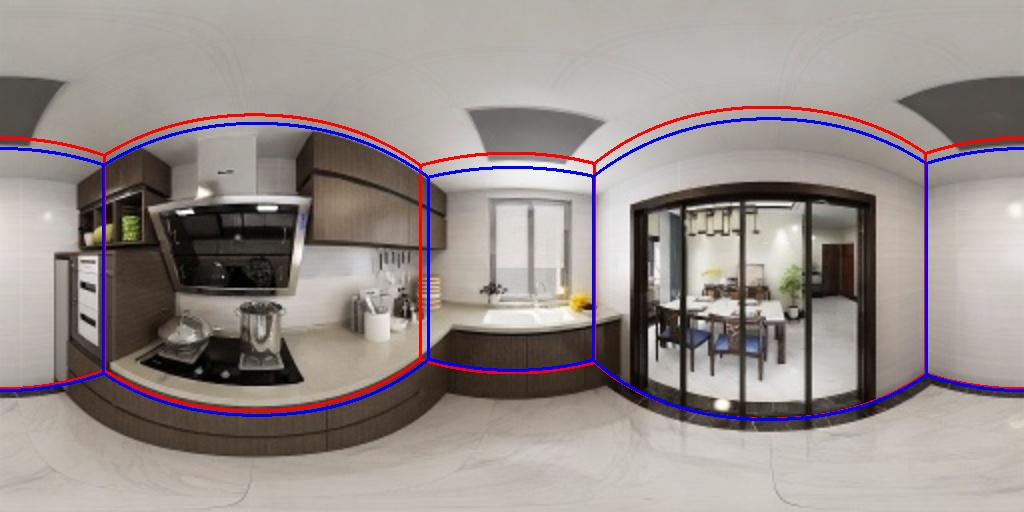}}
    \hfill
    \subfloat{\includegraphics[width=0.24\linewidth]{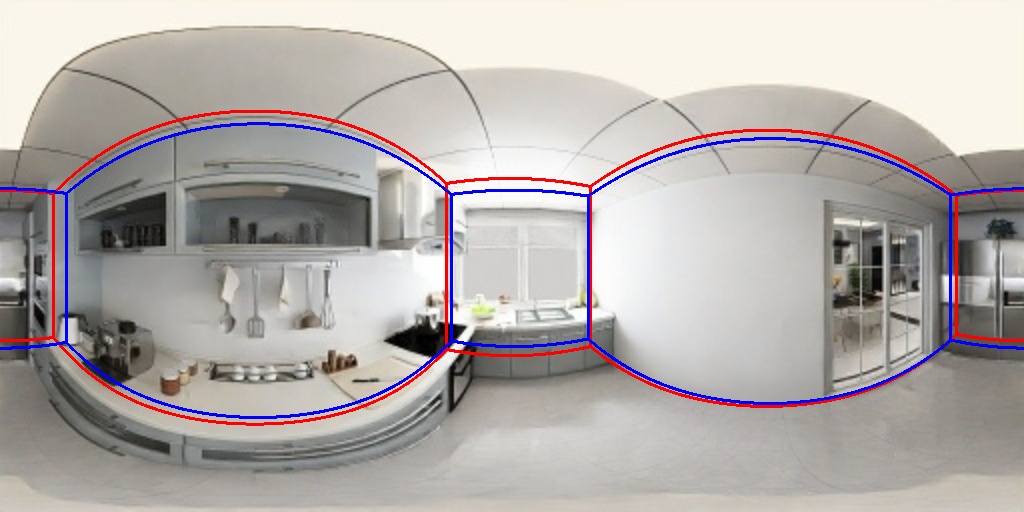}}
    \hfill
    \subfloat{\includegraphics[width=0.24\linewidth]{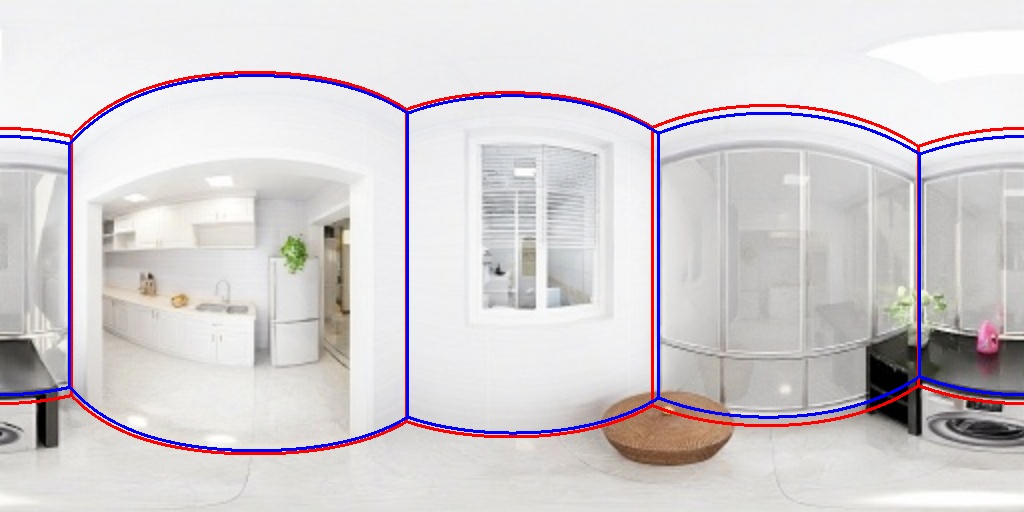}}
    \hfill
    
    \subfloat{\includegraphics[width=0.24\linewidth]{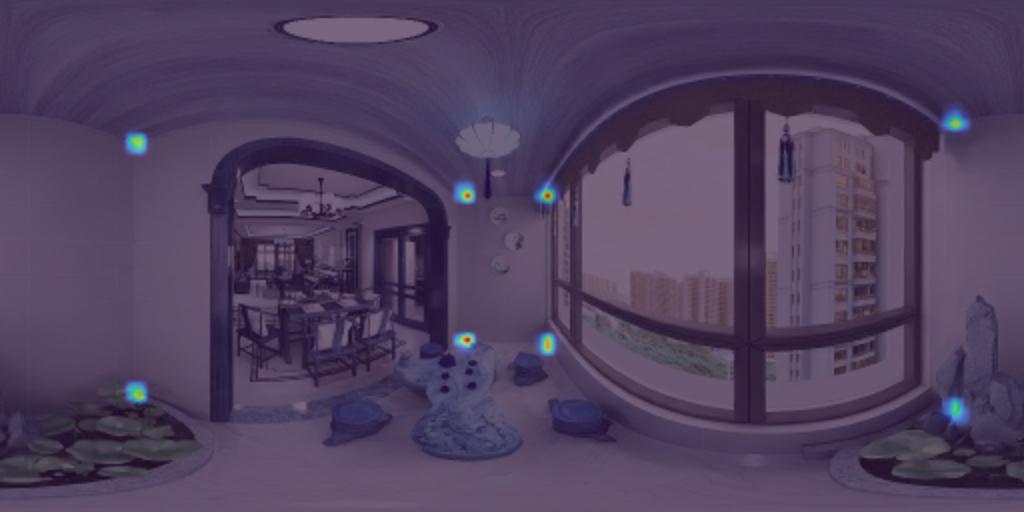}}
    \hfill
    \subfloat{\includegraphics[width=0.24\linewidth]{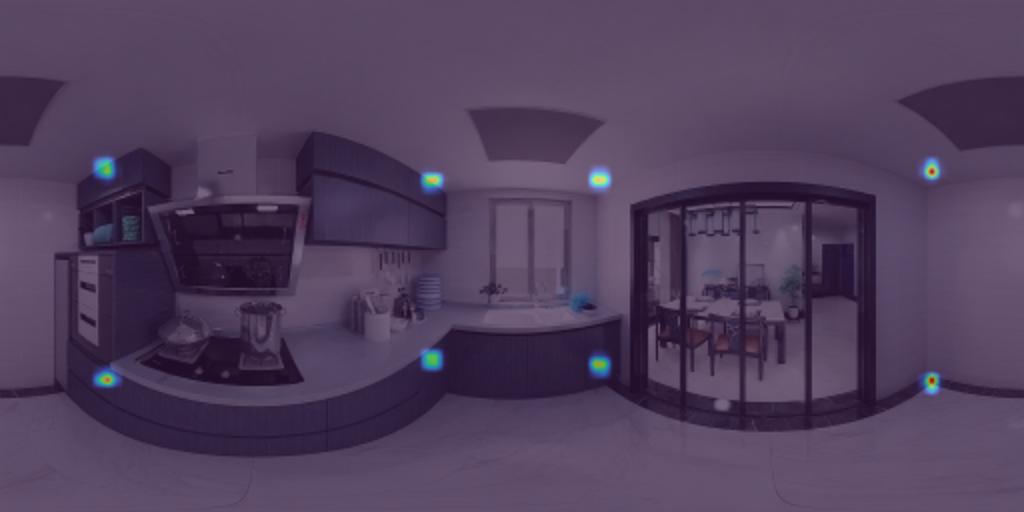}}
    \hfill
    \subfloat{\includegraphics[width=0.24\linewidth]{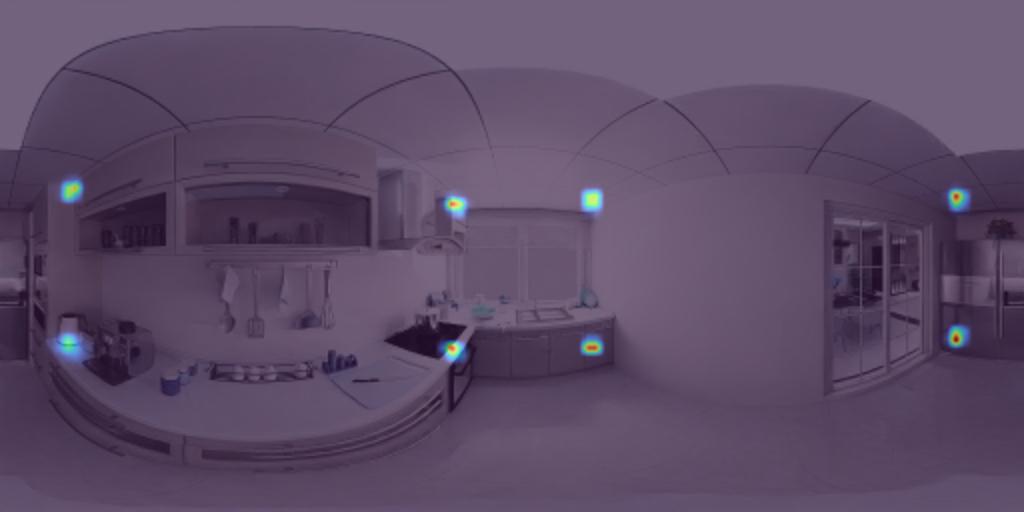}}
    \hfill
    \subfloat{\includegraphics[width=0.24\linewidth]{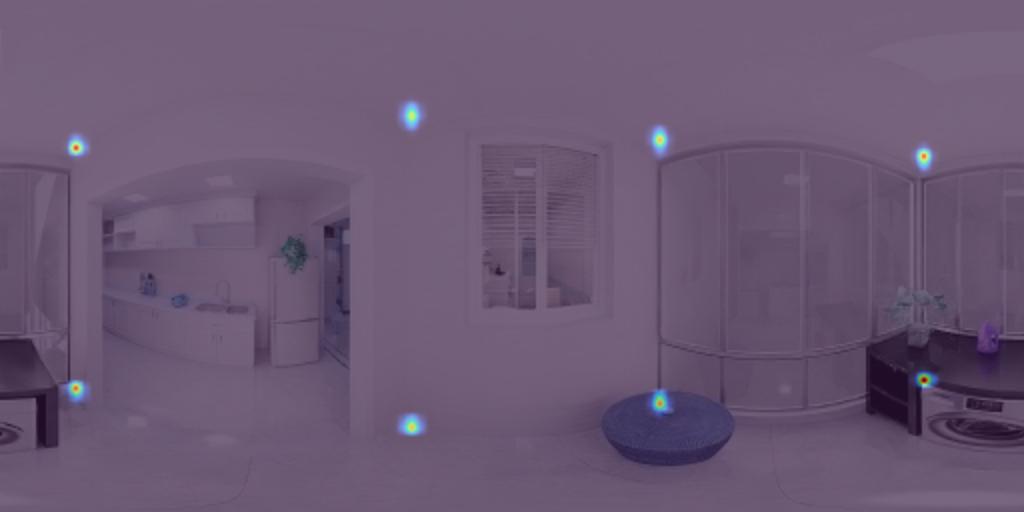}}
    \hfill
    
    \subfloat{\includegraphics[width=0.24\linewidth]{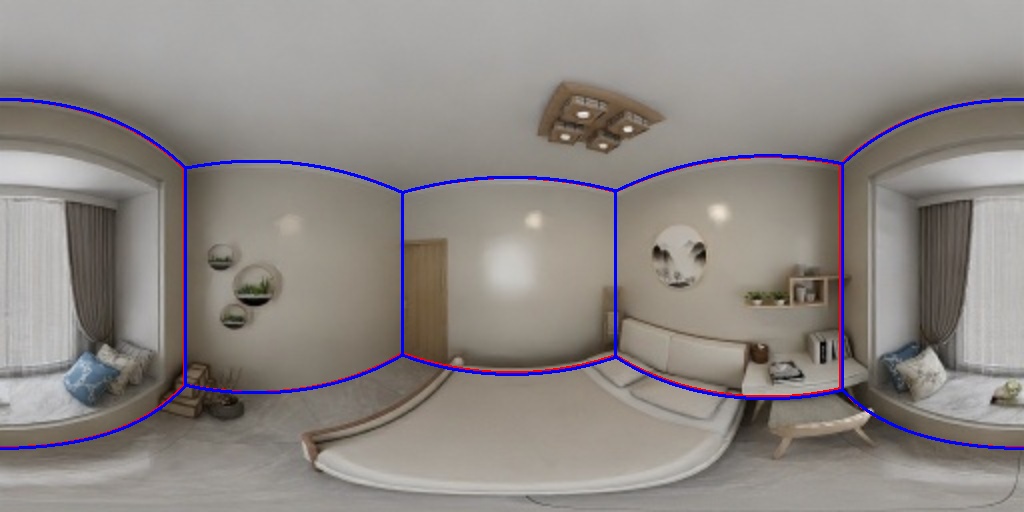}}
    \hfill
    \subfloat{\includegraphics[width=0.24\linewidth]{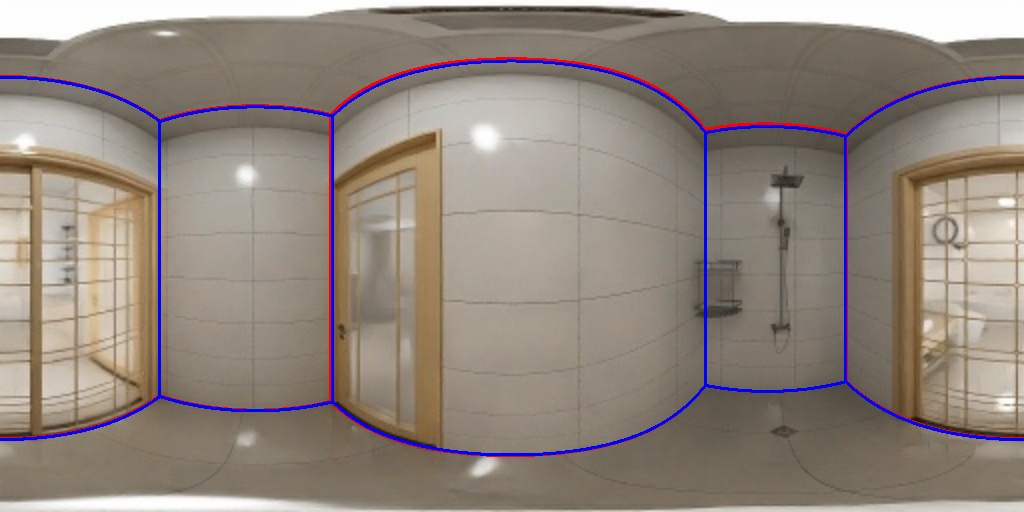}}
    \hfill
    \subfloat{\includegraphics[width=0.24\linewidth]{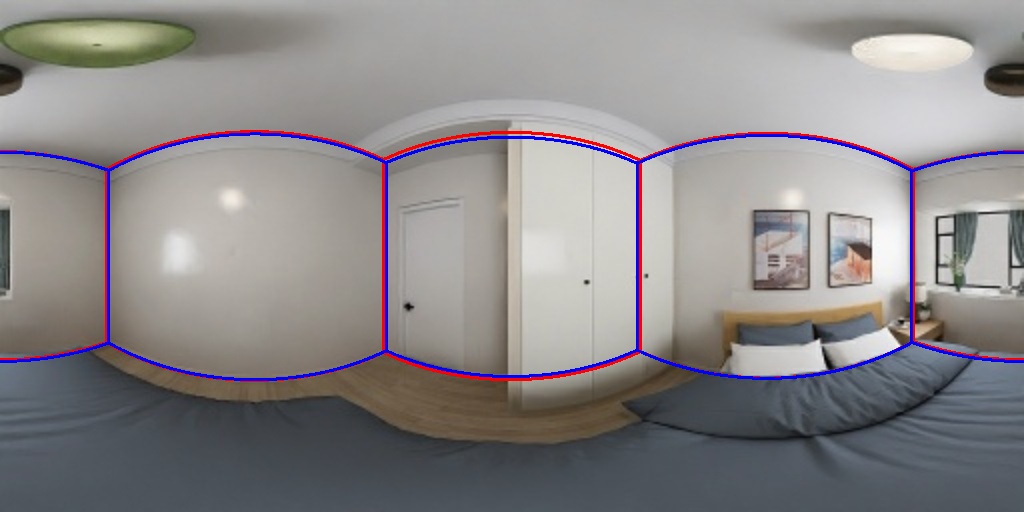}}
    \hfill
    \subfloat{\includegraphics[width=0.24\linewidth]{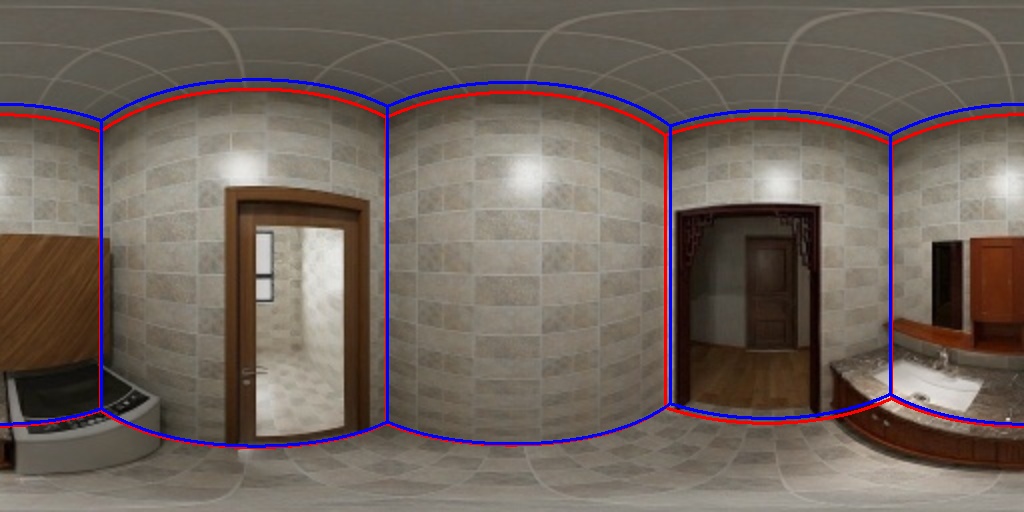}}
    \hfill

    \subfloat{\includegraphics[width=0.24\linewidth]{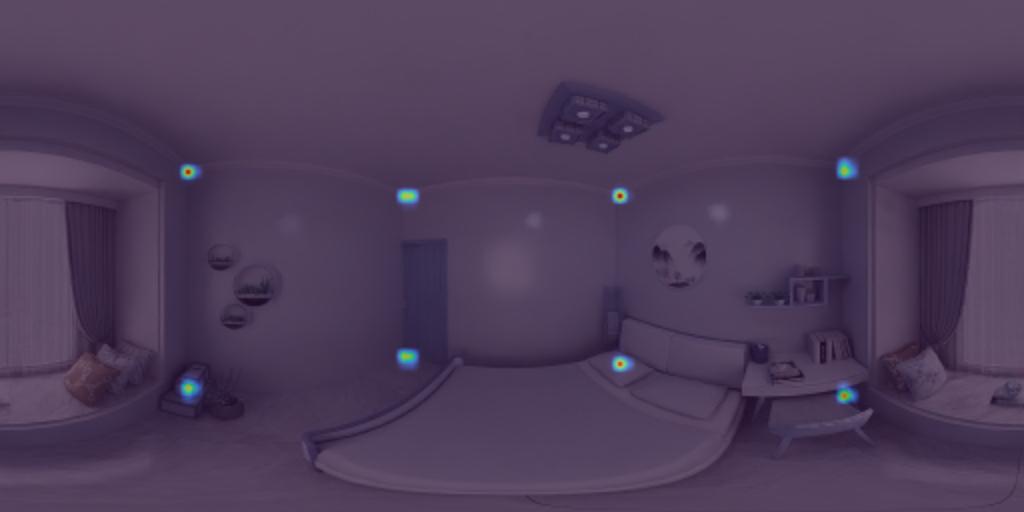}}
    \hfill
    \subfloat{\includegraphics[width=0.24\linewidth]{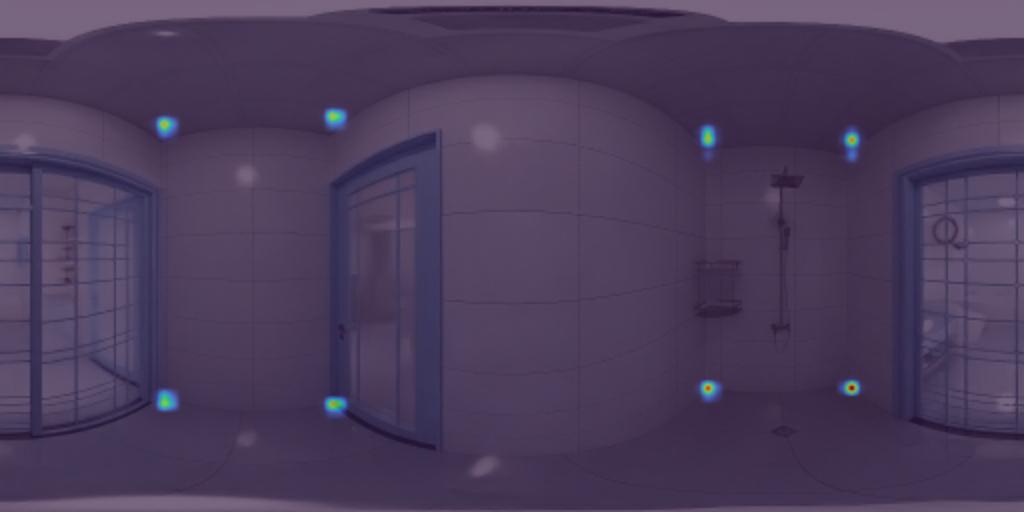}}
    \hfill
    \subfloat{\includegraphics[width=0.24\linewidth]{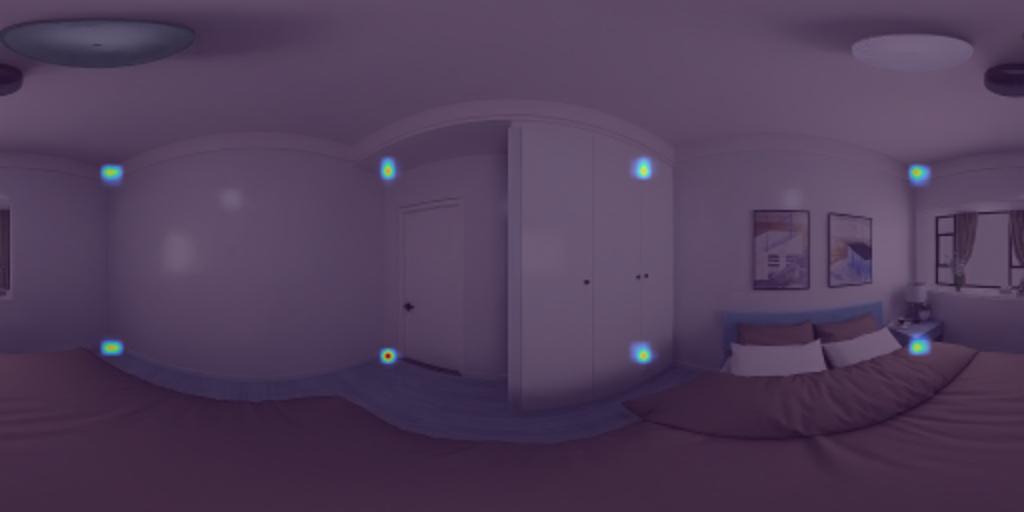}}
    \hfill
    \subfloat{\includegraphics[width=0.24\linewidth]{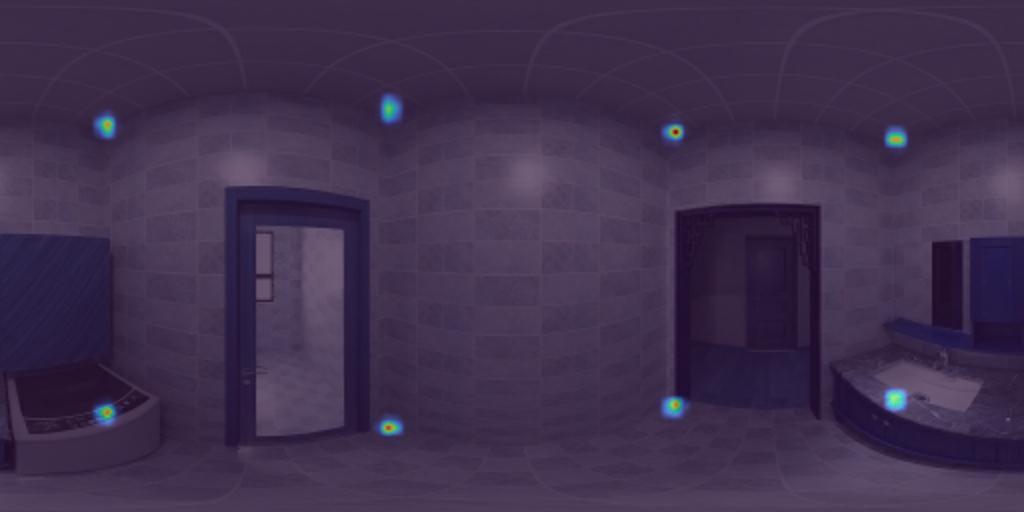}}
    \hfill
    
    \caption{
        Additional qualitative results on the Kujiale dataset.
    }
\label{fig:kuj}
\end{figure*}

In this supplementary material we present additional information regarding runtime and floating point operations, with the data offered in Table~\ref{tab:efficiency}, and illustrated in Figure~\ref{fig:efficiency}.
Apart from the models presented in the main document, we also add efficient CFL models for completeness.
In addition, we provide evaluation results for the Stanford2D3D and PanoContext datasets separately, in Tables~\ref{tab:s2d3d_ablation} and \ref{tab:pc_ablation} respectively.
Further, in Tables~\ref{tab:s2d3d_model_ablation}, \ref{tab:pc_model_ablation}, and \ref{tab:real_model_ablation}, we offer a decomposed model ablation for the Stanford2D3D, the PanoContext, and both datasets (averaged) respectively, where each individual component is ablated (namely, pre-activated bottlenecks, spherical padding, and anti-aliased max pooling).
The pre-activated residual blocks offer the larger gains, followed by the padding and finally, the anti-aliased max pooling.
Nonetheless, each different component is contributing to increased performance, with their combined effect being the most significant as observed by the model without all of these components together.
Figures~\ref{fig:s2d3d}, \ref{fig:pc}, \ref{fig:s3d} and \ref{fig:kuj} present additional qualitative results of our single-shot, end-to-end Manhattan aligned layout estimation model using the joint homography head module in Stanford2D3D, PanoContext, Structured3D and Kujiale datasets respectively.
Finally, Figures~\ref{fig:real_anim} and \ref{fig:syn_anim} present the qualitative samples from the real and synthetic datasets respectively, which are included in the main manuscript in animated 3D views (can only be viewed in recent Adobe Acrobat Reader versions).

\begin{table*}[!htbp]
\centering
\caption{
This table presents model complexity measures (multiply-accumulate giga-operations per inference, millions of parameter counts, runtime performance) as well as accuracy (IoU3D) on real domain datasets.
This table's reported values are used to generate Figure~\ref{fig:efficiency}.
}
\label{tab:efficiency}
\vspace{-0.25cm}
\begin{tabular}{@{}cc|cc|cc|r@{}}
\toprule
Method       & Variant      & MACS            & Parameters     & CPU             & GPU             & \multicolumn{1}{l}{IoU3D} \\ \midrule
SSC          & HG-3          & \first{17.61G}          & \first{6.35M}          & \third{1.78s}           & 0.085s          & \first{85.89\%}          \\
LayoutNet v2 & ResNet18    & 76.12G          & \third{15.57M}         & 11.65s          & \second{0.034s} & 83.83\%                   \\
LayoutNet v2 & ResNet34    & 95.48G          & 25.68M         & 12.97s          & 0.044s        & 84.60\%                   \\
LayoutNet v2 & ResNet50    & 607.43G         & 91.50M         & 34.63s          & 0.130s          & 82.55\%                   \\
DuLa-Net v2  & ResNet18    & 46.76G          & 25.64M         & 4.99s           & \second{0.037s}          & 83.68\%                   \\
DuLa-Net v2  & ResNet34    & 75.79G          & 45.86M         & 6.46s           & 0.049s          & \third{84.93\%}                   \\
DuLa-Net v2  & ResNet50    & 93.53G          & 57.38M         & 7.22s           & 0.072s          & \second{85.19\%}                   \\
HorizonNet   & ResNet18    & \second{23.03G}             & 23.49M            & N/As            & N/As             & 80.43\%                   \\
HorizonNet   & ResNet34    & 42.38G            & 33.59M            & N/As            & N/As             & 80.87\%                   \\
HorizonNet   & ResNet50    & 71.70G          & 81.57M         & 3.21s          & 0.063s          & 82.68\%                   \\
CFL          & EfficientNet & \third{42.19G}          & \second{11.69M}         & \first{0.074s} & \first{0.028s}          & N/A\%   \\                
CFL          & ResNet50 & N/A          & N/A         & \second{0.420s} & 0.052s          & 78.79\% \\
\bottomrule
\end{tabular}
\end{table*}

\begin{table*}[!htbp]
\centering
\caption{Ablation results on the Stanford2D3D dataset.}
\vspace{-0.25cm}
\label{tab:s2d3d_ablation}
\resizebox{\textwidth}{!}{%
\begin{tabular}{|c|cccc|cccccc|cc|}
\hline
Variant &
  \down{CE} &
  \up{IoU2D} &
  \up{IoU3D} &
  \down{PE} &
  \up{$J_5$} &
  \up{$J_{10}$} &
  \up{$J_{15}$} &
  \up{$W_5$} &
  \up{$W_{10}$} &
  \up{$W_{15}$} &
  \down{RMSE} &
  \up{$\delta_1$} \\ \hline
Quasi-Manhattan &
  \second{0.56\%} &
  88.18\% &
  85.16\% &
  1.83\% &
  \first{64.82\%} &
  \second{83.41\%} &
  89.82\% &
  \first{33.55\%} &
  \first{62.32\%} &
  \second{75.81\%} &
  0.1787 &
  95.59\% \\ \hline
w/ Homography (joint) &
  \first{0.51\%} &
  \first{89.83\%} &
  \first{87.80\%} &
  \first{1.62\%} &
  \second{64.27\%} &
  \first{85.07\%} &
  \first{92.70\%} &
  \third{27.65\%} &
  \second{62.02\%} &
  \first{77.36\%} &
  \first{0.1402} &
  \first{98.28\%} \\
w/ Homography (floor) &
  0.59\% &
  \second{89.51\%} &
  \second{87.56\%} &
  \second{1.80\%} &
  54.87\% &
  81.86\% &
  90.27\% &
  23.01\% &
  52.73\% &
  70.35\% &
  \second{0.1474} &
  \second{97.95\%} \\
w/ Homography (ceil) &
  \third{0.58\%} &
  \third{89.04\%} &
  \third{87.04\%} &
  \third{1.82\%} &
  \third{60.29\%} &
  \third{82.74\%} &
  \second{91.81\%} &
  \second{27.88\%} &
  \third{57.52\%} &
  \third{74.04\%} &
  \third{0.1539} &
  \third{97.36\%} \\ \hline
w/o Geodesics &
  0.80\% &
  84.72\% &
  81.73\% &
  2.34\% &
  32.19\% &
  67.70\% &
  87.28\% &
  4.13\% &
  25.15\% &
  49.71\% &
  0.2213 &
  95.31\% \\
w/o Model Adaptation &
  0.62\% &
  87.77\% &
  84.47\% &
  1.85\% &
  59.40\% &
  79.20\% &
  89.60\% &
  25.96\% &
  54.35\% &
  69.76\% &
  0.1815 &
  97.10\% \\
w/o Quasi-Manhattan &
  0.60\% &
  87.50\% &
  84.72\% &
  1.86\% &
  58.30\% &
  79.76\% &
  \third{90.49\%} &
  19.32\% &
  54.79\% &
  71.17\% &
  0.1825 &
  95.97\% \\ \hline
\end{tabular}%
}
\end{table*}
\begin{table*}[!htbp]
\centering
\caption{Ablation results on the PanoContext dataset.}
\label{tab:pc_ablation}
\vspace{-0.25cm}
\resizebox{\textwidth}{!}{%
\begin{tabular}{|c|cccc|cccccc|cc|}
\hline
Variant &
  \down{CE} &
  \up{IoU2D} &
  \up{IoU3D} &
  \down{PE} &
  \up{$J_5$} &
  \up{$J_{10}$} &
  \up{$J_{15}$} &
  \up{$W_5$} &
  \up{$W_{10}$} &
  \up{$W_{15}$} &
  \down{RMSE} &
  \up{$\delta_1$} \\ \hline
Quasi-Manhattan &
  \first{0.53\%} &
  \first{87.63\%} &
  \first{84.89\%} &
  \first{1.65\%} &
  \first{58.25\%} &
  \first{84.67\%} &
  \first{92.22\%} &
  \first{27.67\%} &
  \first{53.62\%} &
  \first{74.06\%} &
  \first{0.1682} &
  96.68\% \\ \hline
w/ Homography (joint) &
  \third{0.63\%} &
  86.95\% &
  \third{83.97\%} &
  \third{1.78\%} &
  45.75\% &
  76.18\% &
  90.80\% &
  \third{14.31\%} &
  43.24\% &
  64.78\% &
  \third{0.1711} &
  \second{97.58\%} \\
w/ Homography (floor) &
  0.76\% &
  \second{87.00\%} &
  \second{84.39\%} &
  2.02\% &
  39.15\% &
  71.46\% &
  87.03\% &
  9.59\% &
  33.49\% &
  56.13\% &
  \second{0.1708} &
  \third{97.09\%} \\
w/ Homography (ceil) &
  0.68\% &
  86.22\% &
  83.47\% &
  1.93\% &
  45.28\% &
  \third{80.42\%} &
  \third{91.04\%} &
  9.12\% &
  \third{45.75\%} &
  \third{65.41\%} &
  0.1803 &
  \first{97.92\%} \\ \hline
w/o Geodesics &
  0.78\% &
  84.08\% &
  80.43\% &
  2.27\% &
  35.38\% &
  72.41\% &
  86.32\% &
  5.19\% &
  27.52\% &
  57.08\% &
  0.2252 &
  95.64\% \\
w/o Model Adaptation &
  0.68\% &
  85.42\% &
  81.41\% &
  2.11\% &
  \second{51.42\%} &
  77.12\% &
  87.50\% &
  \second{19.50\%} &
  \second{46.07\%} &
  63.21\% &
  0.2250 &
  94.13\% \\
w/o Quasi-Manhattan &
  \second{0.61\%} &
  \third{86.99\%} &
  83.46\% &
  \second{1.77\%} &
  \third{50.00\%} &
  \second{80.66\%} &
  \second{91.51\%} &
  13.21\% &
  45.28\% &
  \second{68.08\%} &
  0.1922 &
  96.66\% \\ \hline
\end{tabular}%
}
\end{table*}
\begin{table*}[!htbp]
\centering
\caption{Model ablation results on the Stanford2D3D dataset.}
\label{tab:s2d3d_model_ablation}
\vspace{-0.25cm}
\resizebox{\textwidth}{!}{%
\begin{tabular}{|c|cccc|cccccc|cc|}
\hline
Variant &
  \down{CE} &
  \up{IoU2D} &
  \up{IoU3D} &
  \down{PE} &
  \up{$J_5$} &
  \up{$J_{10}$} &
  \up{$J_{15}$} &
  \up{$W_5$} &
  \up{$W_{10}$} &
  \up{$W_{15}$} &
  \down{RMSE} &
  \up{$\delta_1$} \\ \hline
Quasi-Manhattan &
  \first{0.56\%} &
  \first{88.18\%} &
  \first{85.16\%} &
  \third{1.83\%} &
  \first{64.82\%} &
  \second{83.41\%} &
  \third{89.82\%} &
  \first{33.55\%} &
  \first{62.32\%} &
  \first{75.81\%} &
  \third{0.1787} &
  95.59\% \\
w/o Model Adaptation &
  \third{0.62\%} &
  87.77\% &
  84.47\% &
  1.85\% &
  59.40\% &
  79.20\% &
  89.60\% &
  25.96\% &
  54.35\% &
  69.76\% &
  0.1815 &
  \second{97.10\%} \\ \hline
w/o Pre-activated &
  \second{0.58\%} &
  87.93\% &
  \third{85.09\%} &
  1.86\% &
  \second{63.94\%} &
  \third{83.08\%} &
  \third{89.82\%} &
  28.24\% &
  \third{61.73\%} &
  74.34\% &
  \second{0.1748} &
  96.05\% \\
w/o Padding &
  \first{0.56\%} &
  \second{88.10\%} &
  85.07\% &
  \first{1.73\%} &
  \first{64.82\%} &
  \first{83.85\%} &
  \first{90.71\%} &
  \second{30.60\%} &
  \second{62.02\%} &
  \third{75.22\%} &
  0.1788 &
  \second{96.61\%} \\
w/o Anti-aliasing &
  \first{0.56\%} &
  \third{88.06\%} &
  \second{85.15\%} &
  \second{1.76\%} &
  \third{62.50\%} &
  82.52\% &
  \second{90.60\%} &
  \third{28.32\%} &
  60.03\% &
  \second{75.29\%} &
  \first{0.1721} &
  \first{97.11\%} \\ \hline
\end{tabular}%
}
\end{table*}
\begin{table*}[!htbp]
\centering
\caption{Model ablation results on the PanoContext dataset.}
\label{tab:pc_model_ablation}
\vspace{-0.25cm}
\resizebox{\textwidth}{!}{%
\begin{tabular}{|c|cccc|cccccc|cc|}
\hline
Variant &
  \down{CE} &
  \up{IoU2D} &
  \up{IoU3D} &
  \down{PE} &
  \up{$J_5$} &
  \up{$J_{10}$} &
  \up{$J_{15}$} &
  \up{$W_5$} &
  \up{$W_{10}$} &
  \up{$W_{15}$} &
  \down{RMSE} &
  \up{$\delta_1$} \\ \hline
Quasi-Manhattan &
  \first{0.53\%} &
  \first{87.63\%} &
  \first{84.89\%} &
  \first{1.65\%} &
  \first{58.25\%} &
  \first{84.67\%} &
  \first{92.22\%} &
  \first{27.67\%} &
  \first{53.62\%} &
  \first{74.06\%} &
  \first{0.1682} &
  \second{96.68\%} \\
w/o Model Adaptation &
  0.68\% &
  85.42\% &
  81.41\% &
  2.11\% &
  51.42\% &
  77.12\% &
  87.50\% &
  19.50\% &
  46.07\% &
  63.21\% &
  0.2250 &
  94.13\% \\ \hline
w/o Pre-activated &
  0.62\% &
  \second{87.48\%} &
  \third{83.68\%} &
  \third{1.81\%} &
  \third{55.90\%} &
  79.48\% &
  \third{88.92\%} &
  20.60\% &
  \second{51.10\%} &
  68.40\% &
  \third{0.1793} &
  \third{95.85\%} \\
w/o Padding &
  \third{0.61\%} &
  \third{85.96\%} &
  82.84\% &
  1.96\% &
  \second{56.13\%} &
  \third{82.31\%} &
  87.97\% &
  \third{24.69\%} &
  \third{50.94\%} &
  \second{70.28\%} &
  0.2043 &
  95.18\% \\
w/o Anti-aliasing &
  \second{0.55\%} &
  \second{87.48\%} &
  \second{84.41\%} &
  \second{1.76\%} &
  55.42\% &
  \second{82.78\%} &
  \second{91.98\%} &
  \second{27.52\%} &
  \second{51.10\%} &
  \third{69.50\%} &
  \second{0.1745} &
  \first{96.72\%} \\ \hline
\end{tabular}%
}
\end{table*}
\begin{table*}[!htbp]
\centering
\caption{Average model ablation results on both the real datasets.}
\label{tab:real_model_ablation}
\vspace{-0.25cm}
\resizebox{\textwidth}{!}{%
\begin{tabular}{|c|cccc|cccccc|cc|}
\hline
Variant &
  \down{CE} &
  \up{IoU2D} &
  \up{IoU3D} &
  \down{PE} &
  \up{$J_5$} &
  \up{$J_{10}$} &
  \up{$J_{15}$} &
  \up{$W_5$} &
  \up{$W_{10}$} &
  \up{$W_{15}$} &
  \down{RMSE} &
  \up{$\delta_1$} \\ \hline
Quasi-Manhattan &
  \first{0.55\%} &
  \first{87.90\%} &
  \first{85.02\%} &
  \first{1.74\%} &
  \first{61.54\%} &
  \first{84.04\%} &
  \second{91.02\%} &
  \first{30.61\%} &
  \first{57.97\%} &
  \first{74.93\%} &
  \second{0.1734} &
  \second{96.14\%} \\
w/o Model Adaptation &
  0.65\% &
  86.60\% &
  82.94\% &
  1.98\% &
  55.41\% &
  78.16\% &
  88.55\% &
  22.73\% &
  50.21\% &
  66.49\% &
  0.2033 &
  95.61\% \\ \hline
w/o Pre-activated &
  0.60\% &
  \third{87.70\%} &
  \third{84.39\%} &
  \third{1.84\%} &
  \third{59.92\%} &
  81.28\% &
  \third{89.37\%} &
  24.42\% &
  \third{56.41\%} &
  71.37\% &
  \third{0.1771} &
  \third{95.95\%} \\
w/o Padding &
  \third{0.59\%} &
  87.03\% &
  83.95\% &
  \third{1.84\%} &
  \second{60.48\%} &
  \second{83.08\%} &
  89.34\% &
  \third{27.65\%} &
  \second{56.48\%} &
  \second{72.75\%} &
  0.1916 &
  95.90\% \\
w/o Anti-aliasing &
  \second{0.56\%} &
  \second{87.77\%} &
  \second{84.78\%} &
  \second{1.76\%} &
  58.96\% &
  \third{82.65\%} &
  \first{91.29\%} &
  \second{27.92\%} &
  55.57\% &
  \third{72.40\%} &
  \first{0.1733} &
  \first{96.92\%} \\ \hline
\end{tabular}%
}
\end{table*}


\begin{figure*}[!ht]
    \centering
    \subfloat{\animategraphics[autoplay,palindrome,width=0.32\linewidth]{15}{figs/supp/anim/cfl/color_pred_mesh26_}{90}{179}}
    \hfill
    \subfloat{\animategraphics[autoplay,palindrome,width=0.32\linewidth]{15}{figs/supp/anim/cfl/color_pred_mesh17_}{90}{179}}
    \hfill
    \subfloat{\animategraphics[autoplay,palindrome,width=0.32\linewidth]{15}{figs/supp/anim/cfl/color_pred_mesh28_}{90}{179}}
    \hfill
    
    \subfloat{\animategraphics[autoplay,palindrome,width=0.32\linewidth]{15}{figs/supp/anim/s2d3d/color_pred_mesh33_}{90}{179}}
    \hfill
    \subfloat{\animategraphics[autoplay,palindrome,width=0.32\linewidth]{15}{figs/supp/anim/s2d3d/color_pred_mesh47_}{90}{179}}
    \hfill
    \subfloat{\animategraphics[autoplay,palindrome,width=0.32\linewidth]{15}{figs/supp/anim/s2d3d/color_pred_mesh60_}{90}{179}}
    \hfill
    
    \caption{
        Animated renderings of the 3D qualitative results of the real datasets as presented in the figures of the main manuscript. Top row samples are from PanoContext, bottom row samples are from Stanford2D3D. (animations are only playable in recent Adobe Acrobat Reader versions).
    }
\label{fig:real_anim}
\end{figure*}
\begin{figure*}[!ht]
    \centering
    \subfloat{\animategraphics[autoplay,palindrome,width=0.32\linewidth]{15}{figs/supp/anim/s3d/color_pred_mesh36_}{90}{179}}
    \hfill
    \subfloat{\animategraphics[autoplay,palindrome,width=0.32\linewidth]{15}{figs/supp/anim/s3d/color_pred_mesh441_}{90}{179}}
    \hfill
    \subfloat{\animategraphics[autoplay,palindrome,width=0.32\linewidth]{15}{figs/supp/anim/s3d/color_pred_mesh960_}{90}{179}}
    \hfill
    
    \subfloat{\animategraphics[autoplay,palindrome,width=0.32\linewidth]{15}{figs/supp/anim/kuj/color_pred_mesh5_}{90}{179}}
    \hfill
    \subfloat{\animategraphics[autoplay,palindrome,width=0.32\linewidth]{15}{figs/supp/anim/kuj/color_pred_mesh28_}{90}{179}}
    \hfill
    \subfloat{\animategraphics[autoplay,palindrome,width=0.32\linewidth]{15}{figs/supp/anim/kuj/color_pred_mesh190_}{90}{179}}
    \hfill
    
    \caption{
        Animated renderings of the 3D qualitative results of the synthetic datasets as presented in the figures of the main manuscript. Top rows samples are from Structured3D, bottom row samples are from Kujiale. (animations are only playable in recent Adobe Acrobat Reader versions).
    }
\label{fig:syn_anim}
\end{figure*}

\end{document}